\pdfoutput=1
\documentclass{article}

\PassOptionsToPackage{table}{xcolor}
\usepackage{hyperref}
\let\klevelAddContentsLine\addcontentsline

\usepackage[preprint]{icml2026}
\newif\ifshowremarks
\showremarksfalse
\newif\ifshownotation
\shownotationfalse
\usepackage{amsmath}
\usepackage{amssymb}
\usepackage{amsthm}
\usepackage{mathtools}
\usepackage{array}
\usepackage{booktabs}
\usepackage{multirow}
\usepackage{longtable}
\usepackage{enumitem}
\usepackage[patch=none]{microtype}
\usepackage[table]{xcolor}
\usepackage{algorithm}
\usepackage{algpseudocode}
\usepackage{thmtools}
\usepackage{thm-restate}
\usepackage{environ}
\usepackage{tikz}
\usetikzlibrary{arrows.meta,positioning,decorations.pathreplacing}
\usepackage{pgfplots}
\pgfplotsset{compat=1.17}

\setlist[itemize]{topsep=3pt,itemsep=2pt,parsep=0pt}
\setlist[enumerate]{topsep=3pt,itemsep=4pt,parsep=0pt,wide=0pt}

\newtheorem{theorem}{Theorem}
\newtheorem{proposition}[theorem]{Proposition}
\newtheorem{lemma}[theorem]{Lemma}
\newtheorem{corollary}[theorem]{Corollary}
\newtheorem{formaltheoreminner}{Theorem}
\newenvironment{formaltheorem}[1]{\renewcommand{\theformaltheoreminner}{\ref*{#1}}\formaltheoreminner}{\endformaltheoreminner}
\newtheorem{formalpropositioninner}{Proposition}
\newenvironment{formalproposition}[1]{\renewcommand{\theformalpropositioninner}{\ref*{#1}}\formalpropositioninner}{\endformalpropositioninner}
\theoremstyle{definition}
\newtheorem{definition}{Definition}
\newtheorem{assumption}{Assumption}
\newtheorem{remark}{Remark}
\ifshowremarks\else
  \RenewEnviron{remark}[1][]{}
\fi

\providecommand{\todo}[1]{}
\renewcommand{\todo}[1]{\textcolor{red}{#1}}

\newcommand{\factNcases}{24}
\newcommand{\factNconverged}{18}
\newcommand{\factNgoal}{16}

\newcommand{\factTMax}{3{,}000{,}000}
\newcommand{\factStallPct}{2}

\newcommand{\factNKsixteen}{8}
\newcommand{\factMaxGoalKsixteen}{0.34}

\newcommand{\factNeFive}{18}
\newcommand{\factUnrollWins}{17}
\newcommand{\factImplicitLower}{10}
\newcommand{\factImplicitMaxMarginPct}{56}

\newcommand{\factHardRatioMin}{1.3}
\newcommand{\factHardRatioMax}{1.8}
\newcommand{\factRobKsixteenPass}{15}
\newcommand{\factRobKsixteenTotal}{32}

\newcommand{\factRobKsixteenGoal}{30}
\newcommand{\factRobKsixteenMiss}{2}
\newcommand{\factRobNearMissGoal}{0.426}

\newcommand{\factNAdjusted}{8}
\newcommand{\factMemRatioMin}{27}
\newcommand{\factMemRatioMax}{44}

\newcommand{\factTimeBudget}{90}
\newcommand{\factNScreenRuns}{934}
\newcommand{\factWreg}{0.01}
\newcommand{\factWvel}{0.3}
\newcommand{\factBandHalfGoal}{13}
\newcommand{\factBandEighthGoal}{9}
\newcommand{\factReachCorrKfour}{1.43}
\newcommand{\factReachAttempts}{96}
\newcommand{\factReachZigKfour}{5.97}
\newcommand{\factReachZigKeight}{2.68}

\usepackage{prettyref}
\usepackage{refcount}
\MakeRobust\prettyref
\makeatletter
\newcommand{\klevel@secword}[1]{\begingroup\edef\klevel@num{\getrefnumber{#1}}\expandafter\klevel@secword@\klevel@num\relax\endgroup}
\def\klevel@secword@#1#2\relax{\ifcat a\noexpand#1Appendix\else Section\fi}
\newrefformat{sec}{\klevel@secword{#1}~\ref{#1}}
\newrefformat{sub}{\klevel@secword{#1}~\ref{#1}}
\makeatother
\newrefformat{fig}{Figure~\ref{#1}}
\newrefformat{tab}{Table~\ref{#1}}
\newrefformat{thm}{Theorem~\ref{#1}}
\newrefformat{lem}{Lemma~\ref{#1}}
\newrefformat{cor}{Corollary~\ref{#1}}
\newrefformat{prop}{Proposition~\ref{#1}}
\newrefformat{def}{Definition~\ref{#1}}
\newrefformat{ass}{Assumption~\ref{#1}}
\newrefformat{rem}{Remark~\ref{#1}}
\newrefformat{alg}{Algorithm~\ref{#1}}
\newrefformat{eq}{\eqref{#1}}
\newrefformat{icml}{\eqref{#1}}
\newrefformat{ns}{Assumption~\ref{ass:ns}.\ref{#1}}
\newrefformat{nscb}{Assumption~\ref{ass:nscb}.\ref{#1}}
\newrefformat{A}{Assumption~\ref{ass:A}.\ref{#1}}
\newrefformat{cblay}{\ref{#1}}
\newrefformat{id}{\ref{#1}}
\newrefformat{sp}{\ref{#1}}
\newrefformat{os}{\ref{#1}}

\usepackage{graphicx}
\usepackage{wrapfig}
\newlength{\resizeEqWidth}
\newcommand{\resizeEq}[2][\dimexpr\linewidth-2.8em\relax]{\setlength{\resizeEqWidth}{#1}\resizebox{\ifdim\width>\resizeEqWidth\resizeEqWidth\else\width\fi}{!}{$\displaystyle #2$}}

\makeatletter
\newcommand{\settag}[1]{\protected@edef\@currentlabel{#1}}
\makeatother

\newcommand{\cX}{\mathcal X}

\newcommand{\cW}{\mathcal W}
\newcommand{\cL}{\mathcal L}
\newcommand{\cG}{\mathcal G}
\newcommand{\cM}{\mathcal M}
\newcommand{\R}{\mathbb R}

\newcommand{\PiX}{\operatorname{proj}}
\DeclareMathOperator*{\argmin}{arg\,min}

\newcommand{\Phib}{\bar\Phi_\rho}

\newcommand{\Flo}{F_{\rm lo}}
\newcommand{\Gphi}{\cG^{\varphi}}

\newcommand{\cT}{\mathcal T}
\newcommand{\Ecal}{\mathfrak E}

\newcommand{\hnabla}{\hat\nabla}

\newcommand{\Ball}{\mathbb B}
\newcommand{\cE}{\mathcal E}
\newcommand{\pos}{\mathrm{pos}}
\newcommand{\vel}{\mathrm{vel}}
\newcommand{\SelMat}{\mathbf S}

\icmltitlerunning{A Matrix-Free Single-Loop Method for \texorpdfstring{$K$}{K}-Level Control}

\begin{document}

\twocolumn[
  \icmltitle{PSK: A Matrix-Free Single-Loop Method for \texorpdfstring{$K$}{K}-Level Physics-Based Control}

  \begin{icmlauthorlist}
    \icmlauthor{Zherong Pan}{meta}
  \end{icmlauthorlist}

  \icmlaffiliation{meta}{Meta Reality Labs}
  \icmlcorrespondingauthor{Zherong Pan}{zrpan@meta.com}
  \icmlkeywords{multilevel optimization, physics-based control, first-order methods, multi-robot navigation}

  \vskip 0.3in
]

\printAffiliationsAndNotice{}

\begin{abstract}
  Physics-based control is central to robot learning, but differentiating through implicit physics steps typically requires Hessian or Karush--Kuhn--Tucker (KKT) linear solves; we ask whether gradient steps suffice.  Each simulation step minimizes an energy under many contact and collision constraints.  Without constraints, we view control over $K$ steps as $K$-level optimization with strongly convex lower levels, recast as one min--max problem.  Our Penalized Single-loop $K$-level method (PSK) uses only gradients and a projection, updates all levels in parallel, and reaches an $\epsilon$-stationary control in $O(\epsilon^{-3})$ iterations, with a $K$-independent exponent but a $K$-dependent constant.  Contact-rich constraints, however, typically violate a constraint qualification (CQ).  We replace each constraint by a smooth wall that acts on a band of constant width.  Under mild assumptions that need no CQ, every exact solution of the banded lower levels is then feasible and banded Fritz--John, and PSK keeps the $O(\epsilon^{-3})$ complexity.  Without a CQ that is uniform in the band width, however, a shrinking band loses the $K$-independent exponent: with width $\epsilon$ and fixed $K\geq2$, deterministic zero-respecting methods need $\Omega(\epsilon^{-(K+3)/4})$ calls on a hard instance whose control dimension grows with $1/\epsilon$.  We prove our assumptions for smoothed kinematic and dynamic multi-robot navigation among segment obstacles.  In navigation experiments, PSK, tuned per case, reaches every goal with sixteen levels and, at equal time, reaches a lower loss than tuned unrolling in \factUnrollWins{} of \factNeFive{} cases, while implicit differentiation is lower in \factImplicitLower{}.

\end{abstract}

\section{Introduction}\label{sec:intro}

Many machine-learning and robotics problems optimize controls over chains of physical-system states: a robot's actions change states through a physics model.  Differentiable physics passes gradients through the simulator to train controls, policies, and physical parameters end to end \citep{degrave2019differentiable,belbuteperes2018end,hu2020difftaichi,freeman2021brax,werling2021fast}.  It underpins model-based reinforcement learning \citep{deisenroth2011pilco,janner2019when,hafner2020dream,hansen2022temporal,xu2022accelerated}, physical world models \citep{battaglia2016interaction,sanchezgonzalez2020learning,li2019learning,ha2018world}, and real-to-sim parameter fitting \citep{ramos2019bayessim,jatavallabhula2021gradsim,heiden2021neuralsim,lelidec2021differentiable}. Each physics step is itself an optimization problem: implicit integration minimizes an energy, often under contact or collision constraints.  Newton-type solvers solve a Hessian or Karush--Kuhn--Tucker (KKT) linear system at every step, but local projections, accelerated fixed-point iterations, and gradient descent reach simulation accuracy without one \citep{macklin2016xpbd,wang2015chebyshev,wang2016descent}.  Control, however, needs each state's derivative with respect to the controls, which implicit-function and adjoint methods obtain from a Hessian or KKT solve \citep{amos2017optnet,amos2018differentiable,geilinger2020add,du2021diffpd,howell2022dojo}.  Thus matrix-free simulation need not yield matrix-free control.  We ask:
\begin{center}
\fbox{\parbox{0.92\columnwidth}{\centering\emph{Can we solve physics-based control with gradient steps alone, without building or inverting any matrix?}}}
\end{center}

We give an affirmative answer for a constant-band surrogate by reformulating physics-based control as a $K$-level optimization problem.  We model the $K$ simulation steps as strongly convex constrained problems driven by one global control, with stage-dependent energies and shared constraints.  Our contributions follow.
\begin{itemize}[leftmargin=1.25em]
\item \textbf{A matrix-free method.}  We replace each hard constraint by a smooth wall that acts on a band of constant width, and cast the resulting surrogate as one min--max problem.  We propose the Penalized Single-loop $K$-level method (PSK), which uses only gradients and a projection onto the control set.  It builds no Hessian, Jacobian, or linear system, and its stages update in parallel (\prettyref{sec:results}).
\item \textbf{A $K$-independent exponent for constant bands.}  We design mild stage assumptions that require no constraint qualification (CQ) and under which every exact surrogate state is feasible and banded Fritz--John.  Under these assumptions, PSK reaches an $\epsilon$-stationary control of the surrogate in $O(\epsilon^{-3})$ iterations for every $K$; the exponent is independent of $K$, although the constant depends on $K$ (\prettyref{sec:results}).
\item \textbf{A $K$-dependent lower bound for shrinking bands.}  A CQ rarely holds for contact-rich physics.  Without a CQ that is uniform in the band width, we prove a lower bound: when the band width shrinks with the target accuracy $\epsilon$, every deterministic zero-respecting method, including PSK, needs $\Omega(\epsilon^{-(K+3)/4})$ oracle calls on a hard instance for each fixed $K\geq2$.  Hence the exponent grows linearly with $K$.  The instance meets all other assumptions (\prettyref{prop:lbcbbase}), but its two constraint normals become parallel as the band shrinks (\prettyref{sec:shrink}).
\item \textbf{Practical assumptions and experiments.}  Our assumptions hold for kinematic and dynamic multi-robot navigation among segment obstacles with smoothed normals.  On four scenes, PSK reaches every goal with $K=16$ and, at equal time, reaches a lower loss than tuned unrolling in \factUnrollWins{} of \factNeFive{} cases, while implicit differentiation is lower in \factImplicitLower{} (\prettyref{sec:application} and \ref{sec:experiments}).
\end{itemize}

\section{Related Work}\label{sec:related}

\paragraph{Optimization methods.} PSK is inspired by bilevel optimization and its recent matrix-free methods.  Value-gap penalties track two lower-level solutions \citep{kwon2023fully,kwon2024penalty} or solve the resulting structured min--max problem \citep{lu2024firstorder}, and extensions cover general, minimax, or linearly constrained lower levels under, respectively, existing multipliers, a uniform Slater condition, or a full-rank constraint matrix \citep{yao2024overcoming,shen2026penalty,shen2025singleloop}; single-step methods instead track adjoints under local regularity \citep{suonpera2023linearly,suonpera2025singleloop,suonpera2026nonsmooth}.  Most of this work treats two levels; multilevel methods unroll every level \citep{sato2021gradient}, combine monotone maps \citep{shafiei2021trilevel}, or interleave inner and reduced-adjoint estimates \citep{dizon2024differential}, and \citet{zhang2026distributed} reach $O(\epsilon^{-3})$ for stochastic $K$-level problems with unconstrained strongly convex levels, using Hessian- and Jacobian-vector products.  Multilevel compositional methods reach level-independent sample-complexity exponents for nested explicit maps \citep{balasubramanian2022stochastic,jiang2022optimal}, but query each level's Jacobian, which for argmin levels needs a matrix solve.  PSK extends the value-gap penalty to $K$ levels with shared hard constraints that depend on the previous state, handles them by a constant band without a linear independence CQ (LICQ), strict complementarity, or assumed multipliers, and needs no adjoint, dual variable, or second-order oracle.

Multi-block splitting methods, such as the alternating direction method of multipliers \citep{boyd2011distributed,hong2017linear,lin2015sublinear,aspelmeier2016local,raguet2013generalized,bastianello2020distributed,zhang2021online,yuan2025admm}, and min--max methods \citep{bot2023alternating,xu2023unified,zhang2025iterative,shi2026nonsmooth,tapiariera2026convergence,hu2024singleloop} alternate updates among the blocks of one flat problem.  In PSK, each block instead solves a stage problem driven by the previous one, so $K$ enters our upper bound through its constant and our lower bound through its exponent.

\paragraph{Lower bounds.} Our hard instance builds on the smooth zero chain of \citet{carmon2020lower}.  Related oracle lower bounds treat nonsmooth stationarity \citep{zhang2020complexity,kornowski2022oracle,jordan2022complexity,kornowski2026complexity} and nonconvex--Polyak--{\L}ojasiewicz min--max problems \citep{pan2026lower}, and two-level problems with an unconstrained convex lower level \citep{ji2023lower,chen2024finding}.  Our lower bound shows that, without a uniform CQ, the exponent grows with $K$.

\paragraph{Differentiable physics simulation and control.} Many simulators recast a time step as an optimization problem: they minimize an incremental potential \citep{gast2015optimization,karamouzas2017implicit,li2020incremental}, alternate local projections with a global solve \citep{bouaziz2014projective}, project constraints \citep{muller2007position,macklin2016xpbd}, or solve a convex contact problem \citep{todorov2012mujoco}.  Control over many steps follows three routes.  Unrolled simulation backpropagates through every step and inner iteration \citep{degrave2019differentiable,hu2020difftaichi,freeman2021brax}, so its memory grows with the number of unrolled steps and truncation biases its gradient; short-horizon policy learning mitigates this \citep{xu2022accelerated}.  Implicit differentiation applies the implicit-function theorem or the adjoint method to each step's optimality conditions \citep{amos2017optnet,belbuteperes2018end,geilinger2020add,werling2021fast,li2022diffcloth,howell2022dojo,lecleach2023single}, which needs a Hessian or KKT solve per step, even with reused Cholesky factors \citep{du2021diffpd} or Riccati recursions \citep{amos2018differentiable}.  Trajectory optimization keeps the states as variables and imposes the dynamics as constraints or penalties \citep{posa2014direct,mordatch2012discovery}.  PSK also keeps the states as variables, but enforces each step's optimality by a value-gap penalty that needs only first derivatives, avoiding matrix solves and unrolling.  To our knowledge, PSK is the first gradient-only single-loop method for constrained $K$-level problems with a proved $K$-independent exponent for constant bands; for shrinking bands, the exponent grows with $K$.

\section{Problem Statement}\label{sec:problem}
\ifshownotation See \prettyref{sec:notation} for a notation table.  \fi We control a physical system through $K$ simulation steps.  Fix a depth $K\geq1$, a control set $\cX\subset\R^{n_x}$, an initial state $y_0\in\R^{n_y}$, and a number $N_h\geq0$ of hard constraints.  For each stage $i$, fix a selection matrix $\SelMat_i\in\R^{n_{u,i}\times n_x}$.  Stage $i$ reads the selected control $\SelMat_i x$ and the previous state $y_{i-1}$.  We write the stage parameter as $a_i:=(x,y_{i-1})$ and use a stage energy $g_i(x,p,y)$ that depends on $x$ only through $\SelMat_i x$. Our $K$-level problem is
\begin{equation}\label{icml:klevel}
 \begin{aligned}
 \min_{x\in\cX}\ \ & F(x,y_1,\ldots,y_K)\\
 \text{s.t.}\ \ & y_i=s_i^B(a_i),\;\;
 s_i^B(a):=\begin{aligned}[t]
 &\argmin_{y\in\R^{n_y}}\ g_i(a,y)\\
 &\text{s.t. } h_l(p,y)\leq0.
 \end{aligned}
 \end{aligned}
\end{equation}
Here $a=(x,p)$ is a generic stage parameter with incoming state $p$, $1\leq i\leq K$ indexes the stages, and $1\leq l\leq N_h$ indexes the constraints.  $F$ is the task loss, $g_i$ the stage-$i$ energy, and the $h_l$ are shared hard constraints, such as non-penetration.  We write $g_i(a,y)=g_i(x,p,y)$ and $s_i^B(x,p):=s_i^B(a)$ for $a=(x,p)$.  The superscript $B$ marks the original hard-constrained step; \prettyref{sec:results} softens these constraints with a cubic wall, and the superscript $C$ marks the resulting surrogate.  In a physics model, each level (stage) is one implicit time step.  Under the conditions of \prettyref{sec:results}, each $g_i(a,\cdot)$ is strongly convex and every $h_l(p,\cdot)$ is convex, so $s_i^B(a)$ is unique whenever its feasible set is nonempty.

\paragraph{Nested formulation.} Differentiable simulation removes the states by nesting the steps.  It writes $\widehat y_i^B(x)$ for the state reached by the constrained chain and solves
\begin{equation}\label{eq:traj}
 \begin{aligned}
 \min_{x\in\cX}\quad
 &\varphi^B(x):=F(x,\widehat y_1^B(x),\ldots,\widehat y_K^B(x)),\\
 &\widehat y_0^B(x):=y_0,\qquad
 \widehat y_i^B(x):=s_i^B(x,\widehat y_{i-1}^B(x)).
 \end{aligned}
\end{equation}
A gradient method on $\varphi^B$ needs the derivative of every $\widehat y_i^B$ with respect to $x$.  Since each $\widehat y_i^B$ solves an optimization problem, implicit-function and adjoint methods obtain this derivative by solving a Hessian or KKT linear system at every step (\prettyref{sec:related}), which is the matrix computation we want to avoid.  These derivative formulas also hold only at exact step solutions, which iterative solvers do not reach in finite time, so the computed derivative is biased.  In \prettyref{sec:results}, we keep the states $y_i$ as variables and rewrite \eqref{icml:klevel} as a min--max problem that needs only gradients.

\subsection{Motivating problem: multi-robot navigation}\label{sub:motivating}

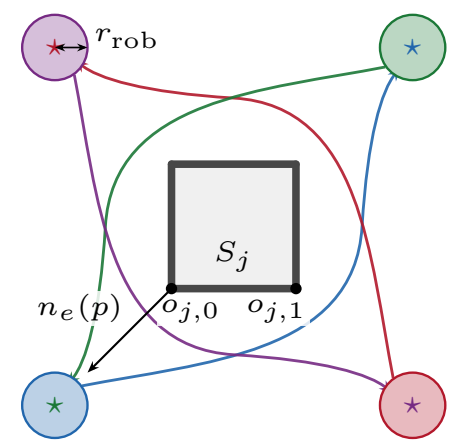
\begin{wrapfigure}{r}{0.46\columnwidth}
 \vspace{-14pt}
 \centering
 \resizebox{0.44\columnwidth}{!}{\begingroup
\definecolor{navA}{HTML}{2166AC}
\definecolor{navB}{HTML}{B2182B}
\definecolor{navC}{HTML}{1B7837}
\definecolor{navD}{HTML}{762A83}
\begin{tikzpicture}[
  x=0.07\columnwidth,
  y=0.07\columnwidth,
  traj/.style={line width=0.6pt,opacity=0.9,-{Stealth[length=3pt]}},
  obstacle/.style={line width=1.6pt,line cap=round,black!72},
  goal/.style={font=\scriptsize},
  lab/.style={font=\tiny,fill=white,fill opacity=0.85,text opacity=1,inner sep=0.6pt}
]
  \path[use as bounding box] (-0.15,-0.15) rectangle (6.15,6.15);

  \fill[black!6] (2.2,2.2) rectangle (3.8,3.8);
  \draw[obstacle] (2.2,2.2) -- (3.8,2.2);
  \draw[obstacle] (3.8,2.2) -- (3.8,3.8);
  \draw[obstacle] (3.8,3.8) -- (2.2,3.8);
  \draw[obstacle] (2.2,3.8) -- (2.2,2.2);

  \draw[traj,navA] (1.05,0.95) .. controls (2.6,1.25) and (4.55,1.35) .. (4.65,3.0)
    .. controls (4.72,4.15) and (4.90,4.75) .. (5.15,5.05);
  \draw[traj,navB] (5.05,1.05) .. controls (4.75,2.6) and (4.65,4.55) .. (3.0,4.65)
    .. controls (1.85,4.72) and (1.25,4.90) .. (0.95,5.15);
  \draw[traj,navC] (4.95,5.05) .. controls (3.4,4.75) and (1.45,4.65) .. (1.35,3.0)
    .. controls (1.28,1.85) and (1.10,1.25) .. (0.85,0.95);
  \draw[traj,navD] (0.95,4.95) .. controls (1.25,3.4) and (1.35,1.45) .. (3.0,1.35)
    .. controls (4.15,1.28) and (4.75,1.10) .. (5.05,0.85);

  \foreach \x/\y/\c in {0.7/0.7/navA, 5.3/0.7/navB, 5.3/5.3/navC, 0.7/5.3/navD}
    \filldraw[fill=\c!30,draw=\c,line width=0.5pt] (\x,\y) circle (0.42);
  \foreach \x/\y/\c in {5.3/5.3/navA, 0.7/5.3/navB, 0.7/0.7/navC, 5.3/0.7/navD}
    \node[goal,text=\c] at (\x,\y) {$\star$};

  \node[font=\tiny,above] at (3.0,2.2) {$S_j$};
  \fill (2.2,2.2) circle (1.2pt);
  \fill (3.8,2.2) circle (1.2pt);
  \node[lab,anchor=north] at (2.45,2.1) {$o_{j,0}$};
  \node[lab,anchor=north] at (3.55,2.1) {$o_{j,1}$};
  \draw[-{Stealth[length=2.5pt]},line width=0.45pt] (2.2,2.2) -- (1.12,1.12);
  \node[lab,anchor=east] at (1.62,1.95) {$n_e(p)$};
  \draw[{Stealth[length=2.2pt]}-{Stealth[length=2.2pt]},line width=0.35pt] (0.7,5.3) -- (1.12,5.3);
  \node[lab,anchor=west] at (1.16,5.42) {$r_{\rm rob}$};
\end{tikzpicture}
\endgroup}
 \vspace{-8pt}
 \caption{Four robots (discs of radius $r_{\rm rob}$) swap corners around four segment obstacles; stars mark goals.}
 \label{fig:navproblem}
 \vspace{-10pt}
\end{wrapfigure}

Our motivating example is a team of robots that must reach their goals without collisions (\prettyref{fig:navproblem}).  Each simulation step resolves all contacts at once by solving a small constrained optimization problem.  Such optimization-based updates are standard in crowd simulation and multi-robot navigation \citep{karamouzas2017implicit,vandenberg2011reciprocal,wang2017safety}.
Let $N_{\rm rob}$ robots of radius $r_{\rm rob}>0$ move in the plane $\R^2$.  Obstacle $j$, $1\leq j\leq N_{\rm obs}$, is the segment $S_j=\operatorname{conv}\{o_{j,0},o_{j,1}\}$.  The robot positions are $p=(p_1,\ldots,p_{N_{\rm rob}})$.  The entity pairs (edges) that must stay collision-free are $\cE:=\cE_{\rm p}\cup\cE_{\rm o}$, where
\[
 \begin{aligned}
 \cE_{\rm p}&:=\{\{r,r'\}:1\leq r<r'\leq N_{\rm rob}\},\\
 \cE_{\rm o}&:=\{(r,j):1\leq r\leq N_{\rm rob},\ 1\leq j\leq N_{\rm obs}\}.
 \end{aligned}
\]

\paragraph{Constraints.} Each step linearizes the distances at the previous positions $p$, which we assume to be collision-free.  Linearization is only one simple choice: our assumptions (\prettyref{ass:ns}) require only that each constraint be smooth and convex in the next state, so more general nonlinear constraints are allowed.  For a pair $e=\{r,r'\}$, let $n_e(p)$ be the unit vector from $p_{r'}$ to $p_r$.  For an obstacle pair $e=(r,j)$, let $n_e(p)$ be the unit vector from the nearest point of $S_j$ to $p_r$.  The constraints are
\begin{equation}\label{icml:exactcon}
 \resizeEq{\begin{aligned}
 h_e(p,y)&:=2r_{\rm rob}-\langle n_e(p),y_r-y_{r'}\rangle, && e=\{r,r'\}\in\cE_{\rm p},\\
 h_{e,k}(p,y)&:=2r_{\rm rob}-\langle n_e(p),y_r-o_{j,k}\rangle, && e=(r,j)\in\cE_{\rm o},\ k\in\{0,1\}.
 \end{aligned}}
\end{equation}

Every constraint is affine in the next positions $y$.  A pair constraint keeps two robots $2r_{\rm rob}$ apart along their current normal.  The two obstacle constraints keep both endpoints, and hence the whole segment, $2r_{\rm rob}$ behind the robot along its normal.  Thus every feasible $y$ is collision-free.  There are $N_h=|\cE_{\rm p}|+2|\cE_{\rm o}|$ constraints, which we index by $l$. We consider two variants of this problem. They share the constraints \eqref{icml:exactcon} but differ in the stage energies $g_i$: in the kinematic model the state holds only the robot positions, while in the dynamic model it also holds their velocities and the robots have mass.  Both are standard: reciprocal collision avoidance plans with a kinematic model that ignores dynamics \citep{vandenberg2011reciprocal}, while implicit crowd integrators and multi-robot safety filters use second-order models with mass or acceleration inputs \citep{karamouzas2017implicit,wang2017safety}.

\paragraph{Kinematic model.} The state holds only the robot positions, so $n_y=2N_{\rm rob}$ and $y^{\pos}:=y$.  The global control $x$ stacks the robots' target positions at every stage, and $\SelMat_i x\in\R^{n_y}$ selects the target positions for stage $i$.  With $\mu>0$, we set $g_i(x,p,y):=\tfrac{\mu}{2}\|y-\SelMat_i x\|^2$.  Thus each step moves the robots as close as possible to the selected targets.

\paragraph{Dynamic model.} The state holds positions and velocities, $p=(p^{\pos},p^{\vel})$ and $y=(y^{\pos},y^{\vel})$, so $n_y=4N_{\rm rob}$.  The selected vector $\SelMat_i x$ is the stacked force used at stage $i$.  For a mass $m_{\rm mass}>0$ and a time step $\Delta t>0$,
\begin{equation}\label{eq:dynobjective}
 \resizeEq{\begin{aligned}
 g_i(x,p,y)&:=\frac{m_{\rm mass}}2
 \left\|p^{\vel}-\frac{y^{\pos}-p^{\pos}}{\Delta t}\right\|^2\\
 &\quad+\frac12
 \left\|y^{\vel}-\frac{y^{\pos}-p^{\pos}}{\Delta t}\right\|^2
 -(\SelMat_i x)^\top y^{\pos}.
 \end{aligned}}
\end{equation}
The first term penalizes a change of velocity, weighted by the mass.  The second term makes the new velocity match the displacement.  The last term is the work of the force.  Without constraints, the minimizer is the semi-implicit Euler step.  The constraints are the same as in the kinematic model and act on the position blocks, that is, on $(p^{\pos},y^{\pos})$.

\paragraph{Task loss.} Write $p^{\rm goal}:=(p_1^{\rm goal},\ldots,p_{N_{\rm rob}}^{\rm goal})$.  Robot $r$ should end near $p_r^{\rm goal}$.  With $(y_K^{\pos})_r$ the final position of robot $r$ and $\delta_F>0$, we use the pseudo-Huber loss
\begin{equation}\label{icml:navF}
 \resizeEq{F(x,y_1,\ldots,y_K)=\delta_F^2\sum_{r=1}^{N_{\rm rob}}\Bigl(\sqrt{1+\|(y_K^{\pos})_r-p_r^{\rm goal}\|^2/\delta_F^2}-1\Bigr).}
\end{equation}
In practice, $F$ may also contain regularizers, such as a trajectory-smoothness term (\prettyref{sec:app_experiments}).  The loss above is smooth and nonnegative with a bounded and Lipschitz gradient, so it meets \prettyref{ns:F}.  In the kinematic model, the targets $\SelMat_i x$ need not equal the goal: the optimizer chooses them so that the robots end near $p^{\rm goal}$ after $K$ constrained steps while reducing regularizers such as the trajectory-smoothness term.

\section{PSK: Penalized Single-loop \texorpdfstring{$K$}{K}-level Method}\label{sec:results}

This section develops PSK in three steps.  \prettyref{sub:smooth} treats steps without hard constraints: it rewrites the nested problem \eqref{eq:traj} as a min--max problem, states PSK, and proves its $O(\epsilon^{-3})$ rate.  \prettyref{sub:band} adds hard constraints through a smooth wall on a band of constant width and shows that the same rate holds and that the exact surrogate states are feasible.  \prettyref{sec:shrink} gives a lower bound for deterministic zero-respecting methods: when the band shrinks with the target accuracy, the exponent of $1/\epsilon$ grows linearly with $K$ if no CQ holds uniformly in the band width.

\subsection{Matrix-free min--max reformulation}\label{sub:smooth}

We first treat steps without hard constraints, $N_h=0$.  This analysis is the foundation of the constrained case: \prettyref{sub:band} replaces each constraint by a smooth wall in the stage energy and then applies the same method.  Set $f_i:=g_i$ and define the exact chain
\begin{equation}\label{eq:stage}
\begin{aligned}
 \widehat y_0(x)&:=y_0,\;
 \widehat y_i(x):=s_i(x,\widehat y_{i-1}(x)),\\
 s_i(a)&:=\argmin_{y\in\R^{n_y}}f_i(a,y).
\end{aligned}
\end{equation}
Here $s_i(x,p):=s_i(a)$ for $a=(x,p)$, as in \prettyref{sec:problem}, and $\varphi(x):=F(x,\widehat y_1(x),\ldots,\widehat y_K(x))$. We assume, uniformly over the finite family, that $f_i(a,\cdot)$ is $\mu$-strongly convex, that the gradients of $f_i$ and $F$ are Lipschitz, that each $f_i$ is $C^2$ with a Lipschitz Hessian (used only in the analysis), that $F$ is bounded below with a bounded gradient, and that $\cX$ is compact and convex with a computable projection.  These conditions are standard for bilevel and penalty methods.  \prettyref{ass:A} in \prettyref{sec:P} states them precisely and defines what it means for them to hold globally.
We keep the states as variables
\begin{equation}\label{eq:w}
 \begin{aligned}
 w&:=(x,y_1,\ldots,y_K)\in\cW
 :=\cX\times\R^{Kn_y},\\
 a_i(w)&=(x,y_{i-1}),\\
 F(w)&:=F(x,y_1,\ldots,y_K).
 \end{aligned}
\end{equation}
Thus $w$ is a trajectory and $a_i(w)$ is the stage-$i$ parameter of \prettyref{sec:problem}.  For the iterates, write $w_t=(x_t,y_{1,t},\ldots,y_{K,t})$, set $y_{0,t}:=y_0$, and let $b_t=(b_{1,t},\ldots,b_{K,t})$ be the stage trackers.  For copies $z=(z_1,\ldots,z_K)$ of the states and a penalty parameter $\rho>0$, define
\begin{equation}\label{eq:saddle}
 \resizeEq{\cL_\rho(w,z):=F(w)+\rho\sum_{i=1}^K
 \bigl[f_i(a_i(w),y_i)-f_i(a_i(w),z_i)\bigr].}
\end{equation}
Our min--max problem minimizes $\max_z\cL_\rho(w,z)$ over $w\in\cW$.  For fixed $w$, the maximum over $z_i$ is attained at $z_i^\star(w)=s_i(a_i(w))$, and its value is $-\rho f_i^\star(a_i(w))$, where $f_i^\star(a):=\min_yf_i(a,y)$.  Thus the inner maximum is the penalty function $\Phi_\rho(w):=\max_z\cL_\rho(w,z)=F(w)+\rho\sum_{i=1}^K R_i(a_i(w),y_i)$, where $R_i(a,y):=f_i(a,y)-f_i^\star(a)$.  Each gap is nonnegative and vanishes exactly when $y_i=s_i(a_i(w))$, so all gaps vanish exactly when $w$ is a true trajectory.  Thus the min--max problem is a penalty form of the chain in \eqref{eq:stage}, which is heterogeneous: $f_i$ may differ across stages. It contains no implicit map, and its gradients need only first-order oracle calls.
For a $C^1$ function $\mathfrak f$ on a nonempty closed convex set $\mathcal Z$ and $\alpha>0$, define
\begin{equation}\label{eq:G}
 \mathcal G_{\alpha,\mathcal Z}^{\mathfrak f}(u)
 :=\alpha^{-1}\|u-\PiX_{\mathcal Z}
 (u-\alpha\nabla\mathfrak f(u))\|.
\end{equation}
Here $\PiX_{\mathcal Z}$ is the Euclidean projection onto $\mathcal Z$, so this is the length of one projected gradient step over its step size; it is zero exactly at projected stationary points.  For any function $\mathfrak f$ of the control, write $\cG^{\mathfrak f}_\alpha:=\mathcal G^{\mathfrak f}_{\alpha,\cX}$.  We measure progress on the implicit loss by $\Gphi_\alpha(x)$.  At an interior control, $\Gphi_\alpha(x)=\|\nabla\varphi(x)\|$ for all sufficiently small $\alpha$.  Fix a reference step $\bar\alpha>0$.  A control is $\epsilon$-stationary if $\cG^\varphi_{\bar\alpha}(x)\leq\epsilon$.

\begin{algorithm}[t]
\caption{PSK}\label{alg:psk}
\small
\begin{algorithmic}[1]
\Require $\{f_i\}_{i=1}^K$, $\gamma>0$, $m\geq1$, $\kappa_0>0$;
$\rho_t,\alpha_t>0$; arbitrary $w_0\in\cW$ and
$b_0=(b_{1,0},\ldots,b_{K,0})$; $T\geq1$
\For{$t=0,\ldots,T-1$}
  \For{$i=1,\ldots,K$ \textbf{in parallel}}
    \State $b_{i,t}^{(0)}\gets b_{i,t}$
    \For{$k=0,\ldots,m-1$}
      \State $\xi_{i,t}^{(k)}\gets
      \nabla_yf_i(a_i(w_t),b_{i,t}^{(k)})$
      \State $b_{i,t}^{(k+1)}\gets
      b_{i,t}^{(k)}-\gamma\xi_{i,t}^{(k)}$
    \EndFor
    \State $b_{i,t+1}\gets b_{i,t}^{(m)}$
  \EndFor
  \State $\hnabla_t\gets\nabla_w\cL_{\rho_t}(w_t,b_{t+1})$
  \State $w_{t+1}\gets\PiX_\cW(w_t-\alpha_t\hnabla_t)$
  \State {\footnotesize $C_t\gets\bigl[2\alpha_t^{-2}\|w_{t+1}-w_t\|^2+2\kappa_0\rho_t^2\mu^{-2}\sum_i\|\xi_{i,t}^{(0)}\|^2\bigr]^{1/2}$}
\EndFor
\State \Return $w_{t^\star}$,
$t^\star\in\argmin_{\lfloor T/2\rfloor\leq t<T}C_t$
\end{algorithmic}
\end{algorithm}

\paragraph{Algorithm.} PSK (\prettyref{alg:psk}) stores a trajectory $w_t$ and one tracker $b_{i,t}$ per stage; each tracker follows the exact solution $s_i(a_i(w_t))$.  Its inputs are the stage family $\{f_i\}_{i=1}^K$, the inner step $\gamma$, the inner budget $m$, the certificate weight $\kappa_0$, and the schedules $\rho_t$ and $\alpha_t$; \prettyref{sub:band} reuses it with $f_i=f_i^C$ for all $i$.  The constant $\mu$ is the uniform strong-convexity modulus, and $\PiX_\cW$ projects the control block onto $\cX$ and leaves the states unchanged. In each iteration, every tracker first takes $m$ gradient steps on its own stage, starting from its previous value.  Then $w$ takes one projected gradient step along $\hnabla_t$, the gradient of $\cL_{\rho_t}(\cdot,b_{t+1})$, in which the trackers replace the copies; \prettyref{sec:alg} lists its blocks. If every tracker were exact, $\hnabla_t$ would be the gradient $\nabla\Phi_{\rho_t}(w_t)$ of the penalty function. The certificate $C_t$ combines the length of the outer step with the first tracker gradients $\xi_{i,t}^{(0)}$ and adds no oracle call; the method returns the iterate with the smallest $C_t$ in the last half of the run.  All stages read the same $w_t$, so they run in parallel.
Without constraints, take $\gamma\leq L_{yy}^{-1}$, where $L_{yy}$ uniformly bounds the Lipschitz constants of $\nabla_yf_i$.  Let $L_{ay}$ uniformly bound the Lipschitz constants of $\nabla_yf_i$ in $a$, set $q:=(1-\gamma\mu)^{1/2}$ and $\widetilde q:=q^m$, and use $\kappa_0=(K+1)(L_{ay}\widetilde q)^2$ from \eqref{eq:kappa0}.  Choose $\rho_0>0$, $0<c_\alpha\leq c_\alpha^{\max}$, and $0<\beta\leq\beta_{\max}$, with the step and growth caps of \eqref{eq:amax} and \eqref{eq:betamax} in \prettyref{sec:alg}.  Use $\rho_t=\rho_0(1+\beta t)^{1/3}$ and $\alpha_t={c_\alpha}/{\rho_t}$, so the penalty parameter grows slowly and the outer step shrinks with it.  A constant $B_0$ set by the initial trajectory and trackers and the constants above, and the rate map $C^\star$, are given in \eqref{eq:PsiUB} and \eqref{eq:Cs}.  The main result reads as follows.

\begin{theorem}[PSK convergence ($N_h=0$), informal]
\label{thm:main}
Suppose the smoothness, strong-convexity, and boundedness conditions above hold at every point, and run \prettyref{alg:psk} with the schedule above. Then $\mathcal G_{\bar\alpha}^{\varphi}(x_{t^\star}) \le C^\star(B_0)T^{-1/3}$ for every $T\ge1$ and every $\bar\alpha\ge\alpha_{t^\star}$.
\end{theorem}

Each iteration uses $O(mK)$ first-order calls, so, for fixed data, $O(\epsilon^{-3})$ iterations give an $\epsilon$-stationary control for every $\bar\alpha\geq c_\alpha/\rho_0$.  The exponent does not depend on $K$.  The constant $C^\star(B_0)$ does, through the depth sums $W_i=\sum_{k=0}^{K-i}M^k$ of $M=L_{ay}/\mu$ (\prettyref{sec:alg}).

\paragraph{Sketch of proof.} We follow the Lyapunov analysis of \citet{kwon2024penalty} and extend it from two levels to $K$.  Our $K$-stage Lyapunov candidate $\Psi_t$ in \eqref{eq:PsiUB} adds one tracking error $\|b_{i,t}-s_i(a_i(w_t))\|^2$ per stage to the penalty function $\Phi_{\rho_t}(w_t)$.  Each iteration decreases $\Psi_t$ up to a drift that is summable for $\rho_t\sim t^{1/3}$, and an implicit-function argument turns the resulting PSK certificate into a bound on $\Gphi$ (formal statement in \prettyref{sec:statements}, proof in \prettyref{sec:mainresult}).  Like \citet{kwon2024penalty}, and unlike fixed-penalty methods \citep{lu2024firstorder}, PSK grows $\rho_t$ during the run instead of choosing it from the target accuracy, so it can be stopped at any time.  It also selects its output by the computable certificate $C_t$ rather than at random.

\subsection{Hard constraints with a constant band}\label{sub:band}

We now return to \eqref{icml:klevel} with $N_h>0$ constraints (\prettyref{sec:ns}).  The constrained stage maps $s_i^B$ of \prettyref{sec:problem} are not smooth where constraints become active, so we replace each constraint by a smooth wall.  Set
\begin{equation}\label{eq:nsslack}
 \begin{aligned}
 \operatorname{sl}(p,y)&:=\min_l\{-h_l(p,y)\}.
 \end{aligned}
\end{equation}
The slack $\operatorname{sl}(p,y)$ is the smallest constraint margin.  For a band width $d>0$ and height $\Gamma>0$ (both fixed in \eqref{eq:nscbdata} below), the $C^2$ cubic wall is
\begin{equation}\label{eq:nswall}
  \resizeEq{
 \psi_{d,\Gamma}(u):=
 \begin{cases}
 0,&u>d,\\
 \Gamma(1-u/d)^3,&0\leq u\leq d,\\
 \Gamma(1-3u/d+3u^2/d^2),&u<0.
 \end{cases}}
\end{equation}
It is zero when the margin $u$ exceeds $d$, rises smoothly inside the band, and grows quadratically under violation.  For $1\le i\le K$, define
\begin{equation}\label{eq:nscbsurrogate}
\begin{aligned}
 f_i^C(x,p,y)
 &:=g_i(x,p,y)
 +\sum_{l=1}^{N_h}\psi_{d,\Gamma}(-h_l(p,y)),\\
 s_i^C(x,p)&:=\argmin_y f_i^C(x,p,y),
\end{aligned}
\end{equation}
and the exact chain with its loss
\begin{equation}\label{eq:nscbchain}
\begin{aligned}
 \widehat y_0^C(x)&:=y_0,\;
 \widehat y_i^C(x):=s_i^C(x,\widehat y_{i-1}^C(x)),\\
 \varphi^C(x)&:=F\bigl(x,\widehat y_1^C(x),\ldots,
 \widehat y_K^C(x)\bigr).
\end{aligned}
\end{equation}
Every surrogate stage energy is smooth and strongly convex, so PSK can run on the family $(f_i^C)_{i=1}^K$, but two questions remain: whether the family's smoothness constants stay finite along the run, and whether its exact states are feasible.  We first define the exact-state certificate.

\begin{definition}[Banded Fritz--John certificate]\label{def:nscert}
Fix a stage $i$.  A feasible $y$ is a $d$-banded Fritz--John point at $(x,p)$ if there is $\vartheta\in\R_+^{N_h+1}$, with objective weight $\vartheta_0$, such that
\[
 \vartheta_0>0,\quad \sum_{l=0}^{N_h}\vartheta_l=1,\quad
 \vartheta_0\nabla_yg_i+\sum_{l=1}^{N_h}\vartheta_l\nabla_yh_l=0,
\]
and $\vartheta_l\max\{-h_l(p,y)-d,0\}=0$ for $1\leq l\leq N_h$. Derivatives of $g_i$ and $h_l$ are taken at $(x,p,y)$.
\end{definition}

To answer both questions, we design two assumptions on the problem data.  The first collects the smoothness and boundedness conditions commonly used in bilevel analysis; since the walls have no global derivative bounds, it bounds derivatives only on bounded sets.
\begin{assumption}[Base constrained-stage class]\label{ass:ns}\leavevmode
\begin{enumerate}
\item[(i)]\settag{i}\label{ns:X} $\cX$ is nonempty, convex, and compact.

\item[(ii)]\settag{ii}\label{ns:F} $F$ is $C^1$, $\nabla F$ is $L_F$-Lipschitz, $F\geq\Flo> -\infty$, and $\|\nabla F\|\leq G_F$.

\item[(iii)]\settag{iii}\label{ns:G} The stages are the first $K$ members of a possibly infinite sequence $(g_i)_{i\geq1}$.  Every $g_i\in C^2$, and $g_i(x,p,\cdot)$ is $\mu$-strongly convex with one $\mu>0$ for all $i\geq1$.  On every bounded convex set $\mathcal D\subseteq\R^{n_x+n_y}\times\R^{n_y}$, there are finite constants $L_{yy,\mathcal D}^g,L_{ay,\mathcal D}^g,L_{aa,\mathcal D}^g, L_{f,\mathcal D}^g,L_{H,\mathcal D}^g$ such that, for every $i\geq1$ and $v=(a,y),\widetilde v=(\widetilde a,\widetilde y)\in\mathcal D$,
\[
\begin{gathered}
 \mu I\preceq\nabla^2_{yy}g_i(v)\preceq L_{yy,\mathcal D}^gI,
 \qquad \|\partial_a\nabla_yg_i(v)\|\leq L_{ay,\mathcal D}^g,\\
 \|\nabla^2_{aa}g_i(v)\|\leq L_{aa,\mathcal D}^g,
 \qquad \|\nabla^2g_i(v)\|\leq L_{f,\mathcal D}^g,\\
 \|\nabla^2g_i(v)-\nabla^2g_i(\widetilde v)\|
 \leq L_{H,\mathcal D}^g\|v-\widetilde v\|.
\end{gathered}
\]
The oracle returns the first derivatives of the requested $g_i$.

\item[(iv)]\settag{iv}\label{ns:H1} Each $h_l\in C^2$ and $\nabla^2h_l$ is Lipschitz on every bounded convex set.  The oracle returns $h_l,\nabla_ph_l$, and $\nabla_yh_l$.

\item[(v)]\settag{v}\label{ns:H2} Each $h_l(p,\cdot)$ is convex.
\end{enumerate}
\end{assumption}

\begin{remark}[Finite prefixes and all-stage envelopes]\label{rem:nsguniform}
For fixed $K$, \prettyref{ns:G} is no stronger than the usual finite-family condition: any admissible family $(g_i)_{i=1}^K$ can be continued by repeating $g_K$, and its envelope is the maximum of finitely many stagewise bounds.  The all-stage quantifier matters only for a cross-depth claim: the same $\mu$ and finite derivative and Hessian-Lipschitz envelopes must then work for every $i\geq1$, rather than being re-maximized after each appended stage.
\end{remark}

\begin{figure}[t]
 \centering
 \resizebox{\columnwidth}{!}{\begingroup
\definecolor{cbTube}{HTML}{2166AC}
\definecolor{cbCore}{HTML}{B2182B}
\definecolor{cbBall}{HTML}{1B7837}
\begin{tikzpicture}[
  tube/.style={draw=cbTube,line width=0.6pt,fill=cbTube!6},
  core/.style={draw=cbCore,line width=0.6pt,fill=cbCore!10},
  ball/.style={draw=cbBall,line width=0.5pt,dashed,fill=cbBall!8,fill opacity=0.7},
  pt/.style={circle,fill=black,inner sep=0pt,minimum size=3pt},
  gpt/.style={circle,fill=black!50,inner sep=0pt,minimum size=3pt},
  map/.style={-{Stealth[length=4pt]},line width=0.6pt},
  lab/.style={font=\scriptsize,inner sep=1pt},
  panel/.style={draw=black!25,line width=0.4pt,rounded corners=2pt}
]
  \draw[panel] (-1.55,-2.05) rectangle (1.55,2.05);
  \draw[panel] (2.65,-2.05) rectangle (7.35,2.05);
  \node[lab,anchor=north west,text=black!60] at (-1.5,2.0) {previous state};
  \node[lab,anchor=north west,text=black!60] at (2.7,2.0) {next state};

  \draw[tube] (0,0) ellipse (1.25 and 1.35);
  \draw[core] (0.15,-0.15) ellipse (0.6 and 0.62);
  \node[lab,text=cbTube] at (0,1.05) {$\mathcal Y_{i-1}$};
  \node[lab,text=cbCore] at (0.15,0.25) {$\mathcal S_{i-1}$};

  \draw[tube] (5,0) ellipse (2.0 and 1.75);
  \draw[core] (4.55,0.3) ellipse (0.62 and 0.65);
  \node[lab,text=cbTube] at (5.35,1.45) {$\mathcal Y_i$};
  \node[lab,text=cbCore] at (4.3,0.68) {$\mathcal S_i$};

  \draw[ball] (4.7,-0.1) circle (0.62);
  \draw[ball] (6.0,0.55) circle (0.62);
  \draw[draw=cbCore,line width=0.6pt] (4.55,0.3) ellipse (0.62 and 0.65);

  \node[pt,label={[lab]below:$p$}] (p) at (0.25,-0.3) {};
  \node[gpt,label={[lab,text=black!60]left:$p'$}] (pp) at (-0.55,0.75) {};
  \node[pt] (s) at (4.7,-0.1) {};
  \node[lab,anchor=south] at (4.72,-0.05) {$s_i^C(x,p)$};
  \node[gpt] (ss) at (6.0,0.55) {};
  \node[lab,anchor=north,text=black!60] at (6.0,0.5) {$s_i^C(x,p')$};

  \draw[line width=0.45pt,cbBall] (s.center) -- ++(-30:0.62)
    node[lab,pos=0.75,below=1pt,text=cbBall] {$r_{\rm in}$};

  \draw[map] (p) -- (s);
  \draw[map,black!50] (pp) to[bend left=22] (ss);

  \node[pt,label={[lab]below:$\bar y$}] (ybar) at (3.0,-1.7) {};
  \node[lab,anchor=south west] at (3.08,-1.68)
    {$\langle\nabla_y g_i,\bar y-s_i^C\rangle-\mu\|\bar y-s_i^C\|^2\le\Upsilon_{\rm cb}$};
  \draw[line width=0.5pt,dash pattern=on 2pt off 1.5pt] (ybar) -- (s);
\end{tikzpicture}
\endgroup}
 \caption{\prettyref{ass:nscb} at stage $i$ for a fixed control $x\in\cX$.  Both panels show the state space $\R^{n_y}$, and the tubes are nested, $\mathcal Y_{i-1}\subseteq\mathcal Y_i$.  \prettyref{nscb:F} gives a state $\bar y$ with slack at least $\sigma_*$; the dashed line joins it to the exact step.  The state $\bar y$ need not lie in a tube.  \prettyref{nscb:XT} then puts the exact step $s_i^C(x,p)$ from a core state $p$ into the next core.  \prettyref{nscb:NI} keeps the ball of radius $r_{\rm in}$ around $s_i^C(x,p')$ inside $\mathcal Y_i$ for every tube state $p'$, including core states.}
 \label{fig:cb}
\end{figure}
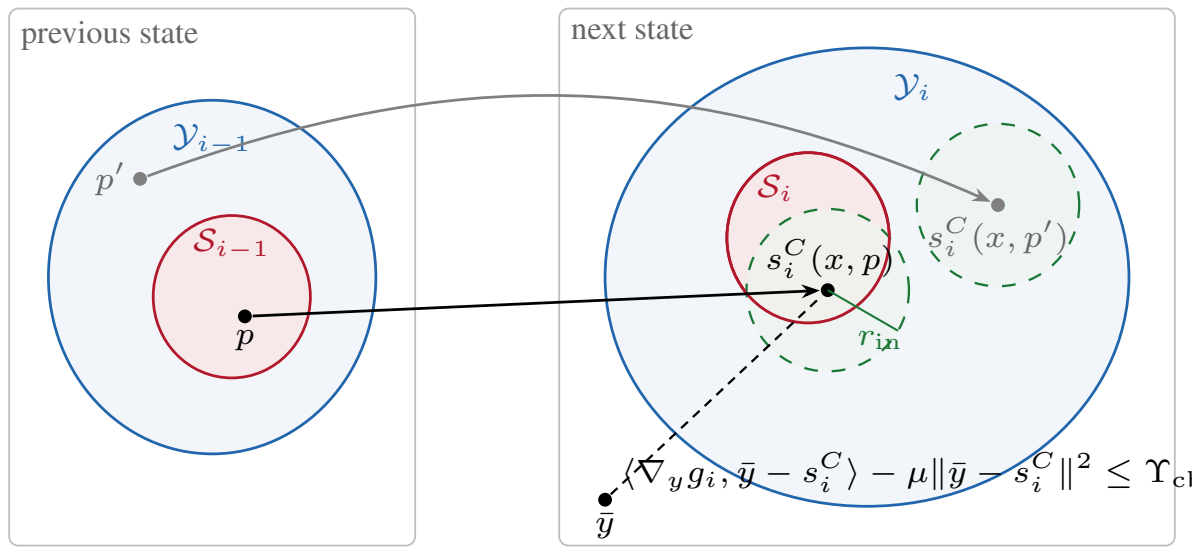

\prettyref{ass:ns} alone is not enough: the wall forces grow without bound under violation, so the surrogate family has no global smoothness constants, and nothing yet keeps its exact states feasible.  The second assumption lets one band width $d$ work at every stage (\prettyref{fig:cb}).  It uses exact cores $\mathcal S_i$, which hold the exact surrogate states, and numerical tubes $\mathcal Y_i$, which hold the iterates.  \prettyref{nscb:F} asks that, from every core state, a comparison point $\bar y$ with slack $\sigma_*>d$ be reachable within the force bound $\Upsilon_{\rm cb}$.  This caps the wall force at the exact state, so the exact state has slack at least $\tau_{\rm cb}$ (cb: constant band) and is feasible.  \prettyref{nscb:XT} and~\prettyref{nscb:NI} carry this into the next core and tube, so all iterates stay in one compact box where the surrogate family has finite smoothness constants, though they need not be feasible.  Formally, fix $0<\tau_{\rm cb}<\sigma_*$, $r_{\rm in}>0$, cores $\mathcal S_0,\ldots,\mathcal S_K$, and compact convex tubes with
\begin{equation}\label{eq:nscbnested}
 \mathcal Y_0\subseteq\cdots\subseteq\mathcal Y_K,
 \qquad \mathcal S_i\subseteq\mathcal Y_i\quad(0\leq i\leq K).
\end{equation}
Let $\Upsilon_{\rm cb}>0$ be a force bound and set $d:=(\tau_{\rm cb}+\sigma_*)/2$ and $b_{\rm cb}:=\tau_{\rm cb}/d$.  For $N_h>0$, define
\begin{equation}\label{eq:nscbdata}
 \Lambda_{\rm cb}:=\frac{\Upsilon_{\rm cb}}{\sigma_*-d},\qquad
 \Gamma_{\rm cb}:=\frac{d\Upsilon_{\rm cb}}
 {3(\sigma_*-d)(1-b_{\rm cb})^2}.
\end{equation}
For $N_h=0$, set $\Lambda_{\rm cb}=0$ and choose any fixed $\Gamma_{\rm cb}>0$.  We use $\Gamma=\Gamma_{\rm cb}$ in \eqref{eq:nscbsurrogate}.

\begin{assumption}[Finite-horizon constant-band class]
\label{ass:nscb}
Assume \prettyref{ass:ns} and the fixed data above.
\begin{enumerate}
\item[$(0_K)$]\settag{$0_K$}\label{nscb:0} $y_0\in\mathcal S_0$, and every $\mathcal S_i$ is compact.

\item[$(\mathrm F_K)$]\settag{$\mathrm F_K$}\label{nscb:F} For every $1\leq i\leq K$ and $(x,p)\in\cX\times\mathcal S_{i-1}$, there is $\bar y$ such that $\operatorname{sl}(p,\bar y)\geq\sigma_*$ and
\[
 \langle\nabla_y g_i(x,p,\bar y),
 \bar y-s_i^C(x,p)\rangle
 -\mu\|\bar y-s_i^C(x,p)\|^2
 \leq\Upsilon_{\rm cb}.
\]

\item[$(\mathrm{XT}_K)$]\settag{$\mathrm{XT}_K$}\label{nscb:XT} For every $1\leq i\leq K$ and $(x,p)\in\cX\times\mathcal S_{i-1}$, $\operatorname{sl}(p,s_i^C(x,p))\geq\tau_{\rm cb}$ implies $s_i^C(x,p)\in\mathcal S_i$.

\item[$(\mathrm{NI}_K)$]\settag{$\mathrm{NI}_K$}\label{nscb:NI} For every $1\leq i\leq K$ and $(x,p)\in\cX\times\mathcal Y_{i-1}$,
\begin{equation}\label{eq:nscbonestep}
 \Ball_{r_{\rm in}}(s_i^C(x,p))\subseteq\mathcal Y_i.
\end{equation}
\end{enumerate}
\end{assumption}

\begin{algorithm}[t]
\caption{Bounded-anchor warm start}\label{alg:nswarm}
\small
\begin{algorithmic}[1]
\Require $x_0\in\cX$, $y_{0,0}:=y_0$, $\gamma$, and $m_0$
\For{$i=1,\ldots,K$}
  \State $y_{i,0}^{(0)}\gets y_0$
  \For{$k=0,\ldots,m_0-1$}
    \State $y_{i,0}^{(k+1)}\gets y_{i,0}^{(k)}-
    \gamma\nabla_y f_i^C(x_0,y_{i-1,0},y_{i,0}^{(k)})$
  \EndFor
  \State $y_{i,0}\gets y_{i,0}^{(m_0)}$
  \State $b_{i,0}\gets y_{i,0}^{(m_0)}$
\EndFor
\State \Return $w_0=(x_0,y_{1,0},\ldots,y_{K,0})$ and $b_0$
\end{algorithmic}
\end{algorithm}

\paragraph{Method.} For constrained stages, PSK needs a start near the exact chain.  \prettyref{alg:nswarm} builds one stage by stage: it starts every stage from the bounded anchor $y_0$ and takes $m_0$ gradient steps on its surrogate with step $\gamma$.  Starting from its output, run \prettyref{alg:psk} with $f_i=f_i^C$, inner step $\gamma$, weight $\kappa_0$, a fixed inner budget $m$, and the schedule $\rho_t:=\rho_0(1+\beta t)^{1/3}$ and $\alpha_t:=c_\alpha/\rho_t$.  \prettyref{sec:nsschedule} defines $\gamma$, the warm-start budget $m_0$, the schedule constants $\rho_0,\beta,c_\alpha$, and $B_{\rm cb}$, and \prettyref{sec:nsstatements} defines $C_{\rm cb}^\star$.

\begin{theorem}[PSK convergence ($N_h>0$), informal]
\label{thm:nsrecover}
Let \prettyref{ass:nscb} hold, and run PSK with the family $\{f_i^C\}_{i=1}^K$, the bounded-anchor warm start, and the schedule above. Then $\cG^{\varphi^C}_{\bar\alpha}(x_{t^\star})=O(T^{-1/3})$ for every $T\geq1$ and every $\bar\alpha\geq\alpha_{t^\star}$.  For every stage $i$, every exact surrogate state is feasible and a $d$-banded Fritz--John point (\prettyref{def:nscert}).
\end{theorem}

\paragraph{Sketch of proof.} Only the strong-convexity modulus $\mu$ of $(f_i^C)_{i=1}^K$ is global.  The wall force $-\psi'_{d,\Gamma}$ grows linearly with the violation, and \prettyref{ns:G} and~\prettyref{ns:H1} bound the derivatives of $g_i$ and $h_l$ only on bounded sets, so the Lipschitz bounds on $\nabla f_i^C$ and $\nabla^2f_i^C$, such as $L_{yy}$ and $L_{ay}$, hold only on bounded sets, and \prettyref{thm:main} does not apply directly.  \prettyref{ass:nscb} removes this obstacle by induction.  Over the stages, \prettyref{nscb:F} and~\prettyref{nscb:XT} keep each exact state in its core with slack at least $\tau_{\rm cb}$ and a bounded total wall force.  Over the iterations, the warm start, the Lyapunov budget, and a large $\rho_0$ keep every residual below $r_{\rm in}$, so \prettyref{nscb:NI} keeps every iterate, tracker, and proof segment in one compact box with finite smoothness constants.  The proof of \prettyref{thm:main} then applies by specializing the corresponding results of \prettyref{sec:smooth} to $(f_i^C)_{i=1}^K$ (formal statement in \prettyref{sec:nsstatements}, proof in \prettyref{sec:nsproof}).

\paragraph{No constraint qualification.} \prettyref{ass:nscb} contains no CQ.  Physical systems often have many active contacts at once, which makes contact forces non-unique~\citep{stewart2000rigid}, and complementarity formulations of contact violate the Mangasarian--Fromovitz CQ at every feasible point~\citep{scheel2000mathematical,posa2014direct}.  Without a CQ, exact multipliers may not exist, so \prettyref{thm:nsrecover} certifies banded Fritz--John points.  Simulators also activate contact over a positive band~\citep{li2020incremental,todorov2012mujoco,muller2007position,macklin2016xpbd}.

\subsection{Can the band shrink?}\label{sec:shrink}
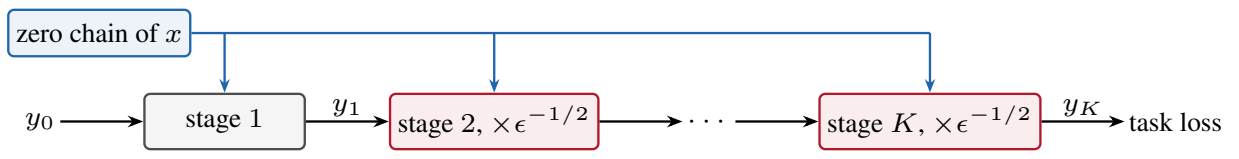
\begin{figure}[t]
 \centering
 \resizebox{\columnwidth}{!}{\begingroup
\definecolor{lbChain}{HTML}{2166AC}
\definecolor{lbAmp}{HTML}{B2182B}
\begin{tikzpicture}[
  node distance=0.75cm,
  box/.style={draw=black!70,line width=0.6pt,rounded corners=1.5pt,
    minimum width=1.45cm,minimum height=0.5cm,inner sep=2pt,font=\scriptsize},
  amp/.style={box,draw=lbAmp,fill=lbAmp!8},
  chain/.style={box,draw=lbChain,fill=lbChain!8,minimum width=0pt,minimum height=0.42cm},
  flow/.style={-{Stealth[length=4pt]},line width=0.6pt},
  feed/.style={-{Stealth[length=3.5pt]},line width=0.6pt,lbChain},
  lab/.style={font=\scriptsize,inner sep=1pt}
]
  \node[lab] (y0) at (0,0) {$y_0$};
  \node[box,fill=black!4,right=of y0] (s1) {stage $1$};
  \node[amp,right=of s1] (s2) {stage $2$, $\times\epsilon^{-1/2}$};
  \node[lab,right=of s2] (dots) {$\cdots$};
  \node[amp,right=of dots] (sK) {stage $K$, $\times\epsilon^{-1/2}$};
  \node[lab,right=of sK] (F) {task loss};

  \draw[flow] (y0) -- (s1);
  \draw[flow] (s1) -- node[lab,above,midway] {$y_1$} (s2);
  \draw[flow] (s2) -- (dots);
  \draw[flow] (dots) -- (sK);
  \draw[flow] (sK) -- node[lab,above,midway] {$y_K$} (F);

  \node[chain,anchor=east] (zc) at ([xshift=-0.3cm]s1.north |- 0,0.8) {zero chain of $x$};
  \draw[line width=0.6pt,lbChain] (zc.east) -- (sK.north |- zc.east);
  \foreach \s in {s1,s2,sK}
    \draw[feed] (\s.north |- zc.east) -- (\s.north);
\end{tikzpicture}
\endgroup}
 \caption{The relay behind the lower bound.  Each primitive call reveals at most one new coordinate of the zero chain; every stage after the first multiplies the gradient by order $\epsilon^{-1/2}$ between two nearly parallel walls.}
 \label{fig:lb}
\end{figure}

Ideally, one exponent $p$ in an $\epsilon^{-p}$ iteration count would work for every depth $K$; \prettyref{thm:nsrecover} gives $p=3$ for a fixed band.  Solving each constrained lower-level problem to accuracy $\epsilon$, however, means a band of width $\epsilon$.  Without a CQ that holds uniformly in the band width, no such $p$ then exists for multi-block zero-respecting methods: the exponent of $1/\epsilon$ is $(K+3)/4$, which grows linearly with $K$.  The key idea is a relay.  Two shared constraints have normals that become parallel as the band width shrinks to $\epsilon$, so no CQ holds uniformly in $\epsilon$.  A small shift of the incoming state moves the exact solution, squeezed between the two walls, about $\epsilon^{-1/2}$ times farther, so each stage after the first amplifies the gradient by this factor (\prettyref{fig:lb}).  We feed the relay with the zero chain of \citet{carmon2020lower} (\prettyref{sec:lbzerochain}), whose length, and hence the control dimension, grows with $1/\epsilon$.  The bound covers deterministic multi-block zero-respecting methods, which query and output, in each block of the trajectory, only coordinates that earlier oracle replies revealed (\prettyref{def:lbmodel}).  It counts primitive calls: the value and first derivatives of one function, or one projection onto $\cX$.

\begin{theorem}[Lower bound for multi-block zero-respecting methods, informal]\label{thm:lbpsk}
Fix $K\geq2$, and for small $\epsilon>0$ use the band width $\epsilon$.
\begin{enumerate}
\item[(i)] PSK, with any parameters, is a deterministic multi-block zero-respecting method.
\item[(ii)] There is an instance satisfying \prettyref{ass:ns} and \ref{ass:nscb} with $d=\epsilon$, whose smoothness constants do not depend on $\epsilon$, on which every such method needs at least $c_K\epsilon^{-(K+3)/4}$ primitive calls to return a control $x$ with $\cG^{\varphi^C}_{\bar\alpha}(x)\le\epsilon$, where $c_K>0$ depends only on $K$.
\end{enumerate}
\end{theorem}

\paragraph{Sketch of proof.} As in \citet{carmon2020lower}, each primitive call reveals at most one new coordinate of the zero chain, and the residual stays above $\epsilon$ until the last one is revealed.  The relay amplifies the gradient by order $\epsilon^{-1/2}$ per stage after the first, so a deeper relay lets a smaller and longer chain keep the residual above $\epsilon$.  Balancing the two gives a chain length, and hence a call count, of order $\epsilon^{-(K+3)/4}$.  The hard instance satisfies \prettyref{ass:ns} and \ref{ass:nscb} at every fixed band width (\prettyref{prop:lbcbbase}); \prettyref{sec:lb} gives the proof.
\section{Motivating Problem Validation}\label{sec:application}

We verify \prettyref{ass:nscb} for the navigation problem of \prettyref{sub:motivating} (\prettyref{sec:instances}), so \prettyref{thm:nsrecover} applies to both its models.  \prettyref{ns:H1} needs smooth constraints, but the normals $n_e(p)$ in \eqref{icml:exactcon} have kinks, so we smooth them (\prettyref{sec:instances}).  The smoothed constraints stay affine in $y$ and never looser than the true clearance, so feasible states stay collision-free; the results below, and our experiments, use these constraints.

\begin{proposition}[Kinematic navigation, informal]\label{prop:nsrobot}
Under \prettyref{ns:X} and~\prettyref{ns:F}, if every robot starts more than $2r_{\rm rob}$ away from all other robots and obstacles, then for small enough smoothing the kinematic model satisfies \prettyref{ass:nscb} for every $K\geq1$.  If $\sup_{K\geq1}\max_{i\leq K}\|\SelMat_i\|<\infty$ and $\sup_{K\geq1}\max_{i\leq K,x\in\cX}\|\SelMat_i x\|<\infty$, its cores, tubes, wall data, and \prettyref{ns:G} bounds can be $K$-independent.
\end{proposition}

\begin{proposition}[Dynamic navigation, informal]\label{prop:dynrobot}
Under the same conditions, the dynamic model satisfies \prettyref{ass:nscb} for every finite $K$, with cores and tubes that grow with $K$.
\end{proposition}

\paragraph{Sketch of proof.} The key step is \prettyref{nscb:F}: from every core state $p$, the robots must reach, within the force bound $\Upsilon_{\rm cb}$, a bounded state whose slack exceeds the band width.  By linear-programming duality~\citep{schrijver1986theory}, the best common slack of the linearized constraints is a stress-weighted average of the edge clearances, which exceeds the clearance of $p$ unless $p$ is a jam, where the active normals balance and no motion increases all active slacks, as for a robot at the center of a triangle of walls.  The jam set is semialgebraic, so it splits into finitely many smooth pieces of constant clearance~\citep{bochnak1998real,vandendries1998tame}.  Placing the core clearance below these finitely many levels and using compactness give a uniform slack gap and a bounded comparison point that survive small smoothing (\prettyref{lem:nsjam}).  For \prettyref{nscb:XT} and~\prettyref{nscb:NI}, strong convexity keeps every exact step near its unconstrained center.  In the kinematic model this center is the target $\SelMat_ix$, so the radii do not grow with depth.  In the dynamic model it is the Euler step $p^{\pos}+\Delta t\,p^{\vel}+(\Delta t)^2\SelMat_ix/m_{\rm mass}$, so the radii grow at every stage; this is unavoidable, since a constant force accelerates a free robot (\prettyref{sec:inst_main}).

\section{Experiments}
\label{sec:experiments}

\begin{figure}[t]
 \centering
 \includegraphics[width=0.92\columnwidth]{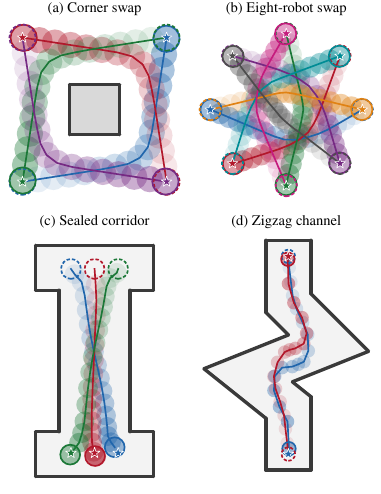}
 \caption{PSK with the dynamic model and $K=16$, started at rest, on (a)~the corner swap, (b)~the eight-robot swap, (c)~the sealed corridor, which fits at most two robots side by side, and (d)~the zigzag channel.  Dashed rings: starts; stars: goals; filled discs: exact surrogate positions, darker at later stages; solid rings: final positions, each covering its goal.}
 \label{fig:gallery}
\end{figure}

We run PSK on the four scenes of \prettyref{fig:gallery} with both models and $K\in\{4,8,16\}$, for \factNcases{} cases.  Every run starts from a neutral control (robots hold still or apply zero force) plus small noise, so no route is given.  PSK parameters are tuned per case, and wall heights stay below $\Gamma_{\rm cb}$, so neither guarantee of \prettyref{thm:nsrecover} applies, and we check feasibility directly.  A run stops when ten times as many iterations improve its best exact chain loss $\varphi^C$ by less than \factStallPct\%, or after \factTMax{} iterations, and reports its best feasible checkpoint.  We compare PSK on our surrogate with unrolled differentiable simulation \citep{degrave2019differentiable,hu2020difftaichi} with $5$, $20$, or $50$ tuned gradient steps per stage, and implicit differentiation \citep{amos2017optnet,geilinger2020add} with Newton solves and adjoints; every method gets \factTimeBudget{} seconds from the same start, including PSK's initialization.  With $K=16$, every robot reaches its goal in all \factNKsixteen{} cases (final distance at most \factMaxGoalKsixteen{} $<r_{\rm rob}$), and so do \factRobKsixteenGoal{} of \factRobKsixteenTotal{} reruns with other noise draws (\prettyref{tab:app_cases} and \ref{tab:app_robust}).  Every primary run keeps its exact surrogate states feasible at every checkpoint from $10^3$ iterations onward, and \factNconverged{} of the \factNcases{} runs stop by the plateau rule before the cap.  On the \factNeFive{} reachable cases, PSK reaches a lower exact chain loss than the best tuned unrolling depth in \factUnrollWins{}, while the deepest (50-step) unrolling stores \factMemRatioMin--\factMemRatioMax$\times$ as much peak data per iteration as PSK.  Implicit differentiation reaches a lower loss in \factImplicitLower{} of them (by up to \factImplicitMaxMarginPct\%), but on the two corridor cases with $K=16$ its loss is \factHardRatioMin--\factHardRatioMax$\times$ that of PSK (\prettyref{tab:app_e5}).  With fewer stages, a multi-start search finds no feasible chain to the goals in the corridor ($K=4$) or the zigzag channel ($K=4,8$), and on the reachable corridor with $K=8$, PSK stops short of the goals.  \prettyref{sec:app_experiments} adds details and ablations.

\section{Conclusion}
\label{sec:conclusion}
PSK is a matrix-free, single-loop method for $K$-level physics-based control.  With a constant band and under mild assumptions that need no CQ, every exact surrogate state is feasible and banded Fritz--John.  PSK then reaches an $\epsilon$-stationary control of the surrogate in $O(\epsilon^{-3})$ iterations, with a $K$-independent exponent but a $K$-dependent constant.  Without a uniform CQ, a shrinking band makes the exponent for deterministic zero-respecting methods grow linearly with $K$ on a hard instance.  Our assumptions hold for smoothed multi-robot navigation among segment obstacles.  In experiments, PSK reaches every goal with sixteen levels and, at equal time, a lower loss than tuned unrolling in \factUnrollWins{} of \factNeFive{} cases; implicit differentiation is lower in \factImplicitLower{}.  Open directions include lower bounds with disjoint stage controls, contact-rich dynamics, and adaptive bands.

\clearpage
\section*{Impact Statement}

This work may help design reliable optimization tools for coordinated robot motion.  Failures in modeling or optimization could still cause unsafe motion in real systems, so deployment should use independent validation, physical safety limits, and human oversight.

\section*{AI Usage Disclosure}
AI assistants were used to revise the prose, to help revise the proofs, and to check the proofs adversarially.  We first drafted the smooth-stage theory, the constant-band analysis, and the lower bound (\prettyref{sec:smooth}, \ref{sec:ns}, and \ref{sec:lb}) under an earlier version of \prettyref{ass:nscb} that was stronger than the current one.  We could not prove that our practical instance, the multi-agent navigation problem of \prettyref{sec:application}, satisfies that version.  GPT-5.6-Sol helped us find the key missing piece: the uniform jam-regeneration result (\prettyref{lem:nsjam}) and a weakened \prettyref{ass:nscb}, the version stated in this paper, which the navigation instances satisfy.

\bibliography{refs}
\bibliographystyle{icml2026}

\newpage
\appendix
\onecolumn
\let\addcontentsline\klevelAddContentsLine
\setcounter{tocdepth}{2}
\renewcommand{\contentsname}{Contents of the Appendix}
\tableofcontents
\newpage
This appendix collects the supporting smooth-stage theory and proofs, the constant-band construction and localization argument, the depth-dependent lower bound, the navigation specializations, and the experiment details\ifshownotation, and a notation table\fi. The organization follows these parts in order.

\section{Smooth stages}\label{sec:smooth}

This appendix treats stages without hard constraints, as in \prettyref{sub:smooth}.  The stages use a finite family $(f_i)_{i=1}^K$; \prettyref{sub:smooth} takes $f_i=g_i$, and \prettyref{sec:ns} takes $f_i=f_i^C$. The product space $\cW$ of \eqref{eq:w} has its Euclidean norm.  Under \prettyref{ass:A} below, $s_i$ is well defined.  Recall from \prettyref{sub:smooth} the stage value and gap
\begin{equation}\label{eq:gap}
 f_i^\star(a)=\min_y f_i(a,y),\qquad
 R_i(a,y)=f_i(a,y)-f_i^\star(a)\geq0,
\end{equation}
and define the solution residual
\begin{equation}\label{eq:ri}
 r_i(w):=\|y_i-s_i(a_i(w))\|.
\end{equation}
Recall from \prettyref{sub:smooth} that eliminating $z$ in \eqref{eq:saddle} gives the penalty function
\begin{equation*}
 \Phi_\rho(w)=\max_z\cL_\rho(w,z).
\end{equation*}
\prettyref{prop:identity} shows that $\Phi_\rho$ is finite and gives its explicit form.  With the indicator $\iota_{\mathcal Z}$ of a set $\mathcal Z$ ($0$ on $\mathcal Z$ and $+\infty$ off it), its constrained extension is
\begin{equation}\label{eq:Phibar}
 \Phib:=\Phi_\rho+\iota_\cW.
\end{equation}
Since $\iota_\cW(w)=\iota_\cX(x)$, this indicator is the only source of infinite values.  \prettyref{lem:value} gives $\partial\Phib=\nabla\Phi_\rho+N_\cW$, where $N_{\mathcal Z}$ is the normal cone of a closed convex set $\mathcal Z$.  With \eqref{eq:G}, the projected-gradient residual of the penalty function on $\cW$ is
\begin{equation}\label{icml:Grho}
 \cG_\alpha^\rho(w):=\mathcal G_{\alpha,\cW}^{\Phi_\rho}(w).
\end{equation}
It is a nonnegative scalar residual.  Since $\PiX_\cW$ acts as $\PiX_\cX$ on $x$ and as the identity on the states,
\begin{equation}\label{icml:Gprod}
 (\cG_\alpha^\rho(w))^2
 =\alpha^{-2}\|x-\PiX_\cX(x-\alpha\nabla_x\Phi_\rho(w))\|^2
 +\sum_{i=1}^K\|\nabla_{y_i}\Phi_\rho(w)\|^2.
\end{equation}
Consequently, $\|\nabla_{y_i}\Phi_\rho(w)\|\le\cG_\alpha^\rho(w)$ for every $i$, and $\alpha^{-1}\|x-\PiX_\cX(x-\alpha\nabla_x\Phi_\rho(w))\|\le\cG_\alpha^\rho(w)$.  The residual is zero exactly at projected stationary points of $\Phi_\rho$ on $\cW$.

\subsection{Assumptions}\label{sec:P}

Fix smooth stage energies $f_i:\mathbb R^{n_x+n_y}\times\mathbb R^{n_y}\to\mathbb R$ for $1\le i\le K$, and fix a convex numerical domain $\mathcal U\subseteq\mathbb R^{n_x+n_y}\times\mathbb R^{n_y}$.

\begin{assumption}[Uniform smooth strongly convex stages]\label{ass:A}
\leavevmode
\begin{enumerate}[label=\textnormal{(\roman*)},ref=\roman*]
\item\label{A:X} $\mathcal X$ is nonempty, convex, and compact, and $\PiX_{\mathcal X}$ is computable.

\item\label{A:f} For every $1\le i\le K$, the function $f_i$ is globally $C^1$. For all $(a,y),(\widetilde a,\widetilde y)\in\mathcal U$,
\begin{subequations}\label{eq:La}
\begin{align}
\|\nabla_y f_i(a,y)-\nabla_y f_i(a,\widetilde y)\|
&\le L_{yy}\|y-\widetilde y\|,\label{eq:Ly}\\
\|\nabla_y f_i(a,y)-\nabla_y f_i(\widetilde a,y)\|
&\le L_{ay}\|a-\widetilde a\|,\notag\\
\|\nabla_a f_i(a,y)-\nabla_a f_i(\widetilde a,y)\|
&\le L_{aa}\|a-\widetilde a\|,\notag\\
\|\nabla_a f_i(a,y)-\nabla_a f_i(a,\widetilde y)\|
&\le L_{ay}\|y-\widetilde y\|.\label{eq:Lay2}
\end{align}
\end{subequations}
Moreover, $\nabla f_i$ is $L_f$-Lipschitz on $\mathcal U$. For $a=(x,y')$, where $y'$ is the incoming state $p$ of \prettyref{sec:problem}, write $\nabla_a f_i=(\nabla_x f_i,\nabla_{y'}f_i)$.

\item\label{A:sc} For every $i$ and every $a$, the map $f_i(a,\cdot)$ is $\mu$-strongly convex on $\mathbb R^{n_y}$, with $\mu>0$.

\item\label{A:F} $F$ is $C^1$, $\nabla F$ is $L_F$-Lipschitz, $F\ge \Flo> -\infty$, and $\|\nabla F\|\le G_F$.

\item\label{A:oracle} The first-derivative oracles for every $f_i$ and for $F$, and $\PiX_{\mathcal X}$, are available at every requested point.

\item\label{A:C2} Every $f_i$ is globally $C^2$, and $\nabla^2 f_i$ is $L_H$-Lipschitz on $\mathcal U$.  Write
\begin{equation}\label{eq:HJ}
H_i(a,y):=\nabla^2_{yy}f_i(a,y),
\qquad
J_i(a,y):=\partial_a\nabla_y f_i(a,y),
\end{equation}
and use superscripts $x$ and $y'$ for the parameter sub-blocks.
\end{enumerate}
The uniform bounds obey $L_{ay}>0$, $L_{aa},L_f,L_F,G_F,L_H\ge0$, and $0<\mu<L_{yy}$.
\end{assumption}
By \prettyref{A:sc} and~\prettyref{A:f}, $\mu I\preceq H_i\preceq L_{yy}I$ and $\|J_i\|\le L_{ay}$ on $\mathcal U$ (for $\mathcal U$ with nonempty interior, as in this paper; the bounds extend to its closure by continuity).  We say that \prettyref{ass:A} holds globally if it holds with $\mathcal U=\R^{n_x+n_y}\times\R^{n_y}$, that is, if every upper bound in it holds at every point.

\begin{remark}[Scope of \prettyref{ass:A}]\label{rem:assA}
The state space is not compact, and $F$ need not be convex.  The set $\mathcal U$ localizes only the numerical upper bounds; strong convexity and $C^2$ regularity remain global for every stage.  Global strong convexity gives existence, uniqueness, stationarity, and lower gap bounds even when $\mathcal U$ is local.
\end{remark}

\begin{remark}[First-order method]\label{rem:whatadded}
\prettyref{A:C2} supports the implicit-function and conversion arguments for every stage.  \prettyref{alg:psk} uses only first derivatives and $\PiX_\cX$.  No $H_i$, $J_i$, Hessian, or Jacobian solve is computed.
\end{remark}

\subsection{Parameter Schedule}\label{sec:alg}

One inner tracker step at the parameter $a$ is
\begin{equation}\label{eq:T}
 \mathsf T_{i,a}(y):=y-\gamma\nabla_yf_i(a,y).
\end{equation}
Let $a_{i,t}:=a_i(w_t)$.  The direction $\hnabla_t=\nabla_w\cL_{\rho_t}(w_t,b_{t+1})$ of \prettyref{alg:psk} has blocks
\begin{equation}\label{icml:blocks}
\begin{aligned}
\hnabla_t^x
&=\nabla_xF(w_t)+\rho_t\sum_{i=1}^K
 [\nabla_xf_i(a_{i,t},y_{i,t})
  -\nabla_xf_i(a_{i,t},b_{i,t+1})],\\
\hnabla_t^{y_i}
&=\nabla_{y_i}F(w_t)+\rho_t\nabla_yf_i(a_{i,t},y_{i,t})\\
&\quad+\rho_t[
\nabla_{y'}f_{i+1}(a_{i+1,t},y_{i+1,t})
-\nabla_{y'}f_{i+1}(a_{i+1,t},b_{i+1,t+1})],
\quad i<K,\\
\hnabla_t^{y_K}
&=\nabla_{y_K}F(w_t)+\rho_t\nabla_yf_K(a_{K,t},y_{K,t}).
\end{aligned}
\end{equation}

\paragraph{Constants.} All constants below are fixed before the run.  They come in three groups: the stage and tracker constants, the schedule constants, and the conversion constants.  The later groups are built from the first. The first group, the stage and tracker constants, depends only on $K$, the bounds of \prettyref{ass:A}, and the tracker settings $(m,\gamma)$.  Fix $m\ge1$ and an inner step $0<\gamma\le L_{yy}^{-1}$, define the constants below, and recall $q$ and $\widetilde q$ from \prettyref{sub:smooth}
\begin{equation}\label{eq:rhofree}
 M:=\frac{L_{ay}}\mu,
 \qquad L_\star:=L_{aa}+L_{ay}M,
 \qquad \chi:=\frac{L_{yy}}\mu,
 \qquad q=(1-\gamma\mu)^{1/2},
 \qquad \widetilde q=q^m<1,
\end{equation}
\begin{equation}\label{eq:depthsums}
 W_i:=\sum_{k=0}^{K-i}M^k\qquad(1\le i\le K),
\end{equation}
\begin{equation}\label{eq:kappa0}
 \kappa_0:=(K+1)(L_{ay}\widetilde q)^2.
\end{equation}
The factor $M$ bounds the Lipschitz modulus of the stage map, so $M^{j-i}$ transfers a perturbation from stage $i$ to stage $j$, and $W_i$ sums these transfers over $j\ge i$.  The upper-triangular matrix $(M^{j-i}\mathbf1_{i\le j})_{i,j=1}^K$ has every row and column sum at most $W_1$, so its spectral norm is at most $W_1$.  These constants do not depend on the penalty parameter.

The second group, the schedule constants, also depends on the initial penalty parameter.  Fix $\rho_0>0$ and choose
\begin{equation}\label{eq:amax}
 0<c_\alpha\le c_\alpha^{\max}:=
 \min\left\{
 \frac1{2(L_F/\rho_0+K(L_f+L_\star))},
 \frac{1-\widetilde q}{8M\sqrt{K\kappa_0}}
 \right\}.
\end{equation}
Then define
\begin{equation}\label{eq:rhoconsts}
 c_\Psi:=\frac{4c_\alpha\kappa_0}{1-\widetilde q}.
\end{equation}
Further define
\begin{equation}\label{eq:omega}
 \omega_1:=L_{yy}\left[
 \frac{4W_1^2}{\mu^2c_\alpha^2}+2(1+KM^2)
 \right],
 \qquad \omega_2:=\frac{4L_{yy}W_1^2\kappa_0}{\mu^2},
 \qquad \omega_3:=\frac{2L_{yy}W_1^2G_F^2}{\mu^2},
\end{equation}
and choose
\begin{equation}\label{eq:betamax}
 0<\beta\le\beta_{\max}:=
 \min\left\{
 \frac{1-\widetilde q}{4},
 \frac{c_\alpha\mu^2}{4L_{yy}W_1^2},
 \frac1{4c_\alpha\omega_1}
 \right\}.
\end{equation}
The third group, the conversion constants, enters the rate constant $C^\star$ through the conversion step of the proof; besides the first group, it also uses $L_F$ and $L_H$.  Define
\begin{equation}\label{eq:convconsts}
\begin{gathered}
 \mathfrak H_0:=0,
 \qquad \mathfrak H_i:=W_i+M\mathfrak H_{i-1},
 \qquad \mathfrak H_2^{\rm all}:=\left(\sum_{i=1}^K\mathfrak H_i^2\right)^{1/2},\\
 D_i^{\rm der}:=(1+M)\mathfrak H_{i-1}+W_i,
 \qquad
 \vartheta_i^{\rm der}:=\frac{(1+M)L_HD_i^{\rm der}}\mu,\\
 \mathfrak B_i:=\sum_{j=i}^{K-1}M^{j-i}\vartheta_{j+1}^{\rm der}W_{j+1}
 \quad(1\le i\le K),\\
 c_{\rm adj}:=\sum_{i=1}^KMW_i,
 \qquad c_\varphi:=1+c_{\rm adj},
 \qquad c_{\rm b1}:=\frac{c_\varphi L_F\mathfrak H_2^{\rm all}}\mu,\\
 c_{\rm b2}:=\frac1\mu\sum_{i=1}^K
 \left(M\mathfrak B_i+\vartheta_i^{\rm der}W_i\right).
\end{gathered}
\end{equation}
The smooth run uses
\[
 \rho_t=\rho_0(1+\beta t)^{1/3},
 \qquad \alpha_t=\frac{c_\alpha}{\rho_t}.
\]
The schedule grows slowly.  Indeed,
\begin{align}
 \frac{\rho_{t+1}}{\rho_t}
 &=\left(\frac{1+\beta(t+1)}{1+\beta t}\right)^{1/3}\notag\\
 &=\left(1+\frac{\beta}{1+\beta t}\right)^{1/3}\notag\\
 &\le1+\frac{\beta}{3(1+\beta t)}
 &&\text{(concave Bernoulli inequality)}\notag\\
 &\le1+\frac\beta3\le1+\beta .\label{eq:ratio}
\end{align}
The registers $(w_0,b_0)$ may be initialized before \prettyref{alg:psk}; the initialization is not an algorithmic update.

The projection in \prettyref{alg:psk} acts as $\PiX_\cX$ on $x$ and as the identity on every state block.  All blocks of $\hnabla_t$ are computed from $w_t$ before $w_t$ is overwritten.

\begin{remark}[Work and timing]\label{rem:pgstep}
One iteration uses $mK$ tracker gradients, one $\nabla F$, two parameter-gradient calls per stage, one further state-gradient call per stage, and one projection onto $\cX$.  Its work is $O(mK)$.  No dual variable, inner stopping test, or target accuracy is used.
\end{remark}

\subsection{What PSK achieves}\label{sec:statements}

For $u\in\cW$, write
\[
 u=(u_x,u_{y_1},\ldots,u_{y_K}),
 \qquad u_y=(u_{y_1},\ldots,u_{y_K}),
\]
and set $u_{y_0}:=y_0$.  Define the direction error
\begin{equation}\label{eq:epsdef}
 \delta_t:=\hnabla_t-\nabla\Phi_{\rho_t}(w_t).
\end{equation}
Set
\begin{equation}\label{eq:errs}
 e_{i,t}:=\|b_{i,t}-s_i(a_i(w_t))\|,
 \qquad
 \widehat e_{i,t+1}:=\|b_{i,t+1}-s_i(a_i(w_t))\|,
 \qquad
 \Delta w_t:=w_{t+1}-w_t,
 \qquad \Delta w_t^{y_0}:=0.
\end{equation}
As for $u$, but with superscripts, the blocks of $\Delta w_t$ and $\delta_t$ are $\Delta w_t^x,\Delta w_t^{y_i}$ and $\delta_t^x,\delta_t^{y_i}$, with $\Delta w_t^y=(\Delta w_t^{y_1},\ldots,\Delta w_t^{y_K})$ and $\delta_t^y=(\delta_t^{y_1},\ldots,\delta_t^{y_K})$.  Also write
\[
 \Phi_t:=\Phi_{\rho_t},
 \qquad
 \Xi_t:=\sum_{i=1}^Ke_{i,t}^2,
 \qquad
 \Ecal_t:=\cG_{\alpha_t}^{\rho_t}(w_t).
\]
Define
\begin{equation}\label{eq:PsiUB}
 \Psi_t:=\Phi_t(w_t)+c_\Psi\rho_t\Xi_t,
 \qquad
 U_t:=\rho_t\left(
 \frac{\|\Delta w_t\|^2}{8c_\alpha}+c_\alpha\kappa_0\Xi_t
 \right),
 \qquad
 B_0:=\Psi_0-\Flo+\frac{2\omega_3}{\rho_0}.
\end{equation}
For $B\ge0$, define the rate maps, with $c_\varphi,c_{\rm b1},c_{\rm b2}$ from \eqref{eq:convconsts},
\begin{equation}\label{eq:Cs}
\begin{aligned}
 C_E(B)&:=\left(\frac{32\,2^{1/3}\rho_0B}{c_\alpha}\right)^{1/2},
\qquad
 C_\rho:=\frac{2^{1/3}}{\rho_0\beta^{1/3}},\\
 C^\star(B)&:=c_\varphi\chi C_E(B)
 +C_\rho\left[
 c_{\rm b1}(\chi C_E(B)+G_F)
 +c_{\rm b2}(\chi C_E(B)+G_F)^2
 \right].
\end{aligned}
\end{equation}
$C_E$, $C_\rho$, and $C^\star$ also depend on $(\rho_0,c_\alpha,\beta)$ and on the problem constants; we suppress these arguments.

We say that $(\mathrm H_\rho)$ holds if the conditions
\begin{equation}\label{eq:Hrho}
\begin{gathered}
 m\ge1,
 \qquad 0<\gamma\le L_{yy}^{-1},
 \qquad 0<c_\alpha\le c_\alpha^{\max},
 \qquad 0\le\beta\le\beta_{\max},
 \qquad \rho_0>0,\\
 \rho_t\le\rho_{t+1}\le(1+\beta)\rho_t,
 \qquad \alpha_t=\frac{c_\alpha}{\rho_t}
 \qquad(t\ge0),
\end{gathered}
\end{equation}
hold, and \prettyref{alg:psk} starts from arbitrary $w_0\in\cW$ and $b_{i,0}\in\R^{n_y}$.

\begin{formaltheorem}{thm:main}[PSK convergence ($N_h=0$)]
Let \prettyref{ass:A} hold globally for $\{f_i\}_{i=1}^K$.  Run \prettyref{alg:psk} from arbitrary $w_0\in\cW$ and $b_0$, with $m\ge1$, $0<\gamma\le L_{yy}^{-1}$, $\kappa_0$ from \eqref{eq:kappa0}, $\rho_0>0$, $c_\alpha$ and $\beta$ as in \eqref{eq:amax} and \eqref{eq:betamax}, and the schedule $\rho_t=\rho_0(1+\beta t)^{1/3}$, $\alpha_t=c_\alpha/\rho_t$.  Then, with $B_0$ and $C^\star$ from \eqref{eq:PsiUB} and \eqref{eq:Cs}, for every $T\ge1$ the output index $t^\star$ of \prettyref{alg:psk} with horizon $T$ satisfies
\[
 \Gphi_{\bar\alpha}(x_{t^\star})\le C^\star(B_0)T^{-1/3}
 \qquad\text{for every $\bar\alpha\ge\alpha_{t^\star}$.}
\]
\end{formaltheorem}

The proof of \prettyref{thm:main} rests on the following Lyapunov inequality, proved in \prettyref{sec:mainresult}.

\begin{theorem}[Lyapunov inequality]\label{thm:lyap}
Let \prettyref{ass:A} hold globally for $\{f_i\}_{i=1}^K$, and assume $(\mathrm H_\rho)$.  Then, for every $t\ge0$,
\begin{gather*}
 \Psi_{t+1}\le\Psi_t-U_t
 +\omega_3\frac{\rho_{t+1}-\rho_t}{\rho_t^2},
 \qquad
 \Ecal_t^2\le\frac{16}{c_\alpha}\rho_tU_t,
 \qquad
 \sum_{i=1}^KR_i(a_i(w_t),y_{i,t})\le\frac{B_0}{\rho_t},\\
 \sum_{t\ge0}U_t\le B_0.
\end{gather*}
\end{theorem}

\begin{remark}[Reading the rate]\label{rem:complexity}
For fixed data and $K$, any fixed $m\ge1$ gives $O(\epsilon^{-3})$ oracle work. Only the stopping horizon depends on the requested accuracy.
\end{remark}

\subsection{Proofs}\label{sec:proof}

\paragraph{Idea of the proof.} The trackers follow the exact stage solutions while one outer step decreases the penalty function.  Slow growth of the penalty parameter lets this decrease pay for the drift of the penalty function, and the resulting budget controls the certificate $C_t$.  A half-window argument balances the penalized residual $\sqrt{\rho_T/T}$ with the conversion bias $1/\rho_T$.  The last step turns penalized stationarity into stationarity of the implicit loss $\varphi$: the multipliers of the penalty function satisfy almost the same backward recursion as the exact adjoints of $\varphi$, and their difference is of order $E+1/\rho$, where $E$ is the penalized residual.

\paragraph{Global and local bounds.} This appendix and \prettyref{sec:ns} differ in the numerical domain $\mathcal U$ of \prettyref{ass:A}.  \prettyref{thm:lyap} and \ref{thm:main} assume that \prettyref{ass:A} holds globally (\prettyref{sec:P}), that is, with $\mathcal U=\R^{n_x+n_y}\times\R^{n_y}$, so every upper bound in it holds at every point.  The wall surrogates of \prettyref{sec:ns} have no global bounds, so that appendix works on a compact box, which serves as a bounded $\mathcal U$.  For this reason, every intermediate result below that uses the upper bounds is stated for a general convex $\mathcal U$, together with the points and segments that must lie in $\mathcal U$; the exception is \prettyref{lem:bdd}, which uses global bounds.  When \prettyref{ass:A} holds globally, these containments hold automatically.  \prettyref{sec:ns} checks them and then reuses the same results.

\paragraph{Hats and bars.} In this appendix (and where \prettyref{sec:ns} reuses its lemmas), a hat marks a quantity on the exact chain at the control $x$: $\widehat y_i(x)$, $\widehat w(x)$, $\widehat a_i$, and the matrices and adjoints $\widehat A_i,\widehat B_i,\widehat H_i,\widehat J_i,\widehat Q_i,\widehat p_i$ evaluated there; the argument $x$ of hatted quantities is suppressed when $x$ is fixed.  A bar marks an average along the segment from $s_i(a_i(w))$ to $y_i$ at the current trajectory $w$: $\bar H_i$, $\bar J_i$, the averaged matrix $\bar Q_i=\bar J_i^\top\bar H_i^{-1}$, and the multipliers $\bar p_i=-\rho\nabla_yf_i(a_i,y_i)=-\rho\bar H_i(y_i-s_i(a_i))$ built from them.  Two hats belong to the algorithm instead: $\hnabla_t$ is the PSK direction, and $\widehat e_{i,t+1}$ is the tracker error after the inner loop.  The bars in $\Phib$, $\bar\alpha$, $\bar r_\rho$, and $\bar e_{i,t}$ have their own meanings.  Elsewhere, hats and bars are defined where they appear.

\subsubsection{Auxiliary lemmas}\label{sec:auxlemmas}

The results in this part are standard or have short proofs.  Each one that uses the upper bounds is stated for a general convex $\mathcal U$.

\begin{proposition}[Stage maps, saddle elimination, and gap bounds]\label{prop:identity}
Let \prettyref{A:f}, \prettyref{A:sc}, and~\prettyref{A:F} hold.  Then:
\begin{enumerate}[label=\textnormal{(\roman*)}]
\item\label{id:map} For every $i$ and $a$, $f_i(a,\cdot)$ has the unique minimizer $s_i(a)$, and
\begin{equation}\label{eq:stationarity}
 \nabla_y f_i(a,s_i(a))=0,
 \qquad f_i^\star(a)=f_i(a,s_i(a))\in\mathbb R.
\end{equation}
\item\label{id:primal} For every $\rho>0$, $\Phi_\rho$ is real-valued and
\begin{equation}\label{eq:primal}
 \Phi_\rho(w)=\max_z\cL_\rho(w,z)
 =F(w)+\rho\sum_{i=1}^K R_i(a_i(w),y_i),
 \qquad z_i^\star(w)=s_i(a_i(w)).
\end{equation}
\item\label{id:gap} If $\mathfrak f$ is proper and $c$-strongly convex with minimizer $y^\star$, then
\begin{equation*}
 \mathfrak f(y)\ge \mathfrak f(y^\star)+\frac c2\|y-y^\star\|^2.
\end{equation*}
Consequently,
\begin{equation}\label{eq:errorbound}
 R_i(a,y)\ge\frac\mu2\|y-s_i(a)\|^2.
\end{equation}
\item\label{id:gapup} If $\{a\}\times\operatorname{conv}\{s_i(a),y\}\subseteq \mathcal U$, then
\begin{equation*}
 R_i(a,y)\le\frac{L_{yy}}2\|y-s_i(a)\|^2.
\end{equation*}
\end{enumerate}
\end{proposition}

\begin{proof}
For part~\prettyref{id:map}, $f_i(a,\cdot)$ is $C^1$ and $\mu$-strongly convex on $\R^{n_y}$, so $f_i(a,y)\ge f_i(a,0)+\langle\nabla_yf_i(a,0),y\rangle+\frac\mu2\|y\|^2$ for every $y$.  The right side tends to $+\infty$ as $\|y\|\to\infty$, so $f_i(a,\cdot)$ is coercive and, being continuous, has a minimizer; strict convexity makes it unique.  First-order optimality gives \eqref{eq:stationarity}.

For part~\prettyref{id:primal}, the maximization in \eqref{eq:saddle} separates over the blocks $z_i$, and by part~\prettyref{id:map} each block maximum is $-\rho f_i^\star(a_i(w))$; it is finite and attained only at $z_i=s_i(a_i(w))$.  This gives \eqref{eq:primal}.

Part~\prettyref{id:gap} is the strong-convexity inequality \citep[equation~(3.13)]{bubeck2015convex} at the point $y^\star$ with the subgradient $0\in\partial\mathfrak f(y^\star)$.  Taking $\mathfrak f=f_i(a,\cdot)$ and $c=\mu$ gives \eqref{eq:errorbound}.

For part~\prettyref{id:gapup}, the descent lemma \citep[Lemma~3.4]{bubeck2015convex} gives
\begin{equation*}
 f_i(a,y)\le f_i(a,s_i(a))+\langle\nabla_yf_i(a,s_i(a)),y-s_i(a)\rangle+\frac{L_{yy}}2\|y-s_i(a)\|^2,
\end{equation*}
and its proof uses \eqref{eq:Ly} only on the segment from $s_i(a)$ to $y$, which lies in $\mathcal U$ by hypothesis.  Use \eqref{eq:stationarity} and subtract $f_i^\star(a)=f_i(a,s_i(a))$.
\end{proof}

\begin{remark}[Reading the gap bounds]
The lower bound is global because strong convexity is global.  The upper bound uses only the set $\{a\}\times\operatorname{conv}\{s_i(a),y\}$.
\end{remark}

\begin{lemma}[Stage contraction]\label{lem:gd}
Let \prettyref{ass:A} hold.  Fix a stage $i$, a parameter $a$, and $0<\gamma\le L_{yy}^{-1}$, and let $\mathsf T_{i,a}$ be the tracker step \eqref{eq:T}.  If the upper bound \eqref{eq:Ly} holds globally, then
\begin{equation*}
 \|\mathsf T_{i,a}(y)-s_i(a)\|
 \le q\|y-s_i(a)\|,
 \qquad q=(1-\gamma\mu)^{1/2},
\end{equation*}
for every $y$. By \prettyref{A:C2}, the same conclusion holds locally whenever $\{a\}\times\operatorname{conv}\{s_i(a),y\}\subseteq \mathcal U$.
\end{lemma}

\begin{proof}
For the global claim, put $s=s_i(a)$, $u=y-s$, and $v=\nabla_yf_i(a,y)$, and note that $\nabla_yf_i(a,s)=0$ by \eqref{eq:stationarity}.  Since $f_i(a,\cdot)$ is $L_{yy}$-smooth and $\mu$-strongly convex on $\R^{n_y}$, \citet[Lemma~3.11]{bubeck2015convex} gives
\[
 \langle v,u\rangle\ge\frac{\mu L_{yy}}{\mu+L_{yy}}\|u\|^2+\frac1{\mu+L_{yy}}\|v\|^2.
\]
As in the proof of \citet[Theorem~3.12]{bubeck2015convex},
\begin{equation*}
 \|\mathsf T_{i,a}(y)-s\|^2
 =\|u\|^2-2\gamma\langle v,u\rangle+\gamma^2\|v\|^2
 \le\Bigl(1-\frac{2\gamma\mu L_{yy}}{\mu+L_{yy}}\Bigr)\|u\|^2
 +\gamma\Bigl(\gamma-\frac2{\mu+L_{yy}}\Bigr)\|v\|^2
 \le(1-\gamma\mu)\|u\|^2,
\end{equation*}
because $\gamma\le L_{yy}^{-1}\le2/(\mu+L_{yy})$ and $2L_{yy}/(\mu+L_{yy})\ge1$.

For the local claim, \prettyref{A:C2} and \eqref{eq:stationarity} give $\nabla_yf_i(a,y)=\bar H(y-s_i(a))$ with
\[
 \bar H:=\int_0^1H_i\bigl(a,s_i(a)+\theta(y-s_i(a))\bigr)\,d\theta
\]
(at the generic parameter $a$).  The matrix $\bar H$ is symmetric, as an average of symmetric Hessians, and $\mu I\preceq\bar H\preceq L_{yy}I$ by the segment hypothesis and the bounds stated after \prettyref{ass:A}.  Since $\gamma\le L_{yy}^{-1}$, the eigenvalues of the symmetric matrix $I-\gamma\bar H$ lie in $[0,1-\gamma\mu]$, so $\|\mathsf T_{i,a}(y)-s_i(a)\|=\|(I-\gamma\bar H)(y-s_i(a))\|\le(1-\gamma\mu)\|y-s_i(a)\|\le q\|y-s_i(a)\|$.
\end{proof}

\begin{lemma}[The stage map is Lipschitz]\label{lem:lip}
Let \prettyref{ass:A} hold.  Fix a stage $i$ and parameters $a,a'$.  If
\begin{equation*}
 \operatorname{conv}\{a,a'\}\times\{s_i(a)\}
 \ \cup\ \{a'\}\times\operatorname{conv}\{s_i(a),s_i(a')\}
 \subseteq \mathcal U,
\end{equation*}
then
\begin{equation}\label{eq:lip}
 \|s_i(a)-s_i(a')\|\le M\|a-a'\|,
 \qquad M=\frac{L_{ay}}\mu.
\end{equation}
\end{lemma}

\begin{proof}
Set $s=s_i(a)$ and $s'=s_i(a')$.  By \eqref{eq:stationarity}, $\nabla_yf_i(a,s)=\nabla_yf_i(a',s')=0$.  Strong monotonicity at parameter $a'$ gives the first line below, the two stationarity identities give the second, and Cauchy--Schwarz with the $L_{ay}$ bound of \eqref{eq:La} on the fixed-output parameter segment $\operatorname{conv}\{a,a'\}\times\{s\}$ in the hypothesis gives the third.
\begin{align*}
 \mu\|s-s'\|^2
 &\le\langle\nabla_yf_i(a',s)-\nabla_yf_i(a',s'),s-s'\rangle\\
 &=\langle\nabla_yf_i(a',s)-\nabla_yf_i(a,s),s-s'\rangle\\
 &\le L_{ay}\|a-a'\|\,\|s-s'\|.
\end{align*}
If $s\ne s'$, divide by $\|s-s'\|$; otherwise \eqref{eq:lip} is immediate.
\end{proof}

\begin{remark}[Local stage-map comparisons]
The two sets in the hypothesis are $\operatorname{conv}\{a,a'\}\times\{s_i(a)\}$, which contains the graph point $(a,s_i(a))$, and $\{a'\}\times\operatorname{conv}\{s_i(a),s_i(a')\}$, which contains the graph point $(a',s_i(a'))$.  Global strong monotonicity supplies the outgoing-state comparison.
\end{remark}

\begin{lemma}[Danskin, smoothness, and the computable residual]\label{lem:value}
Let \prettyref{ass:A} hold.  Then:
\begin{enumerate}[label=\textnormal{(\roman*)}]
\item\label{sp:danskin} For every $i$, the value $f_i^\star$ is $C^1$ and
\begin{equation}\label{eq:danskin}
 \nabla f_i^\star(a)=\nabla_af_i(a,s_i(a)).
\end{equation}
\item\label{sp:Lv} If $\operatorname{conv}\{a,a'\}\times\{s_i(a)\}$ and $\{a'\}\times\operatorname{conv}\{s_i(a),s_i(a')\}$ lie in $\mathcal U$, then
\begin{equation*}
 \|\nabla f_i^\star(a)-\nabla f_i^\star(a')\|
 \le L_\star\|a-a'\|,
 \qquad L_\star=L_{aa}+L_{ay}M.
\end{equation*}
\item\label{sp:smooth} For every $\rho>0$, $\Phi_\rho$ is $C^1$ and, with $a_i=a_i(w)$,
\begin{equation}\label{eq:gradPhi}
\begin{aligned}
 \nabla_x\Phi_\rho(w)
 &=\nabla_xF(w)+\rho\sum_{i=1}^K
 [\nabla_xf_i(a_i,y_i)-\nabla_xf_i(a_i,s_i(a_i))],\\
 \nabla_{y_i}\Phi_\rho(w)
 &=\nabla_{y_i}F(w)+\rho\nabla_yf_i(a_i,y_i)\\
 &\quad+\rho[\nabla_{y'}f_{i+1}(a_{i+1},y_{i+1})
 -\nabla_{y'}f_{i+1}(a_{i+1},s_{i+1}(a_{i+1}))]
 \quad(i<K),\\
 \nabla_{y_K}\Phi_\rho(w)
 &=\nabla_{y_K}F(w)+\rho\nabla_yf_K(a_K,y_K).
\end{aligned}
\end{equation}
Let $\mathcal I$ be a line segment in trajectory space.  If for every $u,v\in\mathcal I$ and every $i$,
\begin{equation}\label{eq:valuecontain}
\begin{gathered}
 \operatorname{conv}\{(a_i(u),u_{y_i}),(a_i(v),v_{y_i})\}\subseteq \mathcal U,\\
 \operatorname{conv}\{a_i(u),a_i(v)\}\times\{s_i(a_i(u))\}\subseteq \mathcal U,\\
 \{a_i(v)\}\times\operatorname{conv}\{s_i(a_i(u)),s_i(a_i(v))\}\subseteq \mathcal U,
\end{gathered}
\end{equation}
then $\|\nabla\Phi_\rho(u)-\nabla\Phi_\rho(v)\| \le(L_F+\rho K(L_f+L_\star))\|u-v\|$.
\item\label{sp:stat} For every $w\in\cW$,
\begin{equation*}
 \partial\Phib(w)=\nabla\Phi_\rho(w)+N_\cW(w),
 \qquad N_\cW(w)=N_\cX(x)\times\{0\}^K.
\end{equation*}
\item\label{sp:res} The scalar residual $\cG_\alpha^\rho$ is continuous and
\begin{equation*}
 \cG_\alpha^\rho(w)=0\quad\Longleftrightarrow\quad0\in\partial\Phib(w).
\end{equation*}
\end{enumerate}
\end{lemma}

\begin{proof}
For part~\prettyref{sp:danskin}, global strong monotonicity gives $\mu\|s_i(a')-s_i(a)\|\le\|\nabla_yf_i(a',s_i(a))-\nabla_yf_i(a,s_i(a))\|$, so global $C^1$ regularity makes $s_i$, and hence $f_i^\star=f_i(\cdot,s_i(\cdot))$, continuous.  By \prettyref{prop:identity}\prettyref{id:gap}, the set $\{y:f_i(a,y)\le c\}$ lies in the ball of radius $(2(c-f_i^\star(a))_+/\mu)^{1/2}$ around $s_i(a)$.  Since $s_i$ and $f_i^\star$ are continuous, for $a$ near any fixed point these balls lie in one bounded set, so $f_i(a,y)$ is level-bounded in $y$ locally uniformly in $a$ \citep[Definition~1.16]{rockafellar1998variational}.  Since $f_i$ is $C^1$ and the minimizer set is the singleton $\{s_i(a)\}$, \citet[Theorem~10.58]{rockafellar1998variational} shows that $f_i^\star$ is strictly differentiable at every $a$ with $\partial f_i^\star(a)=\{\nabla_af_i(a,s_i(a))\}$.  This is \eqref{eq:danskin} by \citet[Theorem~9.18]{rockafellar1998variational}, and \citet[Corollary~9.19]{rockafellar1998variational} gives $f_i^\star\in C^1$.

For part~\prettyref{sp:Lv}, write $s=s_i(a)$ and $s'=s_i(a')$.  The two sets in the hypothesis of part~\prettyref{sp:Lv} justify \eqref{eq:La} at the compared points and form the hypothesis of \prettyref{lem:lip}, which gives $\|s-s'\|\le M\|a-a'\|$.  Hence
\begin{align*}
 \|\nabla f_i^\star(a)-\nabla f_i^\star(a')\|
 &\le\|\nabla_af_i(a,s)-\nabla_af_i(a',s)\|
 +\|\nabla_af_i(a',s)-\nabla_af_i(a',s')\|\\
 &\le L_{aa}\|a-a'\|+L_{ay}\|s-s'\|\\
 &\le(L_{aa}+L_{ay}M)\|a-a'\|.
\end{align*}

For part~\prettyref{sp:smooth}, \prettyref{prop:identity} gives
\begin{equation*}
 \Phi_\rho(w)=F(w)+\rho\sum_{i=1}^K[f_i(a_i(w),y_i)-f_i^\star(a_i(w))].
\end{equation*}
Differentiate this identity and use \eqref{eq:danskin} to obtain \eqref{eq:gradPhi}.  For a $C^1$ function $\mathfrak f$ whose gradient is $L_{\mathfrak f}$-Lipschitz and an affine coordinate selector $\mathsf E_i$ whose linear part, also written $\mathsf E_i$, has $\|\mathsf E_i\|\le1$,
\begin{equation*}
 \|\mathsf E_i^\top\nabla\mathfrak f(\mathsf E_iu)-\mathsf E_i^\top\nabla\mathfrak f(\mathsf E_iv)\|
 \le\|\mathsf E_i\|^2L_{\mathfrak f}\|u-v\|
 \le L_{\mathfrak f}\|u-v\|.
\end{equation*}
Apply this with $\mathsf E_iu=(a_i(u),u_{y_i})$ to each lifted $f_i$ term, and with $\mathsf E_iu=a_i(u)$, combined with part~\prettyref{sp:Lv}, to each lifted $f_i^\star$ term.  The containments \eqref{eq:valuecontain} justify every comparison.  Summation gives
\begin{equation*}
 \|\nabla\Phi_\rho(u)-\nabla\Phi_\rho(v)\|
 \le\bigl(L_F+\rho K(L_f+L_\star)\bigr)\|u-v\|.
\end{equation*}

For part~\prettyref{sp:stat}, $\Phi_\rho$ is $C^1$ by part~\prettyref{sp:smooth} and $\iota_\cW$ is finite at $w$, so \citet[Exercises~8.8(c) and~8.14]{rockafellar1998variational} give $\partial\Phib(w)=\nabla\Phi_\rho(w)+N_\cW(w)$.  The product form follows from \citet[Proposition~6.41]{rockafellar1998variational}, because $\cW=\cX\times\mathbb R^{Kn_y}$ has closed factors and the normal cone to $\mathbb R^{Kn_y}$ is $\{0\}$.

For part~\prettyref{sp:res}, first let $w\in\cW$.  Then $\cG_\alpha^\rho(w)=0$ means $\PiX_\cW(w-\alpha\nabla\Phi_\rho(w))=w$.  Since $\cW$ is closed and convex, \citet[Proposition~6.17]{rockafellar1998variational} makes this equivalent to $-\alpha\nabla\Phi_\rho(w)\in N_\cW(w)$, hence to $-\nabla\Phi_\rho(w)\in N_\cW(w)$ because $N_\cW(w)$ is a cone, and hence to $0\in\partial\Phib(w)$ by part~\prettyref{sp:stat}.  If $w\notin\cW$, both sides fail: the projection lies in $\cW$ and so differs from $w$, which gives $\cG_\alpha^\rho(w)\ne0$, and $\partial\Phib(w)$ is empty because $\Phib(w)=+\infty$.  Continuity follows from continuity of $\nabla\Phi_\rho$, the projection, and the norm.
\end{proof}

\begin{lemma}[The outer step minimizes an explicit model]\label{lem:model}
Let \prettyref{ass:A} hold, and run one iteration of \prettyref{alg:psk} at penalty parameter $\rho>0$ and step $\alpha>0$.  Put
\begin{equation*}
 \cM_t(u):=\Phi_\rho(w_t)+\langle\hnabla_t,u-w_t\rangle
 +\frac1{2\alpha}\|u-w_t\|^2+\iota_\cW(u).
\end{equation*}
Then $\cM_t$ is $1/\alpha$-strongly convex, and
\begin{equation}\label{eq:wproj}
 w_{t+1}=\argmin_u\cM_t(u)=\PiX_\cW(w_t-\alpha\hnabla_t).
\end{equation}
Moreover, $\hnabla_t=\nabla\Phi_\rho(w_t)+\delta_t$, and, with $a_i=a_i(w_t)$,
\begin{align}
 \delta_t^x
 &=\rho\sum_{i=1}^K[\nabla_xf_i(a_i,s_i(a_i))-
 \nabla_xf_i(a_i,b_{i,t+1})],
 \qquad
 \delta_t^{y_K}=0,\label{eq:epsx}\\
 \delta_t^{y_i}
 &=\rho[\nabla_{y'}f_{i+1}(a_{i+1},s_{i+1}(a_{i+1}))-
 \nabla_{y'}f_{i+1}(a_{i+1},b_{i+1,t+1})]
 \quad(i<K).\label{eq:epsy}
\end{align}
\end{lemma}

\begin{proof}
Here $\rho_t=\rho$ and $\alpha_t=\alpha$.  The indicator is convex and the quadratic is $1/\alpha$-strongly convex.  Completing the square gives
\[
 \cM_t(u)=\frac1{2\alpha}\|u-(w_t-\alpha\hnabla_t)\|^2+\iota_\cW(u)+\Phi_\rho(w_t)-\frac\alpha2\|\hnabla_t\|^2,
\]
so $\cM_t$ is minimized exactly at the projection of $w_t-\alpha\hnabla_t$ onto the nonempty, closed, convex set $\cW$, which is unique.  This is \eqref{eq:wproj}.  By \eqref{eq:epsdef}, $\delta_t=\hnabla_t-\nabla\Phi_\rho(w_t)$.  Compare the block formulas \eqref{icml:blocks} with \eqref{eq:gradPhi}.  The only replacement is $s_i(a_i(w_t))\mapsto b_{i,t+1}$ in the parameter-gradient differences.  Subtraction gives \eqref{eq:epsx} and \eqref{eq:epsy}.
\end{proof}

\begin{remark}[Local points used by the model]
The model identity is algebraic.  A local numerical estimate of its error terms uses the current point $(a_i(w_t),y_{i,t})$, the exact graph point $(a_i(w_t),s_i(a_i(w_t)))$, the final tracker point $(a_i(w_t),b_{i,t+1})$, and the vertical segments joining the exact point to the other two.
\end{remark}

\begin{lemma}[Projected residual versus step size]\label{lem:mono}
Let $\mathcal Z$ be nonempty, closed, and convex.  For fixed $x,v$, the function $\alpha^{-1}\|x-\PiX_\mathcal Z(x-\alpha v)\|$ is nonincreasing on $(0,\infty)$.
\end{lemma}

\begin{proof}
Let $0<\alpha<\alpha'$, and put $u=x-\PiX_\mathcal Z(x-\alpha v)$ and $u'=x-\PiX_\mathcal Z(x-\alpha'v)$.  The two projection inequalities are
\begin{align*}
 \langle u-\alpha v,u-u'\rangle&\le0,\\
 \langle u'-\alpha'v,u'-u\rangle&\le0.
\end{align*}
Multiply the first by $\alpha'$, the second by $\alpha$, and add.
\begin{equation*}
 \alpha'\|u\|^2+\alpha\|u'\|^2
 \le(\alpha+\alpha')\langle u,u'\rangle
 \le(\alpha+\alpha')\|u\|\|u'\|,
 \qquad
 (\alpha'\|u\|-\alpha\|u'\|)(\|u\|-\|u'\|)\le0.
\end{equation*}
If $\alpha'\|u\|<\alpha\|u'\|$, the product forces $\|u\|\ge\|u'\|$, and then $\alpha'\|u\|<\alpha\|u'\|\le\alpha\|u\|$, that is, $(\alpha'-\alpha)\|u\|<0$, which is impossible since $\alpha'>\alpha$ and $\|u\|\ge0$.  Hence $\|u'\|/\alpha'\le\|u\|/\alpha$.
\end{proof}

\subsubsection{Proof of the main result}\label{sec:mainresult}

We first bound one iteration and the penalty drift; together, these bounds give the Lyapunov inequality of \prettyref{thm:lyap}.  We then convert penalized stationarity into a bound on $\Gphi$ and prove \prettyref{thm:main}.

\paragraph{One iteration.}

This block bounds the descent, the direction error, and the tracker errors of one iteration at a fixed penalty parameter.  In the global case the containments below hold automatically; \prettyref{sec:ns} checks them for the band surrogates.  One iteration uses the following local containments, each required for every $1\le i\le K$.
\begin{subequations}\label{eq:onestepcontain}
\begin{align}
 &\{a_i(w_t)\}\times\operatorname{conv}\{s_i(a_i(w_t)),b_{i,t}^{(k)}\}
 \subseteq \mathcal U\quad(0\le k\le m),
 \label{eq:onestepcontain-tracker}\\
 &\{a_i(w_t)\}\times\operatorname{conv}\{s_i(a_i(w_t)),y_{i,t}\}\subseteq \mathcal U,\notag\\
 &\operatorname{conv}\{a_i(w_t),a_i(w_{t+1})\}\times\{s_i(a_i(w_t))\}
 \subseteq \mathcal U,
 \label{eq:onestepcontain-parameter}\\
 &\{a_i(w_{t+1})\}\times
 \operatorname{conv}\{s_i(a_i(w_t)),s_i(a_i(w_{t+1}))\}\subseteq \mathcal U
 .
 \label{eq:onestepcontain-solution}
\end{align}
\end{subequations}

\begin{lemma}[One iteration]\label{lem:onestep}
Let \prettyref{ass:A} hold. Run one iteration of \prettyref{alg:psk} at penalty parameter $\rho>0$ and step $\alpha>0$, and assume \eqref{eq:onestepcontain}.  Use the local form of \prettyref{lem:gd} when $\mathcal U$ is not global. Then:
\begin{enumerate}[label=\textnormal{(\roman*)}]
\item\label{os:i} $\langle\hnabla_t,\Delta w_t\rangle\le-\alpha^{-1}\|\Delta w_t\|^2$;
\item\label{os:ii} $\|\delta_t\|^2\le\kappa_0\rho^2\Xi_t$;
\item\label{os:iii} $\widehat e_{i,t+1}\le\widetilde qe_{i,t}$, $e_{i,t+1}\le\widetilde qe_{i,t}+M\|(\Delta w_t^x,\Delta w_t^{y_{i-1}})\|$, and
\begin{equation}\label{eq:track3}
 \Xi_{t+1}\le\widetilde q\,\Xi_t+
 \frac{KM^2}{1-\widetilde q}\|\Delta w_t\|^2;
\end{equation}
\item\label{os:iv} $\cG_\alpha^\rho(w_t)\le\alpha^{-1}\|\Delta w_t\|+\|\delta_t\|$.
\end{enumerate}
\end{lemma}

\begin{proof}[Proof of \prettyref{lem:onestep}]
By \prettyref{lem:model}, the model $\cM_t$ is $1/\alpha$-strongly convex and is minimized by $w_{t+1}$.  Hence \prettyref{prop:identity}\prettyref{id:gap}, with $1/\alpha$ in place of $c$, gives
\[
 \cM_t(w_t)\ge\cM_t(w_{t+1})+\frac1{2\alpha}\|\Delta w_t\|^2.
\]
Since $w_t\in\cW$, we have $\cM_t(w_t)=\Phi_\rho(w_t)$.  Since $w_{t+1}\in\cW$, we have $\cM_t(w_{t+1})=\Phi_\rho(w_t)+\langle\hnabla_t,\Delta w_t\rangle+\frac1{2\alpha}\|\Delta w_t\|^2$.  Substituting both values gives $\langle\hnabla_t,\Delta w_t\rangle\le-\alpha^{-1}\|\Delta w_t\|^2$, which is part~\prettyref{os:i}.

For part~\prettyref{os:ii}, \eqref{eq:onestepcontain-tracker} allows every tracker step, so $m$ applications of \prettyref{lem:gd} give $\widehat e_{i,t+1}\le\widetilde qe_{i,t}$.  The case $k=m$ of that containment contains the segment from $s_i(a_i(w_t))$ to $b_{i,t+1}$, so \eqref{eq:Lay2}, \eqref{eq:epsx}, and \eqref{eq:epsy} give
\begin{equation*}
 \|\delta_t^x\|\le\rho L_{ay}\widetilde q\sum_{i=1}^Ke_{i,t},
 \qquad
 \|\delta_t^{y_i}\|\le\rho L_{ay}\widetilde q\,e_{i+1,t}\quad(i<K),
 \qquad \delta_t^{y_K}=0.
\end{equation*}
Cauchy--Schwarz in the first bound gives
\begin{equation}\label{eq:epsbound}
 \|\delta_t^x\|^2\le K(L_{ay}\widetilde q)^2\rho^2\Xi_t,
\end{equation}
while squaring and summing the remaining bounds gives
\begin{equation}\label{eq:epsbound2}
 \sum_{i=1}^{K-1}\|\delta_t^{y_i}\|^2
 \le(L_{ay}\widetilde q)^2\rho^2\Xi_t.
\end{equation}
Adding \eqref{eq:epsbound} and \eqref{eq:epsbound2} and using \eqref{eq:kappa0} proves part~\prettyref{os:ii}.

For part~\prettyref{os:iii}, define the map motion
\begin{equation*}
 \eta_{i,t}^{\rm map}:=
 \|s_i(a_i(w_t))-s_i(a_i(w_{t+1}))\|.
\end{equation*}
The triangle inequality and tracker contraction give the first two lines below.  The containments \eqref{eq:onestepcontain-parameter} and \eqref{eq:onestepcontain-solution} justify \prettyref{lem:lip}, which gives the third
\begin{align*}
 e_{i,t+1}
 &\le\widehat e_{i,t+1}+\eta_{i,t}^{\rm map}\\
 &\le\widetilde qe_{i,t}+\eta_{i,t}^{\rm map},\\
 \eta_{i,t}^{\rm map}
 &\le M\|(\Delta w_t^x,\Delta w_t^{y_{i-1}})\|.
\end{align*}
For $c\in[0,1)$ and $e,\eta\ge0$,
\[
 (ce+\eta)^2\le ce^2+\frac{\eta^2}{1-c},
\]
because the difference of the right-hand side and the left-hand side is $c(\eta-(1-c)e)^2/(1-c)$.  With $c=\widetilde q$,
\begin{equation}\label{eq:track2}
 e_{i,t+1}^2\le\widetilde q e_{i,t}^2+
 \frac{(\eta_{i,t}^{\rm map})^2}{1-\widetilde q}.
\end{equation}
Moreover, since $\Delta w_t^{y_0}=0$,
\begin{align}
 \sum_{i=1}^K(\eta_{i,t}^{\rm map})^2
 &\le M^2\sum_{i=1}^K
 (\|\Delta w_t^x\|^2+\|\Delta w_t^{y_{i-1}}\|^2)\notag\\
 &\le M^2(K\|\Delta w_t^x\|^2+\|\Delta w_t^y\|^2)
 \le KM^2\|\Delta w_t\|^2.
 \label{eq:tausum}
\end{align}
Summing \eqref{eq:track2} over $i$ and using \eqref{eq:tausum} proves \eqref{eq:track3}.

Finally, the scalar triangle inequality and projection nonexpansiveness give
\[
\begin{aligned}
 \alpha\cG_\alpha^\rho(w_t)
 &=\|w_t-\PiX_\cW(w_t-\alpha\nabla\Phi_\rho(w_t))\|\\
 &\le\|w_t-w_{t+1}\|
 +\|\PiX_\cW(w_t-\alpha\hnabla_t)
 -\PiX_\cW(w_t-\alpha\nabla\Phi_\rho(w_t))\|\\
 &\le\|\Delta w_t\|+\alpha\|\delta_t\|.
\end{aligned}
\]
Division by $\alpha$ proves part~\prettyref{os:iv}.
\end{proof}

\begin{lemma}[One step at a frozen penalty]\label{lem:descent}
Let \prettyref{ass:A} hold. Run one iteration at penalty parameter $\rho>0$ and step $\alpha>0$, assume \eqref{eq:onestepcontain}, and let $\mathcal I=[w_t,w_{t+1}]$.  Assume, for every $u,v\in\mathcal I$, the three stagewise containments in \eqref{eq:valuecontain}.  If $\alpha(L_F+\rho K(L_f+L_\star))\le1/2$, then
\begin{equation*}
 \Phi_\rho(w_{t+1})
 \le\Phi_\rho(w_t)-\frac1{2\alpha}\|\Delta w_t\|^2
 +\alpha\|\delta_t\|^2.
\end{equation*}
For $\alpha=c_\alpha/\rho$, the last two coefficients are $-\rho/(2c_\alpha)$ and $c_\alpha/\rho$.
\end{lemma}

\begin{proof}
Put $L=L_F+\rho K(L_f+L_\star)$.  \prettyref{lem:value}\prettyref{sp:smooth}, whose hypothesis \eqref{eq:valuecontain} holds for all $u,v\in\mathcal I$, shows that $\nabla\Phi_\rho$ is $L$-Lipschitz on $\mathcal I$.  This gives the descent lemma on $\mathcal I$:
\[
\begin{aligned}
 &\Phi_\rho(w_{t+1})-\Phi_\rho(w_t)-\langle\nabla\Phi_\rho(w_t),\Delta w_t\rangle\\
 &\quad=\int_0^1\langle\nabla\Phi_\rho(w_t+\theta\Delta w_t)-\nabla\Phi_\rho(w_t),\Delta w_t\rangle\,d\theta
 \le\int_0^1L\theta\|\Delta w_t\|^2\,d\theta=\frac L2\|\Delta w_t\|^2.
\end{aligned}
\]
By \eqref{eq:epsdef} with $\rho_t=\rho$, $\nabla\Phi_\rho(w_t)=\hnabla_t-\delta_t$, and \prettyref{lem:onestep}\prettyref{os:i} supplies the projected-step inequality. Therefore
\begin{align*}
 \Phi_\rho(w_{t+1})
 &\le\Phi_\rho(w_t)+\langle\nabla\Phi_\rho(w_t),\Delta w_t\rangle
 +\frac L2\|\Delta w_t\|^2\\
 &=\Phi_\rho(w_t)+\langle\hnabla_t,\Delta w_t\rangle
 -\langle\delta_t,\Delta w_t\rangle+\frac L2\|\Delta w_t\|^2\\
 &\le\Phi_\rho(w_t)-\frac1\alpha\|\Delta w_t\|^2
 +\|\delta_t\|\|\Delta w_t\|+\frac L2\|\Delta w_t\|^2\\
 &\le\Phi_\rho(w_t)-\left(\frac3{4\alpha}-\frac L2\right)\|\Delta w_t\|^2
 +\alpha\|\delta_t\|^2\\
 &\le\Phi_\rho(w_t)-\frac1{2\alpha}\|\Delta w_t\|^2
 +\alpha\|\delta_t\|^2.
\end{align*}
The fourth line uses $\|\delta_t\|\|\Delta w_t\|\le\alpha\|\delta_t\|^2+\|\Delta w_t\|^2/(4\alpha)$, and the last uses $\alpha L\le1/2$.
\end{proof}

\paragraph{Penalty drift.} When the penalty parameter grows from $\rho_t$ to $\rho_{t+1}$, the penalty function at $w_{t+1}$ grows by $(\rho_{t+1}-\rho_t)\sum_iR_i(a_i(w_{t+1}),y_{i,t+1})$, and \prettyref{prop:identity}\prettyref{id:gapup} bounds each $R_i$ through the stage residual $r_i$.  \prettyref{lem:allgaps} and \ref{lem:resid} bound the $r_i$ terms, and \prettyref{lem:drift} then bounds the drift.

\begin{lemma}[All $K$ stage gaps]\label{lem:allgaps}
Let \prettyref{ass:A} hold.  Let $\rho>0$ and $w\in\cW$.  Assume $\{a_i(w)\}\times\operatorname{conv}\{s_i(a_i(w)),y_i\}\subseteq \mathcal U$ for every $i$.  Put
\begin{equation*}
 \eta_i^\Phi:=\|\nabla_{y_i}\Phi_\rho(w)\|,
 \qquad \eta_i^F:=\|\nabla_{y_i}F(w)\|.
\end{equation*}
Then
\begin{equation*}
 r_i(w)\le\frac1{\rho\mu}\sum_{j=i}^KM^{j-i}(\eta_j^\Phi+\eta_j^F).
\end{equation*}
For any $\alpha>0$, with $E:=\cG_\alpha^\rho(w)$ and $\bar r_\rho:=(E+G_F)/(\rho\mu)$,
\begin{equation}\label{eq:allgapscoarse}
 \eta_i^\Phi\le E,
 \qquad r_i(w)\le\bar r_\rho W_i.
\end{equation}
\end{lemma}

\begin{proof}
Write $a_i=a_i(w)$, $s_{i,w}=s_i(a_i)$, and $r_i=r_i(w)$.  Strong monotonicity, $\nabla_yf_i(a_i,s_{i,w})=0$ from \eqref{eq:stationarity}, and Cauchy--Schwarz give the first two lines below; dividing by $r_i$ gives the third (which is trivial if $r_i=0$).
\begin{align*}
 \mu r_i^2
 &\le\langle\nabla_yf_i(a_i,y_i)-\nabla_yf_i(a_i,s_{i,w}),y_i-s_{i,w}\rangle\\
 &\le\|\nabla_yf_i(a_i,y_i)\|r_i,\\
 \mu r_i&\le\|\nabla_yf_i(a_i,y_i)\|.
\end{align*}
For $i<K$, the hypothesis $\{a_{i+1}(w)\}\times\operatorname{conv}\{s_{i+1}(a_{i+1}(w)),y_{i+1}\}\subseteq \mathcal U$ and \eqref{eq:Lay2} give
\begin{equation*}
 \|\nabla_{y'}f_{i+1}(a_{i+1},y_{i+1})-
 \nabla_{y'}f_{i+1}(a_{i+1},s_{i+1,w})\|
 \le L_{ay}r_{i+1}.
\end{equation*}
Solve \eqref{eq:gradPhi} for $\rho\nabla_yf_i(a_i,y_i)$, take norms, and use $\mu r_i\le\|\nabla_yf_i(a_i,y_i)\|$.  At the last stage, this gives $\rho\mu r_K\le \eta_K^\Phi+\eta_K^F$.  For $i<K$, the last display also bounds the successor bracket, so
\begin{align*}
 \rho\mu r_i
 &\le \eta_i^\Phi+\eta_i^F+\rho L_{ay}r_{i+1},\\
 r_i&\le\frac{\eta_i^\Phi+\eta_i^F}{\rho\mu}+Mr_{i+1}.
\end{align*}
Backward induction from $i=K$ gives
\begin{equation*}
 r_i\le\frac1{\rho\mu}\sum_{j=i}^KM^{j-i}(\eta_j^\Phi+\eta_j^F).
\end{equation*}
The state blocks are unprojected, so \eqref{icml:Gprod} gives $\eta_i^\Phi\le E$.  Also $\eta_i^F\le G_F$.  Hence
\begin{equation*}
 r_i\le\frac{E+G_F}{\rho\mu}\sum_{j=i}^KM^{j-i}
 =\bar r_\rho W_i.
\end{equation*}
\end{proof}

\begin{lemma}[Stage residuals at the current iterate]\label{lem:resid}
Let \prettyref{ass:A} hold. Run one iteration at penalty parameter $\rho>0$ with $\alpha=c_\alpha/\rho$, and assume \eqref{eq:onestepcontain}.  Then
\begin{equation}\label{eq:residbound}
 \sum_{i=1}^Kr_i(w_t)^2
 \le\frac{2W_1^2}{\mu^2}
 \left(\frac{2\|\Delta w_t\|^2}{c_\alpha^2}
 +2\kappa_0\Xi_t+\frac{G_F^2}{\rho^2}\right).
\end{equation}
\end{lemma}

\begin{proof}
Put $c_j=\eta_j^\Phi+\eta_j^F$ at $(w_t,\rho)$, $\mathbf c:=(c_j)_{j=1}^K$, $\mathbf r:=(r_i(w_t))_{i=1}^K$, and $\mathbf T:=(M^{j-i}\mathbf1_{i\le j})_{i,j=1}^K$. Since $w_t\in\cW$ and the second line of \eqref{eq:onestepcontain} is the segment hypothesis of \prettyref{lem:allgaps} at $w=w_t$, that lemma gives the componentwise bound $0\le\mathbf r\le(\rho\mu)^{-1}\mathbf T\mathbf c$.  Every row sum of $\mathbf T$ is some $W_i\le W_1$, and every column sum is $\sum_{k=0}^{j-1}M^k\le W_1$.  Thus
\begin{align*}
 \|\mathbf T\mathbf c\|_2&\le\|\mathbf T\|_2\|\mathbf c\|_2
 \le\sqrt{\|\mathbf T\|_\infty\|\mathbf T\|_1}\,\|\mathbf c\|_2
 \le W_1\|\mathbf c\|_2,\\
 \sum_ir_i(w_t)^2&\le\frac{W_1^2}{\rho^2\mu^2}\sum_jc_j^2.
\end{align*}
The state update is unprojected.  \prettyref{lem:model} therefore gives
\begin{equation*}
 \nabla_y\Phi_\rho(w_t)=-\alpha^{-1}\Delta w_t^y-\delta_t^y.
\end{equation*}
Together with \prettyref{lem:onestep}\prettyref{os:ii}, this gives
\begin{align*}
 \sum_j(\eta_j^\Phi)^2
 &\le2\alpha^{-2}\|\Delta w_t^y\|^2+2\|\delta_t^y\|^2\\
 &\le\frac{2\rho^2}{c_\alpha^2}\|\Delta w_t\|^2
 +2\kappa_0\rho^2\Xi_t,\\
 \sum_j(\eta_j^F)^2&\le G_F^2,\\
 \sum_jc_j^2
 &\le2\sum_j(\eta_j^\Phi)^2+2\sum_j(\eta_j^F)^2\\
 &\le\frac{4\rho^2}{c_\alpha^2}\|\Delta w_t\|^2
 +4\kappa_0\rho^2\Xi_t+2G_F^2.
\end{align*}
Insert this into the bound on $\sum_ir_i(w_t)^2$ above.  The factor $\rho^2$ cancels in the first two terms, the last term becomes $G_F^2/\rho^2$, and \eqref{eq:residbound} follows.
\end{proof}

\begin{lemma}[Penalty drift]\label{lem:drift}
Let \prettyref{ass:A} hold. Run one iteration at penalty parameter $\rho>0$ with $\alpha=c_\alpha/\rho$, let $\rho'\ge\rho$, and assume \eqref{eq:onestepcontain}.  Assume also, for every $i$,
\begin{equation}\label{eq:driftcontain}
 \{a_i(w_{t+1})\}\times\operatorname{conv}\{s_i(a_i(w_{t+1})),y_{i,t+1}\}\subseteq \mathcal U.
\end{equation}
Then
\begin{equation*}
 \Phi_{\rho'}(w_{t+1})-\Phi_\rho(w_{t+1})
 \le(\rho'-\rho)
 \left(\omega_1\|\Delta w_t\|^2+\omega_2\Xi_t+\frac{\omega_3}{\rho^2}\right).
\end{equation*}
\end{lemma}

\begin{proof}
By \eqref{eq:primal}, the difference equals $(\rho'-\rho)\sum_iR_i(a_i(w_{t+1}),y_{i,t+1})$.  \prettyref{prop:identity}\prettyref{id:gapup} and \eqref{eq:driftcontain} then give
\begin{align*}
 \Phi_{\rho'}(w_{t+1})-\Phi_\rho(w_{t+1})
 &=(\rho'-\rho)\sum_iR_i(a_i(w_{t+1}),y_{i,t+1})\\
 &\le(\rho'-\rho)\frac{L_{yy}}2
 \sum_ir_i(w_{t+1})^2.
\end{align*}
Note that $a_i(w_{t+1})-a_i(w_t)=(\Delta w_t^x,\Delta w_t^{y_{i-1}})$.  The triangle inequality gives the first line below.  The containments \eqref{eq:onestepcontain-parameter} and \eqref{eq:onestepcontain-solution}, together with \prettyref{lem:lip}, bound the last term in it and give the second line
\begin{align*}
 r_i(w_{t+1})
 &\le\|y_{i,t+1}-y_{i,t}\|+\|y_{i,t}-s_i(a_i(w_t))\|
 +\|s_i(a_i(w_t))-s_i(a_i(w_{t+1}))\|\\
 &\le r_i(w_t)+\|\Delta w_t^{y_i}\|
 +M\|(\Delta w_t^x,\Delta w_t^{y_{i-1}})\|,\\
 r_i(w_{t+1})^2
 &\le2r_i(w_t)^2+2\left(\|\Delta w_t^{y_i}\|
 +M\|(\Delta w_t^x,\Delta w_t^{y_{i-1}})\|\right)^2\\
 &\le2r_i(w_t)^2+4\|\Delta w_t^{y_i}\|^2
 +4M^2\|(\Delta w_t^x,\Delta w_t^{y_{i-1}})\|^2.
\end{align*}
After summation over $i$, using $\Delta w_t^{y_0}=0$,
\begin{equation*}
 \sum_ir_i(w_{t+1})^2
 \le2\sum_ir_i(w_t)^2+4(1+KM^2)\|\Delta w_t\|^2.
\end{equation*}
Insert \eqref{eq:residbound} and multiply by $L_{yy}/2$.  The coefficients are
\begin{align*}
 \omega_1&=L_{yy}\left[
 \frac{4W_1^2}{\mu^2c_\alpha^2}+2(1+KM^2)\right],\\
 \omega_2&=\frac{4L_{yy}W_1^2\kappa_0}{\mu^2},\\
 \omega_3&=\frac{2L_{yy}W_1^2G_F^2}{\mu^2}.
\end{align*}
These are exactly \eqref{eq:omega}.
\end{proof}

\paragraph{Certificate and Lyapunov inequality.} This block links the certificate $C_t$ to the penalized residual $\cG_\alpha^\rho(w_t)$ and proves \prettyref{thm:lyap}, stated in \prettyref{sec:statements}.

\begin{lemma}[The certificate]\label{lem:cert}
Let \prettyref{ass:A} hold. Run one iteration at penalty parameter $\rho>0$ and step $\alpha>0$, and assume \eqref{eq:onestepcontain}. Let $\xi_{i,t}^{(0)}=\nabla_yf_i(a_i(w_t),b_{i,t})$ and let $C_t$ be the certificate computed by \prettyref{alg:psk}.  Then, with $\kappa_0$ from \eqref{eq:kappa0},
\begin{equation}\label{eq:certbound}
 (\cG_\alpha^\rho(w_t))^2
 \le2\alpha^{-2}\|\Delta w_t\|^2+2\kappa_0\rho^2\Xi_t
 \le C_t^2.
\end{equation}
If $\alpha=c_\alpha/\rho$, then also
\begin{equation}\label{eq:certupper}
 C_t^2\le\frac{16\chi^2}{c_\alpha}\rho U_t.
\end{equation}
\end{lemma}

\begin{proof}[Proof of \prettyref{lem:cert}]
Here $\rho_t=\rho$ and $\alpha_t=\alpha$.  \prettyref{lem:onestep}\prettyref{os:iv}, followed by $(a+b)^2\le2a^2+2b^2$ and \prettyref{lem:onestep}\prettyref{os:ii}, gives $(\cG_\alpha^\rho(w_t))^2\le2\alpha^{-2}\|\Delta w_t\|^2+2\kappa_0\rho^2\Xi_t$ in \eqref{eq:certbound}.  The case $k=0$ of \eqref{eq:onestepcontain-tracker} contains the segment from $s_i(a_i(w_t))$ to $b_{i,t}$.  Strong monotonicity and the local upper bound \eqref{eq:Ly} on that segment give
\[
 \mu e_{i,t}\le\|\xi_{i,t}^{(0)}\|\le L_{yy}e_{i,t}.
\]
(Both bounds use $\nabla_yf_i(a_i(w_t),s_i(a_i(w_t)))=0$ from \eqref{eq:stationarity}.  The upper bound is then \eqref{eq:Ly} on that segment.  The lower bound is immediate when $e_{i,t}=0$; otherwise it follows by pairing the gradient with $b_{i,t}-s_i(a_i(w_t))$, using strong monotonicity and Cauchy--Schwarz, and dividing by $e_{i,t}$.)  Hence, for $\bar e_{i,t}:=\mu^{-1}\|\xi_{i,t}^{(0)}\|$, $e_{i,t}\le\bar e_{i,t}\le\chi e_{i,t}$.  The lower bound gives $\Xi_t\le\mu^{-2}\sum_i\|\xi_{i,t}^{(0)}\|^2$, and substitution gives $2\alpha^{-2}\|\Delta w_t\|^2+2\kappa_0\rho^2\Xi_t\le C_t^2$ in \eqref{eq:certbound}, using the definition of $C_t$ in \prettyref{alg:psk}.

Conversely, using $\alpha=c_\alpha/\rho$, $\bar e_{i,t}\le\chi e_{i,t}$, and $\chi=L_{yy}/\mu>1$ (so the $\|\Delta w_t\|^2$ term may also be multiplied by $\chi^2$),
\begin{align*}
 C_t^2
 &\le\chi^2\left(
 \frac{2\rho^2}{c_\alpha^2}\|\Delta w_t\|^2
 +2\kappa_0\rho^2\Xi_t\right)\\
 &\le\chi^2\left(
 \frac{2\rho^2}{c_\alpha^2}\|\Delta w_t\|^2
 +16\kappa_0\rho^2\Xi_t\right)\\
 &=\chi^2\frac{16}{c_\alpha}\rho
 \left[\rho\left(\frac{\|\Delta w_t\|^2}{8c_\alpha}
 +c_\alpha\kappa_0\Xi_t\right)\right]\\
 &=\frac{16\chi^2}{c_\alpha}\rho U_t.
\end{align*}
This is \eqref{eq:certupper}.
\end{proof}

\begin{lemma}[Global compact sublevels]\label{lem:bdd}
Let \prettyref{ass:A} hold globally, and let $\rho>0$.  Put $s_{\max}:=\max_{1\le i\le K}\max_{x\in\cX}\|s_i(x,0)\|$.  For $c\ge\Flo$, define, for this fixed $\rho$, $r_{\rm sub}(c):=[2(c-\Flo)/(\rho\mu)]^{1/2}$.  If $w\in\cW$ and $\Phib(w)\le c$, then
\begin{equation}\label{eq:sublevel}
 r_i(w)\le r_{\rm sub}(c),
 \qquad
 \|y_i\|\le M^i\|y_0\|
 +\bigl(r_{\rm sub}(c)+s_{\max}\bigr)\sum_{k=0}^{i-1}M^k.
\end{equation}
Hence every sublevel set of $\Phib$ is compact.
\end{lemma}

\begin{proof}
\prettyref{lem:lip} makes every $x\mapsto s_i(x,0)$ continuous, so compactness of $\cX$ and finiteness of the stage family give $s_{\max}<\infty$.  By \eqref{eq:primal} and \eqref{eq:errorbound}, $\Phi_\rho\ge F\ge\Flo$, so the sublevel set is empty when $c<\Flo$; let $c\ge\Flo$.  If $\Phib(w)\le c$, then
\begin{equation*}
 \rho R_i(a_i(w),y_i)
 \le\rho\sum_{j=1}^KR_j(a_j(w),y_j)
 =\Phi_\rho(w)-F(w)
 \le c-\Flo.
\end{equation*}
The lower gap bound \eqref{eq:errorbound} gives $r_i(w)\le r_{\rm sub}(c)$.  \prettyref{lem:lip}, with global bounds and $a=(x,y_{i-1})$, $a'=(x,0)$, together with $\|y_i\|\le r_i(w)+\|s_i(a_i(w))\|$, gives
\begin{align*}
 \|s_i(x,y_{i-1})\|&\le \|s_i(x,0)\|+M\|y_{i-1}\|
 \le s_{\max}+M\|y_{i-1}\|,\\
 \|y_i\|&\le r_{\rm sub}(c)+s_{\max}+M\|y_{i-1}\|.
\end{align*}
Unrolling this recurrence from $y_0$ yields \eqref{eq:sublevel}. Since also $x\in\cX$ and $\cX$ is compact, the sublevel set is bounded.  It is closed because $\cW$ is closed and $F$, $f_i$, and $s_i$, and therefore $f_i^\star$ and $\Phi_\rho$, are continuous.  Being closed and bounded in finite dimension, it is compact.
\end{proof}

\begin{remark}[Global compactness]
Local upper moduli do not control stage-map growth outside their domains.  The compact-sublevel conclusion therefore belongs only to the global theory.
\end{remark}

\begin{proof}[Proof of \prettyref{thm:lyap}]
By the global hypothesis in \prettyref{thm:lyap}, all upper bounds in \prettyref{ass:A} hold globally, so every lemma above applies without checking its containments.  Moreover,
\begin{equation*}
 \alpha_t\bigl(L_F+\rho_tK(L_f+L_\star)\bigr)
 =c_\alpha\left(\frac{L_F}{\rho_t}+K(L_f+L_\star)\right)
 \le c_\alpha\left(\frac{L_F}{\rho_0}+K(L_f+L_\star)\right)
 \le\frac12
\end{equation*}
by $\rho_t\ge\rho_0$, which holds because the schedule in \eqref{eq:Hrho} is nondecreasing, and by \eqref{eq:amax}.  \prettyref{lem:descent} and \ref{lem:drift}, together with \prettyref{lem:onestep}\prettyref{os:ii} and~\ref{os:iii}, give
\begin{align*}
 \Psi_{t+1}
 &=\Phi_t(w_{t+1})
 +\bigl[\Phi_{t+1}(w_{t+1})-\Phi_t(w_{t+1})\bigr]
 +c_\Psi\rho_{t+1}\Xi_{t+1}\\
 &\le\Phi_t(w_t)-\frac{\rho_t}{2c_\alpha}\|\Delta w_t\|^2
 +\frac{c_\alpha}{\rho_t}\|\delta_t\|^2
 +(\rho_{t+1}-\rho_t)\Bigl(\omega_1\|\Delta w_t\|^2+\omega_2\Xi_t
 +\frac{\omega_3}{\rho_t^2}\Bigr)\\
 &\quad+c_\Psi\rho_{t+1}\Bigl(\widetilde q\,\Xi_t
 +\frac{KM^2}{1-\widetilde q}\|\Delta w_t\|^2\Bigr)\\
 &\le\Phi_t(w_t)\\
 &\quad+\left[-\frac{\rho_t}{2c_\alpha}
 +(\rho_{t+1}-\rho_t)\omega_1
 +\frac{c_\Psi\rho_{t+1}KM^2}{1-\widetilde q}\right]\|\Delta w_t\|^2\\
 &\quad+\left[c_\alpha\kappa_0\rho_t
 +(\rho_{t+1}-\rho_t)\omega_2
 +c_\Psi\rho_{t+1}\widetilde q\right]\Xi_t\\
 &\quad+\omega_3\frac{\rho_{t+1}-\rho_t}{\rho_t^2}.
\end{align*}
Since $0\le\rho_{t+1}-\rho_t\le\beta\rho_t$ and $\rho_{t+1}\le(1+\beta)\rho_t\le2\rho_t$ (here $\beta\le\beta_{\max}\le(1-\widetilde q)/4<1$ by \eqref{eq:betamax}), and using $\beta\le1/(4c_\alpha\omega_1)$ from \eqref{eq:betamax} and $c_\Psi$ from \eqref{eq:rhoconsts}, the coefficient of $\|\Delta w_t\|^2$, divided by $\rho_t$, obeys
\begin{align*}
 -\frac1{2c_\alpha}+\beta\omega_1
 +\frac{2c_\Psi KM^2}{1-\widetilde q}
 &\le-\frac1{2c_\alpha}+\frac1{4c_\alpha}
 +\frac{8c_\alpha\kappa_0KM^2}{(1-\widetilde q)^2}\\
 &\le-\frac1{2c_\alpha}+\frac1{4c_\alpha}+\frac1{8c_\alpha}\\
 &=-\frac1{8c_\alpha}.
\end{align*}
The bound $c_\alpha\le(1-\widetilde q)/(8M\sqrt{K\kappa_0})$ from \eqref{eq:amax} gives $8c_\alpha\kappa_0KM^2/(1-\widetilde q)^2\le1/(8c_\alpha)$.  Next, consider the coefficient of $\Xi_t$; we bound its excess over $c_\Psi\rho_t$, divided by $\rho_t$.  By \eqref{eq:betamax}, $\beta\le c_\alpha\mu^2/(4L_{yy}W_1^2)$, so $\beta\omega_2\le c_\alpha\kappa_0$, and $\beta\le(1-\widetilde q)/4$.  By \eqref{eq:rhoconsts}, $2c_\alpha\kappa_0=c_\Psi(1-\widetilde q)/2$.  Hence
\begin{align*}
 c_\alpha\kappa_0+\beta\omega_2
 +c_\Psi[(1+\beta)\widetilde q-1]
 &\le2c_\alpha\kappa_0+c_\Psi\widetilde q
 +\frac{c_\Psi(1-\widetilde q)}4-c_\Psi\\
 &=\frac{c_\Psi(1-\widetilde q)}2+c_\Psi\widetilde q
 +\frac{c_\Psi(1-\widetilde q)}4-c_\Psi\\
 &=-\frac{c_\Psi(1-\widetilde q)}4\\
 &=-c_\alpha\kappa_0.
\end{align*}
Thus
\[
 \Psi_{t+1}\le\Psi_t-U_t+
 \omega_3\frac{\rho_{t+1}-\rho_t}{\rho_t^2}.
\]

For the budget, the drift term satisfies
\begin{equation*}
 \frac{\rho_{t+1}-\rho_t}{\rho_t^2}
 =\frac{\rho_{t+1}}{\rho_t}
 \frac{\rho_{t+1}-\rho_t}{\rho_t\rho_{t+1}}
 =\frac{\rho_{t+1}}{\rho_t}
 \left(\frac1{\rho_t}-\frac1{\rho_{t+1}}\right)
 \le(1+\beta)\left(\frac1{\rho_t}-\frac1{\rho_{t+1}}\right).
\end{equation*}
Hence
\begin{equation*}
 \sum_{t=0}^{T-1}\frac{\rho_{t+1}-\rho_t}{\rho_t^2}
 \le(1+\beta)\left(\frac1{\rho_0}-\frac1{\rho_T}\right)
 \le\frac{1+\beta}{\rho_0}
 \le\frac2{\rho_0}.
\end{equation*}
Summing the Lyapunov inequality, and using $\Psi_T\ge\Phi_{\rho_T}(w_T)\ge F(w_T)\ge\Flo$ (by $\Xi_T\ge0$, \eqref{eq:primal}, and \eqref{eq:errorbound}), gives
\begin{equation*}
 \sum_{t<T}U_t
 \le\Psi_0-\Psi_T+\frac{2\omega_3}{\rho_0}
 \le\Psi_0-\Flo+\frac{2\omega_3}{\rho_0}
 =B_0.
\end{equation*}
Letting $T\to\infty$ proves the summable budget.  By \eqref{eq:certbound} in \prettyref{lem:cert} and the definition of $U_t$ in \eqref{eq:PsiUB},
\begin{equation*}
 \Ecal_t^2
 \le2\alpha_t^{-2}\|\Delta w_t\|^2+2\kappa_0\rho_t^2\Xi_t
 \le\frac{16}{c_\alpha}\rho_tU_t.
\end{equation*}

Finally,
\begin{equation*}
 \Phi_{\rho_0}(w_t)
 \le\Phi_{\rho_t}(w_t)
 \le\Psi_t
 \le\Psi_0+\frac{2\omega_3}{\rho_0}
 =B_0+\Flo.
\end{equation*}
The first inequality uses $\rho_t\ge\rho_0$ and $R_i\ge0$, and the second uses $\Xi_t\ge0$.  The third is the summed inequality above with $t$ in place of $T$, after dropping $\sum_{s<t}U_s\ge0$.  \prettyref{lem:bdd} makes this sublevel set compact, so the sequence $(w_t)$ is bounded.  Also
\begin{equation*}
 \rho_t\sum_iR_i(a_i(w_t),y_{i,t})
 =\Phi_{\rho_t}(w_t)-F(w_t)
 \le B_0+\Flo-\Flo
 =B_0.
\end{equation*}
Dividing by $\rho_t>0$ gives $\sum_iR_i(a_i(w_t),y_{i,t})\le B_0/\rho_t$.
\end{proof}

\paragraph{Conversion to the implicit loss.} It remains to compare $\nabla_x\Phi_\rho$ with $\nabla\varphi$.  We write $\nabla\varphi$ in adjoint form (\prettyref{lem:dphi}), write $\nabla_x\Phi_\rho$ in the same form with multipliers (\prettyref{lem:mult}), and compare the two (\prettyref{lem:conv}).

\begin{lemma}[The implicit gradient in adjoint form]\label{lem:dphi}
Let \prettyref{ass:A} hold.  Define $\widehat w(x):=(x,\widehat y_1(x),\ldots,\widehat y_K(x))$ and $\widehat a_i:=(x,\widehat y_{i-1}(x))$.  Then every $s_i$ and every $\widehat y_i$ is $C^1$.  At every exact graph point in $\mathcal U$,
\begin{equation}\label{eq:Ds}
 D_as_i(a)=-H_i(a,s_i(a))^{-1}J_i(a,s_i(a)),
 \qquad \|D_as_i(a)\|\le M.
\end{equation}
If $(\widehat a_i,\widehat y_i)\in \mathcal U$ for all $i$, define
\[
 [\widehat A_i\ \widehat B_i]:=D_as_i(\widehat a_i).
\]
For $X_i:=D_x\widehat y_i(x)$,
\begin{equation*}
 X_0=0,
 \qquad
 X_i=\widehat A_i+\widehat B_iX_{i-1},
 \qquad
 X_i=\sum_{j=1}^i
 \widehat B_i\widehat B_{i-1}\cdots\widehat B_{j+1}\widehat A_j,
 \qquad
 \|X_i\|\le\sum_{j=1}^iM^j,
\end{equation*}
where $\widehat A_i$ and $\widehat B_i$ are the $x$ and $y'$ blocks, respectively. Put
\[
 \widehat Q_i:=J_i(\widehat a_i,\widehat y_i)^\top
 H_i(\widehat a_i,\widehat y_i)^{-1}
\]
and define
\begin{equation}\label{eq:adj}
 \widehat p_K:=\nabla_{y_K}F(\widehat w),
 \qquad
 \widehat p_i:=\nabla_{y_i}F(\widehat w)
 -\widehat Q_{i+1}^{y'}\widehat p_{i+1}\quad(i<K).
\end{equation}
Then
\begin{equation}\label{eq:gradphi}
 \nabla\varphi(x)=\nabla_xF(\widehat w)-
 \sum_{i=1}^K\widehat Q_i^x\widehat p_i,
 \qquad
 \|\widehat p_i\|\le\sum_{j=i}^KM^{j-i}\|\nabla_{y_j}F(\widehat w)\|
 \le G_FW_i.
\end{equation}
\end{lemma}

\begin{proof}
Global strong convexity gives $H_i\succeq\mu I$ at every point.  At a graph point in $\mathcal U$, $\|J_i\|\le L_{ay}$.  The implicit function theorem applied to $\nabla_yf_i(a,s_i(a))=0$ gives
\begin{align*}
 0&=J_i(a,s_i(a))+H_i(a,s_i(a))D_as_i(a),\\
 D_as_i(a)&=-H_i(a,s_i(a))^{-1}J_i(a,s_i(a)),\\
 \|D_as_i(a)\|&\le\frac{L_{ay}}\mu=M.
\end{align*}
Each local implicit-function solution $y(a)$ satisfies $\nabla_yf_i(a,y(a))=0$, so by strong convexity it equals the unique minimizer $s_i(a)$; hence $s_i$ coincides near every point with a $C^1$ map and is $C^1$ everywhere.  We need every $\widehat y_i$ to be $C^1$ because $\varphi=F\circ\widehat w$ must be $C^1$: the residual $\cG^\varphi_\alpha$ in \eqref{eq:G} is defined only for $C^1$ functions, and the chain rule below needs $D_x\widehat y_{i-1}$.  We prove this claim by induction on $i$.  The map $\widehat y_0\equiv y_0$ is constant.  If $\widehat y_{i-1}$ is $C^1$, then $\widehat y_i(x)=s_i(x,\widehat y_{i-1}(x))$ is the composition of the $C^1$ map $x\mapsto(x,\widehat y_{i-1}(x))$ with the $C^1$ map $s_i$, so it is $C^1$.  The chain rule gives $X_i=\widehat A_i+\widehat B_iX_{i-1}$, and induction on $i$, with $X_0=0$ and the empty product equal to the identity, gives
\begin{align*}
 X_i
 &=\widehat A_i+\widehat B_i\widehat A_{i-1}
 +\widehat B_i\widehat B_{i-1}X_{i-2}
 =\sum_{j=1}^i
 \widehat B_i\widehat B_{i-1}\cdots\widehat B_{j+1}\widehat A_j,\\
 \|X_i\|
 &\le\sum_{j=1}^iM^{i-j}M
 =\sum_{j=1}^iM^j.
\end{align*}
By \eqref{eq:Ds} and the symmetry of $H_i$, $\widehat A_i^\top=-\widehat Q_i^x$ and $\widehat B_i^\top=-\widehat Q_i^{y'}$.  Since $\varphi=F\circ\widehat w$ and every $\widehat y_i$ is $C^1$ with $D_x\widehat y_i=X_i$, the chain rule gives
\begin{equation*}
 \nabla\varphi(x)=\nabla_xF(\widehat w)+\sum_{i=1}^KX_i^\top\nabla_{y_i}F(\widehat w).
\end{equation*}
It remains to rewrite the sum.  By \eqref{eq:adj} and $\widehat B_{i+1}^\top=-\widehat Q_{i+1}^{y'}$, $\nabla_{y_i}F(\widehat w)=\widehat p_i-\widehat B_{i+1}^\top\widehat p_{i+1}$ for $i<K$, and $\nabla_{y_K}F(\widehat w)=\widehat p_K$.  Below, the successor term $\widehat B_{i+1}^\top\widehat p_{i+1}$ is simply absent at $i=K$; equivalently, put $\widehat p_{K+1}=0$.  Using also $X_0=0$ and $X_i-\widehat B_iX_{i-1}=\widehat A_i$,
\begin{align*}
 \sum_{i=1}^KX_i^\top\nabla_{y_i}F(\widehat w)
 &=\sum_iX_i^\top(\widehat p_i-\widehat B_{i+1}^\top\widehat p_{i+1})\\
 &=\sum_iX_i^\top\widehat p_i-
 \sum_{i=2}^K(\widehat B_iX_{i-1})^\top\widehat p_i\\
 &=\sum_i(X_i-\widehat B_iX_{i-1})^\top\widehat p_i\\
 &=\sum_i\widehat A_i^\top\widehat p_i
 =-\sum_i\widehat Q_i^x\widehat p_i.
\end{align*}
Substituting this into the chain-rule formula gives \eqref{eq:gradphi}.  At $i=K$, $\|\widehat p_K\|=\|\nabla_{y_K}F(\widehat w)\|\le\|\nabla F(\widehat w)\|\le G_F=G_FW_K$ by \prettyref{A:F}.  For $i<K$, $\|\widehat Q_{i+1}^{y'}\|=\|\widehat B_{i+1}\|\le\|D_as_{i+1}(\widehat a_{i+1})\|\le M$ by \eqref{eq:Ds}, and backward induction gives
\begin{equation*}
 \|\widehat p_i\|
 \le\|\nabla_{y_i}F(\widehat w)\|+M\|\widehat p_{i+1}\|
 \le\sum_{j=i}^KM^{j-i}\|\nabla_{y_j}F(\widehat w)\|
 \le G_FW_i.
\end{equation*}
\end{proof}

\begin{lemma}[Exact multiplier identities]\label{lem:mult}
Under \prettyref{ass:A}, let $\rho>0$ and $w\in\cW$.  Assume $\{a_i(w)\}\times\operatorname{conv}\{s_i(a_i(w)),y_i\}\subseteq \mathcal U$ for every $i$.  Put, with $a_i=a_i(w)$,
\begin{equation}\label{eq:bars}
\begin{aligned}
 \bar H_i&:=\int_0^1H_i(a_i,s_i(a_i)+\theta(y_i-s_i(a_i)))\,d\theta,\\
 \bar J_i&:=\int_0^1J_i(a_i,s_i(a_i)+\theta(y_i-s_i(a_i)))\,d\theta,\\
 \bar Q_i&:=\bar J_i^\top\bar H_i^{-1},
 \qquad \bar p_i:=-\rho\nabla_yf_i(a_i,y_i).
\end{aligned}
\end{equation}
Then $\mu I\preceq\bar H_i\preceq L_{yy}I$, $\|\bar J_i\|\le L_{ay}$, $\|\bar Q_i\|\le M$, and
\begin{equation}\label{eq:multid}
\begin{aligned}
 \nabla_x\Phi_\rho(w)
 &=\nabla_xF(w)-\sum_{i=1}^K\bar Q_i^x\bar p_i,\\
 \nabla_{y_i}\Phi_\rho(w)
 &=\nabla_{y_i}F(w)-\bar p_i-\bar Q_{i+1}^{y'}\bar p_{i+1}\quad(i<K),\\
 \nabla_{y_K}\Phi_\rho(w)
 &=\nabla_{y_K}F(w)-\bar p_K.
\end{aligned}
\end{equation}
Moreover, for $\alpha>0$ and $E=\cG_\alpha^\rho(w)$,
\begin{equation}\label{eq:gbound}
 \mu\rho r_i(w)\le\|\bar p_i\|
 \le L_{yy}\rho r_i(w)
 \le\frac{L_{yy}}\mu(E+G_F)W_i.
\end{equation}
\end{lemma}

\begin{proof}
The fundamental theorem of calculus along the segment from $s_i(a_i)$ to $y_i$ gives the two identities below; the second uses that the mixed block $\partial_y\nabla_af_i$ equals $J_i^\top$, by symmetry of the Hessian of the $C^2$ function $f_i$.
\begin{align*}
 \nabla_yf_i(a_i,y_i)-\nabla_yf_i(a_i,s_i(a_i))
 &=\bar H_i(y_i-s_i(a_i)),\\
 \nabla_af_i(a_i,y_i)-\nabla_af_i(a_i,s_i(a_i))
 &=\bar J_i^\top(y_i-s_i(a_i)).
\end{align*}
The matrix $\bar H_i$ is symmetric, as an average of symmetric matrices.  The segment hypothesis and the bounds stated after \prettyref{ass:A}, which extend to the closure of $\mathcal U$ by continuity, give the spectral and norm bounds; in particular $\|\bar Q_i\|\le\|\bar J_i\|\,\|\bar H_i^{-1}\|\le L_{ay}/\mu=M$.  Since $\nabla_yf_i(a_i,s_i(a_i))=0$ by \eqref{eq:stationarity}, the first identity gives $\bar p_i=-\rho\bar H_i(y_i-s_i(a_i))$, so
\begin{equation*}
 \rho[\nabla_af_i(a_i,y_i)-\nabla_af_i(a_i,s_i(a_i))]
 =\rho\bar J_i^\top(y_i-s_i(a_i))
 =-\bar J_i^\top\bar H_i^{-1}\bar p_i
 =-\bar Q_i\bar p_i.
\end{equation*}
Substitute this identity and $\rho\nabla_yf_i(a_i,y_i)=-\bar p_i$ into \eqref{eq:gradPhi}.  This proves \eqref{eq:multid}.  The spectral bounds give
\begin{equation*}
 \mu\rho r_i(w)
 \le\rho\|\bar H_i(y_i-s_i(a_i))\|
 =\|\bar p_i\|
 \le L_{yy}\rho r_i(w).
\end{equation*}
Applying \eqref{eq:allgapscoarse} yields $L_{yy}\rho r_i(w)\le(L_{yy}/\mu)(E+G_F)W_i$ in \eqref{eq:gbound}.
\end{proof}

We use the conversion constants of \eqref{eq:convconsts}.

\begin{lemma}[Implicit conversion]\label{lem:conv}
Under \prettyref{ass:A}, let $w=(x,y_1,\ldots,y_K)\in\cW$, $\rho,\alpha>0$, and $E:=\cG_\alpha^\rho(w)$.  Recall $\widehat a_i=(x,\widehat y_{i-1}(x))$ from \prettyref{lem:dphi}.  Assume, for every $i$, the conversion rectangle
\begin{equation}\label{eq:convrect}
 \operatorname{conv}\{a_i(w),\widehat a_i\}
 \times\operatorname{conv}\{y_i,s_i(a_i(w)),\widehat y_i(x)\}
 \subseteq \mathcal U.
\end{equation}
Then
\begin{equation}\label{eq:convgrad}
 \|\nabla\varphi(x)-\nabla_x\Phi_\rho(w)\|
 \le c_{\rm adj}E+
 \frac{c_{\rm b1}(E+G_F)+c_{\rm b2}(E+G_F)^2}{\rho},
\end{equation}
and
\begin{equation}\label{eq:convproj}
 \Gphi_\alpha(x)
 \le c_\varphi E+
 \frac{c_{\rm b1}(E+G_F)+c_{\rm b2}(E+G_F)^2}{\rho}.
\end{equation}
\end{lemma}

\begin{proof}
Write $a_i=a_i(w)$ and $s_{i,w}=s_i(a_i)$, and recall $\bar r_\rho=(E+G_F)/(\rho\mu)$ from \prettyref{lem:allgaps}.  The rectangle \eqref{eq:convrect} contains every set used below: $\{a_i\}\times\operatorname{conv}\{y_i,s_{i,w}\}$, $\operatorname{conv}\{a_i,\widehat a_i\}\times\{s_{i,w}\}$, $\{\widehat a_i\}\times\operatorname{conv}\{s_{i,w},\widehat y_i(x)\}$, and every segment joining a point of $\{a_i\}\times\operatorname{conv}\{y_i,s_{i,w}\}$ to $(\widehat a_i,\widehat y_i(x))$. \prettyref{lem:allgaps} gives $r_i(w)\le\bar r_\rho W_i$. The multiplier recursion \eqref{eq:multid}, with $\|\nabla_{y_i}\Phi_\rho(w)\|\le E$ by \eqref{icml:Gprod} and $\|\bar Q_{i+1}^{y'}\|\le\|\bar Q_{i+1}\|\le M$ by \prettyref{lem:mult}, gives
\begin{align*}
 \|\bar p_i\|
 &\le E+\|\nabla_{y_i}F(w)\|+M\|\bar p_{i+1}\|\\
 &\le\sum_{j=i}^KM^{j-i}(E+\|\nabla_{y_j}F(w)\|)\\
 &\le(E+G_F)W_i,
\end{align*}
where the successor term is absent at $i=K$.

Put $\eta_0^{\rm traj}=0$ and $\eta_i^{\rm traj}:=\|y_i-\widehat y_i(x)\|$.  Since $a_i-\widehat a_i=(0,y_{i-1}-\widehat y_{i-1}(x))$, we have $\|a_i-\widehat a_i\|=\eta_{i-1}^{\rm traj}$.  Also $\widehat y_i(x)=s_i(\widehat a_i)$, and \prettyref{lem:lip}, whose two sets lie in the rectangle, gives $\|s_i(a_i)-s_i(\widehat a_i)\|\le M\eta_{i-1}^{\rm traj}$.  Hence
\begin{align*}
 \eta_i^{\rm traj}
 &\le\|y_i-s_i(a_i)\|+
 \|s_i(a_i)-s_i(\widehat a_i)\|\\
 &\le r_i(w)+M\eta_{i-1}^{\rm traj}\\
 &\le\bar r_\rho W_i+M\bar r_\rho\mathfrak H_{i-1}\\
 &=\bar r_\rho\mathfrak H_i,\\
 \|w-\widehat w(x)\|
 &=\left(\sum_i(\eta_i^{\rm traj})^2\right)^{1/2}
 \le\bar r_\rho\mathfrak H_2^{\rm all}.
\end{align*}
Define the maximum distance of the barred stage-$i$ segment from the exact derivative point by
\[
 \eta_i^{\rm der}:=\max_{\theta\in[0,1]}
 \|(a_i,s_i(a_i)+\theta(y_i-s_i(a_i)))-(\widehat a_i,\widehat y_i)\|.
\]
Then, using again $\|a_i-\widehat a_i\|=\eta_{i-1}^{\rm traj}$ and \prettyref{lem:lip},
\begin{align*}
 \eta_i^{\rm der}
 &\le\|a_i-\widehat a_i\|+
 \|s_i(a_i)-s_i(\widehat a_i)\|+r_i(w)\\
 &\le\eta_{i-1}^{\rm traj}+M\eta_{i-1}^{\rm traj}+r_i(w)\\
 &\le\bar r_\rho[(1+M)\mathfrak H_{i-1}+W_i]\\
 &=\bar r_\rho D_i^{\rm der}.
\end{align*}
Let $\widehat H_i=H_i(\widehat a_i,\widehat y_i)$ and $\widehat J_i=J_i(\widehat a_i,\widehat y_i)$.  The Hessian-Lipschitz bound of \prettyref{A:C2} gives
\begin{align*}
 \|\bar H_i-\widehat H_i\|&\le L_H\eta_i^{\rm der},\\
 \|\bar J_i-\widehat J_i\|&\le L_H\eta_i^{\rm der}.
\end{align*}
Using $\bar H_i^{-1}-\widehat H_i^{-1}=\bar H_i^{-1}(\widehat H_i-\bar H_i)\widehat H_i^{-1}$,
\begin{align*}
 \bar Q_i-\widehat Q_i
 &=(\bar J_i-\widehat J_i)^\top\bar H_i^{-1}
 +\widehat J_i^\top(\bar H_i^{-1}-\widehat H_i^{-1}),\\
 \|\bar Q_i-\widehat Q_i\|
 &\le\frac{L_H\eta_i^{\rm der}}\mu
 +L_{ay}\frac{L_H\eta_i^{\rm der}}{\mu^2}\\
 &=\frac{(1+M)L_H\eta_i^{\rm der}}\mu\\
 &\le\bar r_\rho\vartheta_i^{\rm der}.
\end{align*}

We next compare the multipliers $\bar p_i$ of \eqref{eq:bars} at the iterate $w$ with the exact adjoints $\widehat p_i$ of \eqref{eq:adj} at $\widehat w(x)$.  Put $\bar p_{K+1}=\widehat p_{K+1}=0$, and read the products $\bar Q_{K+1}^{y'}\bar p_{K+1}$ and $\widehat Q_{K+1}^{y'}\widehat p_{K+1}$ as zero.  Rearranging the state-block identities in \eqref{eq:multid} shows that the $\bar p_i$ satisfy the backward recursion
\begin{equation*}
 \bar p_i=\nabla_{y_i}F(w)-\bar Q_{i+1}^{y'}\bar p_{i+1}-\nabla_{y_i}\Phi_\rho(w)
 \qquad(1\le i\le K).
\end{equation*}
It has the same form as \eqref{eq:adj}, except that it is evaluated at $w$ instead of $\widehat w(x)$, uses the averaged matrices $\bar Q_{i+1}$ instead of $\widehat Q_{i+1}$, and carries the extra term $-\nabla_{y_i}\Phi_\rho(w)$, whose norm is at most $E$.
For $\eta_i^{\rm adj}:=\|\widehat p_i-\bar p_i\|$, subtraction from \eqref{eq:adj} gives the display below.  In its second line, $\|\widehat Q_{i+1}^{y'}\|\le\|\widehat Q_{i+1}\|\le\|\widehat J_{i+1}\|\,\|\widehat H_{i+1}^{-1}\|\le L_{ay}/\mu=M$, because a sub-block has norm at most that of the full matrix and $(\widehat a_{i+1},\widehat y_{i+1})$ lies in the rectangle, hence in $\mathcal U$; the same sub-block bound gives $\|(\widehat Q_{i+1}-\bar Q_{i+1})^{y'}\|\le\|\widehat Q_{i+1}-\bar Q_{i+1}\|$.
\begin{align*}
 \eta_i^{\rm adj}
 &=\bigl\|\nabla_{y_i}F(\widehat w)-\nabla_{y_i}F(w)
 -\widehat Q_{i+1}^{y'}(\widehat p_{i+1}-\bar p_{i+1})
 -(\widehat Q_{i+1}-\bar Q_{i+1})^{y'}\bar p_{i+1}
 +\nabla_{y_i}\Phi_\rho(w)\bigr\|\\
 &\le L_F\|\widehat w-w\|+M\eta_{i+1}^{\rm adj}
 +\|\widehat Q_{i+1}-\bar Q_{i+1}\|\,\|\bar p_{i+1}\|+E\\
 &\le L_F\bar r_\rho\mathfrak H_2^{\rm all}+E
 +M\eta_{i+1}^{\rm adj}
 +\bar r_\rho(E+G_F)\vartheta_{i+1}^{\rm der}W_{i+1}.
\end{align*}
The successor terms are absent at $i=K$.  Backward substitution yields
\begin{equation*}
 \eta_i^{\rm adj}
 \le W_i\bigl(L_F\bar r_\rho\mathfrak H_2^{\rm all}+E\bigr)
 +\bar r_\rho(E+G_F)\mathfrak B_i.
\end{equation*}

Subtract the first identity in \eqref{eq:multid} from \eqref{eq:gradphi}, and use $\|\widehat Q_i^x\|\le\|\widehat Q_i\|\le M$ as above.  Then
\begin{align*}
 \|\nabla\varphi(x)-\nabla_x\Phi_\rho(w)\|
 &\le L_F\|\widehat w-w\|+
 \sum_i\|\widehat Q_i\|\eta_i^{\rm adj}
 +\sum_i\|\widehat Q_i-\bar Q_i\|\,\|\bar p_i\|\\
 &\le L_F\bar r_\rho\mathfrak H_2^{\rm all}
 +\sum_iM\left[W_i(L_F\bar r_\rho\mathfrak H_2^{\rm all}+E)
 +\bar r_\rho(E+G_F)\mathfrak B_i\right]
 +\bar r_\rho(E+G_F)\sum_i\vartheta_i^{\rm der}W_i\\
 &=c_{\rm adj}E+c_\varphi L_F\bar r_\rho\mathfrak H_2^{\rm all}
 +\bar r_\rho(E+G_F)\sum_i
 (M\mathfrak B_i+\vartheta_i^{\rm der}W_i)\\
 &=c_{\rm adj}E+
 \frac{c_{\rm b1}(E+G_F)+c_{\rm b2}(E+G_F)^2}{\rho}.
\end{align*}
This is \eqref{eq:convgrad}. For the projected conclusion, \eqref{icml:Gprod} and projection nonexpansiveness give
\begin{align*}
 \alpha\Gphi_\alpha(x)
 &=\|x-\PiX_\cX(x-\alpha\nabla\varphi(x))\|\\
 &\le\|x-\PiX_\cX(x-\alpha\nabla_x\Phi_\rho(w))\|
 +\|\PiX_\cX(x-\alpha\nabla_x\Phi_\rho(w))
 -\PiX_\cX(x-\alpha\nabla\varphi(x))\|\\
 &\le\alpha E+\alpha\|\nabla\varphi(x)-\nabla_x\Phi_\rho(w)\|.
\end{align*}
Divide by $\alpha$ and use \eqref{eq:convgrad}.
\end{proof}

\begin{remark}
If $x$ is interior to $\cX$, the right side also bounds $\|\nabla\varphi(x)\|$ for all sufficiently small $\alpha>0$, because projection is then the identity.  In particular, when $E=0$,
\begin{equation}\label{eq:bias}
 \Gphi_\alpha(x)\le\frac{c_{\rm b1}G_F+c_{\rm b2}G_F^2}{\rho}.
\end{equation}
\end{remark}

\begin{remark}[Conversion geometry]
The rectangle \eqref{eq:convrect} contains more than the endpoints.  It contains $\{a_i(w)\}\times\operatorname{conv}\{y_i,s_i(a_i(w))\}$, $\operatorname{conv}\{a_i(w),\widehat a_i\}\times\{s_i(a_i(w))\}$, $\{\widehat a_i\}\times\operatorname{conv}\{s_i(a_i(w)),\widehat y_i(x)\}$, and every segment joining a point of the first set to $(\widehat a_i,\widehat y_i(x))$.
\end{remark}

\begin{proof}[Proof of \prettyref{thm:main}]
The schedule is nondecreasing and satisfies \eqref{eq:Hrho} by \eqref{eq:ratio}.  Therefore \prettyref{thm:lyap} applies.  Since $U_t\ge0$ and $\sum_tU_t\le B_0$ by \prettyref{thm:lyap}, we have $B_0\ge0$.  The half-window $\cT_T=\{\lfloor T/2\rfloor,\ldots,T-1\}$ satisfies $|\cT_T|=T-\lfloor T/2\rfloor=\lceil T/2\rceil\ge T/2$. Thus some $t'\in\cT_T$ has $U_{t'}\le2B_0/T$. By \eqref{eq:betamax}, $\beta\le\beta_{\max}\le(1-\widetilde q)/4<1$.  For $t<T$, this gives $1+\beta t\le1+\beta(T-1)\le T<2T$ and $\rho_t\le\rho_0(2T)^{1/3}$. For $t\in\cT_T$, we have $t\ge\lfloor T/2\rfloor\ge(T-1)/2$, and $\beta<1\le2$ gives
\begin{align*}
 1+\beta t
 &\ge1+\frac{\beta(T-1)}2
 =\frac{\beta T}{2}+1-\frac\beta2
 \ge\frac{\beta T}{2},\\
 \rho_t^{-1}
 &\le\frac{2^{1/3}}{\rho_0\beta^{1/3}}T^{-1/3}
 =C_\rho T^{-1/3}.
\end{align*}
\prettyref{lem:cert} and the choice of $t^\star$ give
\begin{align*}
 \Ecal_{t^\star}^2
 &\le C_{t^\star}^2\\
 &\le C_{t'}^2\\
 &\le\frac{16\chi^2}{c_\alpha}\rho_{t'}U_{t'}\\
 &\le\frac{16\chi^2}{c_\alpha}\rho_0(2T)^{1/3}\frac{2B_0}{T}\\
 &=\chi^2\frac{32\,2^{1/3}\rho_0B_0}{c_\alpha}T^{-2/3}\\
 &=\chi^2C_E(B_0)^2T^{-2/3}.
\end{align*}
Hence $\Ecal_{t^\star}\le\chi C_E(B_0)T^{-1/3}$ and also $\Ecal_{t^\star}\le\chi C_E(B_0)$. The global bounds make \eqref{eq:convrect} automatic.  Apply \eqref{eq:convproj} of \prettyref{lem:conv} at the output.  Its right side is nondecreasing in $E$, so upper bounds on $E=\Ecal_{t^\star}$ may be substituted.  In the linear term, use $\Ecal_{t^\star}\le\chi C_E(B_0)T^{-1/3}$; inside the bracket, use $\Ecal_{t^\star}\le\chi C_E(B_0)$ and $\rho_{t^\star}^{-1}\le C_\rho T^{-1/3}$, which holds because $t^\star\in\cT_T$.  This gives
\begin{align*}
 \Gphi_{\alpha_{t^\star}}(x_{t^\star})
 &\le c_\varphi\chi C_E(B_0)T^{-1/3}
 +C_\rho T^{-1/3}
 \left[c_{\rm b1}(\chi C_E(B_0)+G_F)
 +c_{\rm b2}(\chi C_E(B_0)+G_F)^2\right]\\
 &=C^\star(B_0)T^{-1/3}.
\end{align*}
\prettyref{lem:mono} extends this bound to every $\bar\alpha\ge\alpha_{t^\star}$.  Since $\alpha_{t^\star}\le c_\alpha/\rho_0$, every $\bar\alpha\ge c_\alpha/\rho_0$ is included.
\end{proof}

\begin{remark}[A horizon-free certificate]\label{rem:runningmin}
Define
\[
 D_t:=c_\varphi C_t+
 \frac{c_{\rm b1}(C_t+G_F)+c_{\rm b2}(C_t+G_F)^2}{\rho_t}.
\]
\prettyref{lem:cert} and \ref{lem:conv} give $\Gphi_{\alpha_t}(x_t)\le D_t$.  The running minimum of $D_t$ therefore gives a computable stopping rule with $O(1)$ extra memory.
\end{remark}

\ifshowremarks
\paragraph{Limits of the analysis.}
\fi

\begin{remark}[The exponent supplied by this argument]\label{rem:envelope}
For any admissible nondecreasing schedule, the same proof gives an output in the last half with
\begin{equation}\label{eq:envelope}
 \Gphi_{\alpha_{t^\star}}(x_{t^\star})
 \lesssim\sqrt{\frac{\rho_TB_0}{T}}+\frac1{\rho_{\lfloor T/2\rfloor}}.
\end{equation}
For $\rho_t=\rho_0(1+\beta t)^{1/3}$, the two terms have the same $T$-order and give the $T^{-1/3}$ rate.  This is a limit of the displayed penalty argument, not a lower bound for the problem class.
\end{remark}

\begin{remark}[The smoothness term]\label{rem:smoothterm}
The last-state identity in \eqref{eq:gradPhi} contains $\rho\nabla_yf_K(a_K,y_K)$.  Strong convexity makes its monotonicity grow linearly with $\rho$, so a first-order outer step must have size $O(1/\rho)$.  This produces the first term in \eqref{eq:envelope}.
\end{remark}

\begin{remark}[The bias term is attained]\label{rem:biasex}
Take $K=1$, $f(x,y',y)=\tfrac12(y-x)^2$, and $\cX=[-1,1]$.  Let $F(x,y)=-x/2+y^2/2$ on a neighbourhood of $[-2,2]^2$, with a bounded-below smooth extension having bounded Lipschitz gradient.  Then $s(x,y')=x$ and $R(a,y)=\tfrac12(y-x)^2$.  For $\rho>1$, $y=1/2$ and $x=1/2+1/(2\rho)$.  It is interior and has $\cG_\alpha^\rho=0$, while $\nabla\varphi(x)=x-1/2=1/(2\rho)$.  Thus the $O(1/\rho)$ term in \eqref{eq:bias} is attained.
\end{remark}

\begin{remark}[Why the projected residual is necessary]\label{rem:rawgrad}
Take $K=1$, $\cX=[0,1]$, $f(x,y',y)=y^2/2$, and $F(x,y)=\arctan x+\pi/2$.  Then $s\equiv0$ and $|\varphi'(x)|=1/(1+x^2)\ge1/2$ on $\cX$, but the constrained minimizer $x=0$ has $\Gphi_\alpha(0)=0$.  The projected residual, not the raw gradient norm, is therefore the correct stationarity measure on a constrained control set.
\end{remark}

\section{Hard constraints with a constant band}\label{sec:ns}\label{sec:nsmodel}

This appendix proves \prettyref{thm:nsrecover}: under \prettyref{ass:nscb}, PSK started by \prettyref{alg:nswarm} and run on the surrogate family $(f_i^C)_{i=1}^K$ keeps the $T^{-1/3}$ rate of \prettyref{thm:main}, and every exact surrogate state is feasible and a $d$-banded Fritz--John point.  The proof reuses the results of \prettyref{sec:smooth} with $f_i=f_i^C$.

The surrogate family has no global smoothness constants, because the wall forces grow without bound under violation.  The key is therefore to keep the exact solutions, the iterates, the trackers, and the segments between them in compact local sets on which the constants of \prettyref{sec:smooth} are finite.  Besides the exact cores $\mathcal S_i$ and numerical tubes $\mathcal Y_i$ of \prettyref{ass:nscb}, we use the following local sets.  Define
\[
 R_{{\rm tube},K}:=\max_{y\in\mathcal Y_K}\|y\|,
 \qquad D_K:=\operatorname{diam}\mathcal Y_K.
\]

Let $\mathcal D_{\rm cb}$ be the compact convex box
\begin{equation}\label{eq:nscbbox}
 \mathcal D_{\rm cb}:=
 \cX\times\Ball_{R_{{\rm tube},K}+1}(0)
 \times\Ball_{R_{{\rm tube},K}+D_K+1}(0).
\end{equation}

Define the stagewise fibers and the trajectory tube
\[
 \mathcal B_i:=\{(x,p,y):x\in\cX,\ p\in\mathcal Y_{i-1},\
 \|y-s_i^C(x,p)\|\le r_{\rm in}\},
\]
\[
 \mathcal W_{\rm cb}:=\{w=(x,y_1,\ldots,y_K)\in\cW:
 (x,y_{i-1},y_i)\in\mathcal B_i\ \text{for every }i\}.
\]

Finally put
\[
 \mathcal W_{\rm box}:=\cX\times\mathcal Y_1\times\cdots\times\mathcal Y_K.
\]

These sets are related as follows.  By \eqref{eq:nscbnested}, $\mathcal S_i\subseteq\mathcal Y_i\subseteq\mathcal Y_K\subseteq\Ball_{R_{{\rm tube},K}}(0)$, so $\cX\times\mathcal Y_K\times\mathcal Y_K\subseteq\mathcal D_{\rm cb}$.  If $w\in\mathcal W_{\rm cb}$, then $y_{i-1}\in\mathcal Y_{i-1}$ and $\|y_i-s_i^C(x,y_{i-1})\|\le r_{\rm in}$ for every $i$, so \prettyref{nscb:NI} gives $y_i\in\mathcal Y_i$; hence $\mathcal W_{\rm cb}\subseteq\mathcal W_{\rm box}$.  \prettyref{lem:nslayered} collects the containments used in the proofs.  The extra radius $D_K$ in the last factor of \eqref{eq:nscbbox} holds the warm-start segments of \prettyref{lem:nsboxwarm}.

For a stage $i$, a parameter $a=(x,p)$, and wall data $d,\Gamma>0$, we call the minimizer of $g_i(a,\cdot)+\sum_l\psi_{d,\Gamma}(-h_l(p,\cdot))$ the exact center at $a$; with the data of \eqref{eq:nscbdata}, it is $s_i^C(a)$.  If $N_h=0$, we set $\operatorname{sl}\equiv+\infty$ and treat every sum over $l$ as $0$ (the empty-family convention).

\subsection{Parameter Schedule}\label{sec:nsschedule}

\paragraph{Constants.} On $\mathcal D_{\rm cb}$ choose the \prettyref{sec:smooth} bounds $L_{yy},L_{ay},L_{aa},L_f,L_H$ uniformly over $1\le i\le K$; they are finite because $\mathcal D_{\rm cb}$ is compact and $g_i$, $h_l$, and $\psi_{d,\Gamma_{\rm cb}}$ are $C^2$ with Hessians that are Lipschitz on bounded sets (\prettyref{ns:G}, \prettyref{ns:H1}, and \prettyref{lem:nswall}).  Use the specialization
\begin{equation}\label{eq:nsinstantiate}
 \mathcal U:=\mathcal D_{\rm cb},
 \qquad
 f_i:=f_i^C,
 \qquad
 s_i:=s_i^C
 \quad(1\le i\le K).
\end{equation}
Every constant below is the \prettyref{sec:smooth} object with the same name, evaluated under \eqref{eq:nsinstantiate} with the values of $\gamma,c_\alpha,\beta,\rho_0$ chosen here; as in \prettyref{sec:smooth}, these constants may depend on $K$. Fix $m\ge1$, set $\gamma:=1/(2L_{yy})$, and take $\kappa_0$ from \eqref{eq:kappa0}. Form $q$ and $\widetilde q$, and set
\begin{equation}\label{eq:nscalpha}
 c_\alpha:=
 \min\left\{\frac1{2(L_F+K(L_f+L_\star))},
 \frac{1-\widetilde q}{8M\sqrt{K\kappa_0}}\right\}.
\end{equation}
Form $c_\Psi,\omega_1,\omega_2,\omega_3$ by \eqref{eq:rhoconsts} and \eqref{eq:omega}, and set $\beta:=\beta_{\max}$, the largest value allowed by \eqref{eq:betamax} under this specialization.

Put
\[
 F_{\rm hi}:=\max_{w\in\mathcal W_{\rm box}}F(w),
 \qquad B_{\rm cb}:=F_{\rm hi}-\Flo+2,
\]
where $B_{\rm cb}$ is the localization budget, and set
\begin{equation}\label{eq:nslocalization}
 r_{\rm cb}:=\frac{r_{\rm in}}{8(1+M)},
\end{equation}
and
\begin{equation}\label{eq:nsrho0}
 \rho_0:=\max\left\{1,2\omega_3,
 \frac{2B_{\rm cb}}{\mu r_{\rm cb}^2},
 \frac{B_{\rm cb}}{c_\Psi r_{\rm cb}^2},
 \frac{c_\alpha G_F}{r_{\rm cb}}\right\}.
\end{equation}
Since $\rho_0\ge1$, the choice \eqref{eq:nscalpha} satisfies \eqref{eq:amax}.  Finally set
\begin{equation}\label{eq:nswarmradius}
 r_{\rm warm}:=\min\left\{r_{\rm cb},
 (K\rho_0L_{yy})^{-1/2},
 (2c_\Psi\rho_0K)^{-1/2}\right\},
\end{equation}
\begin{equation}\label{eq:nswarmsteps}
 m_0:=\left\lceil
 \frac{\log\max\{1,D_K/r_{\rm warm}\}}{\log(1/q)}
 \right\rceil.
\end{equation}
Every quantity is fixed before the outer run and is finite for the fixed horizon. \prettyref{alg:nswarm} uses only surrogate gradients and makes no oracle call for an exact stage solution.  \prettyref{lem:nsboxwarm} proves containment of every warm-start segment before stage contraction is applied. Starting from the output of \prettyref{alg:nswarm}, run \prettyref{alg:psk} with $f_i=f_i^C$ ($1\le i\le K$), the step $\gamma$ and weight $\kappa_0$ above, and the schedule $\rho_t=\rho_0(1+\beta t)^{1/3}$, $\alpha_t=c_\alpha/\rho_t$ of \prettyref{sub:band}.  The fixed inner budget $m$ is independent of $t$, $T$, and the target accuracy. For the outer run write
\[
 r_{i,t}:=r_i(w_t)=\|y_{i,t}-s_i^C(a_i(w_t))\|,
 \qquad e_{i,t}^{(k)}:=\|b_{i,t}^{(k)}-s_i^C(a_i(w_t))\|,
 \qquad e_{i,t}:=e_{i,t}^{(0)},
 \qquad w_t(\theta):=w_t+\theta\Delta w_t.
\]

\subsection{What warm-started PSK achieves}\label{sec:nsstatements}

Under \eqref{eq:nsinstantiate}, $\Phi_\rho$ denotes the \prettyref{sec:smooth} object for the family $(f_i^C)_{i=1}^K$, and stage $i$ uses the gap $R_i$ associated with $f_i^C$.  The projected implicit residual is
\begin{equation}\label{eq:nsGphi}
 \cG_\alpha^{\varphi^C}(x)=\mathcal G^{\varphi^C}_{\alpha,\cX}(x)
 \quad\text{from \eqref{eq:G}}.
\end{equation}

For $x\in\cX$ and $1\le i\le K$, the wall multipliers (wall forces) of the exact chain are
\begin{equation}\label{icml:nsmult}
 \lambda_{i,l}(x):=-\psi'_{d,\Gamma_{\rm cb}}
 (-h_l(\widehat y_{i-1}^C(x),\widehat y_i^C(x))).
\end{equation}

\begin{definition}[Approximate KKT point]\label{def:nskkt}
Fix a stage $i$ and $\varepsilon_{\rm KKT}\geq0$.  A feasible $y$ is an $\varepsilon_{\rm KKT}$-KKT point at $(x,p)$ if some $\lambda\in\R_+^{N_h}$ satisfies
\[
 \nabla_yg_i+\sum_{l=1}^{N_h}\lambda_l\nabla_yh_l=0,
 \qquad \lambda_l|h_l(p,y)|\leq\varepsilon_{\rm KKT}\quad(1\leq l\leq N_h),
\]
with derivatives of $g_i$ and $h_l$ taken at $(x,p,y)$.
\end{definition}

\begin{formaltheorem}{thm:nsrecover}[PSK convergence ($N_h>0$)]
Let \prettyref{ass:nscb} hold.  Use the surrogate family $(f_i^C)_{i=1}^K$ and the constants $\gamma,\kappa_0,c_\alpha,\beta,\rho_0,m_0$ of \prettyref{sec:nsschedule}: the bounded-anchor warm start with $m_0$ steps, the PSK settings $\gamma$ and $\kappa_0$, a fixed inner budget, and the schedule $\rho_t=\rho_0(1+\beta t)^{1/3}$, $\alpha_t=c_\alpha/\rho_t$.  For every $T\geq1$, if $t^\star$ minimizes $C_t$ over $\lfloor T/2\rfloor\leq t<T$, then, with $C_{\rm cb}^\star$ the map $C^\star$ of \eqref{eq:Cs} evaluated with the constants of \prettyref{sec:nsschedule},
\begin{equation}\label{eq:nsmainrate}
 \cG_{\bar\alpha}^{\varphi^C}(x_{t^\star})
 \leq C_{\rm cb}^\star(B_{\rm cb})T^{-1/3}
 \qquad(\bar\alpha\geq\alpha_{t^\star}).
\end{equation}
With $\lambda_{i,l}$ from \eqref{icml:nsmult}, for every $x\in\cX$ and every $1\le i\le K$, the exact stage-$i$ surrogate state satisfies
\[
 \operatorname{sl}(\widehat y_{i-1}^C(x),\widehat y_i^C(x))
 \geq\tau_{\rm cb},
 \qquad
 \sum_l\lambda_{i,l}(x)\leq\Lambda_{\rm cb},
\]
and hence is feasible.  It is a $d$-banded Fritz--John point at $(x,\widehat y_{i-1}^C(x))$ with objective weight at least $(1+\Lambda_{\rm cb})^{-1}$ (\prettyref{def:nscert}) and a $4\Gamma_{\rm cb}/9$-KKT point at $(x,\widehat y_{i-1}^C(x))$ (\prettyref{def:nskkt}).
\end{formaltheorem}
All rate constants may depend on $K$, through the depth factors $M^{j-i}$ and $W_i$.  The iterate sequence does not use $T$ or a target accuracy; only the post-run choice of $t^\star$ uses $T$.

\subsection{Proofs}\label{sec:nsproof}

\paragraph{Idea of the proof.} The proof has two parts.  The first concerns the exact chain and runs by induction on the stage.  For a core state $p$, \prettyref{nscb:F} supplies $\bar y$.  Pairing the stationarity condition of $f_i^C$ with $\bar y-s_i^C(x,p)$ bounds $\sum_l\lambda_l(\sigma_*-u_l)$ by $\Upsilon_{\rm cb}$, where $u_l$ is the margin of constraint $l$ and $\lambda_l$ its wall force, so the total force is at most $\Lambda_{\rm cb}$.  The wall height $\Gamma_{\rm cb}$ in \eqref{eq:nscbdata} is chosen so that a margin below $\tau_{\rm cb}$ would force one multiplier above $\Lambda_{\rm cb}$.  Hence the exact state has slack at least $\tau_{\rm cb}$, and \prettyref{nscb:XT} places it in the next core.  A wall force is positive only for margins below $d$, and $\lambda_lu_l\le4\Gamma_{\rm cb}/9$, so the exact state is a $d$-banded Fritz--John point and a $4\Gamma_{\rm cb}/9$-KKT point.  The second part concerns the numerical run.  Its main step is an induction over all iterations $t\ge0$ on the invariant $\mathbb L_t$ (\prettyref{lem:nsinvariant}).  The warm start puts $w_0$ in $\mathcal W_{\rm cb}$ with $B_0\le B_{\rm cb}$, which gives $\mathbb L_0$.  Assume $\mathbb L_t$.  Its budget clause bounds the Lyapunov candidate $\Psi_t$ by $\Flo+B_{\rm cb}$, which, by the choice of $\rho_0$, gives $r_{i,t},e_{i,t}\le r_{\rm cb}$.  Then $w_t$, the trackers, and the outer segment $[w_t,w_{t+1}]$ stay in $\mathcal W_{\rm cb}$, and \prettyref{lem:nslayered} places every set in \eqref{eq:onestepcontain}, \eqref{eq:valuecontain}, \eqref{eq:driftcontain}, and \eqref{eq:convrect} in $\mathcal D_{\rm cb}$.  The one-step, descent, residual, and drift lemmas of \prettyref{sec:smooth} then give the Lyapunov inequality \eqref{eq:nslyap}, which yields $\mathbb L_{t+1}$.  Finally, the certificate, conversion, and half-window steps of the proof of \prettyref{thm:main} give the rate with $B_{\rm cb}$ in place of $B_0$.

\begin{lemma}[Strict-comparison multiplier cap]\label{lem:strictcomparison}
Fix a stage $i$, a parameter $(x,p)$, and $d,\Gamma>0$, and let $s$ be the exact center at $(x,p)$.  Put $u_l:=-h_l(p,s)$ and $\lambda_l:=-\psi'_{d,\Gamma}(u_l)$.  Suppose $\lambda_l\ge0$, $\lambda_l>0\Longrightarrow u_l<d$, and, for some $\sigma>d$ and $\Upsilon\ge0$,
\[
 \sum_l\lambda_l(\sigma-u_l)\le\Upsilon.
\]
Then
\[
 \sum_l\lambda_l\le\frac{\Upsilon}{\sigma-d},
 \qquad
 \lambda_lu_l\le\frac{\Upsilon d}{\sigma-d}
 \quad\text{for every }l.
\]
\end{lemma}

\begin{proof}
For every $l$ with $\lambda_l>0$, the hypothesis $\lambda_l>0\Rightarrow u_l<d$ and $\sigma>d$ give $\sigma-u_l>\sigma-d$; hence $\lambda_l(\sigma-u_l)\ge\lambda_l(\sigma-d)$ for every $l$ (both sides vanish when $\lambda_l=0$).  Summing over $l$ gives
\[
 (\sigma-d)\sum_l\lambda_l
 \le\sum_l\lambda_l(\sigma-u_l)\le\Upsilon,
\]
which proves the sum bound.  Moreover, if $\lambda_l>0$, then $u_l<d$ and
\[
 \lambda_lu_l\le d\lambda_l
 \le d\sum_j\lambda_j\le\frac{\Upsilon d}{\sigma-d};
\]
for $\lambda_l=0$ the same conclusion is immediate.
\end{proof}
\prettyref{lem:nscbforce} supplies the force inequality for the surrogate stages.

\begin{lemma}[Cubic wall]\label{lem:nswall}
Let $\lambda(u):=-\psi'_{d,\Gamma}(u)$.  Then $\psi_{d,\Gamma}\in C^2(\R)$ is convex and nonincreasing (so $\lambda\ge0$), $\psi_{d,\Gamma}''$ is $6\Gamma/d^3$-Lipschitz, and
\begin{align}
 \max_{u\ge0}u\lambda(u)&=\frac{4\Gamma}{9},\notag\\
 \lambda(u)&=\frac{3\Gamma}{d}+\frac{6\Gamma(-u)}{d^2}\quad(u<0).
 \label{eq:nswalldata-negative}
\end{align}
If $h$ is $C^2$ and $\psi_h:=\psi_{d,\Gamma}(-h)$, then
\[
 \nabla\psi_h=\lambda(-h)\nabla h,
 \qquad
 \nabla^2\psi_h=\psi_{d,\Gamma}''(-h)\nabla h\nabla h^{\!\top}
 +\lambda(-h)\nabla^2h.
\]
If $h(p,\cdot)$ is convex, then $\psi_h(p,\cdot)$ is convex.
\end{lemma}

\begin{proof}
On $0\le u\le d$, put $v=u/d$.  Direct differentiation of \eqref{eq:nswall} gives
\[
 \lambda(u)=\frac{3\Gamma}{d}(1-v)^2,
 \qquad u\lambda(u)=3\Gamma v(1-v)^2,
 \qquad \frac{\mathrm d}{\mathrm d v}[v(1-v)^2]=(1-v)(1-3v).
\]
Thus the maximum on this branch occurs at $v=1/3$ and equals $4\Gamma/9$; it is zero on the flat branch.  On $u<0$, differentiation gives \eqref{eq:nswalldata-negative}, and then $u\lambda(u)<0$.  The function values and the first two derivatives of adjacent branches agree at $u=0$ and at $u=d$; in particular $\psi_{d,\Gamma}''(0)=6\Gamma/d^2$ and $\psi_{d,\Gamma}''(d)=0$.  More explicitly,
\[
 \psi_{d,\Gamma}''(u)=\begin{cases}
 6\Gamma/d^2,&u<0,\\
 6\Gamma(1-u/d)/d^2,&0\le u\le d,\\
 0,&u>d,
 \end{cases}
\]
so $\psi_{d,\Gamma}''$ is continuous, nonnegative, and $6\Gamma/d^3$-Lipschitz.  Hence $\psi_{d,\Gamma}$ is convex and, since $\psi_{d,\Gamma}'=-\lambda\le0$, nonincreasing.  The chain and product rules give the two derivative identities.  If $h(p,\cdot)$ is convex, then both terms of the $y$-Hessian are positive semidefinite, because $\psi_{d,\Gamma}''\ge0$ and $\lambda\ge0$; this proves convexity of $\psi_h(p,\cdot)$.
\end{proof}

\begin{lemma}[Force from a finite-horizon comparison point]\label{lem:nscbforce}
Let \prettyref{ns:G}, \prettyref{ns:H1}, and \prettyref{ns:H2} hold, let $N_h>0$, fix a stage $i$ and $(x,p)\in\cX\times\R^{n_y}$, and put
\[
 s=s_i^C(x,p),\qquad u_l=-h_l(p,s),\qquad \lambda_l=-\psi'_{d,\Gamma_{\rm cb}}(u_l).
\]
Suppose there is $\bar y$ such that
\[
 \operatorname{sl}(p,\bar y)\ge\sigma_*,\qquad
 \langle\nabla_yg_i(x,p,\bar y),\bar y-s\rangle-\mu\|\bar y-s\|^2\le\Upsilon_{\rm cb}.
\]
Then
\[
 \sum_l\lambda_l(\sigma_*-u_l)\le\Upsilon_{\rm cb},
 \qquad \sum_l\lambda_l\le\Lambda_{\rm cb},
 \qquad \operatorname{sl}(p,s)\ge\tau_{\rm cb}.
\]
\end{lemma}

\begin{proof}
By \prettyref{ns:G}, \prettyref{ns:H1}, \prettyref{ns:H2}, and \prettyref{lem:nswall}, $f_i^C(x,p,\cdot)$ is globally finite, $\mu$-strongly convex, and coercive.  Thus $s$ exists uniquely and satisfies exact stationarity.  Strong convexity of $g_i(x,p,\cdot)$ gives
\begin{equation}\label{eq:nscbstrong}
 \langle\nabla_yg_i(x,p,s),\bar y-s\rangle
 \le\langle\nabla_yg_i(x,p,\bar y),\bar y-s\rangle-\mu\|\bar y-s\|^2.
\end{equation}
For each $l$, convexity of $h_l(p,\cdot)$ (\prettyref{ns:H2}) and $h_l(p,\bar y)\le-\sigma_*$ give
\begin{equation}\label{eq:nscbconvex}
 \langle\nabla_yh_l(p,s),\bar y-s\rangle
 \le h_l(p,\bar y)-h_l(p,s)\le-\sigma_*+u_l.
\end{equation}
By \prettyref{ns:H1} and the gradient identity of \prettyref{lem:nswall}, exact stationarity of \eqref{eq:nscbsurrogate} is
\[
 0=\nabla_yg_i(x,p,s)+\sum_l\lambda_l\nabla_yh_l(p,s).
\]
Pair it with $\bar y-s$, use \eqref{eq:nscbstrong} and \eqref{eq:nscbconvex} (the latter multiplied by $\lambda_l\ge0$, which holds by \prettyref{lem:nswall}), and then use the scalar force bound in the displayed premise.  This yields
\[
 \sum_l\lambda_l(\sigma_*-u_l)
 \le\langle\nabla_yg_i(x,p,\bar y),\bar y-s\rangle-\mu\|\bar y-s\|^2\le\Upsilon_{\rm cb}.
\]
By \eqref{eq:nswall}, $\psi_{d,\Gamma_{\rm cb}}$ vanishes on $u\ge d$ (see the display below), so a positive multiplier has $u_l<d$.  Since $0<\tau_{\rm cb}<\sigma_*$ and $d=(\tau_{\rm cb}+\sigma_*)/2$, we have $d<\sigma_*$, and \prettyref{lem:strictcomparison}, applied with $\sigma=\sigma_*$ and $\Upsilon=\Upsilon_{\rm cb}$, therefore gives $\sum_l\lambda_l\le\Upsilon_{\rm cb}/(\sigma_*-d)=\Lambda_{\rm cb}$.  Finally, direct differentiation gives
\[
 -\psi'_{d,\Gamma_{\rm cb}}(u)=\begin{cases}
 0,&u\ge d,\\
 \dfrac{3\Gamma_{\rm cb}}{d}(1-u/d)^2,&0\le u<d,\\
 \dfrac{3\Gamma_{\rm cb}}{d}+\dfrac{6\Gamma_{\rm cb}(-u)}{d^2},&u<0.
 \end{cases}
\]
Suppose some $u_l<b_{\rm cb}d=\tau_{\rm cb}$, and note $0<b_{\rm cb}=\tau_{\rm cb}/d<1$.  If $0\le u_l<\tau_{\rm cb}$, the middle branch applies and $1-u_l/d>1-b_{\rm cb}>0$.  If $u_l<0$, the last branch gives $\lambda_l>3\Gamma_{\rm cb}/d>(3\Gamma_{\rm cb}/d)(1-b_{\rm cb})^2$.  In both cases, \eqref{eq:nscbdata} gives
\[
 \lambda_l>\frac{3\Gamma_{\rm cb}}{d}(1-b_{\rm cb})^2
 =\frac{\Upsilon_{\rm cb}}{\sigma_*-d}=\Lambda_{\rm cb},
\]
contradicting $\lambda_l\le\sum_j\lambda_j\le\Lambda_{\rm cb}$.  Thus every $u_l\ge b_{\rm cb}d=\tau_{\rm cb}$.
\end{proof}

\begin{lemma}[Constant-band wall and finite-horizon exact cores]\label{lem:nscbcore}
Let \prettyref{ass:nscb} hold and use \eqref{eq:nscbdata}. For every $x\in\cX$ and $1\le i\le K$, with $\lambda_{i,l}$ from \eqref{icml:nsmult},
\[
 \widehat y_i^C(x)\in\mathcal S_i,
 \qquad
 \operatorname{sl}(\widehat y_{i-1}^C(x),\widehat y_i^C(x))
 \ge\tau_{\rm cb},
 \qquad
 \sum_l\lambda_{i,l}(x)\le\Lambda_{\rm cb}.
\]
\end{lemma}

\begin{proof}
This paragraph repeats the existence argument of \prettyref{lem:nscbforce} because that lemma assumes $N_h>0$, while the present lemma also covers $N_h=0$.  For fixed $(x,p)$, the wall composition is convex in $y$: $-h_l(p,\cdot)$ is concave by \prettyref{ns:H2}, while $\psi_{d,\Gamma_{\rm cb}}$ is convex and nonincreasing by \prettyref{lem:nswall}.  Since $g_i(x,p,\cdot)$ is $\mu$-strongly convex by \prettyref{ns:G}, every $f_i^C(x,p,\cdot)$ is $\mu$-strongly convex.  Taking a first-order lower bound at any fixed point shows that it dominates an affine function plus $\mu\|y\|^2/2$, and hence is coercive.  Since it is $C^2$ by \prettyref{ns:G}, \prettyref{ns:H1}, and \prettyref{lem:nswall}, hence continuous and globally finite, it has a unique minimizer $s_i^C(x,p)$ at stage $i$.

We prove core membership and the displayed estimates by induction on $i$.  \prettyref{nscb:0} gives the base $\widehat y_0^C(x)=y_0\in\mathcal S_0$.  Assume $\widehat y_{i-1}^C(x)\in\mathcal S_{i-1}$.  When $N_h>0$, \prettyref{nscb:F} supplies the premise of \prettyref{lem:nscbforce} at $(x,\widehat y_{i-1}^C(x))$, which gives both the slack bound $\operatorname{sl}(\widehat y_{i-1}^C(x),\widehat y_i^C(x))\ge\tau_{\rm cb}$ and $\sum_l\lambda_{i,l}(x)\le\Lambda_{\rm cb}$.  When $N_h=0$, the empty-family convention gives infinite slack and the multiplier sum is $0=\Lambda_{\rm cb}$.  In either case \prettyref{nscb:XT} places $\widehat y_i^C(x)=s_i^C(x,\widehat y_{i-1}^C(x))$ directly in $\mathcal S_i$, closing the induction.  In particular, \prettyref{nscb:NI} is not used to prove exact-core membership.
\end{proof}

\begin{lemma}[Generic cubic-wall certificate]\label{lem:cubiccertificate}
Fix a stage $i$ and $(x,p)$, let $d,\Gamma>0$, and let $y$ be a feasible exact center for the cubic-wall surrogate.  Put
\[
 u_l:=-h_l(p,y),\qquad \lambda_l:=-\psi'_{d,\Gamma}(u_l),
\]
and suppose exact stationarity holds, $\lambda_l>0$ implies $0\le u_l<d$, and $\sum_l\lambda_l\le\bar\Lambda$.  Then $y$ is a $d$-banded Fritz--John point at $(x,p)$ with objective weight at least $(1+\bar\Lambda)^{-1}$, and it is a $\max_l\lambda_lu_l$-KKT point there (\prettyref{def:nskkt}), with the maximum interpreted as zero when $N_h=0$.  For the cubic wall,
\[
 \lambda_lu_l\le\frac{4\Gamma}{9}\qquad(1\le l\le N_h).
\]
\end{lemma}

\begin{proof}
By the gradient identity of \prettyref{lem:nswall}, exact stationarity reads $\nabla_yg_i(x,p,y)+\sum_l\lambda_l\nabla_yh_l(p,y)=0$.  Normalize the nonnegative vector $(1,\lambda_1,\ldots,\lambda_{N_h})$ by $1+\sum_l\lambda_l$, which gives $\vartheta:=(1,\lambda_1,\ldots,\lambda_{N_h})/(1+\sum_l\lambda_l)$.  Dividing the stationarity equation by the same factor preserves it, and the objective weight is
\[
 \vartheta_0=\frac{1}{1+\sum_l\lambda_l}
 \ge\frac{1}{1+\bar\Lambda}>0.
\]
If $\lambda_l>0$, the support of the wall derivative gives $u_l<d$, whereas feasibility gives $u_l\ge0$; hence $\vartheta_l\max\{u_l-d,0\}=0$.  If $\lambda_l=0$, then $\vartheta_l=0$ and this condition is trivial.  These are exactly the banded Fritz--John conditions in \prettyref{def:nscert}.  Without normalization, exact stationarity and feasibility are the KKT conditions of \prettyref{def:nskkt}, and their coordinatewise complementarity error is $\lambda_l|h_l(p,y)|=\lambda_lu_l$, because $h_l(p,y)=-u_l\le0$.  Finally, since $u_l\ge0$, \prettyref{lem:nswall} gives $\lambda_lu_l\le\max_{u\ge0}u[-\psi'_{d,\Gamma}(u)]=4\Gamma/9$.  The empty-family convention covers $N_h=0$.
\end{proof}

\begin{lemma}[Exact-state certificates]\label{lem:nsexact}
Under \prettyref{ass:nscb}, for every $x\in\cX$ and $1\le i\le K$, the exact state $\widehat y_i^C(x)$ satisfies
\[
 \operatorname{sl}(\widehat y_{i-1}^C(x),\widehat y_i^C(x))\ge\tau_{\rm cb},
 \qquad \sum_l\lambda_{i,l}(x)\le\Lambda_{\rm cb},
\]
and hence is feasible.  It is a $d$-banded Fritz--John point at $(x,\widehat y_{i-1}^C(x))$ with objective weight at least $(1+\Lambda_{\rm cb})^{-1}$ (\prettyref{def:nscert}), and is a $4\Gamma_{\rm cb}/9$-KKT point there (\prettyref{def:nskkt}).
\end{lemma}

\begin{proof}
Fix $K,x,i$, and write
\[
 p:=\widehat y_{i-1}^C(x),\qquad y:=\widehat y_i^C(x),\qquad
 u_l:=-h_l(p,y),\qquad \lambda_l:=\lambda_{i,l}(x).
\]
\prettyref{lem:nscbcore} gives the displayed slack bound and $\sum_l\lambda_l\le\Lambda_{\rm cb}$; feasibility follows because $\operatorname{sl}(p,y)\ge\tau_{\rm cb}>0$.  Exact stationarity, in the form given by the gradient identity of \prettyref{lem:nswall}, holds because $y=s_i^C(x,p)$.  By \eqref{eq:nswall}, $\psi'_{d,\Gamma_{\rm cb}}\equiv0$ on $u\ge d$, so $\lambda_l>0\Rightarrow u_l<d$, and feasibility gives $u_l\ge0$.  \prettyref{lem:cubiccertificate}, with $\bar\Lambda=\Lambda_{\rm cb}$ and $\Gamma=\Gamma_{\rm cb}$, now gives the stated Fritz--John objective weight.  It also makes $y$ a $\max_l\lambda_lu_l$-KKT point with $\max_l\lambda_lu_l\le4\Gamma_{\rm cb}/9$, including the empty-family case.  Since an $\varepsilon_{\rm KKT}$-KKT point in \prettyref{def:nskkt} is also an $\varepsilon'$-KKT point for every $\varepsilon'\ge\varepsilon_{\rm KKT}$, $y$ is a $4\Gamma_{\rm cb}/9$-KKT point.
\end{proof}

\begin{lemma}[Fixed-parameter warm start]\label{lem:stagewarm}
Fix a function $f$ and a parameter $a$, suppose the local hypotheses of \prettyref{lem:gd} hold for $f$ on the convex set $\mathcal U$, and let $s^\star:=\argmin_yf(a,y)$.  Suppose a convex set $\mathcal Y$, an inward radius $r_{\mathrm{in}}>0$, and a radius $D<\infty$ satisfy
\[
 \Ball_{r_{\mathrm{in}}}(s^\star)\subseteq\mathcal Y,
 \qquad \{a\}\times\Ball_D(s^\star)\subseteq \mathcal U.
\]
Let $c^{(0)}$ satisfy $\|c^{(0)}-s^\star\|\le D$, choose $0<\gamma\le L_{yy}^{-1}$, put $q=(1-\gamma\mu)^{1/2}$, so that $q<1$, and, for any integer $m_0\ge0$, set
\[
 c^{(k+1)}:=c^{(k)}-\gamma\nabla_yf(a,c^{(k)})
 \qquad(0\le k<m_0).
\]
Then, for every $0\le k\le m_0$,
\[
 \|c^{(k)}-s^\star\|\le q^kD,
 \qquad
 \{a\}\times[s^\star,c^{(k)}]\subseteq \mathcal U.
\]
If $q^{m_0}D\le r_{\mathrm{in}}$, then $c^{(m_0)}\in\mathcal Y$.
\end{lemma}

\begin{proof}
At $k=0$, the distance premise and convexity of the ball give $[s^\star,c^{(0)}]\subseteq\Ball_D(s^\star)$, so the lifted segment $\{a\}\times[s^\star,c^{(0)}]$ is contained in $\mathcal U$. Suppose the two conclusions hold at $k<m_0$.  The segment conclusion allows the local part of \prettyref{lem:gd}, which gives, since $q<1$,
\[
 \|c^{(k+1)}-s^\star\|
 \le q\|c^{(k)}-s^\star\|\le q^{k+1}D\le D.
\]
By convexity, $[s^\star,c^{(k+1)}]\subseteq\Ball_D(s^\star)$, and the assumed ball inclusion places its lifted segment in $\mathcal U$.  Induction proves both assertions for every $k\le m_0$.  If $q^{m_0}D\le r_{\rm in}$, the endpoint lies in $\Ball_{r_{\rm in}}(s^\star)\subseteq\mathcal Y$.  Thus the complete gradient path is valid and contracts by the factor of \prettyref{lem:gd}.
\end{proof}

For \prettyref{alg:nswarm}, put $s_{i,0}:=s_i^C(x_0,y_{i-1,0})$ and define the warm-start condition
\[
 \mathbb S_i^{(k)}:\quad
 \left\{\begin{aligned}
 &y_{i-1,0}\in\mathcal Y_{i-1},\\
 &\|y_{i,0}^{(k)}-s_{i,0}\|\le q^kD_K,\\
 &\{(x_0,y_{i-1,0})\}\times[s_{i,0},y_{i,0}^{(k)}]\subseteq\mathcal D_{\rm cb}.
 \end{aligned}\right.
\]

\begin{lemma}[Bounded-anchor warm start]\label{lem:nsboxwarm}
Under \prettyref{ass:nscb} and the data of \prettyref{sec:nsschedule}, every condition $\mathbb S_i^{(k)}$ with $1\le i\le K$ and $0\le k\le m_0$ holds.  Every warm-start segment lies in $\mathcal D_{\rm cb}$ and
\[
 y_{i,0}=b_{i,0}=y_{i,0}^{(m_0)}\in\mathcal Y_i,
 \qquad \|y_{i,0}-s_i^C(x_0,y_{i-1,0})\|\le r_{\rm warm}.
\]
Thus $w_0\in\mathcal W_{\rm cb}$, and the initial Lyapunov budget satisfies $B_0\le B_{\rm cb}$.
\end{lemma}

\begin{proof}
We induct on the stage.  For $i=1$, \prettyref{nscb:0} and \eqref{eq:nscbnested} give $y_{0,0}=y_0\in\mathcal Y_0$.  Assume $y_{i-1,0}\in\mathcal Y_{i-1}$.  \prettyref{nscb:NI} gives $\Ball_{r_{\rm in}}(s_{i,0})\subseteq\mathcal Y_i$.  Nestedness puts the common anchor $y_{i,0}^{(0)}=y_0$ in $\mathcal Y_K$, while the ball inclusion puts $s_{i,0}\in\mathcal Y_i\subseteq\mathcal Y_K$.  Hence $\|y_{i,0}^{(0)}-s_{i,0}\|\le D_K$.  Moreover,
\[
 \{(x_0,y_{i-1,0})\}\times\Ball_{D_K}(s_{i,0})
 \subseteq\mathcal D_{\rm cb}
\]
by \eqref{eq:nscbbox}, because $\|y_{i-1,0}\|\le R_{{\rm tube},K}$ and every $y\in\Ball_{D_K}(s_{i,0})$ has $\|y\|\le\|s_{i,0}\|+D_K\le R_{{\rm tube},K}+D_K$, as $y_{i-1,0},s_{i,0}\in\mathcal Y_K$.  Apply \prettyref{lem:stagewarm} with $a=(x_0,y_{i-1,0})$, $f=f_i^C$, $\mathcal U=\mathcal D_{\rm cb}$, $\mathcal Y=\mathcal Y_i$, $D=D_K$, and $c^{(k)}=y_{i,0}^{(k)}$, with the $\gamma$, $q$, and $m_0$ above, so that $s^\star=s_{i,0}$.  It gives every condition $\mathbb S_i^{(k)}$ and $\|y_{i,0}^{(m_0)}-s_{i,0}\|\le q^{m_0}D_K$. By \eqref{eq:nswarmsteps},
\[
 q^{m_0}D_K
 \le\frac{D_K}{\max\{1,D_K/r_{\rm warm}\}}
 =\min\{D_K,r_{\rm warm}\}
 \le r_{\rm warm}\le r_{\rm in},
\]
where the last step uses $r_{\rm warm}\le r_{\rm cb}\le r_{\rm in}$ from \eqref{eq:nswarmradius} and \eqref{eq:nslocalization}.
The endpoint conclusion of \prettyref{lem:stagewarm} gives $y_{i,0}\in\mathcal Y_i$, closing the stage induction.  The residual bound and these memberships prove $w_0\in\mathcal W_{\rm cb}$.

The final warm segment allows the local upper-gap bound in \prettyref{prop:identity}.  Since $b_{i,0}=y_{i,0}$, one has $e_{i,0}=r_{i,0}\le r_{\rm warm}$.  Hence
\[
 \rho_0\sum_iR_i(a_i(w_0),y_{i,0})
 \le\frac{\rho_0r_{\rm warm}^2}{2}KL_{yy}\le\frac12,
 \qquad
 c_\Psi\rho_0\Xi_0
 \le c_\Psi\rho_0Kr_{\rm warm}^2\le\frac12,
 \qquad
 \frac{2\omega_3}{\rho_0}\le1.
\]
Also $w_0\in\mathcal W_{\rm box}$, so $F(w_0)\le F_{\rm hi}$.  By \eqref{eq:PsiUB} and \eqref{eq:primal}, $\Psi_0=F(w_0)+\rho_0\sum_iR_i(a_i(w_0),y_{i,0})+c_\Psi\rho_0\Xi_0\le F_{\rm hi}+1$.  Using the definition \eqref{eq:PsiUB} of $B_0$ gives
\[
 B_0=\Psi_0-\Flo+\frac{2\omega_3}{\rho_0}
 \le F_{\rm hi}-\Flo+2=B_{\rm cb}.
\]
\end{proof}

\begin{lemma}[Stagewise fibers and the derivative box]\label{lem:nslayered}
Under \prettyref{ass:nscb} and the specialization \eqref{eq:nsinstantiate}, the following properties hold.
\begin{enumerate}[label=\textnormal{(\roman*)}]
\item\label{cblay:fiber} For every $1\le i\le K$, $\mathcal B_i\subseteq\mathcal D_{\rm cb}$; in fact $\Ball_{r_{\rm in}}(s_i^C(x,p))\subseteq\mathcal Y_i$ for $(x,p)\in\cX\times\mathcal Y_{i-1}$.
\item\label{cblay:lip} For every $1\le i\le K$, if $a,a'\in\cX\times\mathcal Y_{i-1}$, the local hypotheses of \prettyref{lem:lip} hold in $\mathcal U=\mathcal D_{\rm cb}$ and $\|s_i^C(a)-s_i^C(a')\|\le M\|a-a'\|$.
\item\label{cblay:chain} A trajectory $w\in\cW$ lies in $\mathcal W_{\rm cb}$ if and only if $r_i(w)\le r_{\rm in}$ for all $i$.  For every $x\in\cX$, the exact chain $(x,\widehat y_1^C(x),\ldots,\widehat y_K^C(x))$ lies in $\mathcal W_{\rm cb}$.
\item\label{cblay:use} If $u,v,w\in\mathcal W_{\rm cb}$, then every set in \eqref{eq:valuecontain}, the conversion rectangle \eqref{eq:convrect}, and the averaged-Hessian segments used by the localized \prettyref{sec:smooth} lemmas lie in $\mathcal D_{\rm cb}$.  Moreover, for every $1\le i\le K$, if $y,\widetilde y\in\Ball_{r_{\rm in}}(s_i^C(a_i(w)))$, then $\{a_i(w)\}\times\operatorname{conv}\{y,\widetilde y\}\subseteq\mathcal D_{\rm cb}$.
\end{enumerate}
\end{lemma}

\begin{proof}
\prettyref{nscb:NI} gives the ball inclusion in part~\prettyref{cblay:fiber}.  The predecessor $p$ lies in $\mathcal Y_{i-1}\subseteq\mathcal Y_K$, and the whole center ball lies in $\mathcal Y_i\subseteq\mathcal Y_K$, so the resulting triple belongs to $\mathcal D_{\rm cb}$. For part~\prettyref{cblay:lip}, both parameters lie in the convex set $\cX\times\mathcal Y_{i-1}$, while \prettyref{nscb:NI} puts both exact centers in the convex set $\mathcal Y_i$.  Thus the parameter hull and center hulls required by \prettyref{lem:lip} lie in $\cX\times\Ball_{R_{{\rm tube},K}}(0)\times\Ball_{R_{{\rm tube},K}}(0) \subseteq\mathcal D_{\rm cb}$, proving the claimed estimate.

The forward implication in part~\prettyref{cblay:chain} is the definition.  Conversely, start with $y_0\in\mathcal S_0\subseteq\mathcal Y_0$.  If $y_{i-1}\in\mathcal Y_{i-1}$ and $r_i(w)\le r_{\rm in}$, then the $i$th triple belongs to $\mathcal B_i$ by definition, and \prettyref{nscb:NI} puts $y_i$ in $\mathcal Y_i$.  Induction proves the reverse implication.  \prettyref{lem:nscbcore} gives exact-chain membership in $\mathcal S_i\subseteq\mathcal Y_i$, with zero residual at each stage.

For part~\prettyref{cblay:use}, every trajectory parameter at stage $i$ belongs to $\cX\times\mathcal Y_{i-1}$, and every trajectory state, exact center, or point in a displayed center ball at that stage belongs to $\mathcal Y_i$ by part~\prettyref{cblay:chain} and \prettyref{nscb:NI}.  Likewise, the exact-chain points satisfy $\widehat a_i\in\cX\times\mathcal Y_{i-1}$ and $\widehat y_i^C(x)\in\mathcal Y_i$ by part~\prettyref{cblay:chain}. All these sets lie in the common convex set $\mathcal Y_K$.  In particular, the two arbitrary points $y,\widetilde y$ in the stated center ball belong to $\mathcal Y_i$, so their convex hull does as well. By convexity and \eqref{eq:nscbnested}, all parameter and state hulls lie in $\cX\times\Ball_{R_{{\rm tube},K}}(0)\times\Ball_{R_{{\rm tube},K}}(0) \subseteq\mathcal D_{\rm cb}$.  The sets in \eqref{eq:valuecontain} and \eqref{eq:convrect} are products or subsets of these hulls.  The averaged-Hessian segments are the line segments along which the \prettyref{sec:smooth} lemmas integrate second derivatives; each joins two points of these hulls, so it lies in them.  This proves every asserted containment.  The second sentence of part~\prettyref{cblay:use} covers the tracker lines of \eqref{eq:onestepcontain} and \eqref{eq:driftcontain}, and part~\prettyref{cblay:lip} covers the parameter and solution lines of \eqref{eq:onestepcontain}, which form the hypothesis of \prettyref{lem:lip}.
\end{proof}

Define the invariant, whose first line is the budget clause and whose second line is the residual clause,
\[
\mathbb L_t:\quad
\left\{\begin{aligned}
 &\Psi_t+\sum_{n<t}U_n
 \le\Psi_0+\omega_3\sum_{n<t}\frac{\rho_{n+1}-\rho_n}{\rho_n^2},\\
 &r_{i,t}\le r_{\rm cb},\qquad e_{i,t}\le r_{\rm cb}\qquad(1\le i\le K).
\end{aligned}\right.
\]

\begin{lemma}[Localization and Lyapunov induction]\label{lem:nsinvariant}
Under \prettyref{ass:nscb} and the preceding choices, $\mathbb L_t$ holds for every $t\ge0$.  Thus
\begin{equation}\label{eq:nslocalbounds}
 e_{i,t}^{(k)}\le q^ke_{i,t}\le r_{\rm cb}
 \qquad(1\le i\le K,\ 0\le k\le m),
\end{equation}
$[w_t,w_{t+1}]\subseteq\mathcal W_{\rm cb}$, $\Psi_t\le\Flo+B_0$, $\sum_{n<t}U_n\le B_0\le B_{\rm cb}$, and
\begin{equation}\label{eq:nslyap}
 \Psi_{t+1}\le\Psi_t-U_t+
 \omega_3\frac{\rho_{t+1}-\rho_t}{\rho_t^2}.
\end{equation}
All containments required by \eqref{eq:onestepcontain}, \eqref{eq:driftcontain}, \eqref{eq:valuecontain}, and \eqref{eq:convrect} hold before the lemmas that need them are applied.
\end{lemma}

\begin{proof}
For $n\ge0$,
\[
 \frac{\rho_{n+1}-\rho_n}{\rho_n^2}
 =\frac{\rho_{n+1}}{\rho_n}
 \left(\frac1{\rho_n}-\frac1{\rho_{n+1}}\right)
 \le(1+\beta)\left(\frac1{\rho_n}-\frac1{\rho_{n+1}}\right),
\]
so, using $\beta=\beta_{\max}\le(1-\widetilde q)/4<1$ from \eqref{eq:betamax},
\begin{equation}\label{eq:nstelescope}
 \sum_{n<t}\frac{\rho_{n+1}-\rho_n}{\rho_n^2}
 \le\frac2{\rho_0}.
\end{equation}
\prettyref{lem:nsboxwarm} gives $B_0\le B_{\rm cb}$.  We induct on $t$.  The budget clause holds at $t=0$ because both sums are empty.  Assume that it holds at time $t$.  By \eqref{eq:nstelescope} and the definition of $B_0$,
\[
 \Psi_t+\sum_{n<t}U_n
 \le\Psi_0+2\omega_3/\rho_0=\Flo+B_0.
\]
Since $U_n\ge0$ and the global lower-gap bound in \prettyref{prop:identity} gives $\Psi_t\ge\Phi_{\rho_t}(w_t)\ge F(w_t)\ge\Flo$,
\[
 \Psi_t\le\Flo+B_0,
 \qquad \sum_{n<t}U_n\le B_0\le B_{\rm cb}.
\]
Global stage strong convexity and the tracker term in $\Psi_t$ then give
\begin{align*}
 \Flo+\frac{\rho_t\mu}{2}\sum_i r_{i,t}^2
 &\le\Phi_{\rho_t}(w_t)\le\Psi_t\le\Flo+B_{\rm cb},\\
 r_{i,t}&\le\sqrt{\frac{2B_{\rm cb}}{\rho_t\mu}}\le r_{\rm cb},\\
 c_\Psi\rho_t\Xi_t&=\Psi_t-\Phi_{\rho_t}(w_t)\le B_{\rm cb},\\
 e_{i,t}&\le\sqrt{\frac{B_{\rm cb}}{c_\Psi\rho_t}}
 \le r_{\rm cb}.
\end{align*}
The last inequalities use $\rho_t\ge\rho_0$ and \eqref{eq:nsrho0}.  Since $r_{\rm cb}\le r_{\rm in}$, \prettyref{lem:nslayered}\prettyref{cblay:chain} now gives $w_t\in\mathcal W_{\rm cb}$.

For each tracker, the case $k=0$ of \eqref{eq:nslocalbounds} was just proved.  If it holds at $k$, then $b_{i,t}^{(k)}\in\Ball_{r_{\rm in}}(s_i^C(a_i(w_t)))\subseteq\mathcal Y_i$. \prettyref{lem:nslayered}\prettyref{cblay:use} places the vertical segment required by \prettyref{lem:gd} in $\mathcal D_{\rm cb}$.  Only then do we apply that lemma, obtaining $e_{i,t}^{(k+1)}\le qe_{i,t}^{(k)}\le q^{k+1}e_{i,t}$. Thus \eqref{eq:nslocalbounds} holds and $\widehat e_{i,t+1}=e_{i,t}^{(m)}\le\widetilde qe_{i,t}\le r_{\rm cb}$.

We next bound the search direction.  The points $y_{i,t}$, $s_i^C(a_i(w_t))$, and $b_{i,t+1}$ lie in the same $r_{\rm in}$-ball.  \prettyref{lem:nslayered}\prettyref{cblay:use} places their vertical segments in $\mathcal D_{\rm cb}$; hence the chosen local derivative bound gives
\begin{align*}
 \|\nabla_af_i^C(a_i(w_t),y_{i,t})-\nabla_af_i^C(a_i(w_t),b_{i,t+1})\|
 &\le2L_fr_{\rm cb},\\
 \|\nabla_yf_i^C(a_i(w_t),y_{i,t})\|&\le L_fr_{\rm cb}.
\end{align*}
Group the terms of \eqref{icml:blocks}, with $f_i=f_i^C$, by stage.  Besides $\nabla F(w_t)$, stage $i$ contributes two terms.  The first is $\rho_t$ times the difference of $\nabla_af_i^C=(\nabla_xf_i^C,\nabla_{y'}f_i^C)$ in the blocks $(x,y_{i-1})$, with the $y_0$ part dropped for $i=1$.  The second is $\rho_t\nabla_yf_i^C(a_i(w_t),y_{i,t})$ in the block $y_i$.  Placing a vector in a subset of the blocks does not increase its norm, so the triangle inequality, $\|\nabla F\|\le G_F$, and the two bounds above give
\begin{align*}
 \|\hnabla_t\|
 &\le\|\nabla F(w_t)\|+\rho_t\sum_{i=1}^K\bigl(
 \|\nabla_af_i^C(a_i(w_t),y_{i,t})-\nabla_af_i^C(a_i(w_t),b_{i,t+1})\|
 +\|\nabla_yf_i^C(a_i(w_t),y_{i,t})\|\bigr)\\
 &\le G_F+\rho_tK(2L_fr_{\rm cb}+L_fr_{\rm cb})
 =G_F+3K\rho_t r_{\rm cb}L_f.
\end{align*}
Projection nonexpansiveness, $\alpha_t=c_\alpha/\rho_t$, \eqref{eq:nscalpha}, and \eqref{eq:nsrho0} yield
\[
 \|\Delta w_t\|
 \le\alpha_t\|\hnabla_t\|
 \le\frac{c_\alpha G_F}{\rho_0}
 +\frac{3KL_f}{2(L_F+K(L_f+L_\star))}r_{\rm cb}
 \le r_{\rm cb}+\frac32r_{\rm cb}<4r_{\rm cb}.
\]

We now establish outer-segment containment before any local one-step, descent, residual, drift, or conversion lemma is used.  We prove by induction on the stage $i$ that, for every $\theta\in[0,1]$, the $y_i$ component of $w_t(\theta)$ lies in $\mathcal Y_i$ within distance $r_{\rm in}$ of $s_i^C(a_i(w_t(\theta)))$.  At stage $i=1$, the predecessor is $y_0\in\mathcal Y_0$.  Suppose the predecessor component of $w_t(\theta')$ belongs to $\mathcal Y_{i-1}$ for every $\theta'\in[0,1]$, and fix $\theta\in[0,1]$.  By convexity of $\cX$, its control component lies in $\cX$, so both parameters compared below lie in $\cX\times\mathcal Y_{i-1}$. \prettyref{lem:nslayered}\prettyref{cblay:lip} allows use of \prettyref{lem:lip} and gives, with $u=w_t(\theta)$,
\[
 \|s_i^C(a_i(w_t))-s_i^C(a_i(u))\|
 \le M\|a_i(w_t)-a_i(u)\|
 \le M\|\Delta w_t\|.
\]
Therefore
\begin{align*}
 \|u_{y_i}-s_i^C(a_i(u))\|
 &\le r_{\rm cb}+(1+M)\|\Delta w_t\|\\
 &<(5+4M)r_{\rm cb}
 \le\frac{5+4M}{8(1+M)}r_{\rm in}\le r_{\rm in}.
\end{align*}
\prettyref{nscb:NI} puts $u_{y_i}$ in $\mathcal Y_i$ for every $\theta\in[0,1]$, closing the stage induction.  Hence $[w_t,w_{t+1}]\subseteq\mathcal W_{\rm cb}$.

All premises of the \prettyref{sec:smooth} lemmas are now available.  \prettyref{lem:model} is algebraic.  \prettyref{lem:onestep} requires \eqref{eq:onestepcontain} with $\mathcal U=\mathcal D_{\rm cb}$.  By \eqref{eq:nslocalbounds} and $r_{\rm cb}\le r_{\rm in}$, every $b_{i,t}^{(k)}$ with $0\le k\le m$ lies in $\Ball_{r_{\rm in}}(s_i^C(a_i(w_t)))$, and so does $y_{i,t}$ because $r_{i,t}\le r_{\rm cb}$.  Since $w_t\in\mathcal W_{\rm cb}$, \prettyref{lem:nslayered}\prettyref{cblay:use} with $y=s_i^C(a_i(w_t))$ and $\widetilde y=b_{i,t}^{(k)}$ or $\widetilde y=y_{i,t}$ gives the tracker line \eqref{eq:onestepcontain-tracker} and the second, unlabeled line of \eqref{eq:onestepcontain}.  The just-proved containment $[w_t,w_{t+1}]\subseteq\mathcal W_{\rm cb}$ gives $w_{t+1}\in\mathcal W_{\rm cb}$, so the definition of $\mathcal W_{\rm cb}$ puts $a_i(w_t)$ and $a_i(w_{t+1})$ in $\cX\times\mathcal Y_{i-1}$.  With $a=a_i(w_t)$ and $a'=a_i(w_{t+1})$, the hypothesis of \prettyref{lem:lip} is the union of \eqref{eq:onestepcontain-parameter} and \eqref{eq:onestepcontain-solution}, so \prettyref{lem:nslayered}\prettyref{cblay:lip} gives both.  Thus \prettyref{lem:onestep} gives
\[
 \Xi_{t+1}\le\widetilde q\Xi_t+
 \frac{KM^2}{1-\widetilde q}\|\Delta w_t\|^2,
 \qquad \|\delta_t\|^2\le\kappa_0\rho_t^2\Xi_t.
\]
Besides \eqref{eq:onestepcontain}, \prettyref{lem:descent} requires \eqref{eq:valuecontain} for all $u,v\in[w_t,w_{t+1}]$.  The just-proved containment $[w_t,w_{t+1}]\subseteq\mathcal W_{\rm cb}$ puts every such $u,v$ in $\mathcal W_{\rm cb}$, so \prettyref{lem:nslayered}\prettyref{cblay:use} puts every set in \eqref{eq:valuecontain} in $\mathcal D_{\rm cb}$.  Moreover $\alpha_t(L_F+\rho_tK(L_f+L_\star))\le c_\alpha(L_F+K(L_f+L_\star))\le1/2$, because $\rho_t\ge\rho_0\ge1$.  Hence
\[
 \Phi_{\rho_t}(w_{t+1})\le\Phi_{\rho_t}(w_t)
 -\frac{\rho_t}{2c_\alpha}\|\Delta w_t\|^2
 +c_\alpha\kappa_0\rho_t\Xi_t.
\]
\prettyref{lem:resid} requires only \eqref{eq:onestepcontain}, verified above, so it gives
\[
 \sum_i r_{i,t}^2\le\frac{2W_1^2}{\mu^2}
 \left(\frac{2\|\Delta w_t\|^2}{c_\alpha^2}+2\kappa_0\Xi_t+
 \frac{G_F^2}{\rho_t^2}\right).
\]
Finally, $w_{t+1}\in\mathcal W_{\rm cb}$ and \prettyref{lem:nslayered}\prettyref{cblay:use} place every set in \eqref{eq:driftcontain} in $\mathcal D_{\rm cb}$; \prettyref{lem:drift} gives
\[
 \Phi_{\rho_{t+1}}(w_{t+1})-\Phi_{\rho_t}(w_{t+1})
 \le(\rho_{t+1}-\rho_t)
 \left(\omega_1\|\Delta w_t\|^2+\omega_2\Xi_t+
 \frac{\omega_3}{\rho_t^2}\right).
\]

Since $w_t\in\mathcal W_{\rm cb}$, \prettyref{lem:nslayered}\prettyref{cblay:use} also places the conversion rectangle \eqref{eq:convrect} at $w=w_t$ in $\mathcal D_{\rm cb}$.

Combining the three displays from \prettyref{lem:onestep}, \ref{lem:descent}, and \ref{lem:drift} with the definition \eqref{eq:PsiUB} of $\Psi_{t+1}$ gives
\begin{align*}
 \Psi_{t+1}
 &\le\Phi_{\rho_t}(w_t)\\
 &\quad+\left[-\frac{\rho_t}{2c_\alpha}+(\rho_{t+1}-\rho_t)\omega_1
 +\frac{c_\Psi\rho_{t+1}KM^2}{1-\widetilde q}\right]\|\Delta w_t\|^2\\
 &\quad+\left[c_\alpha\kappa_0\rho_t
 +(\rho_{t+1}-\rho_t)\omega_2
 +c_\Psi\rho_{t+1}\widetilde q\right]\Xi_t
 +\omega_3\frac{\rho_{t+1}-\rho_t}{\rho_t^2}.
\end{align*}
Since $\rho_{t+1}-\rho_t\le\beta\rho_t$ and $\rho_{t+1}\le2\rho_t$, the displacement coefficient divided by $\rho_t$ obeys
\[
 -\frac1{2c_\alpha}+\beta\omega_1+
 \frac{2c_\Psi KM^2}{1-\widetilde q}
 \le-\frac1{2c_\alpha}+\frac1{4c_\alpha}
 +\frac{8c_\alpha\kappa_0KM^2}{(1-\widetilde q)^2}
 \le-\frac1{8c_\alpha},
\]
where \eqref{eq:rhoconsts}, \eqref{eq:nscalpha}, and \eqref{eq:betamax} were used.  The tracker coefficient beyond $c_\Psi\rho_t$, divided by $\rho_t$, satisfies
\[
 c_\alpha\kappa_0+\beta\omega_2
 +c_\Psi\{(1+\beta)\widetilde q-1\}
 \le2c_\alpha\kappa_0+c_\Psi\widetilde q+
 \frac{c_\Psi(1-\widetilde q)}4-c_\Psi
 =-c_\alpha\kappa_0,
\]
using $\beta\omega_2\le c_\alpha\kappa_0$ and $\beta\le(1-\widetilde q)/4$ from \eqref{eq:betamax}, and $c_\Psi=4c_\alpha\kappa_0/(1-\widetilde q)$ from \eqref{eq:rhoconsts}.  Therefore \eqref{eq:nslyap} holds.  Adding it to the budget clause at time $t$ proves the budget clause at $t+1$; the first part of the proof then gives the residual clause at $t+1$.  This closes the induction and proves every conclusion.
\end{proof}

\begin{lemma}[Localized PSK]\label{lem:nshpsk}
Under the hypotheses of \prettyref{lem:nsinvariant},
\begin{equation}\label{eq:nscertbound}
 \sum_{t\ge0}U_t\le B_0\le B_{\rm cb},
 \qquad \Ecal_t\le C_t,
 \qquad C_t^2\le\frac{16\chi^2}{c_\alpha}\rho_tU_t.
\end{equation}
Every iterate lies in $\mathcal W_{\rm cb}$.  If $\rho>0$, $\alpha>0$, $w=(x,y_1,\ldots,y_K)\in\mathcal W_{\rm cb}$, and $E=\cG_\alpha^\rho(w)$, then
\begin{equation}\label{eq:nsconversion}
 \cG_\alpha^{\varphi^C}(x)\le c_\varphi E+
 \frac{c_{\rm b1}(E+G_F)+c_{\rm b2}(E+G_F)^2}{\rho}.
\end{equation}
\end{lemma}

\begin{proof}
\prettyref{lem:nsinvariant} gives $\sum_{t<N}U_t\le B_0$ for every $N$; monotone convergence of the nonnegative partial sums gives the first claim.  For every $t$, \prettyref{lem:nsinvariant} gives $[w_t,w_{t+1}]\subseteq\mathcal W_{\rm cb}$, so every iterate lies in $\mathcal W_{\rm cb}$; it also supplies \eqref{eq:onestepcontain}.  Since $\alpha_t=c_\alpha/\rho_t$, \eqref{eq:certbound} and \eqref{eq:certupper} of \prettyref{lem:cert}, applied at $(\rho,\alpha)=(\rho_t,\alpha_t)$, give the remaining bounds in \eqref{eq:nscertbound}.

For $w\in\mathcal W_{\rm cb}$, \prettyref{lem:nslayered}\prettyref{cblay:use}, applied with $u=v=w$, places the conversion rectangle \eqref{eq:convrect} in $\mathcal D_{\rm cb}$.  Then \eqref{eq:convproj} of \prettyref{lem:conv} gives \eqref{eq:nsconversion}.
\end{proof}

\begin{proof}[Proof of \prettyref{thm:nsrecover}]
\prettyref{lem:nshpsk} gives the finite budget and certificate.  The window $\mathcal T_T:=\{\lfloor T/2\rfloor,\ldots,T-1\}$ has $|\mathcal T_T|=T-\lfloor T/2\rfloor=\lceil T/2\rceil\ge T/2$ indices.  Hence some $t'\in\mathcal T_T$ satisfies $U_{t'}\le2B_{\rm cb}/T$.  For $t\in\mathcal T_T$, $\beta=\beta_{\max}\le(1-\widetilde q)/4<1$ by \eqref{eq:betamax}, which gives
\[
 1+\beta t\le1+\beta(T-1)<2T,
 \qquad \rho_t\le\rho_0(2T)^{1/3}.
\]
Also $t\ge(T-1)/2$ and $\beta\le1$, so
\begin{align*}
 1+\beta t&\ge1+\frac{\beta(T-1)}2
 =\frac{\beta T}{2}+1-\frac{\beta}2\ge\frac{\beta T}{2},\\
 \rho_t^{-1}&\le\frac{2^{1/3}}{\rho_0\beta^{1/3}}T^{-1/3}
 =C_\rho T^{-1/3}.
\end{align*}
By minimality of $C_{t^\star}$ and \eqref{eq:nscertbound},
\[
 C_{t^\star}^2
 \le C_{t'}^2\le\frac{16\chi^2}{c_\alpha}\rho_{t'}U_{t'}
 \le\frac{16\chi^2}{c_\alpha}\rho_0(2T)^{1/3}
 \frac{2B_{\rm cb}}T
 =\chi^2C_E(B_{\rm cb})^2T^{-2/3}.
\]
Thus $\Ecal_{t^\star}\le\chi C_E(B_{\rm cb})T^{-1/3}$.  Since $w_{t^\star}\in\mathcal W_{\rm cb}$, \eqref{eq:nsconversion} applies, and its right side is nondecreasing in $E$.  Use the sharp bound in its linear term and, since $T\ge1$, the weaker $\Ecal_{t^\star}\le\chi C_E(B_{\rm cb})$ in the reciprocal-penalty bracket.  Then
\begin{align*}
 c_\varphi\Ecal_{t^\star}
 &\le c_\varphi\chi C_E(B_{\rm cb})T^{-1/3},\\
 \rho_{t^\star}^{-1}
 [c_{\rm b1}(\Ecal_{t^\star}+G_F)+c_{\rm b2}(\Ecal_{t^\star}+G_F)^2]
 &\le C_\rho T^{-1/3}
 [c_{\rm b1}(\chi C_E(B_{\rm cb})+G_F)
 +c_{\rm b2}(\chi C_E(B_{\rm cb})+G_F)^2].
\end{align*}
The specialized definition \eqref{eq:Cs} gives \eqref{eq:nsmainrate}.  \prettyref{lem:mono} covers every $\bar\alpha\ge\alpha_{t^\star}$ and hence every $\bar\alpha\ge c_\alpha/\rho_0$.  \prettyref{lem:nsexact} gives the exact-state certificates.
\end{proof}

\begin{remark}[What changes from the smooth proof]\label{rem:nshpsk}
Only the source of localization changes.  \prettyref{lem:nslayered} places every point and segment in $\mathcal D_{\rm cb}$ before a lemma of \prettyref{sec:smooth} is used.  The fixed stage family $(f_i^C)_{i=1}^K$ creates no surrogate-drift term, and feasibility is asserted only for the exact chain.
\end{remark}

\section{Depth-dependent complexity of shrinking-band surrogates}\label{sec:lb}

This appendix proves \prettyref{thm:lbpsk}: PSK is a deterministic multi-block zero-respecting algorithm, and when the band width equals the target accuracy $\epsilon$, every such algorithm needs at least $c_K\epsilon^{-(K+3)/4}$ primitive calls on a class of smooth instances.  The proof reuses the smooth zero chain of \citet{carmon2020lower} and the constant-band objects of \prettyref{sub:band} and \prettyref{sec:ns}: the cubic wall \eqref{eq:nswall}, the surrogate \eqref{eq:nscbsurrogate}, and \prettyref{ass:ns} and \ref{ass:nscb}.

The construction combines two mechanisms.  A zero chain reveals at most one new control coordinate per primitive call, so a zero-respecting algorithm needs at least as many calls as the chain has coordinates.  A $K$-level relay of two nearly parallel walls amplifies the derivative of the terminal loss by a factor of order $\epsilon^{-1/2}$ at every stage after the first, so a smaller and longer chain still keeps the residual above $\epsilon$.  \prettyref{sec:lbzerochain} recalls the zero chain.  \prettyref{sec:lbinstance} builds the relay and the hard instance and proves the amplification.  \prettyref{sec:lbmbzr} defines the algorithm class, shows that PSK belongs to it, and proves \prettyref{thm:lbpsk}.

\subsection{Carmon Zero Chain}\label{sec:lbzerochain}

Our hard instance builds on the first-order zero chain of \citet{carmon2020lower}, who used it to prove lower bounds for finding stationary points of smooth nonconvex functions.  We recall their construction and the properties that we use.  Put
\[
 \Psi_{\rm c}(t):=\begin{cases}
 0,&t\le\tfrac12,\\
 \exp\!\bigl(1-(2t-1)^{-2}\bigr),&t>\tfrac12,
 \end{cases}
 \qquad
 \Phi_{\rm c}(t):=\sqrt e\int_{-\infty}^t e^{-u^2/2}\,du,
\]
and, for $N\ge1$ and $z\in\R^N$, define the Carmon chain
\begin{equation}\label{eq:lbcdhs}
 \bar f_N(z):=-\Psi_{\rm c}(1)\Phi_{\rm c}(z_1)
 +\sum_{i=1}^{N-1}
 \bigl\{\Psi_{\rm c}(-z_i)\Phi_{\rm c}(-z_{i+1})
          -\Psi_{\rm c}(z_i)\Phi_{\rm c}(z_{i+1})\bigr\}.
\end{equation}

\begin{lemma}[Carmon zero chain]\label{lem:lbzerochain}
For every $N\ge1$, the function \eqref{eq:lbcdhs} has the following properties.
\begin{enumerate}[label=\textnormal{(\roman*)}]
\item It is a first-order zero chain: for every integer $0\le k\le N$, if $\operatorname{supp}(z)\subseteq\{1,\ldots,k\}$, then $\operatorname{supp}(\nabla\bar f_N(z))\subseteq\{1,\ldots,k+1\}$, where $\{1,\ldots,0\}:=\varnothing$.
\item For every $z\in\R^N$,
\[
 |\bar f_N(z)-\bar f_N(0)|\le12N.
\]
\item There are absolute constants $L_1,L_2<\infty$ such that
\begin{equation*}
\begin{aligned}
 \max_j|\partial_j\bar f_N(z)|&\le23,
 &\|\nabla\bar f_N(z)\|&\le23\sqrt N,\\
 \|D^2\bar f_N(z)\|_{\rm op}&\le L_1,
 &\|D^3\bar f_N(z)\|_{\rm op}&\le L_2.
\end{aligned}
\end{equation*}
Consequently $\operatorname{Lip}(D^2\bar f_N)\le L_2$.
\item If $|z_i|<1$ for some $i\le N$ and $j\in\{1,\ldots,N\}$ is the least index with $|z_j|<1$, then
\[
 \partial_j\bar f_N(z)<-1.
\]
In particular, such a $j$ exists whenever $z_N=0$.
\end{enumerate}
\end{lemma}

\begin{proof}
Item~(iii) is \citet[Lemma~3]{carmon2020lower}: it bounds every coordinate of the gradient by $23$, which gives $\|\nabla\bar f_N\|\le23\sqrt N$, and it shows that, for each order $k\ge1$, the $k$th derivative of $\bar f_N$ is Lipschitz with a constant at most $\exp(\frac52k\log k+ck)$ for an absolute constant $c$; since $\bar f_N$ is $C^\infty$, the constant for order $k$ also bounds $\|D^{k+1}\bar f_N\|_{\rm op}$, and the constants for $k=1,2$ are the absolute $L_1$ and $L_2$.  For completeness, we prove items~(i), (ii), and~(iv) from three elementary facts.
\begin{itemize}
\item[(F1)] $\Psi_{\rm c}$ is $C^\infty$ \citep[Lemma~1]{carmon2020lower}, and $\Psi_{\rm c}(t)=\Psi_{\rm c}'(t)=0$ for $t\le\frac12$.  On $t>\frac12$, the map $t\mapsto1-(2t-1)^{-2}$ is increasing and tends to $1$ as $t\to\infty$, so $\Psi_{\rm c}$ is nondecreasing on $\R$, $\Psi_{\rm c}'\ge0$, and $0\le\Psi_{\rm c}<e$.  Also $\Psi_{\rm c}(1)=e^{1-1}=1$, so $\Psi_{\rm c}(t)\ge1$ for $t\ge1$.
\item[(F2)] $\Phi_{\rm c}'(t)=\sqrt e\,e^{-t^2/2}>0$ is even, and $0<\Phi_{\rm c}<\sqrt e\int_{\R}e^{-u^2/2}\,du=\sqrt{2\pi e}$.  If $|t|<1$, then $e^{-t^2/2}>e^{-1/2}$, so $\Phi_{\rm c}'(t)>1$.
\item[(F3)] For every $t\in\R$, at most one of $\Psi_{\rm c}(t)$ and $\Psi_{\rm c}(-t)$ is nonzero, because one of $t,-t$ is at most $0\le\frac12$; by (F1), if $|t|\ge1$, the nonzero one is $\Psi_{\rm c}(|t|)\ge1$.
\end{itemize}

\emph{Derivative formula.}  Put $z_0:=1$.  Since $\Psi_{\rm c}(-1)=0$, the first term of \eqref{eq:lbcdhs} equals $\Psi_{\rm c}(-z_0)\Phi_{\rm c}(-z_1)-\Psi_{\rm c}(z_0)\Phi_{\rm c}(z_1)$, so
\[
 \bar f_N(z)=\sum_{i=0}^{N-1}
 \bigl\{\Psi_{\rm c}(-z_i)\Phi_{\rm c}(-z_{i+1})
          -\Psi_{\rm c}(z_i)\Phi_{\rm c}(z_{i+1})\bigr\}.
\]
The coordinate $z_j$ ($1\le j\le N$) appears in the summand $i=j-1$ (as $z_{i+1}$) and, if $j<N$, in the summand $i=j$ (as $z_i$).  Differentiating these two summands gives
\begin{equation}\label{icml:lbchainderiv}
 \partial_j\bar f_N(z)
 =\underbrace{-\Psi_{\rm c}(-z_{j-1})\Phi_{\rm c}'(-z_j)
     -\Psi_{\rm c}(z_{j-1})\Phi_{\rm c}'(z_j)}_{\text{incoming terms}}
 \underbrace{-\Psi_{\rm c}'(-z_j)\Phi_{\rm c}(-z_{j+1})
     -\Psi_{\rm c}'(z_j)\Phi_{\rm c}(z_{j+1})}_{\text{outgoing terms}},
\end{equation}
where the outgoing terms are absent if $j=N$.  By (F1) and (F2), each of the four terms is at most $0$.

\emph{Item~(i).}  Let $\operatorname{supp}(z)\subseteq\{1,\ldots,k\}$ and $k+2\le j\le N$.  Then $j-1\ge k+1\ge1$, so $z_{j-1}=z_j=0$.  Each incoming term in \eqref{icml:lbchainderiv} contains the factor $\Psi_{\rm c}(\pm z_{j-1})=\Psi_{\rm c}(0)=0$, and each outgoing term contains $\Psi_{\rm c}'(\pm z_j)=\Psi_{\rm c}'(0)=0$, by (F1).  Hence $\partial_j\bar f_N(z)=0$ for every $j\ge k+2$.

\emph{Item~(ii).}  At $z=0$, every summand of \eqref{eq:lbcdhs} with $i\ge1$ contains $\Psi_{\rm c}(0)=0$, so $\bar f_N(0)=-\Phi_{\rm c}(0)$ and
\[
 \bar f_N(z)-\bar f_N(0)
 =-\bigl[\Phi_{\rm c}(z_1)-\Phi_{\rm c}(0)\bigr]
 +\sum_{i=1}^{N-1}
 \bigl\{\Psi_{\rm c}(-z_i)\Phi_{\rm c}(-z_{i+1})
          -\Psi_{\rm c}(z_i)\Phi_{\rm c}(z_{i+1})\bigr\}.
\]
By (F2), $\Phi_{\rm c}(z_1)$ and $\Phi_{\rm c}(0)$ lie in $(0,\sqrt{2\pi e})$, so the first bracket has absolute value less than $\sqrt{2\pi e}$.  By (F3), each summand of the sum has at most one nonzero term, and by (F1) and (F2) that term has absolute value less than $e\sqrt{2\pi e}$.  Since $1\le e$ and $e\sqrt{2\pi e}<11.3$,
\[
 |\bar f_N(z)-\bar f_N(0)|
 <\sqrt{2\pi e}\,\bigl(1+(N-1)e\bigr)
 \le e\sqrt{2\pi e}\,N<12N.
\]

\emph{Item~(iv).}  Let $j$ be the least index in $\{1,\ldots,N\}$ with $|z_j|<1$.
\begin{enumerate}[label=\textnormal{(\alph*)}]
\item $|z_{j-1}|\ge1$: for $j\ge2$ this is the minimality of $j$, and for $j=1$ it is the convention $z_0=1$.
\item By (a) and (F3), exactly one incoming term of \eqref{icml:lbchainderiv} is nonzero, and since $\Phi_{\rm c}'$ is even, it equals $-\Psi_{\rm c}(|z_{j-1}|)\,\Phi_{\rm c}'(z_j)$.
\item By (F1) and (a), $\Psi_{\rm c}(|z_{j-1}|)\ge1$.  By (F2) and $|z_j|<1$, $\Phi_{\rm c}'(z_j)>1$.  Hence this incoming term is less than $-1$.
\item The outgoing terms are at most $0$, so $\partial_j\bar f_N(z)$ is at most the incoming term in~(c), and therefore $\partial_j\bar f_N(z)<-1$.
\end{enumerate}
Finally, if $z_N=0$, then $|z_N|<1$, so the least such $j$ exists.
\end{proof}

\subsection{Hard Instance Construction}\label{sec:lbinstance}

Fix the depth $K\ge2$ and the relay constants
\[
 \Gamma:=128,\qquad \sigma_{\rm rel}:=\frac1{100},\qquad
 \mathfrak m:=3,\qquad
 c_{\mathfrak q}:=\frac1{4\sigma_{\rm rel}^{K-2}}.
\]
For a target accuracy $0<\epsilon\le1$, set $\theta:=\sqrt\epsilon$.  To keep the state derivatives globally bounded, define the limiter
\[
 P_\zeta(t):=1-10t^3+15t^4-6t^5\qquad(0\le t\le1),
\]
\[
 \chi_\zeta(\varrho):=
 \begin{cases}
  1,&|\varrho|\le1,\\
  P_\zeta((|\varrho|-1)/4),&1<|\varrho|<5,\\
  0,&|\varrho|\ge5,
 \end{cases}
 \qquad
 \zeta(\varrho):=\int_0^\varrho\chi_\zeta(s)\,ds.
\]
The endpoint identities for the quintic give $\chi_\zeta\in C^2$ and $\zeta\in C^3$.  Direct differentiation gives the global bounds
\begin{equation}\label{eq:lbzetabounds}
 |\zeta|\le3,\qquad |\zeta'|\le1,\qquad
 |\zeta''|\le\frac{15}{32},\qquad
 \operatorname{Lip}(\zeta'')\le\frac{15}{16},
 \qquad \zeta(\varrho)=\varrho\quad(|\varrho|\le1).
\end{equation}
Thus $\zeta'$ is supported on $[-5,5]$, while $\zeta(\varrho)=3\operatorname{sgn}(\varrho)$ for $|\varrho|\ge5$.

The relay acts on states $y=(u,v,\varrho)\in\R^3$, so $n_y=3$.  Fix a control dimension $n_x\ge1$ and a $C^1$ signal $\mathfrak q:\R^{n_x}\to\R$; the signal used for the lower bound is fixed at the end of this subsection.  For an incoming state $p=(p_u,p_v,p_\varrho)$, let $\ell(p):=p_v+p_\varrho$ and set
\begin{equation}\label{eq:lbinstance}
\begin{aligned}
 h_+(p,y)&:=u-2+\theta v+\sigma_{\rm rel}\ell(p),\\
 h_-(p,y)&:=u-2-\theta v-\sigma_{\rm rel}\ell(p),\\
 g(x,p,y)&:=\tfrac12(u-2)^2-\mathfrak m(u-2)+\tfrac12v^2
              +\varrho^2-2\mathfrak q(x)\zeta(\varrho),\\
 F(x,y_1,\ldots,y_K)&:=\tanh(v_K),\qquad y_0:=0.
\end{aligned}
\end{equation}
The relay is diagonal: $\SelMat_i:=I_{n_x}$ and $g_i:=g$ for every $i$, $N_h:=2$, $h_1:=h_+$, and $h_2:=h_-$.  Every stage uses the height $\Gamma=128$ and a common band width $d$ with
\begin{equation}\label{eq:lbwallrule}
 0<d\le1.
\end{equation}
For a stage family $(g_i)_{i=1}^K$, constraints $(h_l)_{l=1}^{N_h}$, and a band width $d$, the surrogate and its exact chain are
\begin{equation}\label{eq:lbwallchain}
\begin{gathered}
 f_{i,d}^C(x,p,y):=g_i(x,p,y)
   +\sum_{l=1}^{N_h}\psi_{d,\Gamma}(-h_l(p,y)),\qquad
 s_{i,d}^C(x,p):=\argmin_y f_{i,d}^C(x,p,y),\\
 \widehat y_{0,d}^C(x):=y_0,
 \qquad
 \widehat y_{i,d}^C(x):=
 s_{i,d}^C(x,\widehat y_{i-1,d}^C(x)),\\
 \varphi_d^C(x):=
 F(x,\widehat y_{1,d}^C(x),\ldots,
     \widehat y_{K,d}^C(x)).
\end{gathered}
\end{equation}
On the relay, $f_{i,d}^C=f_d^C$ and $s_{i,d}^C=s_d^C$ for every $i$, where
\[
 f_d^C(x,p,y):=g(x,p,y)+\sum_{l=1}^{N_h}\psi_{d,\Gamma}(-h_l(p,y)),
 \qquad
 s_d^C(x,p):=\argmin_y f_d^C(x,p,y).
\]

\paragraph{How the relay amplifies.}  The construction is easiest to read on the exact chain.
\begin{itemize}
\item \emph{The limiter.}  In the regime of \prettyref{lem:lbamplifier} below, $|\mathfrak q|\le\theta^2/4\le1$ and, on the exact chain, $|\varrho|\le1$, so only the piece $\zeta(\varrho)=\varrho$ is used, where the $\varrho$-part of $g$ is $\varrho^2-2\mathfrak q(x)\varrho$.  Without saturation, $\nabla^2_{xx}g=-2\varrho\nabla^2\mathfrak q(x)$ would grow with $|\varrho|$, and $\nabla^2g$ would not be globally Lipschitz; the limiter only serves to meet this requirement of the instance class.
\item \emph{Two nearly parallel walls.}  Both constraints are affine, with normals $\nabla_yh_\pm=(1,\pm\theta,0)$.  Their angle is of order $\theta=\sqrt\epsilon$, so they become parallel, and the two active normals nearly linearly dependent, as the band width $d=\theta^2$ shrinks.
\item \emph{One incoming scalar.}  The energy $g$ does not depend on $p$; the previous state enters only through the scalar $\ell(p)$ in the two walls.  The walls do not involve $\varrho$, so $\varrho$ decouples and its minimizer is $\varrho=\mathfrak q(x)$.  Hence stage $i+1$ receives $\ell=v_i+\mathfrak q(x)$: the control enters every stage, and the first stage receives $\ell(y_0)=0$.
\item \emph{The wedge.}  The feasible set is the wedge $u-2\le-|\theta v+\sigma_{\rm rel}\ell|$.  The tilt $-\mathfrak m(u-2)$ pushes $u$ upward into both walls, and both stay engaged, with margins inside the band, at every stage.  The minimizer therefore stays near the tip of the wedge, where $\theta v+\sigma_{\rm rel}\ell=0$, so $v\approx-(\sigma_{\rm rel}/\theta)\ell$.  A small shift of $\ell$ moves $v$ about $\sigma_{\rm rel}/\theta$ times farther; the term $\frac12v^2$ and the finite band reduce this gain only by a bounded factor.
\item \emph{Depth.}  Every stage after the first multiplies the effect of the signal by a factor of order $1/\theta$, and $F=\tanh(v_K)$ reads the result.  The signal is scaled by $c_{\mathfrak q}\theta^K$ so that every incoming scalar stays in $[-\theta^2,\theta^2]$, where this picture is exact.
\end{itemize}
\prettyref{fig:lb} sketches this relay.  The next lemma makes the amplification precise.

\begin{lemma}[Fixed-height depth amplification]\label{lem:lbamplifier}
There are constants $0<\theta_K\le1$ and $\bar c_K>0$, depending only on $K$, with the following property.  Let $n_x\ge1$, let $\mathfrak q\in C^1(\R^{n_x})$, let $0<\theta\le\theta_K$ and $0<d\le\theta^2$, and suppose that $|\mathfrak q|\le c_{\mathfrak q}\theta^K$ on $\R^{n_x}$.  Then $\varphi_d^C\in C^1(\R^{n_x})$, and there is a function $\kappa_d:\R^{n_x}\to\R$ such that
\begin{equation}\label{eq:lbdepthgain}
 \nabla\varphi_d^C(x)=\kappa_d(x)\nabla \mathfrak q(x),\qquad
 |\kappa_d(x)|\ge\bar c_K\theta^{-(K-1)}
 \qquad(x\in\R^{n_x}).
\end{equation}
\end{lemma}

\begin{proof}
\emph{Constants.}  Put $\eta_{\rm rel}:=\frac12$,
\[
 c_0:=\sqrt{\frac{\mathfrak m}{6\Gamma(1+\eta_{\rm rel}^2)}}=\frac1{8\sqrt5},
 \qquad c_T:=(1-\eta_{\rm rel}^2)c_0=\frac3{32\sqrt5},
 \qquad c_{\rm amp}:=\frac1{12\Gamma c_T},\qquad
 \sigma_-:=\frac{\sigma_{\rm rel}}{1+c_{\rm amp}},
\]
and $\bar v:=\sigma_{\rm rel}/2$.  We prove the lemma with
\[
 \theta_K:=\min\left\{1,\frac{\sigma_{\rm rel}}2,\frac{\sigma_-}4\right\},
 \qquad
 \bar c_K:=\operatorname{sech}^2(\bar v)\frac{\sigma_-^{K-1}}{2^{K-2}}.
\]
Fix $n_x$, $\mathfrak q$, $\theta$, and $d$ as in the statement.  Write $A_d:=12\Gamma/d$ and $\bar{\mathfrak g}:=\sigma_{\rm rel}/\theta$.  Since $\theta\le\sigma_{\rm rel}/2$, one has $\bar{\mathfrak g}\ge2$ and
\[
 |\mathfrak q|\le c_{\mathfrak q}\theta^K
 =\frac{\theta^2}{4}\left(\frac{\theta}{\sigma_{\rm rel}}\right)^{K-2}
 \le\frac{\theta^2}{4}\le1.
\]

\emph{One stage.}  The cubic wall is convex: away from its junctions,
\[
 \psi_{d,\Gamma}''(t)=
 \begin{cases}
 0,&t>d,\\
 \dfrac{6\Gamma}{d^2}(1-t/d),&0<t<d,\\
 \dfrac{6\Gamma}{d^2},&t<0,
 \end{cases}
\]
and these values match continuously.  By \eqref{eq:lbzetabounds}, the outgoing $\varrho$-curvature of $g$ satisfies
\[
 2-2\mathfrak q(x)\zeta''(\varrho)\ge2-\frac{30}{32}=\frac{17}{16}>1.
\]
The other two outgoing curvatures equal one, and the walls are composed with affine functions of the outgoing state, so $f_d^C(x,p,\cdot)$ is $1$-strongly convex and has a unique global minimizer.  The walls omit the outgoing $\varrho$-coordinate, and $\varrho\mapsto \varrho^2-2\mathfrak q(x)\zeta(\varrho)$ is strictly convex. Since $|\mathfrak q(x)|\le1$ and $\zeta'(\mathfrak q(x))=1$, its stationarity equation $2\varrho-2\mathfrak q(x)\zeta'(\varrho)=0$ holds at $\varrho=\mathfrak q(x)$, so the minimizer has $\varrho=\mathfrak q(x)$. The wall multipliers are
\begin{equation*}
 -\psi'_{d,\Gamma}(t)=
 \begin{cases}
 0,&t>d,\\
 \dfrac{3\Gamma}{d}(1-t/d)^2,&0\le t\le d,\\
 \dfrac{3\Gamma}{d}+\dfrac{6\Gamma(-t)}{d^2},&t<0.
 \end{cases}
\end{equation*}

Fix the incoming scalar $\ell=\ell(p)$, write the outgoing state as $y=(u,v,\varrho)$, and set
\[
 \xi:=1+\frac{u-2}d,
 \qquad \mathfrak s:=\frac{\theta v+\sigma_{\rm rel}\ell}{d}.
\]
The two wall slacks are $d(1-\xi\mp \mathfrak s)$.  On the middle wall branches, equivalently $0<\xi\pm \mathfrak s<1$, stationarity in $(u,v)$ is equivalent to
\begin{equation}\label{eq:lbscalareq}
\begin{aligned}
 \frac{6\Gamma}{d}(\xi^2+\mathfrak s^2)+d(\xi-1)-\mathfrak m&=0,\\
 v&=-A_d\theta\xi \mathfrak s,\\
 \sigma_{\rm rel}\ell&=\mathfrak s(d+A_d\theta^2\xi).
\end{aligned}
\end{equation}
Let $\xi_b$ be the positive root of
\[
 \frac{6\Gamma}{d}(1+\eta_{\rm rel}^2)\xi^2+d\xi=\mathfrak m+d,
 \qquad \mathfrak s_b:=\eta_{\rm rel}\xi_b.
\]
Since $6\Gamma/d>\mathfrak m$, one has $\xi_b<1$, and hence $(6\Gamma/d)(1+\eta_{\rm rel}^2)\xi_b^2\ge\mathfrak m$, that is, $\xi_b\ge c_0\sqrt d$.  For $|\mathfrak s|\le \mathfrak s_b$, let $\Xi_d(\mathfrak s)$ be the positive root of the first equation in \eqref{eq:lbscalareq}, namely of $(6\Gamma/d)\xi^2+d\xi=\mathfrak m+d-(6\Gamma/d)\mathfrak s^2$.  The right side decreases in $|\mathfrak s|$ and, by the equation for $\xi_b$, is positive at $\mathfrak s_b$.  Hence $\Xi_d$ is well defined, $C^1$, even, nonincreasing in $|\mathfrak s|$, and $\Xi_d(\mathfrak s_b)=\xi_b$.  At $\mathfrak s=0$, the root satisfies $(6\Gamma/d)\Xi_d(0)^2=\mathfrak m+d(1-\Xi_d(0)) \le\mathfrak m+1$.  Therefore, for $|\mathfrak s|\le \mathfrak s_b$,
\begin{equation}\label{eq:lbrootbox}
 |\mathfrak s|\le\eta_{\rm rel}\Xi_d(\mathfrak s),\qquad
 c_0\sqrt d\le\Xi_d(\mathfrak s)
 \le\sqrt{\frac{\mathfrak m+1}{6\Gamma}}\sqrt d
 =\frac1{\sqrt{192}}\sqrt d.
\end{equation}
Implicit differentiation gives $\Xi_d'(\mathfrak s)=-A_d \mathfrak s/(A_d\Xi_d(\mathfrak s)+d)$.  Put $T_d(\mathfrak s):=\Xi_d(\mathfrak s)+\mathfrak s\Xi_d'(\mathfrak s)$.  Since $A_d \mathfrak s^2/(A_d\Xi_d+d)\le \mathfrak s^2/\Xi_d \le\eta_{\rm rel}^2\Xi_d$, \eqref{eq:lbrootbox} gives
\begin{equation}\label{eq:lbTbounds}
 c_T\sqrt d
 \le(1-\eta_{\rm rel}^2)\Xi_d(\mathfrak s)
 \le T_d(\mathfrak s)\le\Xi_d(\mathfrak s).
\end{equation}

Define the odd map $\ell_d(\mathfrak s):=\sigma_{\rm rel}^{-1}\mathfrak s(d+A_d\theta^2\Xi_d(\mathfrak s))$ on $[-\mathfrak s_b,\mathfrak s_b]$.  Then
\[
 \ell_d'(\mathfrak s)=\frac1{\sigma_{\rm rel}}
 \bigl(d+A_d\theta^2T_d(\mathfrak s)\bigr)>0,
\]
and $A_d\xi_b^2\ge2\mathfrak m/(1+\eta_{\rm rel}^2)$ gives
\[
 \ell_d(\mathfrak s_b)
 \ge\frac{2\eta_{\rm rel}\mathfrak m}{\sigma_{\rm rel}(1+\eta_{\rm rel}^2)}\theta^2
 =240\theta^2.
\]
Thus, for $|\ell|\le\theta^2$, $\mathfrak s:=\ell_d^{-1}(\ell)$ is well defined with $|\mathfrak s|<\mathfrak s_b$, and $\xi:=\Xi_d(\mathfrak s)$, $v:=-A_d\theta\xi \mathfrak s$, $u:=2+d(\xi-1)$, $\varrho:=\mathfrak q(x)$ solve \eqref{eq:lbscalareq}.  Moreover,
\[
 0<(1-\eta_{\rm rel})\Xi_d(\mathfrak s)\le\Xi_d(\mathfrak s)\pm \mathfrak s
 \le(1+\eta_{\rm rel})\Xi_d(\mathfrak s)
 \le\frac3{2\sqrt{192}}<1,
\]
so this point lies on the middle branches.  It is therefore stationary, hence the unique minimizer.  Consequently, for $|\ell(p)|\le\theta^2$, the $v$-coordinate of $s_d^C(x,p)$ is
\[
 V_d(\ell(p)),\qquad
 V_d(\ell):=-A_d\theta\,\Xi_d(\mathfrak s)\,\mathfrak s,\quad
 \mathfrak s=\ell_d^{-1}(\ell).
\]
The map $V_d$ is odd and $C^1$ on $[-\theta^2,\theta^2]$, and $V_d(0)=0$.  Differentiating through $\ell_d^{-1}$ gives
\[
 -V_d'(\ell)=
 \frac{A_d\sigma_{\rm rel}\theta T_d(\mathfrak s)}
      {d+A_d\theta^2T_d(\mathfrak s)}
 =\frac{\sigma_{\rm rel}}{\theta}
  \left(1+\frac{d}{A_d\theta^2T_d(\mathfrak s)}\right)^{-1}.
\]
By \eqref{eq:lbTbounds}, $d\le\theta^2$, and $d\le1$,
\[
 \frac{d}{A_d\theta^2T_d(\mathfrak s)}
 \le\frac{d^{3/2}}{12\Gamma c_T\theta^2}
 \le\frac1{12\Gamma c_T}=c_{\rm amp}.
\]
Therefore
\begin{equation}\label{eq:lbstagegain}
 \frac{\sigma_-}{\theta}\le -V_d'(\ell)
 \le\frac{\sigma_{\rm rel}}{\theta}=\bar{\mathfrak g}
 \qquad(|\ell|\le\theta^2).
\end{equation}

\emph{The chain.}  Write $\widehat y_{i,d}^C(x)=(u_i,v_i,\varrho_i)$.  The first stage has $\ell(y_0)=0$, so $v_1=V_d(0)=0$.  Every stage has $\varrho_i=\mathfrak q(x)$, so stage $i\ge2$ has incoming scalar $v_{i-1}+\mathfrak q(x)$.  For a scalar $\mathfrak q$ with $|\mathfrak q|<2c_{\mathfrak q}\theta^K$, define
\[
 v_1(\mathfrak q):=0,\qquad
 v_i(\mathfrak q):=V_d(v_{i-1}(\mathfrak q)+\mathfrak q)\quad(2\le i\le K).
\]
Since $V_d(0)=0$, \eqref{eq:lbstagegain} gives $|V_d(\ell)|\le\bar{\mathfrak g}|\ell|$, and a joint induction on $i$, which uses the next display at each step, gives $|v_i(\mathfrak q)|\le|\mathfrak q|\sum_{j=1}^{i-1}\bar{\mathfrak g}^j$ as long as the previous arguments lie in $[-\theta^2,\theta^2]$.  Because $\bar{\mathfrak g}\ge2$, for $|\mathfrak q|\le c_{\mathfrak q}\theta^K$ and $2\le i\le K$,
\[
 |v_{i-1}(\mathfrak q)+\mathfrak q|
 \le|\mathfrak q|\sum_{j=0}^{i-2}\bar{\mathfrak g}^j
 \le2c_{\mathfrak q}\theta^K\bar{\mathfrak g}^{K-2}=\frac{\theta^2}{2}.
\]
Thus every argument stays in $[-\theta^2,\theta^2]$, every $v_i$ is $C^1$, and $v_i=v_i(\mathfrak q(x))$ for $1\le i\le K$.  For $|\mathfrak q|<2c_{\mathfrak q}\theta^K$ the same estimate gives $|v_{i-1}(\mathfrak q)+\mathfrak q|<4c_{\mathfrak q}\theta^K\bar{\mathfrak g}^{K-2}=\theta^2$, which keeps every argument in the open interval $(-\theta^2,\theta^2)$, so each $v_i$ is $C^1$ on the open interval $|\mathfrak q|<2c_{\mathfrak q}\theta^K$, which contains every value of $\mathfrak q(x)$.  For $|\mathfrak q|\le c_{\mathfrak q}\theta^K$, in particular at $\mathfrak q=\mathfrak q(x)$, the same estimate gives
\[
 |v_K(\mathfrak q)|\le|\mathfrak q|\sum_{j=1}^{K-1}\bar{\mathfrak g}^j\le2c_{\mathfrak q}\theta^K\bar{\mathfrak g}^{K-1}
 =\frac{\sigma_{\rm rel}}2\theta=\bar v\theta\le\bar v.
\]

\emph{The derivative.}  Let $Z_i:=\mathrm d v_i/\mathrm d\mathfrak q$ and $\mathfrak g_i:=-V_d'(v_{i-1}(\mathfrak q)+\mathfrak q)$.  Then
\[
 Z_1=0,\qquad Z_i=-\mathfrak g_i(Z_{i-1}+1),\qquad
 \mathfrak g_i\ge\frac{\sigma_-}{\theta}\ge4,
\]
the last inequality because $\theta\le\sigma_-/4$.  Hence $|Z_2|=\mathfrak g_2\ge\sigma_-/\theta\ge4$.  If $|Z_{i-1}|\ge2$, then
\[
 |Z_i|\ge\mathfrak g_i(|Z_{i-1}|-1)\ge\frac{\sigma_-}{2\theta}|Z_{i-1}|
 \ge2|Z_{i-1}|.
\]
Induction from $i=2$ therefore gives
\[
 |Z_K|\ge\frac{\sigma_-^{K-1}}{2^{K-2}}\theta^{-(K-1)}.
\]
Since $\varphi_d^C(x)=\tanh(v_K(\mathfrak q(x)))$ for every $x$, the map $\varphi_d^C$ is $C^1$ and the chain rule gives \eqref{eq:lbdepthgain} with
\[
 \kappa_d(x):=\operatorname{sech}^2\bigl(v_K(\mathfrak q(x))\bigr)\,
 Z_K(\mathfrak q(x)).
\]
Finally, $|v_K|\le\bar v$ gives $\operatorname{sech}^2(v_K)\ge\operatorname{sech}^2(\bar v)$, so $|\kappa_d(x)|\ge\bar c_K\theta^{-(K-1)}$.
\end{proof}

\paragraph{The signal and the hard instance.}  Fix $\theta_K$ and $\bar c_K$ from \prettyref{lem:lbamplifier}, and put
\[
 c_r:=\frac2{\sqrt{\bar c_K}},\qquad c_{\rm len}:=\frac{c_{\mathfrak q}}{24}.
\]
Set
\begin{equation}\label{eq:lbscaling}
 r_{\rm zc}:=c_r\theta^{(K+1)/2},\qquad
 A_{\rm zc}:=r_{\rm zc}^3,
\end{equation}
\begin{equation}\label{eq:lblength}
 \widetilde N_\theta:=c_{\rm len}\frac{\theta^K}{r_{\rm zc}^3}
 =\frac{c_{\rm len}}{c_r^3}\theta^{-(K+3)/2},\qquad
 N_{\rm zc}:=\lfloor \widetilde N_\theta\rfloor,
 \qquad \cX:=[-1,1]^{N_{\rm zc}}.
\end{equation}
Whenever $N_{\rm zc}\ge1$, take $n_x:=N_{\rm zc}$ and the signal
\[
 \mathfrak q(x):=A_{\rm zc}
 \bigl\{\bar f_{N_{\rm zc}}(x/r_{\rm zc})
             -\bar f_{N_{\rm zc}}(0)\bigr\}.
\]
We denote the relay \eqref{eq:lbinstance} with this signal by $\mathsf I_\epsilon$, the hard instance, and write $\varphi_\epsilon^C$ for its exact chain loss \eqref{eq:lbwallchain} at $d=\epsilon=\theta^2$.  The signal inherits the zero-chain structure: $\partial_j\mathfrak q(x)=(A_{\rm zc}/r_{\rm zc})\,\partial_j\bar f_{N_{\rm zc}}(x/r_{\rm zc})$, $\operatorname{supp}(x/r_{\rm zc})=\operatorname{supp}(x)$, and $A_{\rm zc}/r_{\rm zc}=r_{\rm zc}^2=c_r^2\theta^{K+1}$.  Since $A_{\rm zc}N_{\rm zc}\le A_{\rm zc}\widetilde N_\theta=c_{\rm len}\theta^K$ and $\sqrt{N_{\rm zc}}\le\sqrt{\widetilde N_\theta}$, \prettyref{lem:lbzerochain}(ii) and~(iii) give, for every $x\in\R^{N_{\rm zc}}$,
\begin{equation}\label{icml:lbqbounds}
\begin{gathered}
 |\mathfrak q(x)|\le12A_{\rm zc}N_{\rm zc}\le\frac{c_{\mathfrak q}}{2}\theta^K,
 \qquad
 \|\nabla \mathfrak q(x)\|\le23\sqrt{N_{\rm zc}}\frac{A_{\rm zc}}{r_{\rm zc}}
 \le23\sqrt{c_{\rm len}c_r}\,\theta^{(3K+1)/4},\\
 \|\nabla^2\mathfrak q(x)\|\le L_1\frac{A_{\rm zc}}{r_{\rm zc}^2}=L_1c_r\theta^{(K+1)/2},
 \qquad
 \operatorname{Lip}(\nabla^2\mathfrak q)\le L_2\frac{A_{\rm zc}}{r_{\rm zc}^3}=L_2.
\end{gathered}
\end{equation}
In particular, $\mathfrak q$ is globally $C^{2,1}$, and for every sufficiently small $\theta$, depending only on $K$,
\begin{equation}\label{eq:lbqbox}
 |\mathfrak q|\le1,\qquad \|\nabla \mathfrak q\|\le1,
 \qquad \|\nabla^2\mathfrak q\|\le1,
 \qquad \operatorname{Lip}(\nabla^2\mathfrak q)\le L_2.
\end{equation}

\subsection{Lower Bound of Multi-Block Zero-Respecting Methods}\label{sec:lbmbzr}

This subsection defines the oracle model and the class of multi-block zero-respecting algorithms, shows that warm-started PSK belongs to this class, proves the lower bound for the whole class on the hard instance $\mathsf I_\epsilon$ of \prettyref{sec:lbinstance}, and then proves \prettyref{thm:lbpsk}.

\paragraph{Oracle model.} Fix ambient data $K,n_y,N_h,y_0$.  An \emph{instance} is a tuple
\[
 \mathsf I=(n_x,F,(\SelMat_i,g_i)_{i=1}^K,h_1,\ldots,h_{N_h}),
\]
where $n_x\ge1$, every $\SelMat_i\in\{0,1\}^{n_{u,i}\times n_x}$ is a coordinate-selection matrix, and the maps have signatures
\[
 F:\R^{n_x}\times\R^{Kn_y}\to\R,\qquad
 g_i:\R^{n_x}\times\R^{n_y}\times\R^{n_y}\to\R,\qquad
 h_l:\R^{n_y}\times\R^{n_y}\to\R.
\]
The selector family is known instance data and may be included in $\mathsf{info}(\mathsf I)$ below.  Fix $\alpha_*>0$, write $\cX_{n_x}:=[-1,1]^{n_x}$, and, for $\mathfrak f\in C^1(\R^{n_x})$, define
\begin{equation}\label{icml:Gstar}
 \cG_{*,n_x}^{\mathfrak f}(x):=\mathcal G^{\mathfrak f}_{\alpha_*,\cX_{n_x}}(x)
 \quad\text{from \eqref{eq:G}}.
\end{equation}
We suppress $n_x$ from $\cG_{*,n_x}^{\mathfrak f}$ when the instance dimension is clear.  Vectors are indexed by the blocks of the extended trajectory $(x,y_0,y_1,\ldots,y_K)$, namely
\[
 \mathsf B:=\{x,0,1,\ldots,K\}.
\]
A \emph{query} is a pair $(\mathsf o,\mathsf z)$ of a label
\begin{equation}\label{icml:oracleset}
 \mathsf o\in\mathsf O:=\{F,\cX\}\cup
 \{(g_i,i):1\le i\le K\}\cup
 \{(h_l,i):1\le l\le N_h,\ 1\le i\le K\}
\end{equation}
and a query point $\mathsf z$ whose arguments are assigned to blocks by
\begin{equation}\label{eq:lbsignature}
\begin{array}{lll}
 \mathsf o=F: & \mathsf z=(x,y_1,\ldots,y_K), & \text{blocks }(x,1,\ldots,K),\\
 \mathsf o=(g_i,i): & \mathsf z=(x,p,y), & \text{blocks }(x,i-1,i),\\
 \mathsf o=(h_l,i): & \mathsf z=(p,y), & \text{blocks }(i-1,i),\\
 \mathsf o=\cX: & \mathsf z=v\in\R^{n_x}, & \text{block }x.
\end{array}
\end{equation}
Thus a $(g_i,i)$- or $(h_l,i)$-query treats $(p,y)$ as a candidate pair $(y_{i-1},y_i)$.  A block is a slot of the extended trajectory, not a property of the dummy variables $p,y$.  The reply $\mathsf r=\mathcal O^{\mathsf I}_{\mathsf o}(\mathsf z)$ is
\begin{equation}\label{eq:lboracles}
\begin{aligned}
 \mathcal O^{\mathsf I}_F(x,y_1,\ldots,y_K)
  &:=(F,\nabla_xF,\nabla_{y_1}F,\ldots,\nabla_{y_K}F)
       (x,y_1,\ldots,y_K),\\
 \mathcal O^{\mathsf I}_{g_i,i}(x,p,y)
  &:=(g_i,\nabla_xg_i,\nabla_pg_i,\nabla_yg_i)(x,p,y),\\
 \mathcal O^{\mathsf I}_{h_l,i}(p,y)
  &:=(h_l,\nabla_ph_l,\nabla_yh_l)(p,y),\\
 \mathcal O^{\mathsf I}_\cX(v)&:=\PiX_{\cX_{n_x}}(v).
\end{aligned}
\end{equation}
Every derivative in a reply inherits the block of the argument it differentiates.  The projection reply has block $x$.  Function values are scalars without a block.  Each block occurs at most once in every line of \eqref{eq:lbsignature}.  Hence the block-$\mathsf b$ argument $\mathsf z_{\mathsf b}$ and the block-$\mathsf b$ reply vector $\mathsf r_{\mathsf b}$ are well defined whenever they exist.  Every query counts as one primitive call.

\begin{definition}[Deterministic multi-block zero-respecting algorithms]
\label{def:lbmodel}
A deterministic algorithm $A=(\mathsf{info},(\mathsf Q_\tau)_{\tau\ge1},(\mathsf R_\tau)_{\tau\ge0})$ consists of a parameter map $\mathsf{info}$, query rules $\mathsf Q_\tau$, and output rules $\mathsf R_\tau$.  The map $\mathsf{info}$ assigns to every instance $\mathsf I$ an element $\mathsf{info}(\mathsf I)$ of an arbitrary set, from which $n_x$ can be read off.  The rules may also depend on the fixed data $K,n_y,N_h,y_0,\epsilon,\alpha_*$.  Run on $\mathsf I$, the algorithm generates, for $\tau\ge1$,
\[
 (\mathsf o^\tau,\mathsf z^\tau):=
 \mathsf Q_\tau(\mathsf{info}(\mathsf I);\mathsf r^1,\ldots,\mathsf r^{\tau-1}),
 \qquad
 \mathsf r^\tau:=\mathcal O^{\mathsf I}_{\mathsf o^\tau}(\mathsf z^\tau),
\]
where $\mathsf z^\tau$ has the shape that \eqref{eq:lbsignature} prescribes for $\mathsf o^\tau$.  For $\tau\ge0$, its output after $\tau$ calls is
\[
 x^\tau:=\mathsf R_\tau(\mathsf{info}(\mathsf I);\mathsf r^1,\ldots,\mathsf r^\tau)
 \in\cX_{n_x}.
\]
The revealed supports are
\[
 \mathsf S_x^0:=\varnothing,\qquad
 \mathsf S_j^0:=\operatorname{supp}(y_0)\quad(0\le j\le K),
\]
\[
 \mathsf S_{\mathsf b}^\tau:=\mathsf S_{\mathsf b}^{\tau-1}\cup
 \operatorname{supp}(\mathsf r^\tau_{\mathsf b})
 \qquad(\mathsf b\in\mathsf B,\ \tau\ge1),
\]
with $\operatorname{supp}(\mathsf r^\tau_{\mathsf b}):=\varnothing$ when $\mathsf r^\tau$ has no block-$\mathsf b$ vector.  The algorithm is multi-block zero-respecting if, for every instance $\mathsf I$,
\begin{equation}\label{eq:lbquerysupport}
 \operatorname{supp}(\mathsf z^\tau_{\mathsf b})
 \subseteq\mathsf S_{\mathsf b}^{\tau-1}
 \quad\text{for every $\tau\ge1$ and every block $\mathsf b$ occurring in
 $\mathsf z^\tau$},
\end{equation}
and
\begin{equation}\label{eq:lboutputsupport}
 \operatorname{supp}(x^\tau)\subseteq\mathsf S_x^\tau
 \qquad(\tau\ge0).
\end{equation}
Write $\mathfrak A_{\rm zr}$ for the class of these algorithms.
\end{definition}

\paragraph{PSK as primitive queries.}  Fix wall data $d,\Gamma>0$.  At a triple $(x,p,y)$, the stage-$i$ surrogate \eqref{eq:nscbsurrogate} has
\begin{equation}\label{eq:lbstagecall}
\begin{aligned}
 \nabla_xf_i^C&=\nabla_xg_i,\\
 \nabla_pf_i^C&=\nabla_pg_i-\sum_{l=1}^{N_h}\psi'_{d,\Gamma}(-h_l)\nabla_ph_l,\\
 \nabla_yf_i^C&=\nabla_yg_i-\sum_{l=1}^{N_h}\psi'_{d,\Gamma}(-h_l)\nabla_yh_l,
\end{aligned}
\end{equation}
with $g_i$ and its derivatives at $(x,p,y)$ and $h_l$ and its derivatives at $(p,y)$.  All three partial gradients therefore come from one $(g_i,i)$-query at $(x,p,y)$ and the $N_h$ queries $(h_l,i)$ at $(p,y)$.  We call these $1+N_h$ queries a \emph{stage-$i$ call at $(x,p,y)$}.  The factors $\psi'_{d,\Gamma}(-h_l)$ are functions of scalar replies.

\begin{definition}[PSK as a primitive-query algorithm]\label{def:lbunrolled}
Fix $d,\Gamma>0$, integers $m\ge1$ and $m_0\ge0$, a step $\gamma>0$, and sequences $\rho_t,\alpha_t>0$.  These values, together with $n_x$, form $\mathsf{info}(\mathsf I)$; the map $\mathsf{info}$ is arbitrary.  Set $x_0:=0$ and $y_{0,t}:=y_0$ for $t\ge0$.  The method $A_{\rm PSK}$ issues the following queries, with stages in increasing order of $i$ and inner steps in increasing order of $k$.
\begin{enumerate}[label=\textnormal{(\Alph*)}]
\item[(W)] For $i=1,\ldots,K$: put $y_{i,0}^{(0)}:=y_0$; for $0\le k<m_0$, make a stage-$i$ call at $(x_0,y_{i-1,0},y_{i,0}^{(k)})$ and put $y_{i,0}^{(k+1)}:=y_{i,0}^{(k)}-\gamma\nabla_yf_i^C(x_0,y_{i-1,0},y_{i,0}^{(k)})$; then put $y_{i,0}:=b_{i,0}:=y_{i,0}^{(m_0)}$.
\item[(P$_t$)] For $t=0,1,2,\ldots$:
\begin{enumerate}[label=\textnormal{(\alph*)}]
\item for every $i$, put $b_{i,t}^{(0)}:=b_{i,t}$; for $0\le k<m$, make a stage-$i$ call at $(x_t,y_{i-1,t},b_{i,t}^{(k)})$ and put $b_{i,t}^{(k+1)}:=b_{i,t}^{(k)}-\gamma\nabla_yf_i^C(x_t,y_{i-1,t},b_{i,t}^{(k)})$; then put $b_{i,t+1}:=b_{i,t}^{(m)}$;
\item for every $i$, make stage-$i$ calls at $(x_t,y_{i-1,t},y_{i,t})$ and at $(x_t,y_{i-1,t},b_{i,t+1})$;
\item make one $F$-query at $w_t=(x_t,y_{1,t},\ldots,y_{K,t})$;
\item form $\hnabla_t$ by the block formulas \eqref{icml:blocks} with $f_i=f_i^C$ for every $i$, put $x_{t+1/2}:=x_t-\alpha_t\hnabla_t^x$, make one $\cX$-query at $x_{t+1/2}$, and put
\[
 x_{t+1}:=\PiX_{\cX_{n_x}}(x_{t+1/2}),\qquad
 y_{i,t+1}:=y_{i,t}-\alpha_t\hnabla_t^{y_i}\quad(1\le i\le K).
\]
\end{enumerate}
\end{enumerate}
After $\tau$ calls, the output is $x_t$, where $t$ is the number of completed $\cX$-queries.
\end{definition}

Phase (W) makes $c_{\rm warm}:=(1+N_h)Km_0$ calls, and every phase (P$_t$) makes $c_{\rm ph}:=K(m+2)(1+N_h)+2$ calls.  The quantities in~(d) are available because the block formulas use exactly the partial gradients in \eqref{eq:lbstagecall} returned by~(b), together with the $F$-reply of~(c). \prettyref{alg:psk} acts on the state blocks without projection.  Hence, with the data of \prettyref{sec:nsschedule} (including $m_0$ and $\gamma=1/(2L_{yy})$) and the schedule of \prettyref{sub:band}, $A_{\rm PSK}$ generates exactly the iterates of \prettyref{alg:nswarm} followed by \prettyref{alg:psk}.  The certificate $C_t$ and the choice of $t^\star$ need no further call.

\begin{proposition}[PSK is zero-respecting]\label{prop:lbpskzr}
For every choice of the data in \prettyref{def:lbunrolled}, $A_{\rm PSK}\in\mathfrak A_{\rm zr}$.
\end{proposition}

\begin{proof}
Fix an instance $\mathsf I$ and write $\mathsf S_{\mathsf b}$ for the current revealed supports; they only grow.  A linear combination of vectors, with real coefficients that may depend on scalar replies, is supported in the union of their supports.  We first show that the warm start keeps every register in the right block, and then prove by induction on $t$ the invariant
\[
 (\mathrm J_t)\qquad
 \operatorname{supp}(x_t)\subseteq\mathsf S_x,\qquad
 \operatorname{supp}(y_{i,t})\cup\operatorname{supp}(b_{i,t})
 \subseteq\mathsf S_i\quad(1\le i\le K)
\]
at the start of (P$_t$).  Recall that $\operatorname{supp}(y_0)=\mathsf S_j^0$ for $0\le j\le K$.

\emph{Phase} (W).  We use induction on $i$ and, within stage $i$, on $k$.  A stage-$i$ call at $(x_0,y_{i-1,0},y_{i,0}^{(k)})$ puts its arguments in blocks $x,i-1,i$.  First, $x_0=0$.  Next, $y_{0,0}=y_0$ is supported in $\mathsf S_0^0$, and for $i\ge2$, $y_{i-1,0}=y_{i-1,0}^{(m_0)}$ lies in $\mathsf S_{i-1}$ by the completed stage $i-1$.  Finally, $y_{i,0}^{(0)}=y_0$ is supported in $\mathsf S_i^0$.  Thus \eqref{eq:lbquerysupport} holds for the call.  By \eqref{eq:lbstagecall}, $y_{i,0}^{(k+1)}$ is a linear combination of $y_{i,0}^{(k)}$, the block-$i$ vector $\nabla_yg_i$ of the $(g_i,i)$-reply, and the block-$i$ vectors $\nabla_yh_l$ of the $(h_l,i)$-replies.  Hence it lies in $\mathsf S_i$.  This gives $(\mathrm J_0)$.

\emph{Phase} (P$_t$).  Assume $(\mathrm J_t)$.  The same argument, with $(x_t,y_{i-1,t},b_{i,t}^{(k)})$ in place of $(x_0,y_{i-1,0},y_{i,0}^{(k)})$, shows that every call in~(a) satisfies \eqref{eq:lbquerysupport} and that every $b_{i,t}^{(k)}$, hence $b_{i,t+1}$, lies in $\mathsf S_i$.  The calls in~(b) and the $F$-query in~(c) have arguments $x_t,y_{i-1,t},y_{i,t},b_{i,t+1}$ in blocks $x,i-1,i$, or $x,1,\ldots,K$.  These are already covered.  In~(d), $\hnabla_t^x$ combines $\nabla_xF$, which is block $x$ of the $F$-reply, with vectors $\nabla_xg_i$, which are block $x$ of $(g_i,i)$-replies.  Since $x_{t+1/2}=x_t-\alpha_t\hnabla_t^x$ and $\operatorname{supp}(x_t)\subseteq\mathsf S_x$ by $(\mathrm J_t)$, $\operatorname{supp}(x_{t+1/2})\subseteq\mathsf S_x$, so the $\cX$-query satisfies \eqref{eq:lbquerysupport}, and its block-$x$ reply $x_{t+1}$ lies in $\mathsf S_x$ afterwards.  The direction $\hnabla_t^{y_i}$ combines three kinds of vectors:
\begin{itemize}
\item $\nabla_{y_i}F$, which is block $i$ of the $F$-reply;
\item the stage-$i$ vectors $\nabla_yg_i,\nabla_yh_l$, which are block $i$;
\item for $i<K$, the stage-$(i+1)$ vectors $\nabla_pg_{i+1},\nabla_ph_l$ of \eqref{eq:lbstagecall}, which are block $(i+1)-1=i$.
\end{itemize}
Since $\operatorname{supp}(y_{i,t})\subseteq\mathsf S_i$ by $(\mathrm J_t)$, $y_{i,t+1}=y_{i,t}-\alpha_t\hnabla_t^{y_i}$ is supported in $\mathsf S_i$, and $(\mathrm J_{t+1})$ holds. Every output is either $x_0=0$ or the reply $x_{t+1}$ of a $\cX$-query, so \eqref{eq:lboutputsupport} holds.
\end{proof}

\paragraph{The problem instance class.}  Fix $\Gamma=128$, and use \eqref{eq:lbwallchain} for any tuple $(F,(\SelMat_i,g_i)_{i=1}^K,h_1,\ldots,h_{N_h})$ with the displayed ambient data.

\begin{definition}[The class $\mathcal F_{1,2,2}$]\label{def:lbclass}
For $\Delta,L_F,G_F,L_H^g,L_H^h\ge0$ and $\mu>0$, the class $\mathcal F_{1,2,2}(\Delta,L_F,G_F,\mu,L_H^g,L_H^h)$ consists of all tuples
\[
 \mathsf I=(n_x,F,(\SelMat_i,g_i)_{i=1}^K,h_1,\ldots,h_{N_h})
\]
such that:
\begin{enumerate}[label=\textnormal{(\roman*)}]
\item $n_x\ge1$, $F\in C^1$ is bounded below, $\|\nabla F\|\le G_F$, and $\nabla F$ is globally $L_F$-Lipschitz;
\item for every $1\le i\le K$, $g_i\in C^2$, the map $g_i(x,p,\cdot)$ is $\mu$-strongly convex for all $(x,p)$, and
\[
 \operatorname{Lip}_{\R^{n_x}\times\R^{n_y}\times\R^{n_y}}
 (\nabla^2g_i)\le L_H^g;
\]
\item $h_l\in C^2$, the map $h_l(p,\cdot)$ is convex, and
\[
 \operatorname{Lip}_{\R^{n_y}\times\R^{n_y}}(\nabla^2h_l)
 \le L_H^h\qquad(1\le l\le N_h);
\]
\item for every $0<d\le1$, the exact surrogate chain \eqref{eq:lbwallchain} is well defined and
\[
 \varphi_d^C(0)-
 \inf_{x\in\cX_{n_x}}\varphi_d^C(x)\le\Delta.
\]
\end{enumerate}
The control set of every instance is $\cX:=\cX_{n_x}$.  The fixed ambient data $K,n_y,N_h,y_0,\Gamma$ are suppressed from the notation.  The bounds $\mu,L_H^g,L_H^h$ are uniform over the stages.
\end{definition}

\begin{lemma}[The class satisfies \prettyref{ass:ns}]\label{lem:lbclassns}
Every instance in $\mathcal F_{1,2,2}(\Delta,L_F,G_F,\mu,L_H^g,L_H^h)$, with control set $\cX=\cX_{n_x}$, satisfies \prettyref{ass:ns} with the same $\mu$, $L_F$, and $G_F$.
\end{lemma}

\begin{proof}
For \prettyref{ns:X}, the cube $\cX_{n_x}$ is nonempty, convex, and compact.  For \prettyref{ns:F}, item~(i) of \prettyref{def:lbclass} gives $F\in C^1$, a globally $L_F$-Lipschitz gradient, $\|\nabla F\|\le G_F$, and a finite lower bound $\Flo:=\inf F$.  For \prettyref{ns:G}, continue the stage family by $g_i:=g_K$ for $i>K$; every member is then $C^2$ and $\mu$-strongly convex in $y$ by item~(ii), so $\nabla^2_{yy}g_i\succeq\mu I$.  On a bounded convex set $\mathcal D$, the closure of $\mathcal D$ is compact and each of the finitely many distinct Hessians $\nabla^2g_i$ is continuous, so $\nabla^2_{yy}g_i$, $\partial_a\nabla_yg_i$, $\nabla^2_{aa}g_i$, and $\nabla^2g_i$ are bounded on $\mathcal D$ uniformly in $i\ge1$; this gives $L^g_{yy,\mathcal D}$ (taken at least $\mu$), $L^g_{ay,\mathcal D}$, $L^g_{aa,\mathcal D}$, and $L^g_{f,\mathcal D}$.  The global bound in item~(ii) gives $L^g_{H,\mathcal D}:=L_H^g$, and the $(g_i,i)$-oracle \eqref{eq:lboracles} returns the first derivatives of $g_i$.  For \prettyref{ns:H1}, item~(iii) gives $h_l\in C^2$ with a globally Lipschitz Hessian, and the $(h_l,i)$-oracle returns $h_l,\nabla_ph_l,\nabla_yh_l$.  For \prettyref{ns:H2}, item~(iii) gives convexity of $h_l(p,\cdot)$.
\end{proof}

\begin{definition}[Residual complexity]\label{def:lbcomplexity}
For $A\in\mathfrak A_{\rm zr}$ and an instance $\mathsf I$ of control dimension $n_x$, run the $n_x$-indexed rules of $A$, let $\varphi_\epsilon^C$ be the exact chain loss \eqref{eq:lbwallchain} of $\mathsf I$ at $d=\epsilon$, and define
\[
 \operatorname{Calls}_\epsilon(A,\mathsf I):=\inf\{\tau\ge0:
 \cG_*^{\varphi_\epsilon^C}(x^\tau)\le\epsilon\},
 \qquad \inf\varnothing:=+\infty.
\]
For an algorithm class $\mathfrak A$ and an instance class $\mathcal F$, set
\[
 \operatorname{Calls}_\epsilon(\mathfrak A,\mathcal F):=
 \inf_{A\in\mathfrak A}\sup_{\mathsf I\in\mathcal F}\operatorname{Calls}_\epsilon(A,\mathsf I).
\]
\end{definition}

\paragraph{Frontier and lower bound.}  On the hard instance, each call reveals at most one new control coordinate.

\begin{lemma}[One-coordinate frontier]\label{lem:lbfrontier}
Consider the hard instance $\mathsf I_\epsilon$ of \prettyref{sec:lbinstance}, with $N_{\rm zc}\ge1$.
\begin{enumerate}[label=\textnormal{(\roman*)}]
\item Let $0\le k\le N_{\rm zc}$.  If the block-$x$ argument of a query point is supported in $\{1,\ldots,k\}$, then the block-$x$ vector of its reply, if any, is supported in $\{1,\ldots,\min\{k+1,N_{\rm zc}\}\}$.
\item Every algorithm in $\mathfrak A_{\rm zr}$ satisfies
\[
 \mathsf S_x^\tau\subseteq\{1,\ldots,\min\{\tau,N_{\rm zc}\}\}
 \qquad(\tau\ge0).
\]
Consequently, every output after $\tau<N_{\rm zc}$ calls obeys $x_{N_{\rm zc}}^\tau=0$.
\end{enumerate}
\end{lemma}

\begin{proof}
For~(i), an $(h_l,i)$-query has neither a block-$x$ argument nor a block-$x$ reply.  Since $F=\tanh(v_K)$ does not depend on $x$, an $F$-reply has $\nabla_xF=0$.  The block-$x$ vector of a $(g_i,i)$-reply at $(x,p,y)$, $y=(u,v,\varrho)$, is
\[
 \nabla_xg_i(x,p,y)=\nabla_xg(x,p,y)=-2\zeta(\varrho)\nabla \mathfrak q(x)
 =-2\zeta(\varrho)\frac{A_{\rm zc}}{r_{\rm zc}}
  \nabla\bar f_{N_{\rm zc}}(x/r_{\rm zc}).
\]
Since $\operatorname{supp}(x/r_{\rm zc})=\operatorname{supp}(x)$, \prettyref{lem:lbzerochain}(i) shows that the support of this vector, an element of $\R^{N_{\rm zc}}$, is contained in $\{1,\ldots,k+1\}\cap\{1,\ldots,N_{\rm zc}\}=\{1,\ldots,\min\{k+1,N_{\rm zc}\}\}$.  Projection onto $[-1,1]^{N_{\rm zc}}$ is coordinatewise clipping and maps zero to zero, so its reply is supported in the support of its argument, hence in $\{1,\ldots,k\}\subseteq\{1,\ldots,\min\{k+1,N_{\rm zc}\}\}$.

For~(ii), the claim holds at $\tau=0$ because $\mathsf S_x^0=\varnothing$. Suppose that $\mathsf S_x^{\tau-1}\subseteq\{1,\ldots,\min\{\tau-1,N_{\rm zc}\}\}$.  If query $\tau$ has a block-$x$ argument, then \eqref{eq:lbquerysupport} puts it in this set, and part~(i) with $k=\min\{\tau-1,N_{\rm zc}\}$ gives $\operatorname{supp}(\mathsf r^\tau_x)\subseteq \{1,\ldots,\min\{\tau,N_{\rm zc}\}\}$.  Otherwise it is an $(h_l,i)$-query and $\mathsf r^\tau$ has no block-$x$ vector.  Either way $\mathsf S_x^\tau=\mathsf S_x^{\tau-1}\cup\operatorname{supp}(\mathsf r^\tau_x)\subseteq\{1,\ldots,\min\{\tau,N_{\rm zc}\}\}$.  Finally, \eqref{eq:lboutputsupport} and the bound just proved give $\operatorname{supp}(x^\tau)\subseteq\mathsf S_x^\tau\subseteq\{1,\ldots,\tau\}$ for $\tau<N_{\rm zc}$, so $x^\tau_{N_{\rm zc}}=0$.
\end{proof}

\begin{theorem}[Carmon-style zero-respecting complexity lower bound]
\label{thm:lbmain}
Fix $K\ge2$ and $\alpha_*>0$.  In \prettyref{def:lbclass}, take $n_y=3$, $N_h=2$, $y_0=0$, and $\Gamma=128$.  Set
\[
 L_H^{g,\rm lb}:=6L_2+11.
\]
There are constants $c_K>0$ and $\epsilon_K=\epsilon_K(\alpha_*)>0$ such that, whenever
\[
 \Delta\ge2,\qquad L_F\ge1,\qquad G_F\ge1,\qquad 0<\mu\le1,\qquad
 L_H^g\ge L_H^{g,\rm lb},\qquad L_H^h\ge0,
\]
one has, for every $0<\epsilon\le\epsilon_K$,
\begin{equation}\label{eq:lbmain}
 \operatorname{Calls}_\epsilon\bigl(\mathfrak A_{\rm zr},
 \mathcal F_{1,2,2}(\Delta,L_F,G_F,\mu,L_H^g,L_H^h)\bigr)
 \ge c_K\epsilon^{-(K+3)/4}.
\end{equation}
More precisely, for every $0<\epsilon\le\epsilon_K$, the hard instance $\mathsf I_\epsilon$ is well defined, satisfies $\theta\le\theta_K$ and \eqref{eq:lbqbox}, and belongs to each of these classes; its control dimension satisfies $N_{\rm zc}\ge c_K\epsilon^{-(K+3)/4}$; $\operatorname{Calls}_\epsilon(A,\mathsf I_\epsilon)\ge N_{\rm zc}$ for every $A\in\mathfrak A_{\rm zr}$; and
\begin{equation}\label{eq:lbnostationary}
 x\in\cX,\quad x_{N_{\rm zc}}=0\quad\Longrightarrow\quad
 \cG_*^{\varphi_\epsilon^C}(x)>\epsilon.
\end{equation}
\end{theorem}

\begin{proof}
\emph{Constants.}  Choose $\theta_0\in(0,1]$, depending only on $K$ and $\alpha_*$, such that every $0<\theta\le\theta_0$ satisfies
\[
 \theta\le\theta_K,\qquad
 23\sqrt{c_{\rm len}c_r}\,\theta^{(3K+1)/4}\le1,
 \qquad L_1c_r\theta^{(K+1)/2}\le1,
 \qquad
 \frac{c_{\mathfrak q}}{2}\theta^K\le1,
\]
\[
 r_{\rm zc}\le\frac14,\qquad
 \frac{1-r_{\rm zc}}{\alpha_*}>\theta^2,
 \qquad
 \widetilde N_\theta=\frac{c_{\rm len}}{c_r^3}\theta^{-(K+3)/2}\ge2.
\]
Set
\[
 \epsilon_K:=\theta_0^2,
 \qquad c_K:=\frac{c_{\rm len}}{2c_r^3}.
\]
Fix $0<\epsilon\le\epsilon_K$ and $\theta=\sqrt\epsilon\le\theta_0$.  Then $\lfloor \widetilde N_\theta\rfloor\ge \widetilde N_\theta/2$, so
\[
 N_{\rm zc}\ge\frac{c_{\rm len}}{2c_r^3}
 \theta^{-(K+3)/2}
 =c_K\epsilon^{-(K+3)/4}\ge1,
\]
$\mathsf I_\epsilon$ is well defined, and \eqref{icml:lbqbounds} together with the choice of $\theta_0$ gives \eqref{eq:lbqbox}.

\emph{Class membership.}  We verify membership with constants independent of $\epsilon$ and $N_{\rm zc}$.  Since $F=\tanh(v_K)$, $-1<F<1$, $\|\nabla F\|=\operatorname{sech}^2(v_K)\le1$, and
\[
 \operatorname{Lip}(\nabla F)=\sup_s|\tanh''s|
 =\frac4{3\sqrt3}<1.
\]
The map $\mathfrak q$ is $C^\infty$ and $\zeta\in C^3$, so $g\in C^2$.  By \eqref{eq:lbqbox} and \eqref{eq:lbzetabounds}, for every $x\in\R^{N_{\rm zc}}$ the outgoing $\varrho$-curvature satisfies $2-2\mathfrak q(x)\zeta''(\varrho)\ge17/16$, while the $u$- and $v$-curvatures equal one. Thus $g(x,p,\cdot)$ is $1$-strongly convex.  The affine constraints have zero Hessian.  Because the cubic wall \eqref{eq:nswall} is convex and $-h_l(p,\cdot)$ is affine, every wall composition is convex; hence each surrogate stage energy is $1$-strongly convex and coercive, and every exact chain is well defined.

It remains to control the full Hessian of $g$ globally in both state arguments.  Its only nonconstant part is
\[
 \nabla^2_{(x,\varrho)}\bigl(\varrho^2-2\mathfrak q(x)\zeta(\varrho)\bigr)
 =\begin{bmatrix}
   -2\zeta(\varrho)\nabla^2\mathfrak q(x)&-2\zeta'(\varrho)\nabla \mathfrak q(x)\\
   -2\zeta'(\varrho)\nabla \mathfrak q(x)^\top&2-2\mathfrak q(x)\zeta''(\varrho)
  \end{bmatrix}.
\]
For $z=(x,\varrho)$ and $\widetilde z=(\widetilde x,\widetilde \varrho)$ in $\R^{N_{\rm zc}}\times\R$, let $\Delta_{xx},\Delta_{x\varrho},\Delta_{\varrho\varrho}$ denote the differences of these three blocks.  By \eqref{eq:lbqbox} and~\eqref{eq:lbzetabounds},
\[
\begin{aligned}
 \|\Delta_{xx}\|\le
   (6L_2+2)\|z-\widetilde z\|,
   \qquad
 \|\Delta_{x\varrho}\|\le
   \frac{47}{16}\|z-\widetilde z\|,
   \qquad
 |\Delta_{\varrho\varrho}|\le
   \frac{45}{16}\|z-\widetilde z\|.
\end{aligned}
\]
Indeed, these follow respectively by varying $\nabla^2\mathfrak q$ and $\zeta$, then $\nabla \mathfrak q$ and $\zeta'$, then $\mathfrak q$ and $\zeta''$.  Bounding the operator norm of the block matrix by the sum of its block norms yields
\[
 \operatorname{Lip}_{\R^{N_{\rm zc}}\times\R^3\times\R^3}(\nabla^2g)
 \le6L_2+\frac{171}{16}
 <6L_2+11=L_H^{g,\rm lb}.
\]
Finally, $-1<\varphi_d^C(x)<1$ for every $0<d\le1$, so
\[
 \varphi_d^C(0)-\inf_{x\in\cX}\varphi_d^C(x)<2.
\]
Thus $\mathsf I_\epsilon$, with $\SelMat_i=I_{N_{\rm zc}}$ and $g_i=g$, belongs to every class in the theorem under the displayed parameter inequalities.

\emph{No stationary point with a zero last coordinate.}  Take the width $d=\epsilon=\theta^2$.  Since $|\mathfrak q|\le\frac{c_{\mathfrak q}}2\theta^K$ by \eqref{icml:lbqbounds} and $\theta\le\theta_K$, \prettyref{lem:lbamplifier} applies.  If $x_{N_{\rm zc}}=0$, apply \prettyref{lem:lbzerochain}(iv) to $x/r_{\rm zc}$.  There is an index $j$ with $|x_j|<r_{\rm zc}$ and $|\partial_j\bar f_{N_{\rm zc}}(x/r_{\rm zc})|>1$.  Combining \eqref{eq:lbdepthgain}, $\partial_j\mathfrak q(x)=(A_{\rm zc}/r_{\rm zc})\partial_j\bar f_{N_{\rm zc}}(x/r_{\rm zc})$, and $A_{\rm zc}/r_{\rm zc}=c_r^2\theta^{K+1}$ gives
\begin{equation*}
 |\partial_j\varphi_\epsilon^C(x)|
 \ge\bar c_K\theta^{-(K-1)}\frac{A_{\rm zc}}{r_{\rm zc}}
 =\bar c_Kc_r^2\theta^2=4\epsilon.
\end{equation*}
Projection onto the cube acts coordinatewise, and the norm in \eqref{eq:G} is at least the absolute value of the $j$th coordinate.  Scalar clipping satisfies
\begin{equation*}
 \frac{|\PiX_{[-1,1]}(x_j-\alpha_*\partial_j\varphi_\epsilon^C(x))-x_j|}{\alpha_*}
 \ge\min\left\{|\partial_j\varphi_\epsilon^C(x)|,\frac{1-|x_j|}{\alpha_*}\right\}
 \ge\min\left\{4\epsilon,\frac{1-r_{\rm zc}}{\alpha_*}\right\}.
\end{equation*}
By the choice of $\theta_0$, $(1-r_{\rm zc})/\alpha_*>\epsilon$.  Therefore every $x\in\cX$ with $x_{N_{\rm zc}}=0$ satisfies $\cG_*^{\varphi_\epsilon^C}(x)>\epsilon$, which is \eqref{eq:lbnostationary}.

\emph{Call count.}  By \prettyref{lem:lbfrontier}(ii), every output after fewer than $N_{\rm zc}$ calls has zero last coordinate, so \eqref{eq:lbnostationary} gives $\operatorname{Calls}_\epsilon(A,\mathsf I_\epsilon)\ge N_{\rm zc}\ge c_K\epsilon^{-(K+3)/4}$ for every $A\in\mathfrak A_{\rm zr}$.  Since $\mathsf I_\epsilon$ belongs to the class, taking the supremum over instances and the infimum over algorithms proves \eqref{eq:lbmain}.
\end{proof}

\paragraph{Assumptions of the constant-band theory.}  At every fixed band width, the hard instance satisfies \prettyref{ass:ns} and \ref{ass:nscb}.

\begin{proposition}[Relay membership in the finite-horizon constant-band class]
\label{prop:lbcbbase}
Let $0<\epsilon\le\epsilon_K$, with $\epsilon_K$ from \prettyref{thm:lbmain}, consider the hard instance $\mathsf I_\epsilon$, and fix $0<d\le1$.  In the data of \eqref{eq:nscbdata}, take
\begin{equation}\label{eq:lbcbdata}
\begin{gathered}
 \tau_{\rm cb}:=\frac d2,
 \qquad \sigma_*:=\frac{3d}{2},\qquad b_{\rm cb}:=\frac12,\\
 \Upsilon_{\rm cb}:=48,\qquad \Lambda_{\rm cb}:=\frac{96}{d},
 \qquad \Gamma_{\rm cb}:=128,
\end{gathered}
\end{equation}
and, for every $0\le i\le K$, set
\begin{equation}\label{eq:lbcbsets}
 \mathcal S_i:=\Ball_{16}(0),\qquad
 \mathcal Y_i:=\Ball_{16}(0),\qquad r_{\rm in}:=1.
\end{equation}
Then the nesting conditions \eqref{eq:nscbnested}, the five items of \prettyref{ass:ns} with $\mu=1$, and all four stage-indexed items of \prettyref{ass:nscb} hold for $\cX=[-1,1]^{N_{\rm zc}}$.
\end{proposition}

\begin{proof}
By \prettyref{thm:lbmain}, $\mathsf I_\epsilon\in\mathcal F_{1,2,2}(2,1,1,1,L_H^{g,\rm lb},0)$, with control set $\cX=[-1,1]^{N_{\rm zc}}$, so \prettyref{lem:lbclassns} gives the five items of \prettyref{ass:ns} with $\mu=1$.

The choices in \eqref{eq:lbcbdata} satisfy
\[
 d=\frac{\tau_{\rm cb}+\sigma_*}{2},\qquad
 b_{\rm cb}=\frac{\tau_{\rm cb}}d=\frac12.
\]
Since $\sigma_*-d=d/2$, \eqref{eq:nscbdata} gives
\[
 \frac{\Upsilon_{\rm cb}}{\sigma_*-d}=\frac{96}{d}=\Lambda_{\rm cb},
 \qquad
 \frac{d\Upsilon_{\rm cb}}{3(\sigma_*-d)(1-b_{\rm cb})^2}
 =128=\Gamma_{\rm cb}.
\]
The sets in \eqref{eq:lbcbsets} are compact and convex, satisfy $\mathcal S_i\subseteq\mathcal Y_i$, and form a constant, hence nested, tube. Moreover, $y_0=0\in\mathcal S_0$.  This proves \eqref{eq:nscbnested} and \prettyref{nscb:0}.

Fix a stage $1\le i\le K$ and $(x,p)\in\cX\times\mathcal S_{i-1}$, and put
\[
 \bar y(x,p):=(-4-\sigma_{\rm rel}|\ell(p)|,0,\mathfrak q(x)).
\]
Because $p\in\Ball_{16}(0)$, one has $|\ell(p)|\le16\sqrt2$, and direct substitution gives
\[
 h_+(p,\bar y)=-6-\sigma_{\rm rel}|\ell(p)|
              +\sigma_{\rm rel}\ell(p)\le-6,
\]
\[
 h_-(p,\bar y)=-6-\sigma_{\rm rel}|\ell(p)|
              -\sigma_{\rm rel}\ell(p)\le-6.
\]
Thus $\operatorname{sl}(p,\bar y)\ge6\ge3d/2=\sigma_*$.  Write $s=s_d^C(x,p)$.  The walls vanish at $\bar y$.  Moreover, the last coordinate of $\bar y$ is $\mathfrak q(x)$ and $\zeta'(\mathfrak q(x))=1$ by \eqref{eq:lbqbox} and \eqref{eq:lbzetabounds}, so $\nabla_yg_i(x,p,\bar y)=\nabla_yg(x,p,\bar y)=(-9-\sigma_{\rm rel}|\ell(p)|,0,0)$.  Completing the square gives
\begin{equation*}
 \langle\nabla_yg_i(x,p,\bar y),\bar y-s\rangle-\|\bar y-s\|^2
 \le\frac{(9+\sigma_{\rm rel}|\ell(p)|)^2}{4}<22<48=\Upsilon_{\rm cb}.
\end{equation*}
The stage and pair were arbitrary, so \prettyref{nscb:F} holds with its full stage-indexed quantifiers.

For \prettyref{nscb:XT}, keep $1\le i\le K$ and $(x,p)\in\cX\times\mathcal S_{i-1}$.  The surrogate is $1$-strongly convex ($g$ is $1$-strongly convex in $y$ and each wall composition is convex, as in the proof of \prettyref{thm:lbmain}) and $s$ is stationary, so strong monotonicity at $\bar y$ gives
\[
 \|s-\bar y\|\le\|\nabla_yf_d^C(x,p,\bar y)\|
 =9+\sigma_{\rm rel}|\ell(p)|.
\]
Using $|\mathfrak q(x)|\le1$,
\[
 \|s\|\le14+2\sigma_{\rm rel}|\ell(p)|
 \le14+\frac{32\sqrt2}{100}<15.
\]
Hence $s\in\Ball_{16}(0)=\mathcal S_i$. This conclusion is independent of the slack, so the implication in \prettyref{nscb:XT} holds. Finally, fix $1\le i\le K$ and $(x,p)\in\cX\times\mathcal Y_{i-1}$.  Since $\mathcal Y_{i-1}=\Ball_{16}(0)$, define $\bar y$ as above and set $s:=s_d^C(x,p)$.  The same estimate gives $\|s\|<15$.  Therefore $\Ball_{r_{\rm in}}(s_d^C(x,p))\subseteq\Ball_{16}(0)=\mathcal Y_i$, which proves \prettyref{nscb:NI} with its distinct numerical-tube quantifiers.
\end{proof}

\paragraph{Main result.}  The results above combine into the formal version of \prettyref{thm:lbpsk}.

\begin{formaltheorem}{thm:lbpsk}[Lower bound for multi-block zero-respecting methods]
Fix $K\ge2$ and $\alpha_*>0$, use the ambient data $n_y=3$, $N_h=2$, $y_0=0$, $\Gamma=128$ of \prettyref{thm:lbmain}, let $c_K$ and $\epsilon_K$ be the constants of \prettyref{thm:lbmain}, and let $0<\epsilon\le\epsilon_K$.
\begin{enumerate}[label=\textnormal{(\roman*)}]
\item For every choice of the data in \prettyref{def:lbunrolled}, $A_{\rm PSK}\in\mathfrak A_{\rm zr}$.
\item There is an instance $\mathsf I_\epsilon$ with the following three properties.
\begin{enumerate}[label=\textnormal{(\alph*)}]
\item $\mathsf I_\epsilon\in\mathcal F_{1,2,2}(\Delta,L_F,G_F,\mu,L_H^g,L_H^h)$ whenever $\Delta\ge2$, $L_F\ge1$, $G_F\ge1$, $0<\mu\le1$, $L_H^g\ge L_H^{g,\rm lb}$, and $L_H^h\ge0$; these constants do not depend on $\epsilon$.
\item $\mathsf I_\epsilon$ satisfies \prettyref{ass:ns} and \ref{ass:nscb} with band width $d=\epsilon$ and the data of \prettyref{prop:lbcbbase}.
\item $\operatorname{Calls}_\epsilon(A,\mathsf I_\epsilon)\ge c_K\epsilon^{-(K+3)/4}$ for every $A\in\mathfrak A_{\rm zr}$.
\end{enumerate}
In particular, the class bound \eqref{eq:lbmain} holds.
\end{enumerate}
\end{formaltheorem}

\begin{proof}
Item~(i) is \prettyref{prop:lbpskzr}.  Let $\mathsf I_\epsilon$ be the hard instance of \prettyref{sec:lbinstance}.  \prettyref{thm:lbmain} gives~(a) and~(c).  It also gives $\sqrt\epsilon=\theta\le\theta_K$, and $\theta_K\le1$ by \prettyref{lem:lbamplifier}, so $d=\epsilon\le1$ is admissible in \prettyref{prop:lbcbbase}, which gives~(b).  Since $\mathsf I_\epsilon$ lies in each class in~(a), (c) gives \eqref{eq:lbmain}.
\end{proof}

\section{Motivating instances}\label{sec:instances}

This appendix proves \prettyref{prop:nsrobot} and \ref{prop:dynrobot}: with smoothed constraints, both navigation models of \prettyref{sub:motivating} satisfy \prettyref{ass:nscb}.  We first restate the two models and smooth their constraints.  \prettyref{sec:inst_smooth} collects the properties of the smoothed constraints, and \prettyref{sec:inst_proof} gives the proofs.

\paragraph{Kinematic model.} We restate the kinematic problem of \prettyref{sub:motivating}.  $N_{\rm rob}\ge1$ robots of radius $r_{\rm rob}>0$ move in the plane.  A state stacks the robot positions; as in \prettyref{sec:problem}, $p=(p_1,\dots,p_{N_{\rm rob}})$ is the incoming and $y=(y_1,\dots,y_{N_{\rm rob}})$ the outgoing state of a stage, both in $\R^{n_y}$ with $n_y=2N_{\rm rob}$, and $y_0$ is the initial state.  There are $N_{\rm obs}\ge0$ segment obstacles $S_j=\operatorname{conv}\{o_{j,0},o_{j,1}\}\subset\R^2$ with $o_{j,0}\ne o_{j,1}$.  The edges are the robot pairs $\{r,r'\}\in\cE_{\rm p}$ and the robot--obstacle pairs $(r,j)\in\cE_{\rm o}$, $\cE=\cE_{\rm p}\cup\cE_{\rm o}$, and there are $N_h=|\cE_{\rm p}|+2|\cE_{\rm o}|=N_{\rm rob}(N_{\rm rob}-1)/2+2N_{\rm obs}N_{\rm rob}$ constraints.  The global control $x\in\cX\subset\R^{n_x}$ is shared by the chain.  Stage $i$ reads the selected target positions $\SelMat_ix\in\R^{n_y}$ through a fixed matrix $\SelMat_i$, and, with $\mu>0$, its energy is
\begin{equation}\label{eq:kinenergy}
 g_i(x,p,y)=\frac\mu2\|y-\SelMat_i x\|^2.
\end{equation}
For a nonempty compact convex set $S$, write $\PiX_S$ for the Euclidean projection onto $S$ and $\operatorname{dist}(\cdot,S)$ for the distance to $S$.  At a predecessor $p$ with $p_r\ne p_{r'}$ and $p_r\notin S_j$ for all edges, the exact normals are $n_e(p):=(p_r-p_{r'})/\|p_r-p_{r'}\|$ for $e=\{r,r'\}$ and $n_e(p):=(p_r-\PiX_{S_j}p_r)/\operatorname{dist}(p_r,S_j)$ for $e=(r,j)$, and the exact constraints are
\begin{equation}\label{eq:kinexact}
 \begin{aligned}
 h_e(p,y)&=2r_{\rm rob}-\langle n_e(p),y_r-y_{r'}\rangle,
 &&e=\{r,r'\}\in \cE_{\rm p},\\
 h_{e,k}(p,y)&=2r_{\rm rob}-\langle n_e(p),y_r-o_{j,k}\rangle,
 &&e=(r,j)\in \cE_{\rm o},\ k\in\{0,1\}.
 \end{aligned}
\end{equation}
Equations \eqref{icml:klevel}, \eqref{eq:kinenergy}, \eqref{eq:kinexact}, and the task loss \eqref{icml:navF} define the complete kinematic navigation problem.

\begin{figure}[H]
 \centering
 \includegraphics[width=0.95\textwidth]{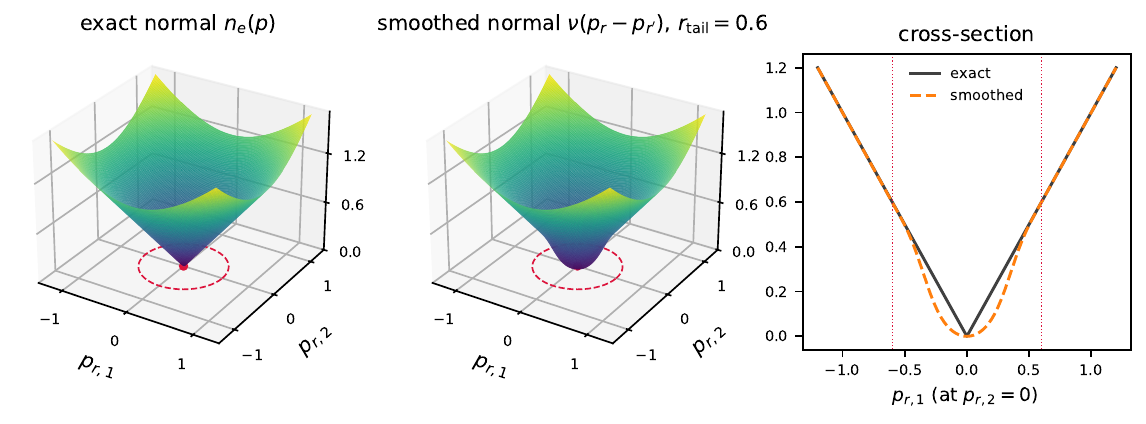}
 \caption{Robot--robot constraint with the other robot at the origin, evaluated at $y=p$: the value $2r_{\rm rob}-h_e(p,p)$ as a function of the robot position $p_r$.  Left: the exact normal $n_e(p)$ gives the cone $\|p_r\|$, with a kink at the origin.  Middle: the smoothed normal $\nu$ of \eqref{icml:nu} ($r_{\rm tail}=0.6$) gives a smooth surface that agrees with the cone outside the dashed circle $\|p_r\|=r_{\rm tail}$.  Right: cross-section along $p_{r,2}=0$.}
 \label{fig:smoothpair}
\end{figure}

\paragraph{Constraint smoothing.} \prettyref{ns:H1} requires every constraint to be $C^2$ in $(p,y)$, with a Hessian that is Lipschitz on bounded sets.  The exact constraints \eqref{eq:kinexact} are affine in $y$, but their normals are not smooth in $p$: the pair normal is undefined when two robots coincide, and the obstacle normal has kinks where the closest point of $S_j$ changes type.  We therefore modify the constraints by smoothing the normals locally.  For robot pairs, fix $r_{\rm tail}>0$ and replace the unit vector $u/\|u\|$ by
\begin{equation}\label{icml:nu}
 \nu(u):=u\left[\|u\|^2+
 \frac{(r_{\rm tail}^2-\|u\|^2)_+^4}{r_{\rm tail}^6}\right]^{-1/2},
\end{equation}
which equals $u/\|u\|$ once $\|u\|\ge r_{\rm tail}$ (\prettyref{lem:nsnormal}).  The smoothed pair constraints are
\begin{equation}\label{eq:nspair}
 h_e(p,y):=2r_{\rm rob}-\langle\nu(p_r-p_{r'}),y_r-y_{r'}\rangle,
 \qquad e=\{r,r'\}\in \cE_{\rm p}.
\end{equation}
\prettyref{fig:smoothpair} illustrates the effect with the other robot fixed at the origin.  At $y=p$, the exact constraint value $2r_{\rm rob}-h_e(p,p)=\|p_r\|$ is a cone with a kink where the two robots coincide.  The smoothed value $\langle\nu(p_r),p_r\rangle$ rounds off the apex inside the disc $\|p_r\|<r_{\rm tail}$ and equals the cone outside it, so the smoothing changes nothing once the robots are $r_{\rm tail}$ apart.

\begin{figure}[H]
 \centering
 \includegraphics[width=0.95\textwidth]{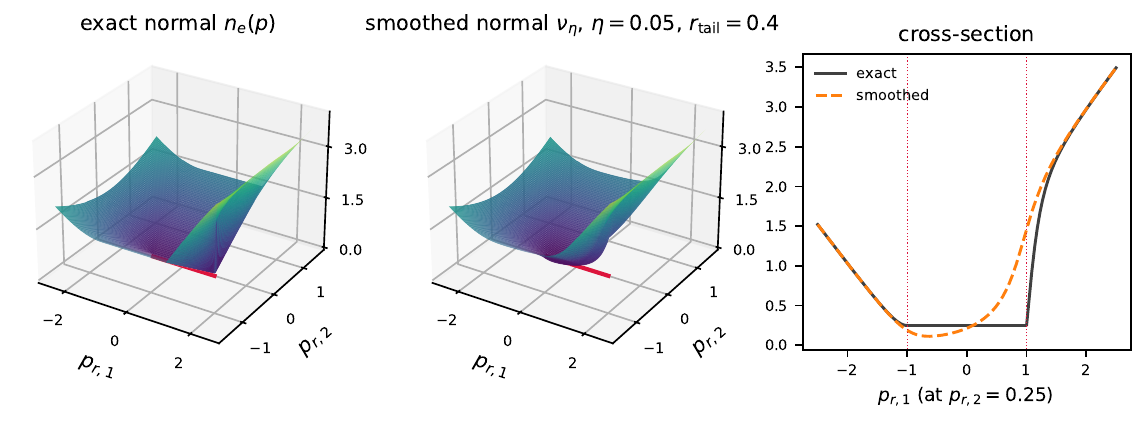}
 \caption{Robot--segment constraint for the segment from $(-1,0)$ to $(1,0)$ (red) and its endpoint $o_{j,0}=(-1,0)$, evaluated at $y=p$: the value $2r_{\rm rob}-h_{e,0}(p,p)$ as a function of the robot position $p_r$.  Left: the exact normal gives a V-shaped valley along the segment and a crease along $p_{r,1}=1$, where the closest point switches from the interior to the endpoint $o_{j,1}$.  Middle: the barrier-smoothed normal $\nu_\eta$ of \eqref{icml:barrier} ($\eta=0.05$, $r_{\rm tail}=0.4$) removes both kinks.  Right: cross-section along $p_{r,2}=0.25$; the exact curve has a kink at $p_{r,1}=1$.}
 \label{fig:smoothobs}
\end{figure}

We next smooth the robot--obstacle constraints.  This is harder: the point of $S_j$ closest to $p_r$ switches between the two endpoints and the interior of the segment, and each switch creates a kink in the exact normal.  We therefore approximate the closest point by a barrier relaxation of the projection onto $S_j$.  Fix $\eta>0$.  For a query point $z\in\R^2$, endpoints $o_0,o_1\in\R^2$ with $o_0\ne o_1$, and $t\in(0,1)$ let
\begin{equation}\label{icml:barrier}
 \begin{aligned}
 \phi(t)&:=\tfrac12\|z-o_0-t(o_1-o_0)\|^2-\eta\log t-\eta\log(1-t),\\
 t_\eta&:=\argmin_{t\in(0,1)}\phi(t),
 \qquad c_\eta:=o_0+t_\eta(o_1-o_0),
 \qquad u_\eta:=z-c_\eta,
 \qquad \nu_\eta(z;o_0,o_1):=\nu(u_\eta).
 \end{aligned}
\end{equation}
The dependence of $\phi$, $t_\eta$, $c_\eta$, and $u_\eta$ on $(z,o_0,o_1)$ is suppressed from the notation.  \prettyref{lem:segsmooth} shows that $t_\eta$ is well defined; the barrier keeps $c_\eta$ inside the segment and makes it a smooth function of $z$.  The smoothed obstacle constraints are
\begin{equation}\label{eq:nsobstacle}
 \begin{aligned}
 h_{e,k}(p,y)&:=2r_{\rm rob}-\langle\nu_\eta(p_r;o_{j,0},o_{j,1}),y_r-o_{j,k}\rangle,
 \qquad e=(r,j)\in \cE_{\rm o},\ k\in\{0,1\}.
 \end{aligned}
\end{equation}
\prettyref{fig:smoothobs} shows the same comparison for the segment from $(-1,0)$ to $(1,0)$ and the constraint of its endpoint $o_{j,0}=(-1,0)$.  With the exact normal, the constraint value at $y=p$ has a V-shaped valley along the segment, where the normal flips sides, and a crease along the line $p_{r,1}=1$, where the closest point switches from the interior of the segment to the endpoint $o_{j,1}$.  The barrier-smoothed normal removes both kinks.  It changes the constraint only near the segment; farther away, $\nu_\eta$ is close to the exact normal (\prettyref{lem:segproj}).

From here on, $h_e$ and $h_{e,k}$ denote the smoothed constraints \eqref{eq:nspair} and \eqref{eq:nsobstacle}.  There are still $N_h$ of them, so $N_h=0$ exactly when $\cE=\varnothing$.  Their slack is $\operatorname{sl}(p,y):=\min_l\{-h_l(p,y)\}$; a minimum over the empty set is $+\infty$, so $\operatorname{sl}\equiv+\infty$ if $N_h=0$, as in \prettyref{sec:nsmodel}.  The slack depends on $\eta$ and $r_{\rm tail}$.  Equations \eqref{icml:klevel}, \eqref{eq:kinenergy}, \eqref{eq:nspair}, \eqref{eq:nsobstacle}, and \eqref{icml:navF} define the complete kinematic navigation problem with smoothed constraints, which satisfies the smoothness requirements \prettyref{ns:H1} and~\prettyref{ns:H2} (\prettyref{cor:segH}).  Our experiments use these smoothed constraints.

\paragraph{Dynamic model.} Let $n_q=2N_{\rm rob}$ and write the predecessor and outgoing states as $p=(p^{\pos},p^{\vel}),y=(y^{\pos},y^{\vel})\in\R^{2n_q}$.  The global control $x\in\R^{n_x}$ is shared, while stage $i$ uses the selected stacked force $\SelMat_i x\in\R^{n_q}$.  Fix a mass $m_{\rm mass}>0$ and a time step $\Delta t>0$.  The selected dynamic stage family is given in \eqref{eq:dynobjective}; here $n_y=2n_q$ and $n_{u,i}=n_q$.  The second term enforces consistency between outgoing velocity and displacement; the first term penalizes the change from the incoming velocity.  The constraints act on the position blocks.  The dynamic model lifts the smoothed constraints to the doubled state through the position blocks,
\begin{equation}\label{eq:dynwalls}
 h_l^{\rm dyn}(p,y):=h_l(p^{\pos},y^{\pos}),
\end{equation}
where $h_l$ runs over the $N_h$ constraints $h_e$ and $h_{e,k}$, and its slack is $\operatorname{sl}(p,y):=\min_l\{-h_l^{\rm dyn}(p,y)\}=\min_l\{-h_l(p^{\pos},y^{\pos})\}$.  The constraints $h_l^{\rm dyn}$ are independent of $(p^{\vel},y^{\vel})$.  Equations \eqref{icml:klevel}, \eqref{eq:dynobjective}, \eqref{eq:dynwalls}, and \eqref{icml:navF} define the complete dynamic navigation problem with smoothed constraints.

\subsection{Properties of the Smoothed Constraints}\label{sec:inst_smooth}

This subsection collects the properties of the smoothed constraints that the proofs use.  We first record some basic facts about convex sets.  For a nonempty compact convex set $S\subset\R^{n_S}$, define
\[
 \sigma_S(n):=\max_{v\in S}\langle n,v\rangle,
 \qquad
 \Sigma_S(n;y):=\langle n,y\rangle-\sigma_S(n)=\min_{v\in S}\langle n,y-v\rangle,
 \qquad
 N_S(y):=\frac{y-\PiX_Sy}{\operatorname{dist}(y,S)}\quad(y\notin S).
\]

\begin{lemma}[Convex-set facts]\label{lem:segcvx}
Let $S\subset\R^{n_S}$ be nonempty, compact, and convex, let $R_S:=\max_{v\in S}\|v\|$, and let $n,n',y\in\R^{n_S}$.
\begin{enumerate}
\item[(i)] $|\Sigma_S(n;y)-\Sigma_S(n';y)|\le\|n-n'\|\max_{v\in S}\|y-v\|\le\|n-n'\|(\|y\|+R_S)$, so $\sigma_S$ is $R_S$-Lipschitz.  If $S=\operatorname{conv}\{v_0,v_1\}$, then $\Sigma_S(n;y)=\min_k\langle n,y-v_k\rangle$ and $\max_{v\in S}\|y-v\|=\max_k\|y-v_k\|$.
\item[(ii)] If $\|n\|\le1$, then $\Sigma_S(n;y)\le \operatorname{dist}(y,S)$.  If $y\notin S$, then $\sigma_S(N_S(y))=\langle N_S(y),\PiX_Sy\rangle$ and $\Sigma_S(N_S(y);y)=\operatorname{dist}(y,S)$.
\item[(iii)] $\operatorname{dist}(\cdot,S)\in C^1(\R^{n_S}\setminus S)$, $\nabla\operatorname{dist}(\cdot,S)=N_S$, and $N_S$ is continuous on $\R^{n_S}\setminus S$.
\item[(iv)] If $S=\operatorname{conv}\{v_0,v_1\}$, then $y\mapsto(\PiX_Sy,\operatorname{dist}(y,S))$ is semialgebraic.
\end{enumerate}
\end{lemma}

\begin{proof}
(i) For $v\in S$, the map $n\mapsto\langle n,y-v\rangle$ is $\|y-v\|$-Lipschitz, and $\Sigma_S(\cdot;y)$ is their pointwise minimum.  Also $\sigma_S=-\Sigma_S(\cdot;0)$.  On $\operatorname{conv}\{v_0,v_1\}$, an affine function attains its minimum and its maximum at a vertex, and the convex function $\|y-\cdot\|$ attains its maximum at a vertex.

(ii) For $v\in S$, $\Sigma_S(n;y)\le\langle n,y-v\rangle\le\|y-v\|$; take the infimum over $v$.  The projection inequality $\langle y-\PiX_Sy,v-\PiX_Sy\rangle\le0$ gives $\langle N_S(y),v\rangle\le\langle N_S(y),\PiX_Sy\rangle$, with equality at $v=\PiX_Sy$.  So $\Sigma_S(N_S(y);y)=\langle N_S(y),y-\PiX_Sy\rangle=\operatorname{dist}(y,S)$.

(iii) Put $\psi:=\frac12\operatorname{dist}(\cdot,S)^2$ and $g(y):=y-\PiX_Sy$.  Then $\psi(y+h)\le\psi(y)+\langle g(y),h\rangle+\frac12\|h\|^2$ and $\psi(y)\le\psi(y+h)-\langle g(y+h),h\rangle+\frac12\|h\|^2$.  Since $\PiX_S$ is $1$-Lipschitz, $\langle g(y+h)-g(y),h\rangle\ge0$, so $|\psi(y+h)-\psi(y)-\langle g(y),h\rangle|\le\frac12\|h\|^2$.  Thus $\psi\in C^1$ with gradient $g$, which is continuous because $\PiX_S$ is $1$-Lipschitz, and $\operatorname{dist}(\cdot,S)=\sqrt{2\psi}$ is $C^1$ where $\psi>0$, with gradient $g/\operatorname{dist}(y,S)=N_S$.

(iv) $(v^{\rm proj},d^{\rm proj})=(\PiX_Sy,\operatorname{dist}(y,S))$ holds exactly when there is $s\in[0,1]$ with $v^{\rm proj}=v_0+s(v_1-v_0)$, $d^{\rm proj}\ge0$ and $(d^{\rm proj})^2=\|y-v^{\rm proj}\|^2$, and $(d^{\rm proj})^2\le\|y-v_0-s'(v_1-v_0)\|^2$ for all $s'\in[0,1]$.  This is a first-order formula, so the graph is semialgebraic by the Tarski--Seidenberg principle~\citep{bochnak1998real}.
\end{proof}

We next show that the smoothing indeed produces smooth constraints: the barrier projection is smooth in the query point (\prettyref{lem:segsmooth}), the mollified normal $\nu$ is smooth (\prettyref{lem:nsnormal}), and hence so are the constraints (\prettyref{cor:segH}).

\begin{lemma}[Smoothness of the barrier projection]\label{lem:segsmooth}
Let $o_0\ne o_1$, $\delta_o:=o_1-o_0$ and $L:=\|\delta_o\|>0$.
\begin{enumerate}
\item[(i)] $\phi''(t)=L^2+\eta t^{-2}+\eta(1-t)^{-2}\ge L^2+8\eta$ on $(0,1)$, and $t_\eta$ is the unique zero of
\begin{equation}\label{eq:segF}
 \phi'(t)=-\langle z-o_0-t\delta_o,\delta_o\rangle-\frac\eta t+\frac\eta{1-t}.
\end{equation}
\item[(ii)] For fixed $(o_0,o_1)$, the maps $z\mapsto t_\eta,c_\eta,u_\eta$ are $C^\infty$ on $\R^2$.
\end{enumerate}
\end{lemma}

\begin{proof}
(i) Differentiate twice.  On $(0,1)$, $t^{-2}+(1-t)^{-2}$ is convex and symmetric about $1/2$, where it equals $8$; so it is at least $8$.  Hence $\phi$ is strictly convex.  Also $\phi\ge-\eta\log t-\eta\log(1-t)\to+\infty$ at both ends.  So the minimizer exists, is unique, and is the unique zero of $\phi'$.

(ii) The right side of \eqref{eq:segF} is real-analytic in $(t,z)\in(0,1)\times\R^2$, and its $t$-derivative is $\phi''\ge8\eta>0$.  Fix $z$.  The implicit function theorem gives a neighborhood $\mathcal V$ of $z$ and a $C^\infty$ map $\tau:\mathcal V\to(0,1)$ that solves \eqref{eq:segF} on $\mathcal V$ and equals $t_\eta$ at $z$.  By uniqueness in (i), $\tau=t_\eta$ on $\mathcal V$.  So $t_\eta$ is $C^\infty$, and so are $c_\eta=o_0+t_\eta\delta_o$ and $u_\eta=z-c_\eta$.
\end{proof}

\begin{lemma}[Exact-tail normal]\label{lem:nsnormal}
The map $\nu$ of \eqref{icml:nu} is $C^3$ on $\R^2$, $\|\nu(u)\|\le1$, and $\nu(u)=u/\|u\|$ for $\|u\|\ge r_{\rm tail}$.
\end{lemma}

\begin{proof}
Put $s=\|u\|^2$ and $\upsilon(s):=s+(r_{\rm tail}^2-s)_+^4/r_{\rm tail}^6$ for $s\in\R$.  On $0\le s\le r_{\rm tail}^2$, $\upsilon(s)/r_{\rm tail}^2=s/r_{\rm tail}^2+(1-s/r_{\rm tail}^2)^4\ge1/2$, since this convex function of $s/r_{\rm tail}^2\in[0,1]$ is smallest at $1-s/r_{\rm tail}^2=4^{-1/3}$, where it equals $1-\tfrac34 4^{-1/3}>1/2$; for $s\ge r_{\rm tail}^2$, $\upsilon(s)=s$.  Thus $\upsilon$ is positive on $[0,\infty)$, and by continuity on an open interval $\mathcal I\supset[0,\infty)$.  The scalar function $(r_{\rm tail}^2-s)_+^4$ is $C^3$ on $\R$, so $\upsilon$ is $C^3$ and $a(s):=\upsilon(s)^{-1/2}$ is $C^3$ on $\mathcal I$.  Hence $\nu(u)=a(\|u\|^2)u$ is $C^3$ on $\R^2$.  Since $\upsilon(s)\ge s$, $\|\nu(u)\|=\sqrt{s/\upsilon(s)}\le1$.  When $\|u\|\ge r_{\rm tail}$, $\upsilon(s)=s$, which gives the exact unit normal.
\end{proof}

\begin{corollary}[Regularity of the smoothed constraints]\label{cor:segH}
Fix $r_{\rm tail}>0$ and $\eta>0$.  Every constraint $h_e$ of \eqref{eq:nspair}, $h_{e,k}$ of \eqref{eq:nsobstacle}, and every lifted constraint $h_l^{\rm dyn}$ of \eqref{eq:dynwalls} is $C^3$ and affine in the outgoing state.  Consequently, both models satisfy \prettyref{ns:H1} and~\prettyref{ns:H2}.
\end{corollary}

\begin{proof}
By \prettyref{lem:nsnormal}, $\nu$ is $C^3$.  For fixed endpoints, $z\mapsto u_\eta$ is $C^\infty$ by \prettyref{lem:segsmooth}(ii), so $\nu_\eta(\cdot;o_{j,0},o_{j,1})=\nu\circ u_\eta$ is $C^3$.  Each $h_e(p,y)$ and $h_{e,k}(p,y)$ is a constant minus the inner product of a $C^3$ function of $p$ with an affine function of $y$, so it is $C^3$ on $\R^{n_y}\times\R^{n_y}$ and affine in $y$.  Each $h_l^{\rm dyn}$ is such a function composed with the linear map $(p,y)\mapsto(p^{\pos},y^{\pos})$, so it has the same properties.  The third derivatives of a $C^3$ function are continuous, hence bounded on the closure of every bounded set; by the mean-value inequality, its Hessian is Lipschitz on every bounded convex set.  The oracle of \prettyref{ns:H1} is available: $t_\eta$ is the unique zero of the increasing function $\phi'$ on $(0,1)$, differentiating $\phi'(t_\eta)=0$ in $z$ gives $\nabla_zt_\eta=\delta_o/\phi''(t_\eta)$ with $\delta_o:=o_{j,1}-o_{j,0}$, so $D_zu_\eta=I-\delta_o\delta_o^\top/\phi''(t_\eta)$, and the chain rule gives $\nabla_ph_l$ and $\nabla_yh_l$ for every constraint.  This gives \prettyref{ns:H1}.  A function affine in the outgoing state is convex in it, which gives \prettyref{ns:H2}.
\end{proof}

We then show that the smoothed constraints stay good approximations of the exact ones.  By \prettyref{lem:nsnormal}, $\nu$ is exact once $\|u\|\ge r_{\rm tail}$.  The next lemma shows that the barrier projection is close to the exact projection and that, far from the segment, its normal is close to the exact normal.

\begin{lemma}[Smoothed segment projection]\label{lem:segproj}
Let $o_0\ne o_1$, $S:=\operatorname{conv}\{o_0,o_1\}$, and $\mathfrak d:=\operatorname{dist}(z,S)$.
\begin{enumerate}
\item[(i)] \emph{One-sided exactness.} $\langle u_\eta,z-v\rangle\ge\|u_\eta\|^2-\eta$ for every $v\in S$.
\item[(ii)] \emph{Closeness.} $\mathfrak d\le\|u_\eta\|$ and $\|c_\eta-\PiX_Sz\|^2\le\eta$.
\item[(iii)] \emph{Normal.}  If $\mathfrak d>0$, then $\bigl\|u_\eta/\|u_\eta\|-N_S(z)\bigr\|\le\sqrt\eta/\mathfrak d$. If $\mathfrak d\ge r_{\rm tail}$, then $\nu_\eta=u_\eta/\|u_\eta\|$ and $\|\nu_\eta-N_S(z)\|\le\sqrt\eta/\mathfrak d$.
\item[(iv)] \emph{Stability.} If $\mathfrak d\ge r_{\rm tail}$, then $|\Sigma_S(\nu_\eta;y)-\Sigma_S(N_S(z);y)|\le(\sqrt\eta/\mathfrak d)\max_k\|y-o_k\|$ for every $y$.
\end{enumerate}
\end{lemma}

\begin{proof}
(i) Put $\delta_o:=o_1-o_0$, so $u_\eta=z-o_0-t_\eta\delta_o$.  Since \eqref{eq:segF} vanishes at $t_\eta$, $\langle u_\eta,\delta_o\rangle=\eta\bigl(\frac1{1-t_\eta}-\frac1{t_\eta}\bigr)$.  Inserting this into $z-o_0=u_\eta+t_\eta\delta_o$ and $z-o_1=u_\eta-(1-t_\eta)\delta_o$ gives
\[
 \langle u_\eta,z-o_0\rangle=\|u_\eta\|^2-\eta+\frac{\eta t_\eta}{1-t_\eta},
 \qquad
 \langle u_\eta,z-o_1\rangle=\|u_\eta\|^2-\eta+\frac{\eta(1-t_\eta)}{t_\eta},
\]
and both are at least $\|u_\eta\|^2-\eta$.  For $v=o_0+s\delta_o$ with $s\in[0,1]$, the value $\langle u_\eta,z-v\rangle$ is affine in $s$, so its minimum over $S$ is attained at $s\in\{0,1\}$.

(ii) $c_\eta\in S$, so $\mathfrak d\le\|u_\eta\|$.  The projection inequality $\langle z-\PiX_Sz,c_\eta-\PiX_Sz\rangle\le0$ gives $\|u_\eta\|^2=\|z-c_\eta\|^2\ge \mathfrak d^2+\|c_\eta-\PiX_Sz\|^2$.  Since $c_\eta-\PiX_Sz=(z-\PiX_Sz)-u_\eta$, item (i) at $v=\PiX_Sz$ gives $\|c_\eta-\PiX_Sz\|^2=\mathfrak d^2-2\langle u_\eta,z-\PiX_Sz\rangle+\|u_\eta\|^2\le \mathfrak d^2-\|u_\eta\|^2+2\eta$.  Adding the two inequalities gives $\|c_\eta-\PiX_Sz\|^2\le\eta$.

(iii) Let $\mathfrak d>0$; then $u_\eta\ne0$ by (ii).  Since $z-\PiX_Sz=\mathfrak d\,N_S(z)$ with $\|N_S(z)\|=1$, expand $\|c_\eta-\PiX_Sz\|^2=\|\mathfrak d\,N_S(z)-u_\eta\|^2$ and use $\bigl\|u_\eta/\|u_\eta\|-N_S(z)\bigr\|^2=2-2\langle u_\eta/\|u_\eta\|,N_S(z)\rangle$:
\[
 \eta\ge\|c_\eta-\PiX_Sz\|^2=(\|u_\eta\|-\mathfrak d)^2+\|u_\eta\|\mathfrak d\,\bigl\|u_\eta/\|u_\eta\|-N_S(z)\bigr\|^2
 \ge \mathfrak d^2\bigl\|u_\eta/\|u_\eta\|-N_S(z)\bigr\|^2,
\]
where the first inequality is (ii) and the last uses $\|u_\eta\|\ge\mathfrak d$.  If $\mathfrak d\ge r_{\rm tail}$, then $\|u_\eta\|\ge r_{\rm tail}$, and \prettyref{lem:nsnormal} gives $\nu_\eta=\nu(u_\eta)=u_\eta/\|u_\eta\|$, so the second bound is the first one.

(iv) By \prettyref{lem:segcvx}(i) for the segment $S$, $|\Sigma_S(n;y)-\Sigma_S(n';y)|\le\|n-n'\|\max_k\|y-o_k\|$.  Take $n=\nu_\eta$ and $n'=N_S(z)$, and use (iii).
\end{proof}

Finally, satisfying the smoothed constraints implies satisfying the true, nonsmooth clearance constraints.  For a pair edge $e=\{r,r'\}$ put $\mathbf B_ey:=y_r-y_{r'}$ and $S_e:=\{0\}$; for an obstacle edge $e=(r,j)$ put $\mathbf B_ey:=y_r$ and $S_e:=S_j$.  The true edge slacks and the true clearance of a state $y$ are
\[
 \operatorname{clr}_e(y):=\operatorname{dist}(\mathbf B_ey,S_e)-2r_{\rm rob},
 \qquad
 \operatorname{clr}(y):=\min_{e\in \cE}\operatorname{clr}_e(y).
\]
So $\operatorname{clr}\equiv+\infty$ if $\cE=\varnothing$.  For a pair edge, $\operatorname{dist}(\mathbf B_ey,S_e)=\|y_r-y_{r'}\|$.  The clearance $\operatorname{clr}$ and the slack $\operatorname{sl}$ measure different constraints.  $\operatorname{clr}(y)$ is the clearance of $y$ with respect to the exact, nonsmooth constraints \eqref{eq:kinexact}: where the exact normals are defined at $y$, it equals the smallest margin of these constraints at $p=y$, by \prettyref{lem:segcvx}(i)--(ii).  $\operatorname{sl}(p,y)$ is the clearance of $y$ with respect to the smoothed constraints \eqref{eq:nspair} and \eqref{eq:nsobstacle}, whose normals are evaluated at the predecessor $p$.  The next lemma shows that $\operatorname{sl}(p,y)\le\operatorname{clr}(y)$, so the smoothed constraints are never looser than the true clearance.

\begin{lemma}[True-clearance transfer]\label{lem:segxt}
For all kinematic states $p,y\in\R^{n_y}$ and $\tau\in\R$, $\operatorname{sl}(p,y)\ge\tau$ implies $\operatorname{clr}(y)\ge\tau$.
\end{lemma}

\begin{proof}
If $\cE=\varnothing$, then $\operatorname{clr}\equiv+\infty$.  Otherwise $\operatorname{sl}(p,y)\ge\tau$ means $-h_e(p,y)\ge\tau$ and $-h_{e,k}(p,y)\ge\tau$ for every constraint.  For a pair edge, $\|\nu\|\le1$ (\prettyref{lem:nsnormal}) and Cauchy--Schwarz give $\|y_r-y_{r'}\|\ge\langle\nu(p_r-p_{r'}),y_r-y_{r'}\rangle\ge2r_{\rm rob}+\tau$.  For an obstacle edge $(r,j)$, the two constraints give $\Sigma_{S_j}(\nu_\eta(p_r;o_{j,0},o_{j,1});y_r)\ge2r_{\rm rob}+\tau$ by \prettyref{lem:segcvx}(i). Since $\|\nu_\eta\|\le1$ (\prettyref{lem:nsnormal}), \prettyref{lem:segcvx}(ii) gives $\operatorname{dist}(y_r,S_j)\ge2r_{\rm rob}+\tau$.  So $\operatorname{clr}_e(y)\ge\tau$ for every $e\in\cE$, and taking the minimum over $e\in\cE$ gives $\operatorname{clr}(y)\ge\tau$.
\end{proof}

\subsection{Proofs}\label{sec:inst_proof}

\begin{figure}[ht]
 \centering
 \resizebox{0.45\textwidth}{!}{\begingroup
\definecolor{jamRob}{HTML}{2166AC}
\definecolor{jamGap}{HTML}{B2182B}
\begin{tikzpicture}[
  wall/.style={line width=1.4pt,line cap=round,black!72},
  rob/.style={draw=jamRob,fill=jamRob!25,line width=0.5pt},
  gap/.style={{Stealth[length=2.6pt]}-{Stealth[length=2.6pt]},line width=0.5pt,jamGap},
  lab/.style={font=\scriptsize,inner sep=1pt},
  x=0.8cm,y=0.8cm
]
  \foreach \dx in {0,4.4} {
    \draw[wall] (\dx,0) -- (\dx+3,0) -- (\dx+1.5,2.6) -- cycle;
    \fill[black!45] (\dx+1.5,0.866) circle (0.9pt);
  }
  \node[lab,text=black!55,anchor=north] at (1.5,0.84) {center};

  \filldraw[rob] (1.376,1.285) circle (0.15);
  \node[lab,anchor=west] at (1.55,1.285) {$p$};
  \draw[gap] (1.246,1.36) -- (0.9,1.56);
  \node[lab,text=jamGap,anchor=east] at (0.86,1.62) {$\tau_{\rm cb}$};

  \draw[-{Implies},double,line width=0.5pt,double distance=1.4pt] (3.25,1.1) -- (4.15,1.1);

  \filldraw[rob] (5.85,0.95) circle (0.15);
  \node[lab,anchor=west] at (6.02,0.95) {$\bar y$};
  \draw[gap] (5.72,1.025) -- (5.174,1.34);
  \node[lab,text=jamGap,anchor=east] at (5.12,1.47) {$\sigma_*>\tau_{\rm cb}$};
\end{tikzpicture}
\endgroup}
 \caption{Left: robot $p$ among three segments with clearance $\tau_{\rm cb}$ (red arrow: the distance to the nearest segment minus $2r_{\rm rob}$).  The only jam is at the center, where the three wall normals balance and no motion improves the clearance.  Right: since $\tau_{\rm cb}$ lies below this jam level, $p$ can move to a comparison point $\bar y$ with slack $\sigma_*>\tau_{\rm cb}$; \prettyref{nscb:F} holds once $\tau_{\rm cb}$ avoids such jams.}
 \label{fig:jam}
\end{figure}
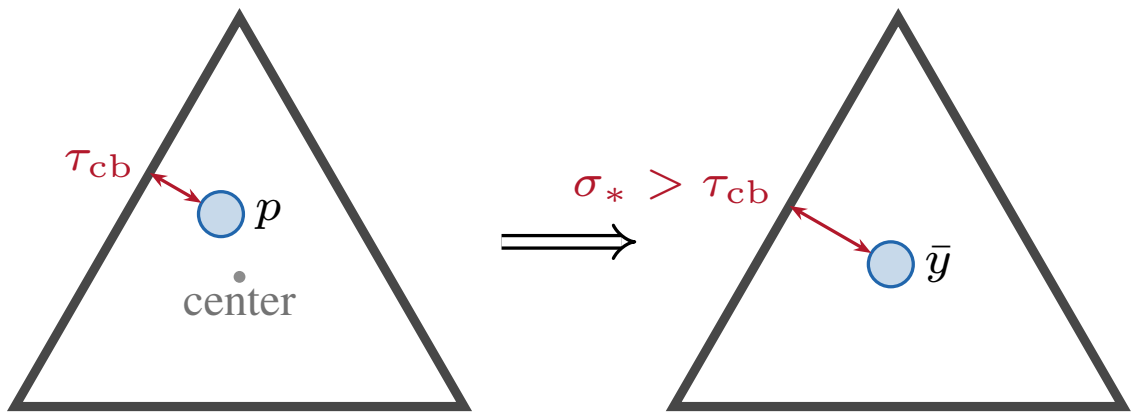

\paragraph{Idea of the proof.} The key item of \prettyref{ass:nscb} is \prettyref{nscb:F}: from every core state, the robots must reach, within a bounded force, a state whose smoothed slack exceeds the band width.  This uses only the geometry of the robots and segments, so \prettyref{sec:inst_aux} proves it once for both models (\prettyref{lem:nsjam}).  \prettyref{fig:jam} shows the obstruction.  A robot $p$ sits inside a triangle of three segments.  From most positions it can step away from its nearest walls, so some $\bar y$ has a larger slack than the clearance of $p$.  At the center, however, the three wall normals balance: every motion that moves the robot away from one wall moves it toward another, so no $\bar y$ increases all three slacks at once.  We call such a state a jam, and its clearance a jam value \eqref{eq:nsjamfinite}.  In general, linear-programming duality \eqref{eq:nsjamdual} shows that the best common slack reachable from $p$ is a stress-weighted average of the edge clearances, so it is at least the clearance of $p$, and larger unless $p$ is a jam.  \prettyref{lem:nsjam} shows that there are only finitely many jam values and puts the core level $\tau_{\rm cb}$ below all of them, so no core state is a jam.  A compactness argument over the normal tuples, which also covers their limits, turns this into a uniform slack gap $\Delta_{\rm jam}$ and a uniform bound $Z_{\rm jam}$ on the comparison points; small smoothing of the constraints keeps half of the gap.  \prettyref{sec:inst_main} then proves the two propositions: strong convexity keeps every exact step near its unconstrained center, which gives the core and tube radii, and \prettyref{lem:segxt} gives \prettyref{nscb:XT}.

\subsubsection{Auxiliary lemmas}\label{sec:inst_aux}

Put $R_O:=\max_{j,k}\|o_{j,k}\|$, with $R_O:=0$ if $N_{\rm obs}=0$; if $N_{\rm obs}\ge1$, then $R_O>0$, since $o_{j,0}\ne o_{j,1}$.  Let $R_e:=\max_{v\in S_e}\|v\|\le R_O$.  Each edge uses one convex set $S_e$, so segments may intersect, touch, overlap, or be collinear; how different segments meet plays no role.  Note that $\cE=\varnothing$ exactly when $N_{\rm rob}=1$ and $N_{\rm obs}=0$.

Put $\mathcal R_{>0}:=\{y:\operatorname{clr}(y)>0\}$ and, for $c>0$, $\mathcal R_{\ge c}:=\{y:\operatorname{clr}(y)\ge c\}\subseteq\mathcal R_{>0}$.  In the statement of \prettyref{lem:nsjam} and in Steps 1 and 4--7 of its proof, the points of these sets are predecessors $p$, whose normals define the constraints of the next step; Step 2 uses only the clearance and writes its points as $y$.  For $y\in\mathcal R_{>0}$ and $e\in \cE$, the clearance gradient and, at a predecessor $p\in\mathcal R_{>0}$, the exact edge normal of \eqref{eq:kinexact} are
\[
 \nabla\operatorname{clr}_e(y):=\mathbf B_e^\top N_{S_e}(\mathbf B_ey),
 \qquad n_e(p)=N_{S_e}(\mathbf B_ep),
\]
so $\nabla\operatorname{clr}_e(p)=\mathbf B_e^\top n_e(p)$.
By \prettyref{lem:segcvx}(iii), $\operatorname{clr}_e$ is $C^1$ on $\mathcal R_{>0}$ with gradient $\nabla\operatorname{clr}_e$.  The jam-value set is
\begin{equation}\label{eq:nsjamfinite}
\resizeEq[\dimexpr\linewidth-5em\relax]{\mathcal J:=\Bigl\{c>0:{}\exists y,\lambda^{\rm jam}\in\R_+^{\cE},\ \operatorname{clr}(y)=c,\ \sum_e\lambda_e^{\rm jam}=1,
 \sum_e\lambda_e^{\rm jam}\nabla\operatorname{clr}_e(y)=0,\ \lambda_e^{\rm jam}(\operatorname{clr}_e(y)-c)=0\ (e\in\cE)\Bigr\}.}
\end{equation}
It contains the positive clearance levels at which the active contact normals can balance, so that no motion increases all active slacks at once.

\begin{lemma}[Uniform jam regeneration for segment obstacles]\label{lem:nsjam}
Let $\sigma_0>0$.  The set $\mathcal J\subset(0,\infty)$ is finite.  Define
\[
 c_+:=\begin{cases}
 \min(\mathcal J\cap(0,\sigma_0]),&
 \mathcal J\cap(0,\sigma_0]\ne\varnothing,\\
 \sigma_0,&\mathcal J\cap(0,\sigma_0]=\varnothing,
 \end{cases}
 \qquad \tau_{\rm cb}:=\frac{c_+}2,
 \qquad \mathfrak d_0:=2r_{\rm rob}+\tau_{\rm cb}.
\]
Then $0<\tau_{\rm cb}\le\sigma_0/2$.  There are $\Delta_{\rm jam}>0$ and $Z_{\rm jam}\in[0,\infty)$, depending only on $N_{\rm rob},r_{\rm rob},(S_j)_j$, and $\tau_{\rm cb}$, with the following properties.
\begin{enumerate}
\item[(i)] Every $p$ with $\operatorname{clr}(p)\ge\tau_{\rm cb}$ admits $\bar y=\bar y(p)$ with
\[
 \|\bar y\|\le Z_{\rm jam},
 \qquad
 \min_{e\in \cE}\bigl\{\Sigma_{S_e}(n_e(p);\mathbf B_e\bar y)-2r_{\rm rob}\bigr\}\ge\tau_{\rm cb}+\Delta_{\rm jam}.
\]
\item[(ii)] Let $0<r_{\rm tail}\le \mathfrak d_0$, and put
\[
 \eta_0:=\Bigl(\frac{\Delta_{\rm jam}\mathfrak d_0}{2(Z_{\rm jam}+R_O)}\Bigr)^2
 \quad(N_{\rm obs}\ge1),
 \qquad \eta_0:=+\infty\quad(N_{\rm obs}=0).
\]
Then $\operatorname{sl}(p,\bar y(p))\ge\tau_{\rm cb}+\Delta_{\rm jam}/2$ for every finite $\eta\in(0,\eta_0]$ and every $p$ with $\operatorname{clr}(p)\ge\tau_{\rm cb}$.
\end{enumerate}
\end{lemma}

\begin{proof}
The proof uses only standard facts of real algebraic and o-minimal geometry: the Tarski--Seidenberg principle, the curve-selection lemma, and one-variable Puiseux expansions~\citep{bochnak1998real}, together with $C^1$ cell decomposition and definable choice~\citep{vandendries1998tame}.  It also uses linear-programming duality~\citep{schrijver1986theory}.

\smallskip\noindent\emph{Step 0 (Empty edge set).} Let $\cE=\varnothing$.  No $\lambda^{\rm jam}\in\R^\varnothing$ has $\sum_e\lambda^{\rm jam}_e=1$, so $\mathcal J=\varnothing$, $c_+=\sigma_0$, and $\tau_{\rm cb}=\sigma_0/2$.  The minima in (i) and (ii) are over empty sets, hence $+\infty$, and $N_{\rm obs}=0$.  So the lemma holds with $\bar y(p):=0$, $Z_{\rm jam}:=0$, and $\Delta_{\rm jam}:=1/2$.  From now on $\cE\ne\varnothing$.

\smallskip\noindent\emph{Step 1 (Notation).} On $\mathcal R_{>0}$, $\operatorname{dist}(\mathbf B_ep,S_e)=\operatorname{clr}_e(p)+2r_{\rm rob}>0$, so $n_e(p)$ is a unit vector.  The \emph{active set} of a state $y$ is $\operatorname{act}(y):=\{e\in \cE:\operatorname{clr}_e(y)=\operatorname{clr}(y)\}$.  The \emph{normal tuple} of $p\in\mathcal R_{>0}$ is $n(p):=(n_e(p))_{e\in \cE}$, a point of the compact product of unit spheres
\[
 \mathcal N:=\{(n_e)_{e\in \cE}:n_e\in\R^2,\ \|n_e\|=1\}.
\]
For $n=(n_e)_{e\in\cE}$ with $n_e\in\R^2$ and $\|n_e\|\le1$ (in particular for $n\in\mathcal N$), $\lambda^{\rm jam}\in\R^{\cE}$ and $\bar y\in\R^{n_y}$ put
\[
 \widehat{\operatorname{clr}}_e[n](\bar y):=\Sigma_{S_e}(n_e;\mathbf B_e\bar y)-2r_{\rm rob}=(\mathbf B_e^\top n_e)^\top \bar y-c_e(n),
 \qquad c_e(n):=2r_{\rm rob}+\sigma_{S_e}(n_e),
\]
\[
 A(n)^\top\lambda^{\rm jam}:=\sum_{e\in \cE}\lambda^{\rm jam}_e\mathbf B_e^\top n_e,
 \qquad \Theta(n,\lambda^{\rm jam}):=-\sum_{e\in \cE}\lambda^{\rm jam}_ec_e(n).
\]
A \emph{stress} at $n$ is a vector $\lambda^{\rm jam}\in\R_+^{\cE}$ with $\sum_e\lambda^{\rm jam}_e=1$ and $A(n)^\top\lambda^{\rm jam}=0$.  Its support is $\cE_+:=\{e:\lambda^{\rm jam}_e>0\}$.  By \prettyref{lem:segcvx}(ii),
\begin{equation}\label{eq:segexact}
 \widehat{\operatorname{clr}}_e[n(p)](p)=\operatorname{clr}_e(p)\qquad(p\in\mathcal R_{>0}).
\end{equation}
The frozen outgoing slack at a predecessor $p\in\mathcal R_{>0}$ is $\widehat{\operatorname{clr}}{}_e^{\,p}:=\widehat{\operatorname{clr}}_e[n(p)]$.  It is affine, and $\widehat{\operatorname{clr}}{}_e^{\,p}(\bar y)=\operatorname{clr}_e(p)+\nabla\operatorname{clr}_e(p)^\top(\bar y-p)$.  By \prettyref{lem:segcvx}(i), $|c_e(n)-c_e(n')|\le R_e\|n_e-n_e'\|$, so $\Theta$ is continuous.

We use the following identity repeatedly.  Let $\lambda^{\rm jam}$ be a stress at $n$, and let $p\in\mathcal R_{>0}$ satisfy $n_e(p)=n_e$ for every $e\in \cE_+$.  By \eqref{eq:segexact}, $\operatorname{clr}_e(p)=(\mathbf B_e^\top n_e)^\top p-c_e(n)$ for $e\in \cE_+$.  Summing against $\lambda^{\rm jam}$ and using $A(n)^\top\lambda^{\rm jam}=0$ gives
\begin{equation}\label{eq:nsjamTheta}
 \Theta(n,\lambda^{\rm jam})=\sum_{e\in \cE_+}\lambda^{\rm jam}_e\operatorname{clr}_e(p)\ \ge\ \operatorname{clr}(p).
\end{equation}
The inequality holds because $\operatorname{clr}_e\ge\operatorname{clr}$ and $\sum_e\lambda^{\rm jam}_e=1$.  In \eqref{eq:nsjamfinite}, the stress condition $\sum_e\lambda_e^{\rm jam}\nabla\operatorname{clr}_e(y)=0$ says that $\lambda^{\rm jam}$ is a stress at $n(y)$, and the complementarity condition $\lambda_e^{\rm jam}(\operatorname{clr}_e(y)-c)=0$ says that $\cE_+\subseteq \operatorname{act}(y)$.

\smallskip\noindent\emph{Step 2 ($\mathcal J$ is finite).} Let $\operatorname{Crit}$ be the set of $y\in\mathcal R_{>0}$ for which some $\lambda^{\rm jam}$ makes $(y,\lambda^{\rm jam})$ satisfy \eqref{eq:nsjamfinite} with $c=\operatorname{clr}(y)$.  Then $\mathcal J=\operatorname{clr}(\operatorname{Crit})$.  Introduce the auxiliary tuples $(v^{\rm proj},d^{\rm proj}):=((v_e^{\rm proj})_{e\in\cE},(d_e^{\rm proj})_{e\in\cE})$, where $(v_e^{\rm proj},d_e^{\rm proj})=(\PiX_{S_e}(\mathbf B_ey),\operatorname{dist}(\mathbf B_ey,S_e))$.  By \prettyref{lem:segcvx}(iv), their graph is semialgebraic.  Then $\operatorname{clr}_e=d_e^{\rm proj}-2r_{\rm rob}$ and $\nabla\operatorname{clr}_e(y)=\mathbf B_e^\top(\mathbf B_ey-v_e^{\rm proj})/d_e^{\rm proj}$.  Hence every condition in \eqref{eq:nsjamfinite} and in the definitions of $\mathcal R_{>0}$ and $\operatorname{act}(y)$ is a first-order formula in $(y,v^{\rm proj},d^{\rm proj},\lambda^{\rm jam})$.  Projecting out $(v^{\rm proj},d^{\rm proj},\lambda^{\rm jam})$ by the Tarski--Seidenberg principle~\citep{bochnak1998real} shows that the following are semialgebraic: $\mathcal R_{>0}$, $\operatorname{Crit}$, the graphs of $\operatorname{clr}$ and $n(\cdot)$ on $\mathcal R_{>0}$, and the active-set strata
\[
 \mathcal R_{>0}^{\mathcal I}:=\{y\in\mathcal R_{>0}:\operatorname{act}(y)=\mathcal I\}\qquad(\varnothing\ne\mathcal I\subseteq \cE).
\]

By the $C^1$ cell-decomposition theorem~\citep[Ch.~7]{vandendries1998tame}, $\mathcal R_{>0}$ is a finite disjoint union of $C^1$ cells with two properties: each cell lies in a single $\mathcal R_{>0}^{\mathcal I}$, and each cell is either contained in $\operatorname{Crit}$ or disjoint from it. Each cell is a connected $C^1$ submanifold of $\R^{n_y}$.  Let $\mathcal Q\subseteq\mathcal R_{>0}^{\mathcal I}\cap\operatorname{Crit}$ be a cell and $y\in\mathcal Q$.  On $\mathcal Q$, every $e\in\mathcal I$ satisfies $\operatorname{clr}_e=\operatorname{clr}$, so $\operatorname{clr}|_{\mathcal Q}$ is $C^1$ and
\[
 \mathrm d(\operatorname{clr}|_{\mathcal Q})(y)[v_{\rm tan}]=\nabla\operatorname{clr}_e(y)^\top v_{\rm tan}
 \qquad(v_{\rm tan}\in T_y\mathcal Q,\ e\in\mathcal I).
\]
Choose $\lambda^{\rm jam}$ for $y$ from \eqref{eq:nsjamfinite}; then $\cE_+\subseteq\operatorname{act}(y)=\mathcal I$.  Multiplying the last display by $\lambda^{\rm jam}_e$, summing over $e\in\mathcal I$, and using $\sum_{e\in\mathcal I}\lambda^{\rm jam}_e=\sum_e\lambda^{\rm jam}_e=1$ (since $\cE_+\subseteq\mathcal I$) gives
\[
 \mathrm d(\operatorname{clr}|_{\mathcal Q})(y)[v_{\rm tan}]=\Big(\sum_e\lambda^{\rm jam}_e\nabla\operatorname{clr}_e(y)\Big)^{\!\top}v_{\rm tan}=0.
\]
So $\operatorname{clr}$ is constant on the connected manifold $\mathcal Q$.  There are finitely many cells, so $\mathcal J$ is finite.

\smallskip\noindent\emph{Step 3 (Linear-programming duality).} Fix $n\in\mathcal N$.  Each $\widehat{\operatorname{clr}}_e[n]$ is affine.  Consider the linear program
\[
 \sup_{\bar y\in\R^{n_y},\ \tau\in\R}\ \tau
 \quad\text{subject to}\quad \widehat{\operatorname{clr}}_e[n](\bar y)\ge\tau\quad(e\in \cE).
\]
It is feasible, and its Lagrangian with multipliers $\lambda^{\rm jam}\in\R_+^{\cE}$ is
\[
 \tau\Big(1-\sum_e\lambda^{\rm jam}_e\Big)+(A(n)^\top\lambda^{\rm jam})^\top \bar y+\Theta(n,\lambda^{\rm jam}).
\]
Its supremum over $(\bar y,\tau)$ is finite exactly when $\lambda^{\rm jam}$ is a stress at $n$, and then it equals $\Theta(n,\lambda^{\rm jam})$.  By the linear-programming duality theorem~\citep[Ch.~7]{schrijver1986theory},
\begin{equation}\label{eq:nsjamdual}
 \sup_{\bar y}\min_{e\in \cE}\widehat{\operatorname{clr}}_e[n](\bar y)
 =\min\{\Theta(n,\lambda^{\rm jam}):\lambda^{\rm jam}\text{ is a stress at }n\},
\end{equation}
where both sides are $+\infty$ if no stress at $n$ exists.

\smallskip\noindent\emph{Step 4 (Limit tuples create no new values).} For $c>0$, let $\overline{\mathcal N}_c$ be the closure of $n(\mathcal R_{\ge c})$ in $\mathcal N$; it is compact.  We claim:

\emph{if $c>0$, $n\in \overline{\mathcal N}_c$, and $\lambda^{\rm jam}$ is a stress at $n$, then there is $\hat p\in\mathcal R_{\ge c}$ with $n_e(\hat p)=n_e$ for every $e\in \cE_+$.}

Granting the claim, \eqref{eq:nsjamTheta} gives
\begin{equation}\label{eq:nsjamlimit}
 \Theta(n,\lambda^{\rm jam})=\sum_{e\in \cE_+}\lambda^{\rm jam}_e\operatorname{clr}_e(\hat p)\ge\operatorname{clr}(\hat p)\ge c.
\end{equation}
Suppose $\Theta(n,\lambda^{\rm jam})=c$.  Then $\operatorname{clr}_e(\hat p)=c=\operatorname{clr}(\hat p)$ for $e\in \cE_+$, and $\sum_e\lambda^{\rm jam}_e\nabla\operatorname{clr}_e(\hat p)=A(n)^\top\lambda^{\rm jam}=0$.  So $(\hat p,\lambda^{\rm jam})$ satisfies \eqref{eq:nsjamfinite}, and $c\in\mathcal J$.  We now prove the claim in four parts.

\emph{(4a) Curve and lift.}  The set $n(\mathcal R_{\ge c})$ is semialgebraic.  By the curve-selection lemma~\citep{bochnak1998real}, there is a continuous semialgebraic arc $\bar n:[0,1)\to\mathcal N$ with $\bar n(0)=n$ and $\bar n(\vartheta)\in n(\mathcal R_{\ge c})$ for $0<\vartheta<1$.  Definable choice~\citep[Ch.~6]{vandendries1998tame} gives a semialgebraic map $\vartheta\mapsto p(\vartheta)\in\mathcal R_{\ge c}$ with $n(p(\vartheta))=\bar n(\vartheta)$.

\emph{(4b) Puiseux expansion.}  Each coordinate of $p(\vartheta)$ is a semialgebraic function of one variable, so, after shrinking the arc to some $[0,\vartheta_1)$, it has a convergent Puiseux expansion near $\vartheta=0$~\citep{bochnak1998real}.  Collecting the negative exponents gives
\[
 p(\vartheta)=\sum_{\iota=1}^J\vartheta^{-\alpha_\iota}v_\iota+p^\infty+o(1)
 \quad(\vartheta\downarrow0),
 \qquad \alpha_1>\cdots>\alpha_J>0,\quad v_\iota\ne0,
\]
with $J\ge0$.  For every edge put $\iota_e:=\min\{\iota:\mathbf B_ev_\iota\ne0\}$, with $\iota_e:=\infty$ if there is no such $\iota$.
\begin{itemize}
\item If $\iota_e<\infty$, then $\|\PiX_{S_e}(\mathbf B_ep(\vartheta))\|\le R_e$, so $\vartheta^{\alpha_{\iota_e}}\bigl(\mathbf B_ep(\vartheta)-\PiX_{S_e}(\mathbf B_ep(\vartheta))\bigr)\to \mathbf B_ev_{\iota_e}$.  Hence $n_e=\mathbf B_ev_{\iota_e}/\|\mathbf B_ev_{\iota_e}\|$.
\item If $\iota_e=\infty$, then $\mathbf B_ep(\vartheta)\to \mathbf B_ep^\infty$ and $\operatorname{dist}(\mathbf B_ep^\infty,S_e)\ge2r_{\rm rob}+c>0$.  By \prettyref{lem:segcvx}(iii), $n_e=N_{S_e}(\mathbf B_ep^\infty)$.
\end{itemize}

\emph{(4c) Stressed edges do not grow.}  We show by induction on $\iota$ that $\mathbf B_ev_\iota=0$ for every $\iota\le J$ and every $e\in \cE_+$.  Suppose this holds for all smaller indices.  Pairing the stress equation with $v_\iota$ gives
\[
 0=(A(n)^\top\lambda^{\rm jam})^\top v_\iota=\sum_{e\in \cE_+}\lambda^{\rm jam}_e\,n_e^\top \mathbf B_ev_\iota.
\]
For $e\in \cE_+$, each term is nonnegative.  If $\mathbf B_ev_\iota=0$, the term is $0$.  If $\mathbf B_ev_\iota\ne0$, then $\iota_e=\iota$ by the induction hypothesis, and (4b) gives $n_e^\top \mathbf B_ev_\iota=\|\mathbf B_ev_\iota\|>0$. So all terms are nonnegative with zero sum.  Since $\lambda^{\rm jam}_e>0$ on $\cE_+$, this forces $\mathbf B_ev_\iota=0$ for every $e\in \cE_+$.

\emph{(4d) Truncation.}  For small $\vartheta'>0$ put $\hat p:=p^\infty+\sum_{\iota=1}^J\vartheta'^{-\alpha_\iota}v_\iota$.
\begin{itemize}
\item If $\iota_e=\infty$, which includes every $e\in \cE_+$, then $\mathbf B_e\hat p=\mathbf B_ep^\infty$.  So $\operatorname{clr}_e(\hat p)\ge c$, and $n_e(\hat p)=n_e$ for $e\in \cE_+$.
\item If $\iota_e<\infty$, then $\|\mathbf B_e\hat p\|\ge\vartheta'^{-\alpha_{\iota_e}}\|\mathbf B_ev_{\iota_e}\|/2$ for small $\vartheta'$, and $\operatorname{clr}_e(\hat p)\ge\|\mathbf B_e\hat p\|-R_e-2r_{\rm rob}\to\infty$.
\end{itemize}
Since $\cE$ is finite, one $\vartheta'$ serves every edge, and then $\hat p\in\mathcal R_{\ge c}$.  This proves the claim.

\smallskip\noindent\emph{Step 5 (The level is not a jam value, and the gap is positive).} By Step~2, $c_+$ is well defined.  If $\mathcal J\cap(0,\sigma_0]\ne\varnothing$, then $\tau_{\rm cb}=c_+/2\in(0,\sigma_0/2]$ lies below the least element of $\mathcal J$.  Otherwise $\tau_{\rm cb}=\sigma_0/2$, and $(0,\sigma_0]$ contains no element of $\mathcal J$.  In both cases $0<\tau_{\rm cb}\le\sigma_0/2$ and $\tau_{\rm cb}\notin\mathcal J$.  Now let
\[
 \mathcal K:=\{(n,\lambda^{\rm jam}):n\in \overline{\mathcal N}_{\tau_{\rm cb}},\ \lambda^{\rm jam}\text{ is a stress at }n\}.
\]
The stress conditions are closed, so $\mathcal K$ is compact, and $\Theta$ is continuous on it.  By \eqref{eq:nsjamlimit}, $\Theta\ge\tau_{\rm cb}$ on $\mathcal K$.  Equality at some point would put $\tau_{\rm cb}$ in $\mathcal J$.  Hence
\[
 \delta_{\rm jam}:=\min\Big\{1,\min_{(n,\lambda^{\rm jam})\in \mathcal K}\big(\Theta(n,\lambda^{\rm jam})-\tau_{\rm cb}\big)\Big\}>0,
\]
with $\delta_{\rm jam}:=1$ if $\mathcal K=\varnothing$.

\smallskip\noindent\emph{Step 6 (Canonical comparison points; proof of (i)).} Put $\Delta_{\rm jam}:=\delta_{\rm jam}/2>0$.  If $\mathcal R_{\ge\tau_{\rm cb}}=\varnothing$, then (i) holds with $Z_{\rm jam}:=0$, so assume $\mathcal R_{\ge\tau_{\rm cb}}\ne\varnothing$.  Fix $n\in \overline{\mathcal N}_{\tau_{\rm cb}}$.  The right side of \eqref{eq:nsjamdual} is at least $\tau_{\rm cb}+\delta_{\rm jam}$, so there is $\bar y_n$ with $\widehat{\operatorname{clr}}_e[n](\bar y_n)\ge\tau_{\rm cb}+\frac34\delta_{\rm jam}$ for all $e\in \cE$.  By \prettyref{lem:segcvx}(i),
\[
 \mathcal O_n:=\{n'\in\mathcal N:\min_e\widehat{\operatorname{clr}}_e[n'](\bar y_n)>\tau_{\rm cb}+\Delta_{\rm jam}\}
\]
is an open neighborhood of $n$.  By compactness, finitely many $\mathcal O_{n_1},\dots,\mathcal O_{n_{N_{\rm cov}}}$ cover $\overline{\mathcal N}_{\tau_{\rm cb}}$.  For $p\in\mathcal R_{\ge\tau_{\rm cb}}$ let
\[
 \Pi^p:=\{\bar y:\widehat{\operatorname{clr}}{}_e^{\,p}(\bar y)\ge\tau_{\rm cb}+\Delta_{\rm jam}\ (e\in \cE)\}.
\]
It is closed and convex.  Some $i$ has $n(p)\in\mathcal O_{n_i}$, so $\bar y_{n_i}\in\Pi^p$.  Let $\bar y(p)$ be the unique point of $\Pi^p$ of least norm, and put $Z_{\rm jam}:=\sup\{\|\bar y(p)\|:p\in\mathcal R_{\ge\tau_{\rm cb}}\}$, with $Z_{\rm jam}:=0$ if $\mathcal R_{\ge\tau_{\rm cb}}=\varnothing$.  Then $\|\bar y(p)\|\le\|\bar y_{n_i}\|$, so $Z_{\rm jam}\le\max_i\|\bar y_{n_i}\|<\infty$.  None of $\Delta_{\rm jam}$, $Z_{\rm jam}$, or $\bar y(\cdot)$ depends on the cover.  This proves (i).

\smallskip\noindent\emph{Step 7 (Smoothed constraints; proof of (ii)).} Let $\operatorname{clr}(p)\ge\tau_{\rm cb}$ and $\bar y:=\bar y(p)$.  For a pair edge $e=\{r,r'\}$, $\|p_r-p_{r'}\|\ge \mathfrak d_0\ge r_{\rm tail}$, so \prettyref{lem:nsnormal} gives $\nu(p_r-p_{r'})=n_e(p)$ and $-h_e(p,\bar y)=\widehat{\operatorname{clr}}{}_e^{\,p}(\bar y)$.  If $N_{\rm obs}=0$, there are no other edges.  For an obstacle edge $e=(r,j)$, $\mathfrak d_e:=\operatorname{dist}(p_r,S_j)\ge \mathfrak d_0\ge r_{\rm tail}$, and by \prettyref{lem:segcvx}(i)
\[
 \min_k\{-h_{e,k}(p,\bar y)\}=\Sigma_{S_j}(\nu_\eta(p_r;o_{j,0},o_{j,1});\bar y_r)-2r_{\rm rob}.
\]
Here $n_e(p)=N_{S_j}(p_r)$, so \prettyref{lem:segproj}(iv) at $z=p_r$ with $\max_k\|\bar y_r-o_{j,k}\|\le Z_{\rm jam}+R_O$ gives
\[
 \min_k\{-h_{e,k}(p,\bar y)\}\ge\widehat{\operatorname{clr}}{}_e^{\,p}(\bar y)-\frac{\sqrt\eta}{\mathfrak d_0}(Z_{\rm jam}+R_O).
\]
With (i) and $\eta\le\eta_0$, $\operatorname{sl}(p,\bar y)\ge\tau_{\rm cb}+\Delta_{\rm jam}-\sqrt\eta(Z_{\rm jam}+R_O)/\mathfrak d_0 \ge\tau_{\rm cb}+\Delta_{\rm jam}/2$.
\end{proof}

\subsubsection{Proofs of the main results}\label{sec:inst_main}

We first prove the formal \prettyref{prop:nsrobot}.  It uses the selector norms $\varsigma_i:=\|\SelMat_i\|$ and the envelopes
\[
 \bar\varsigma_K:=\max_{1\le i\le K}\varsigma_i,
 \qquad R_X:=\max_{x\in\mathcal X}\|x\|,
 \qquad R_{\SelMat,K}:=\max_{1\le i\le K}\max_{x\in\mathcal X}\|\SelMat_i x\|
 \le \bar\varsigma_K R_X.
\]
Across a horizon-indexed family, define the suprema $\bar\varsigma_\infty:=\sup_{K\geq1}\bar\varsigma_K$ and $R_{\SelMat,\infty}:=\sup_{K\geq1}R_{\SelMat,K}$, which are the two quantities assumed finite in the informal \prettyref{prop:nsrobot}.  For prefixes of one nested stage sequence these are equivalently the suprema over $i\geq1$.  Standard $0$--$1$ coordinate selectors have $\varsigma_i\leq1$, hence $\bar\varsigma_\infty\leq1$; a fixed per-block control bound gives $R_{\SelMat,\infty}<\infty$ even when the stacked global-control radius grows with $K$.  For a fixed horizon only $\bar\varsigma_K$ and $R_{\SelMat,K}$ are needed.

\begin{formalproposition}{prop:nsrobot}[Kinematic navigation]
Let \prettyref{ns:X} and~\prettyref{ns:F} hold, let the stages use the energy \eqref{eq:kinenergy} with fixed selectors $\SelMat_i$, and suppose that $\bar\varsigma_\infty<\infty$, $R_{\SelMat,\infty}<\infty$, and $\operatorname{clr}(y_0)\geq\sigma_0>0$.  Let $\tau_{\rm cb},\Delta_{\rm jam},Z_{\rm jam},\eta_0$ be the jam data of \prettyref{lem:nsjam}, set $r_{\rm tail}=r_{\rm rob}$, and take $0<\eta\leq\eta_0$.  With
\[
 \sigma_*:=\tau_{\rm cb}+\frac{\Delta_{\rm jam}}2,
 \qquad d:=\tau_{\rm cb}+\frac{\Delta_{\rm jam}}4,
 \qquad r_{\rm in}:=1,
 \qquad
 \Upsilon_{\rm kin}:=\frac{(\mu(Z_{\rm jam}+R_{\SelMat,\infty})+1)^2}{4\mu},
\]
$\Lambda_{\rm cb},\Gamma_{\rm cb}$ from \eqref{eq:nscbdata} with $\Upsilon_{\rm cb}=\Upsilon_{\rm kin}$, and the radius $R_{\rm kin}$ of \eqref{eq:nsrobotR}, \prettyref{ass:nscb} holds for every $K\geq1$ with
\[
 \mathcal S_i=\mathcal R_{\ge\tau_{\rm cb}}
 \cap\Ball_{R_{\rm kin}}(0),\qquad
 \mathcal Y_i=\Ball_{R_{\rm kin}}(0)
 \qquad(0\le i\le K).
\]
The cores, tubes, wall data, and one-stage base derivative envelopes are independent of $K$ and $\eta$.  Without the hypothesis $\bar\varsigma_\infty,R_{\SelMat,\infty}<\infty$, the same conclusion holds for each fixed $K$ with $\bar\varsigma_\infty$ and $R_{\SelMat,\infty}$ replaced by $\bar\varsigma_K$ and $R_{\SelMat,K}$; these data may then depend on $K$, but not on $\eta$.
\end{formalproposition}

\begin{proof}
\emph{\prettyref{ass:ns}.}  \prettyref{ns:X} and~\prettyref{ns:F} are assumed.  For every stage,
\[
 \nabla^2_{yy}g_i=\mu I,
 \qquad
 \partial_a\nabla_yg_i=[-\mu\SelMat_i\ \ 0].
\]
Valid per-stage constants are
\[
 \mu_i=\mu,
 \quad L_{yy,i}^g=\mu,
 \quad L_{ay,i}^g=\mu\|\SelMat_i\|,
 \quad L_{aa,i}^g=\mu\|\SelMat_i\|^2,
 \quad L_{f,i}^g=\mu(1+\|\SelMat_i\|^2),
 \quad L_{H,i}^g=0.
\]
Using the horizon-indexed supremum $\bar\varsigma_\infty<\infty$ recalled at the start of \prettyref{sec:inst_main} gives finite common bounds over every $i\geq1$; for standard coordinate selectors one may use $L^g_{yy,\mathcal D}=L^g_{ay,\mathcal D}=L^g_{aa,\mathcal D}=\mu$, $L^g_{f,\mathcal D}=2\mu$, and $L^g_{H,\mathcal D}=0$ on every bounded $\mathcal D$. Thus \prettyref{ns:G} holds with the required all-stage envelopes. \prettyref{cor:segH} gives \prettyref{ns:H1} and~\prettyref{ns:H2}.  The fixed data satisfy $0<\tau_{\rm cb}<\sigma_*$ and $d=(\tau_{\rm cb}+\sigma_*)/2$, as \prettyref{ass:nscb} requires.

\emph{Center bound for every predecessor.}  Fix a stage $i\ge1$, a control $x\in\cX$, and any predecessor $p\in\R^{n_y}$; no clearance of $p$ is used.  By \eqref{eq:nswall}, $\psi_{d,\Gamma_{\rm cb}}$ is $C^1$, convex, nonincreasing, and nonnegative, and each $-h_l(p,\cdot)$ is affine.  So $f_i^C(x,p,\cdot)$ is $\mu$-strongly convex and has a unique minimizer $s:=s_i^C(x,p)$.  We bound $s$ by comparing it with the target $u_i:=\SelMat_ix$, at which $g_i(x,p,u_i)=0$ and $\|u_i\|\le R_{\SelMat,\infty}$.

\smallskip\noindent\emph{Step 1 (constraint values at the target).}  For a pair edge $e=\{r,r'\}$,
\[
 \begin{aligned}
 h_e(p,u_i)&=2r_{\rm rob}-\langle\nu(p_r-p_{r'}),(u_i)_r-(u_i)_{r'}\rangle
 \le2r_{\rm rob}+\|(u_i)_r-(u_i)_{r'}\|\\
 &\le2r_{\rm rob}+\sqrt2\bigl(\|(u_i)_r\|^2+\|(u_i)_{r'}\|^2\bigr)^{1/2}
 \le2r_{\rm rob}+\sqrt2R_{\SelMat,\infty}.
 \end{aligned}
\]
The first inequality is Cauchy--Schwarz with $\|\nu\|\le1$ (\prettyref{lem:nsnormal}), the second is $\|a-b\|\le\|a\|+\|b\|\le\sqrt2(\|a\|^2+\|b\|^2)^{1/2}$, and the third uses $\|(u_i)_r\|^2+\|(u_i)_{r'}\|^2\le\|u_i\|^2$.  For an obstacle edge $e=(r,j)$ and $k\in\{0,1\}$, the same steps with $\|\nu_\eta\|\le1$ (since $\nu_\eta=\nu(u_\eta)$) and $\|o_{j,k}\|\le R_O$ give
\[
 h_{e,k}(p,u_i)=2r_{\rm rob}-\langle\nu_\eta(p_r;o_{j,0},o_{j,1}),(u_i)_r-o_{j,k}\rangle
 \le2r_{\rm rob}+\|(u_i)_r\|+\|o_{j,k}\|
 \le2r_{\rm rob}+R_{\SelMat,\infty}+R_O.
\]
Hence every constraint satisfies
\[
 h_l(p,u_i)\le2r_{\rm rob}+\max\{\sqrt2R_{\SelMat,\infty},R_{\SelMat,\infty}+R_O\}=:\bar h.
\]
The bound $\bar h>0$ depends on neither $i$, $x$, nor $p$.

\smallskip\noindent\emph{Step 2 (wall values at the target).}  Every wall argument at $u_i$ satisfies $-h_l(p,u_i)\ge-\bar h$.  Since $\psi_{d,\Gamma_{\rm cb}}$ is nonincreasing, $\psi_{d,\Gamma_{\rm cb}}(-h_l(p,u_i))\le\psi_{d,\Gamma_{\rm cb}}(-\bar h)$, and since $-\bar h<0$, the last case of \eqref{eq:nswall} gives $\psi_{d,\Gamma_{\rm cb}}(-\bar h)=\Gamma_{\rm cb}(1+3\bar h/d+3\bar h^2/d^2)$.  Summing over the $N_h$ constraints and using $g_i(x,p,u_i)=0$ gives
\[
 f_i^C(x,p,u_i)=\sum_{l=1}^{N_h}\psi_{d,\Gamma_{\rm cb}}(-h_l(p,u_i))
 \le N_h\Gamma_{\rm cb}\left(1+\frac{3\bar h}d+\frac{3\bar h^2}{d^2}\right)=:C_{\rm wall}.
\]

\smallskip\noindent\emph{Step 3 (distance to the target).}  The walls are nonnegative and $s$ minimizes $f_i^C(x,p,\cdot)$, so
\[
 \frac\mu2\|s-u_i\|^2=g_i(x,p,s)\le f_i^C(x,p,s)\le f_i^C(x,p,u_i)\le C_{\rm wall},
\]
where the last inequality is Step 2.  Hence $\|s-u_i\|\le\sqrt{2C_{\rm wall}/\mu}$.

\smallskip\noindent\emph{Step 4 (norm of the minimizer).}  By the triangle inequality, $\|u_i\|\le R_{\SelMat,\infty}$, and Step 3,
\begin{equation}\label{eq:nsrobotR}
 \|s\|\le\|u_i\|+\|s-u_i\|\le R_{\SelMat,\infty}+\sqrt{\frac{2C_{\rm wall}}{\mu}}=:R_{\rm ctr},
 \qquad R_{\rm kin}:=\max\{\|y_0\|,\ R_{\rm ctr}+1\}.
\end{equation}
The tube radius $R_{\rm kin}$ covers the initial state and adds the margin $r_{\rm in}=1$ used in \prettyref{nscb:NI}; in particular, $\|s\|\le R_{\rm ctr}<R_{\rm kin}$.
If $N_h=0$, then $C_{\rm wall}=0$ and $s=\SelMat_i x$.

\smallskip\noindent\emph{\prettyref{nscb:0}.} $\operatorname{clr}(y_0)\ge\sigma_0>\sigma_0/2\ge\tau_{\rm cb}$ by the start hypothesis and \prettyref{lem:nsjam}, and $\|y_0\|\le R_{\rm kin}$.  So $y_0\in\mathcal S_0$.  Each $\operatorname{clr}_e$ is continuous, since $\operatorname{dist}(\cdot,S_e)$ is $1$-Lipschitz and $\mathbf B_e$ is linear.  So $\mathcal R_{\ge\tau_{\rm cb}}$ is closed ($\mathcal R_{\ge\tau_{\rm cb}}=\R^{n_y}$ if $\cE=\varnothing$), and every $\mathcal S_i$ is compact.

\smallskip\noindent\emph{\prettyref{nscb:F}.} If $N_h=0$, take $\bar y=s=\SelMat_ix$; then $\operatorname{sl}\equiv+\infty$, the force expression is zero, and the item holds.  Let $N_h>0$, $1\le i\le K$ and $(x,p)\in\cX\times\mathcal S_{i-1}$. \prettyref{lem:nsjam} with $r_{\rm tail}=r_{\rm rob}\le \mathfrak d_0$ supplies $\bar y$ with $\|\bar y\|\le Z_{\rm jam}$ by (i) and $\operatorname{sl}(p,\bar y)\ge\sigma_*$ by (ii).  Then $\|\nabla_yg_i(x,p,\bar y)\|=\mu\|\bar y-\SelMat_i x\|\le\mu(Z_{\rm jam}+R_{\SelMat,\infty})$.  Thus
\[
 \langle\nabla_yg_i(x,p,\bar y),\bar y-s\rangle-\mu\|\bar y-s\|^2
 \le\mu(Z_{\rm jam}+R_{\SelMat,\infty})\|\bar y-s\|-\mu\|\bar y-s\|^2
 \le\frac{\mu(Z_{\rm jam}+R_{\SelMat,\infty})^2}{4}<\Upsilon_{\rm kin}.
\]

\smallskip\noindent\emph{\prettyref{nscb:XT}.} If $\operatorname{sl}(p,s)\ge\tau_{\rm cb}$, \prettyref{lem:segxt} gives $\operatorname{clr}(s)\ge\tau_{\rm cb}$.  With $\|s\|<R_{\rm kin}$, this gives $s\in\mathcal S_i$.

\smallskip\noindent\emph{\prettyref{nscb:NI}.} For every $1\le i\le K$ and $(x,p)\in\cX\times\mathcal Y_{i-1}$, the center bound gives $\Ball_1(s_i^C(x,p))\subseteq\Ball_{R_{\rm ctr}+1}(0)\subseteq\Ball_{R_{\rm kin}}(0)=\mathcal Y_i$.  The tubes are constant, compact, convex, and nested, and $\mathcal S_i\subseteq\mathcal Y_i$.

The finite suprema $R_{\SelMat,\infty}$ and $\bar\varsigma_\infty$ make $\bar h$, $C_{\rm wall}$, $R_{\rm ctr}$, $R_{\rm kin}$, $\Upsilon_{\rm kin}$, $\Lambda_{\rm cb}$ ($=4\Upsilon_{\rm kin}/\Delta_{\rm jam}$ if $N_h>0$), $\Gamma_{\rm cb}$, and the base derivative envelopes independent of depth; none of these quantities depends on $\eta$.  For a fixed $K$ without the suprema hypothesis, the proof uses $\bar\varsigma_\infty$ and $R_{\SelMat,\infty}$ only as bounds on $\|\SelMat_i\|$ and $\|\SelMat_ix\|$ for $1\le i\le K$, so it goes through with $\bar\varsigma_K$ and $R_{\SelMat,K}$, after continuing the family by $g_i:=g_K$ for $i>K$ in \prettyref{ns:G}.
\end{proof}

We next prove the formal \prettyref{prop:dynrobot}, beginning with the conditioning of the dynamic stage objective.

\begin{proposition}[Dynamic conditioning and semi-implicit Euler center]\label{prop:dynconditioning}
For every stage, the outgoing Hessian of \eqref{eq:dynobjective} is
\begin{equation}\label{eq:dynhessian}
 \nabla_{yy}^2g_i=
 \begin{bmatrix}(m_{\rm mass}+1)/(\Delta t)^2&-1/\Delta t\\-1/\Delta t&1\end{bmatrix}\otimes I_{n_q},
\end{equation}
and its smallest eigenvalue is
\begin{equation}\label{eq:dynspectrum}
 \mu_{\rm dyn}:=\frac{(\Delta t)^2+m_{\rm mass}+1-
 \sqrt{(m_{\rm mass}+1-(\Delta t)^2)^2+4(\Delta t)^2}}{2(\Delta t)^2}>0.
\end{equation}
The stage-$i$ unconstrained minimizer $u_i=(u_i^{\pos},u_i^{\vel})$ is
\begin{equation}\label{eq:dyncenter}
 u_i^{\vel}=p^{\vel}+\frac{\Delta t}{m_{\rm mass}}\SelMat_i x,
 \qquad
 u_i^{\pos}=p^{\pos}+\Delta t\,p^{\vel}
 +\frac{(\Delta t)^2}{m_{\rm mass}}\SelMat_i x
 =p^{\pos}+\Delta t\,u_i^{\vel}.
\end{equation}
Moreover, the family $(g_i)_{i=1}^K$, continued by $g_i:=g_K$ for $i>K$, satisfies \prettyref{ns:G} with $\mu=\mu_{\rm dyn}$ and $L_{H,\mathcal D}^g=0$ on every bounded convex set $\mathcal D$.
\end{proposition}

\begin{proof}
Set $A=p^{\vel}-(y^{\pos}-p^{\pos})/\Delta t$ and $B=y^{\vel}-(y^{\pos}-p^{\pos})/\Delta t$.  Direct differentiation gives
\[
 \nabla_{y^{\pos}}g_i=-\frac{m_{\rm mass}A+B}{\Delta t}-\SelMat_i x,
 \qquad \nabla_{y^{\vel}}g_i=B,
\]
and a second derivative gives \eqref{eq:dynhessian}.  The scalar $2\times2$ block has determinant $m_{\rm mass}/(\Delta t)^2>0$ and trace $1+(m_{\rm mass}+1)/(\Delta t)^2>0$, so both its eigenvalues are positive.  The identity $((\Delta t)^2+m_{\rm mass}+1)^2-4m_{\rm mass}(\Delta t)^2=(m_{\rm mass}+1-(\Delta t)^2)^2+4(\Delta t)^2>0$ yields its smaller eigenvalue \eqref{eq:dynspectrum}; the Kronecker factor $I_{n_q}$ does not change the spectrum.  Stationarity gives $B=0$ and then $m_{\rm mass}A=-\Delta t\,\SelMat_i x$, proving \eqref{eq:dyncenter}.

Each $g_i$ is a quadratic polynomial in $(x,p,y)$, so $g_i\in C^2$ and its full Hessian $\nabla^2g_i$ is a constant matrix; in particular, $\nabla^2g_i$ is Lipschitz with constant $0$.  The blocks $\nabla^2_{yy}g_i$, $\partial_a\nabla_yg_i$, and $\nabla^2_{aa}g_i$ are constant submatrices of $\nabla^2g_i$, so their norms are at most $\|\nabla^2g_i\|$.  By \eqref{eq:dynspectrum}, $\mu_{\rm dyn}I\preceq\nabla^2_{yy}g_i$.  The continued family consists of the $K$ functions $g_1,\dots,g_K$ only, so the constants $L_{yy,\mathcal D}^g=L_{ay,\mathcal D}^g=L_{aa,\mathcal D}^g=L_{f,\mathcal D}^g:=\max_{1\le i\le K}\|\nabla^2g_i\|<\infty$ and $L_{H,\mathcal D}^g:=0$ serve every $i\ge1$ and every bounded convex $\mathcal D$.  The oracle returns $\nabla g_i$, which is affine in $(x,p,y)$.
\end{proof}

\begin{formalproposition}{prop:dynrobot}[Dynamic navigation]
Let \prettyref{ns:X} and~\prettyref{ns:F} hold, and let $\operatorname{clr}(y_0^{\pos})\geq\sigma_0>0$.  Let $\tau_{\rm cb},\Delta_{\rm jam},Z_{\rm jam},\eta_0$ be the jam data of \prettyref{lem:nsjam}, set $r_{\rm tail}=r_{\rm rob}$, take $0<\eta\leq\eta_0$, and set
\[
 \sigma_*:=\tau_{\rm cb}+\frac{\Delta_{\rm jam}}2,
 \qquad d:=\tau_{\rm cb}+\frac{\Delta_{\rm jam}}4,
 \qquad r_{\rm in}:=1.
\]
For every fixed $m_{\rm mass},\Delta t>0$ and finite $K$, the stage \eqref{eq:dynobjective}, with the smoothed constraints \eqref{eq:nspair} and \eqref{eq:nsobstacle} lifted to the position blocks by \eqref{eq:dynwalls}, satisfies \prettyref{ass:nscb}.  The dynamic radii for the exact cores and numerical tubes are defined in \eqref{eq:dynexactradii} and \eqref{eq:dynnumradii}.  They may depend on $K$, but none depends on $\eta$.
\end{formalproposition}

\begin{proof}
\emph{\prettyref{ass:ns}.}  \prettyref{ns:X} and~\prettyref{ns:F} are assumed. \prettyref{prop:dynconditioning} proves \prettyref{ns:G} with $\mu=\mu_{\rm dyn}$, and \prettyref{cor:segH} proves \prettyref{ns:H1} and~\prettyref{ns:H2}.  The fixed data satisfy $0<\tau_{\rm cb}<\sigma_*$ and $d=(\tau_{\rm cb}+\sigma_*)/2$.

\emph{Data.}  For $R\ge0$, let
\begin{equation}\label{icml:dynexdata}
 \begin{gathered}
 \mathsf P(R):=(1+\Delta t)R+\frac{(\Delta t)^2}{m_{\rm mass}}R_{\SelMat,K},
 \qquad
 \mathsf V(R):=R+\frac{\Delta t}{m_{\rm mass}}R_{\SelMat,K},\\
 \kappa_{\rm dyn}:=\sqrt{\frac{m_{\rm mass}}{\mu_{\rm dyn}(\Delta t)^2}},\\
 T_{\rm ex}(R):=\sqrt{\mathsf P(R)^2+\mathsf V(R)^2}
 +\kappa_{\rm dyn}\bigl[Z_{\rm jam}+\mathsf P(R)\bigr].
 \end{gathered}
\end{equation}
Define
\begin{equation}\label{eq:dynexactradii}
 \begin{aligned}
 R_{{\rm ex},0}&:=\|y_0\|,\\
 R_{{\rm ex},i}&:=\max\{R_{{\rm ex},i-1},T_{\rm ex}(R_{{\rm ex},i-1})\}&&(1\le i\le K),\\
 \mathcal S_i&:=(\mathcal R_{\ge\tau_{\rm cb}}\times\R^{n_q})\cap\Ball_{R_{{\rm ex},i}}(0)&&(0\le i\le K),
 \end{aligned}
\end{equation}
\begin{equation}\label{eq:dynforceconstants}
 G_{\rm dyn}:=\frac{m_{\rm mass}R_{{\rm ex},K-1}}{\Delta t}
 +\frac{m_{\rm mass}(Z_{\rm jam}+R_{{\rm ex},K-1})}{(\Delta t)^2}+R_{\SelMat,K},
 \qquad
 \Upsilon_{\rm dyn}:=\frac{(G_{\rm dyn}+1)^2}{4\mu_{\rm dyn}},
\end{equation}
set $\Upsilon_{\rm cb}:=\Upsilon_{\rm dyn}$, and define $\Lambda_{\rm cb},\Gamma_{\rm cb}$ by \eqref{eq:nscbdata} (with $\Lambda_{\rm cb}:=0$ and any fixed $\Gamma_{\rm cb}>0$ if $N_h=0$).  For the numerical tube, set
\begin{equation}\label{icml:dynnumdata}
 \bar h(R):=2r_{\rm rob}+\sqrt2\,\mathsf P(R)+R_O,
 \qquad
 \mathsf C(R):=N_h\Gamma_{\rm cb}\left[1+\frac{3\bar h(R)}d+\frac{3\bar h(R)^2}{d^2}\right],
\end{equation}
\begin{equation}\label{eq:dynnumtransfer}
 T_{\rm num}(R):=\sqrt{\mathsf P(R)^2+\mathsf V(R)^2}
 +\sqrt{\frac{2\mathsf C(R)}{\mu_{\rm dyn}}},
\end{equation}
\begin{equation}\label{eq:dynnumradii}
 \begin{aligned}
 R_{{\rm num},0}&:=R_{{\rm ex},0},\\
 R_{{\rm num},i}&:=\max\{R_{{\rm num},i-1},R_{{\rm ex},i}+1,T_{\rm num}(R_{{\rm num},i-1})+1\}
 &&(1\le i\le K),\\
 \mathcal Y_i&:=\Ball_{R_{{\rm num},i}}(0)&&(0\le i\le K).
 \end{aligned}
\end{equation}
The order is: $\tau_{\rm cb},\Delta_{\rm jam},Z_{\rm jam}$ (geometry and $\sigma_0$); $\eta_0$ and $\eta$; $\sigma_*,d,b_{\rm cb}$; $\mu_{\rm dyn},\kappa_{\rm dyn}$; $R_{{\rm ex},0},\dots,R_{{\rm ex},K}$; $G_{\rm dyn},\Upsilon_{\rm dyn}$; $\Lambda_{\rm cb},\Gamma_{\rm cb}$; $\mathsf C$; $R_{{\rm num},0},\dots,R_{{\rm num},K}$.  No item uses a later one, and none uses $\eta$.

\emph{Center bound for every predecessor.}  Fix a stage $1\le i\le K$, a control $x\in\cX$, and a predecessor $p$ with $\|p\|\le R$.  By \eqref{eq:dyncenter} and $\|\SelMat_ix\|\le R_{\SelMat,K}$, the unconstrained minimizer $u_i$ satisfies
\[
 \|u_i^{\vel}\|\le\|p^{\vel}\|+\frac{\Delta t}{m_{\rm mass}}\|\SelMat_ix\|\le\mathsf V(R),
 \qquad
 \|u_i^{\pos}\|\le\|p^{\pos}\|+\Delta t\|p^{\vel}\|+\frac{(\Delta t)^2}{m_{\rm mass}}\|\SelMat_ix\|\le\mathsf P(R),
\]
so $\|u_i\|\le\sqrt{\mathsf P(R)^2+\mathsf V(R)^2}$.  Each wall $\psi_{d,\Gamma_{\rm cb}}(-h_l^{\rm dyn}(p,\cdot))$ is convex and nonnegative, because $\psi_{d,\Gamma_{\rm cb}}$ is convex, nonincreasing, and nonnegative by \eqref{eq:nswall} and $-h_l^{\rm dyn}(p,\cdot)$ is affine.  So $f_i^C(x,p,\cdot)$ is $\mu_{\rm dyn}$-strongly convex and has a unique minimizer $s:=s_i^C(x,p)$, and for every $y\in\R^{2n_q}$,
\begin{equation}\label{eq:dynscmin}
 \tfrac{\mu_{\rm dyn}}2\|s-u_i\|^2\le g_i(x,p,s)-g_i(x,p,u_i)\le f_i^C(x,p,s)-g_i(x,p,u_i)\le f_i^C(x,p,y)-g_i(x,p,u_i).
\end{equation}
The first inequality holds because $g_i(x,p,\cdot)$ is $\mu_{\rm dyn}$-strongly convex (\prettyref{prop:dynconditioning}) with minimizer $u_i$, the second because the walls are nonnegative, and the third because $s$ minimizes $f_i^C(x,p,\cdot)$.

\smallskip\noindent\emph{\prettyref{nscb:0}.} By \prettyref{lem:nsjam}, $\tau_{\rm cb}\le\sigma_0/2<\sigma_0\le\operatorname{clr}(y_0^{\pos})$, and $\|y_0\|=R_{{\rm ex},0}$, so $y_0\in\mathcal S_0$.  Since $\operatorname{clr}$ is continuous, $\mathcal R_{\ge\tau_{\rm cb}}\times\R^{n_q}$ is closed ($\R^{2n_q}$ if $\cE=\varnothing$), so every $\mathcal S_i$ is compact.

\smallskip\noindent\emph{\prettyref{nscb:F}.} If $N_h=0$, then $\operatorname{sl}\equiv+\infty$, and $\bar y:=s$ makes the left side of the item zero.  Let $N_h>0$, $1\le i\le K$, and $(x,p)\in\cX\times\mathcal S_{i-1}$, so $\operatorname{clr}(p^{\pos})\ge\tau_{\rm cb}$ and $\|p\|\le R_{{\rm ex},i-1}$.  Let $\bar y^{\pos}:=\bar y(p^{\pos})$ be the point of \prettyref{lem:nsjam}, and set $\bar y^{\vel}:=(\bar y^{\pos}-p^{\pos})/\Delta t$ and $\bar y:=(\bar y^{\pos},\bar y^{\vel})$.  By \eqref{eq:dynwalls}, $\operatorname{sl}(p,\bar y)=\min_l\{-h_l(p^{\pos},\bar y^{\pos})\}$, so \prettyref{lem:nsjam} with $r_{\rm tail}=r_{\rm rob}\le\mathfrak d_0$ gives $\operatorname{sl}(p,\bar y)\ge\sigma_*$ by (ii) and $\|\bar y^{\pos}\|\le Z_{\rm jam}$ by (i).  Since $\bar y^{\vel}-(\bar y^{\pos}-p^{\pos})/\Delta t=0$, differentiating \eqref{eq:dynobjective} gives
\[
 \nabla_yg_i(x,p,\bar y)=
 \left(-\frac{m_{\rm mass}}{\Delta t}\left[p^{\vel}-\frac{\bar y^{\pos}-p^{\pos}}{\Delta t}\right]-\SelMat_i x,\ 0\right).
\]
The triangle inequality with $\|p^{\pos}\|,\|p^{\vel}\|\le R_{{\rm ex},i-1}\le R_{{\rm ex},K-1}$, $\|\bar y^{\pos}\|\le Z_{\rm jam}$, and $\|\SelMat_ix\|\le R_{\SelMat,K}$ gives $\|\nabla_yg_i(x,p,\bar y)\|\le G_{\rm dyn}$.  Hence, maximizing $G_{\rm dyn}t-\mu_{\rm dyn}t^2$ over $t\ge0$,
\[
 \langle\nabla_yg_i(x,p,\bar y),\bar y-s\rangle-
 \mu_{\rm dyn}\|\bar y-s\|^2
 \le G_{\rm dyn}\|\bar y-s\|-\mu_{\rm dyn}\|\bar y-s\|^2
 \le\frac{G_{\rm dyn}^2}{4\mu_{\rm dyn}}<\Upsilon_{\rm dyn}.
\]

\smallskip\noindent\emph{\prettyref{nscb:XT}.} Let $1\le i\le K$ and $(x,p)\in\cX\times\mathcal S_{i-1}$ with $\operatorname{sl}(p,s)\ge\tau_{\rm cb}$.  \prettyref{lem:segxt} on the position block gives $\operatorname{clr}(s^{\pos})\ge\tau_{\rm cb}$, so it remains to show $\|s\|\le R_{{\rm ex},i}$.  Apply the center bound with $R:=R_{{\rm ex},i-1}$.  If $N_h=0$, then $s=u_i$ and $\|s\|\le\sqrt{\mathsf P(R)^2+\mathsf V(R)^2}\le T_{\rm ex}(R)$.  Otherwise, let $\bar y$ be the point built in \prettyref{nscb:F}.  Since $\operatorname{sl}(p,\bar y)\ge\sigma_*>d$, every wall vanishes at $\bar y$, so $f_i^C(x,p,\bar y)=g_i(x,p,\bar y)$.  Both $\bar y$ and $u_i$ lie on the affine set $\{y:y^{\vel}=(y^{\pos}-p^{\pos})/\Delta t\}$, where $g_i(x,p,\cdot)$ is a quadratic in $y^{\pos}$ with Hessian $(m_{\rm mass}/(\Delta t)^2)I$, minimized at $u_i^{\pos}$ by \eqref{eq:dyncenter}.  Hence
\[
 g_i(x,p,\bar y)-g_i(x,p,u_i)=\frac{m_{\rm mass}}{2(\Delta t)^2}\|\bar y^{\pos}-u_i^{\pos}\|^2,
\]
and \eqref{eq:dynscmin} at $y=\bar y$ gives $\|s-u_i\|\le\kappa_{\rm dyn}\|\bar y^{\pos}-u_i^{\pos}\|\le\kappa_{\rm dyn}(Z_{\rm jam}+\mathsf P(R))$.  In both cases $\|s\|\le\|u_i\|+\|s-u_i\|\le T_{\rm ex}(R)\le R_{{\rm ex},i}$, so $s\in\mathcal S_i$.

\smallskip\noindent\emph{\prettyref{nscb:NI}.} Fix $1\le i\le K$, $x\in\cX$, and $p\in\mathcal Y_{i-1}$; no clearance of $p$ is used.  Apply the center bound with $R:=R_{{\rm num},i-1}$.  By Cauchy--Schwarz with $\|\nu\|,\|\nu_\eta\|\le1$ (\prettyref{lem:nsnormal}),
\[
 h_e(p^{\pos},u_i^{\pos})\le2r_{\rm rob}+\|(u_i^{\pos})_r-(u_i^{\pos})_{r'}\|\le2r_{\rm rob}+\sqrt2\|u_i^{\pos}\|,
 \qquad
 h_{e,k}(p^{\pos},u_i^{\pos})\le2r_{\rm rob}+\|u_i^{\pos}\|+\|o_{j,k}\|,
\]
so $h_l^{\rm dyn}(p,u_i)\le2r_{\rm rob}+\sqrt2\,\mathsf P(R)+R_O=\bar h(R)$ for every constraint.  As in Step 2 of the proof of \prettyref{prop:nsrobot}, each wall at $u_i$ is at most $\psi_{d,\Gamma_{\rm cb}}(-\bar h(R))=\Gamma_{\rm cb}(1+3\bar h(R)/d+3\bar h(R)^2/d^2)$, and summing gives $f_i^C(x,p,u_i)-g_i(x,p,u_i)\le\mathsf C(R)$.  So \eqref{eq:dynscmin} at $y=u_i$ gives $\|s-u_i\|\le\sqrt{2\mathsf C(R)/\mu_{\rm dyn}}$, and $\|s\|\le\|u_i\|+\|s-u_i\|\le T_{\rm num}(R)$.  Hence $\Ball_1(s)\subseteq\Ball_{T_{\rm num}(R)+1}(0)\subseteq\Ball_{R_{{\rm num},i}}(0)=\mathcal Y_i$.  Finally, \eqref{eq:nscbnested} holds: the tubes are compact convex balls with nondecreasing radii, and $\mathcal S_i\subseteq\Ball_{R_{{\rm ex},i}}(0)\subseteq\mathcal Y_i$ since $R_{{\rm num},0}=R_{{\rm ex},0}$ and $R_{{\rm num},i}\ge R_{{\rm ex},i}+1$.
\end{proof}

\paragraph{Depth dependence is unavoidable.} The growth of the dynamic radii with $K$ cannot be removed in general.  Take $N_{\rm rob}=1$ and $N_{\rm obs}=0$, so $N_h=0$, and suppose that some $x\in\cX$ gives the same force $\SelMat_ix=f\ne0$ at every stage.  By \eqref{eq:dyncenter}, the exact chain is the unconstrained semi-implicit Euler recursion, so its velocity after $i$ steps is $y_0^{\vel}+i\,\Delta t\,f/m_{\rm mass}$, whose norm grows linearly in $i$.  \prettyref{nscb:XT} requires $\widehat y_K^C(x)\in\mathcal S_K$, so the radius of $\mathcal S_K$ is at least of order $K\Delta t\|f\|/m_{\rm mass}$, and no family of cores independent of $K$ can serve every horizon.

\section{Experiment details}\label{sec:app_experiments}

This appendix gives the full setup and results behind \prettyref{sec:experiments}.

\paragraph{Setup.} All robots are discs of radius $r_{\rm rob}=0.42$ in the plane.  We use four scenes.  In the \emph{corner swap}, four robots move to the diagonally opposite corners of $[0.7,5.3]^2$ around the square obstacle $[2.2,3.8]^2$.  In the \emph{eight-robot swap}, which has no obstacles, eight robots on a circle of radius $3$ move to the opposite points.  The \emph{sealed corridor} of width $3$ lies between a closed start room and a closed goal room and fits at most two of its three robots side by side, so the outer robots must reorder.  In the closed \emph{zigzag channel} of width $2.8$ with sharp bends, two robots enter from opposite ends and must pass each other.  For the corridor and the channel we also check that every exact robot-motion segment stays inside the polygon.  We use the kinematic model ($\mu=1$) and the dynamic model (mass $1$, time step $0.5$) of \prettyref{sub:motivating}.  The control has one block $\SelMat_i x$ per stage, the control set $\cX$ is a box with a fixed per-block bound, and the band width is $d=0.1$ ($0.08$ for the eight-robot swap).  Every run starts from a neutral control that carries no route information (robots hold their start positions, or all forces are zero), plus i.i.d.\ Gaussian noise on every control entry with the standard deviation $\sigma_{\rm noise}$ of \prettyref{tab:app_schedules}.  The runs minimize the task loss \eqref{icml:navF} plus a trajectory-smoothness term
\[
\begin{gathered}
 R_{\rm reg}=w_{\rm reg}\,\delta_F\Bigl(\sqrt{1+Q_{\rm acc}/\delta_F^2}-1\Bigr)+w_{\rm vel}\,\delta_F\Bigl(\sqrt{1+Q_{\rm vel}/\delta_F^2}-1\Bigr),\\
 Q_{\rm acc}=\sum_{i=2}^{K}\|y_{i-2}^{\pos}-2y_{i-1}^{\pos}+y_i^{\pos}\|^2,\qquad Q_{\rm vel}=\sum_{i=1}^{K}\|y_i^{\pos}-y_{i-1}^{\pos}\|^2,
\end{gathered}
\]
with $y_0^{\pos}$ the start position, $\delta_F=1$, $w_{\rm reg}=\factWreg$, and $w_{\rm vel}=\factWvel$.  Since \eqref{icml:navF} scores only the final state, without $Q_{\rm vel}$ the optimizer uses only the last few stages; the term spreads the motion over all stages.  $F+R_{\rm reg}$ is smooth and bounded below with a bounded and Lipschitz gradient, so \prettyref{ns:F} still holds, with $G_F=\delta_F\sqrt{N_{\rm rob}}+w_{\rm reg}\|\mathbf D_2\|+w_{\rm vel}\|\mathbf D_1\|$ and $L_F=1+(w_{\rm reg}\|\mathbf D_2\|^2+w_{\rm vel}\|\mathbf D_1\|^2)/\delta_F$, where $\mathbf D_2$ and $\mathbf D_1$ are the second- and first-difference operators.  Below, $F$ denotes $F+R_{\rm reg}$, and $\varphi^C$ is the resulting \emph{exact chain loss} \eqref{eq:nscbchain}.  We use it to compare all methods because it depends only on the control: every method's control can be evaluated by solving each stage exactly, so $\varphi^C$ is available and meaningful across all methods.  A run \emph{reaches its goals} if every robot ends within $r_{\rm rob}$ of its goal.  With few stages $K$ this may be impossible: each stage linearizes the constraints at the previous state, which limits how far a robot moves per stage.  A multi-start search finds no feasible chain to the goals in the corridor with $K=4$ or the zigzag channel with $K=4,8$.  This search varies the robot positions at every stage and minimizes the final goal distance under the walls linearized at each previous position, from \factReachAttempts{} different initial paths per case; the best goal errors it finds are \factReachCorrKfour, \factReachZigKfour, and \factReachZigKeight, respectively, all above $r_{\rm rob}$.  The theoretical schedule depends on constants that our analysis only shows to exist, through semialgebraic arguments (\prettyref{sec:instances}).  We therefore chose the PSK parameters per case by a grid search over \factNScreenRuns{} short screening runs in total, which also varied the trajectory-smoothness weights.  Among the parameter settings whose exact states stay feasible and pass the banded Fritz--John check below, we preferred those that reach the goals; \factNAdjusted{} cases were retuned afterwards, for feasibility in long runs or for speed at the comparison budget.  \prettyref{tab:app_schedules} lists the final values. PSK evaluates its output at the checkpoints $T\in\{10^2,3\cdot10^2,10^3,3\cdot10^3,\ldots\}$: it solves the exact surrogate chain at the selected control $x_{t^\star(T)}$ of \prettyref{alg:psk} and records $\varphi^C$ and the minimum hard slack.  A run stops at the first $T\ge10^4$ at which the best $\varphi^C$ recorded up to $T$ is less than $\factStallPct\%$ below the best recorded up to $T/10$, or at the cap $T_{\max}=\factTMax$, and reports the checkpoint control with the lowest $\varphi^C$ among those whose exact chain is feasible.

\begin{table}[htbp]
\centering
\caption{Final PSK parameters per case: inner budget $m$, warm-start budget $m_0$, penalty schedule $\rho_0$ and $\beta$, inner step $\gamma$, outer step scale $c_\alpha$, wall height $\Gamma$, and standard deviation $\sigma_{\rm noise}$ of the Gaussian start noise.}
\label{tab:app_schedules}
\resizebox{\textwidth}{!}{\begin{tabular}{clrrrrrrrrr}
\toprule
scene & model & $K$ & $m$ & $m_0$ & $\rho_0$ & $\beta$ & $\gamma$ & $c_\alpha$ & $\Gamma$ & $\sigma_{\rm noise}$\\
\midrule
\multirow{6}{*}{\shortstack[c]{Corner\\swap}} & \multirow{3}{*}{kinematic} & 4 & 10 & 200 & 0.5 & 0.01 & 0.005 & 0.02 & 0.1 & 0.05\\
 &  & 8 & 10 & 200 & 0.5 & 0.01 & 0.005 & 0.02 & 0.1 & 0.05\\
 &  & 16 & 10 & 200 & 0.5 & 0.01 & 0.005 & 0.03 & 0.1 & 0.05\\
 & \multirow{3}{*}{dynamic} & 4 & 10 & 793 & 0.1 & 0.005 & 0.00404 & 0.005 & 0.2 & 0.05\\
 &  & 8 & 5 & 400 & 0.1 & 0.005 & 0.008 & 0.0025 & 0.1 & 0.05\\
 &  & 16 & 5 & 400 & 0.1 & 0.005 & 0.008 & 0.0025 & 0.1 & 0.05\\
\midrule
\multirow{6}{*}{\shortstack[c]{Eight-robot\\swap}} & \multirow{3}{*}{kinematic} & 4 & 5 & 200 & 1 & 0.01 & 0.003 & 0.015 & 0.1 & 0.3\\
 &  & 8 & 5 & 200 & 1 & 0.01 & 0.003 & 0.015 & 0.1 & 0.3\\
 &  & 16 & 5 & 200 & 1 & 0.01 & 0.003 & 0.015 & 0.1 & 0.3\\
 & \multirow{3}{*}{dynamic} & 4 & 5 & 400 & 0.05 & 0.005 & 0.006 & 0.005 & 0.1 & 0.3\\
 &  & 8 & 5 & 400 & 0.05 & 0.005 & 0.006 & 0.005 & 0.1 & 0.3\\
 &  & 16 & 2 & 400 & 0.075 & 0.0005 & 0.006 & 0.01 & 0.1 & 0.3\\
\midrule
\multirow{6}{*}{\shortstack[c]{Sealed\\corridor}} & \multirow{3}{*}{kinematic} & 4 & 5 & 200 & 1 & 0.002 & 0.0015 & 0.004 & 0.4 & 0.05\\
 &  & 8 & 7 & 275 & 0.1 & 0.002 & 0.00109 & 0.004 & 0.4 & 0.05\\
 &  & 16 & 5 & 200 & 0.1 & 0.001 & 0.0015 & 0.008 & 0.2 & 0.05\\
 & \multirow{3}{*}{dynamic} & 4 & 20 & 1965 & 0.05 & 0.02 & 0.000509 & 0.0015 & 1 & 0.05\\
 &  & 8 & 5 & 500 & 0.05 & 0.02 & 0.002 & 0.002 & 1 & 0.05\\
 &  & 16 & 5 & 500 & 0.05 & 0.05 & 0.002 & 0.003 & 0.3 & 0.05\\
\midrule
\multirow{6}{*}{\shortstack[c]{Zigzag\\channel}} & \multirow{3}{*}{kinematic} & 4 & 20 & 5000 & 0.05 & 0.003 & 0.001 & 0.0005 & 0.15 & 0.15\\
 &  & 8 & 20 & 5000 & 0.05 & 0.003 & 0.001 & 0.000125 & 0.3 & 0.15\\
 &  & 16 & 5 & 200 & 0.3 & 0.002 & 0.0015 & 0.005 & 0.2 & 0.2\\
 & \multirow{3}{*}{dynamic} & 4 & 10 & 1000 & 0.05 & 0.02 & 0.002 & 0.002 & 0.3 & 0.2\\
 &  & 8 & 20 & 1945 & 0.02 & 0.02 & 0.00103 & 0.002 & 0.6 & 0.2\\
 &  & 16 & 3 & 1476 & 0.03 & 0.02 & 0.00176 & 0.0045 & 0.3 & 0.2\\
\bottomrule
\end{tabular}
}
\end{table}

\begin{figure}[p]
\centering
\includegraphics[width=0.85\textwidth]{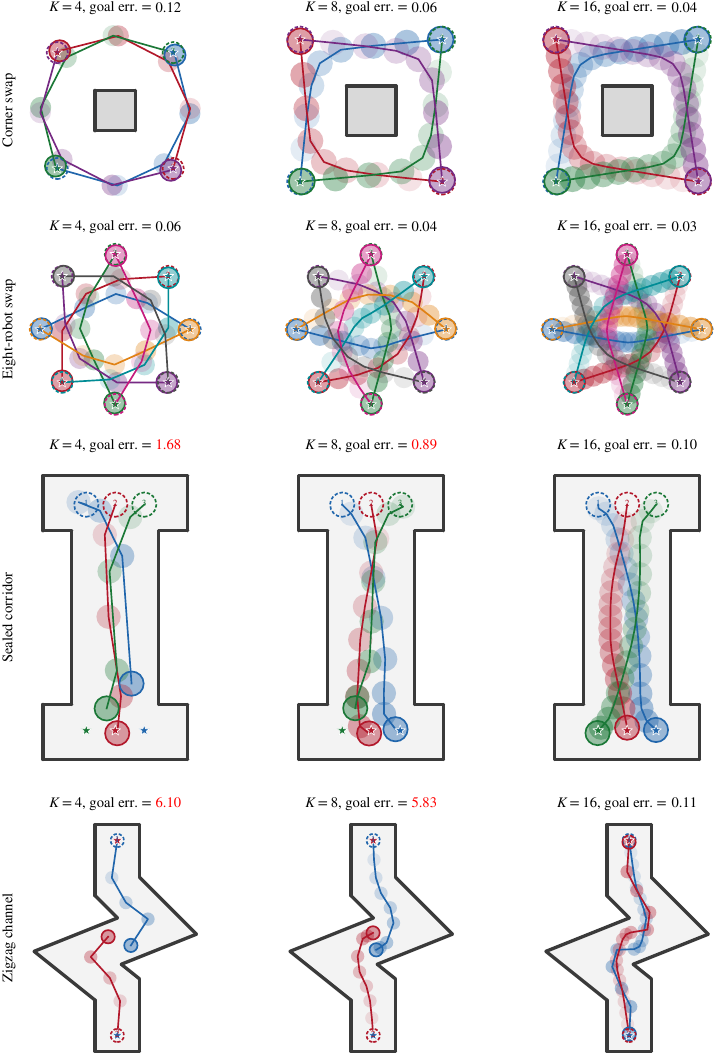}
\caption{All kinematic runs (rows: scenes; columns: $K=4,8,16$).  Same drawing as \prettyref{fig:gallery}; numbers mark the robots at their starts; each panel title gives the goal error, in red if the goal is not reached.}
\label{fig:app_gallery_kin}
\end{figure}

\begin{figure}[p]
\centering
\includegraphics[width=0.85\textwidth]{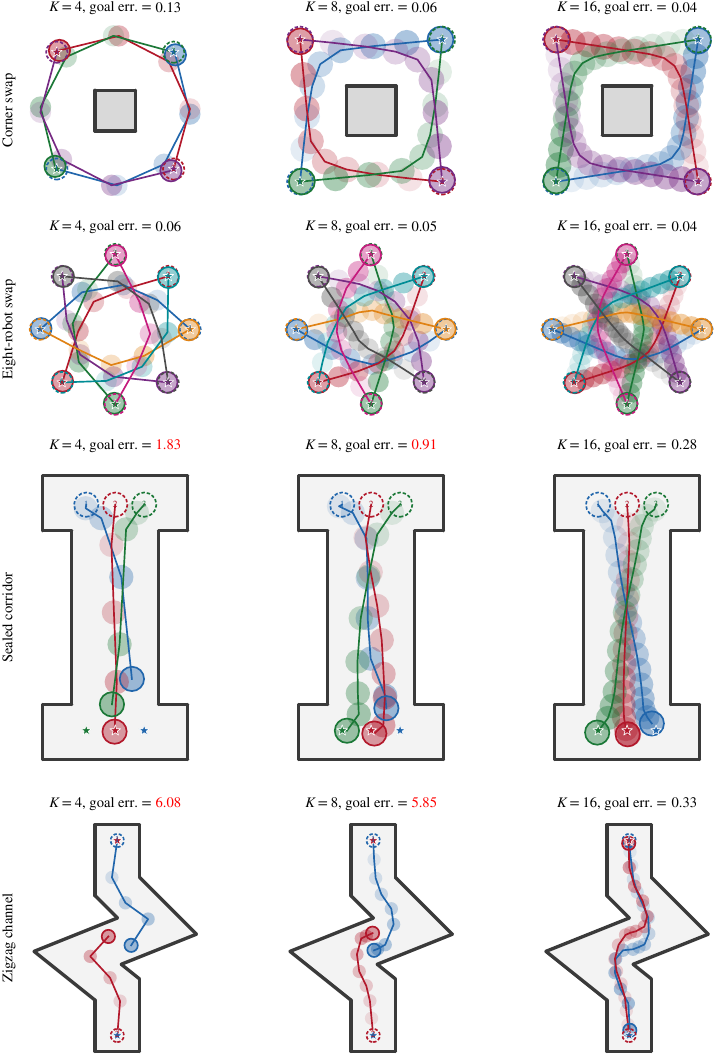}
\caption{All dynamic runs, drawn as in \prettyref{fig:app_gallery_kin} (rows: scenes; columns: $K=4,8,16$).}
\label{fig:app_gallery_dyn}
\end{figure}

\prettyref{tab:app_cases} lists all runs, and \prettyref{fig:app_gallery_kin} and \ref{fig:app_gallery_dyn} show them.  In all \factNKsixteen{} cases with $K=16$, every robot reaches its goal.  We measure goal reaching on the exact chain at the reported control, that is, on the exact solution of every surrogate stage rather than on PSK's inexact iterates, so each reached goal reflects the exactly simulated motion.  Every run keeps its exact chain feasible at every checkpoint from $10^3$ until it stops, and every reported control passes the banded Fritz--John check of \prettyref{thm:nsrecover}.  This check solves every stage of the exact chain by damped Newton steps to gradient norm at most $10^{-10}$, sets the wall multipliers $-\psi'_{d,\Gamma}(-h_l)$ of \eqref{icml:nsmult}, with the practical height $\Gamma$ in place of $\Gamma_{\rm cb}$, which vanish outside the band, and tests feasibility and the Fritz--John stationarity residual.  By \prettyref{lem:cubiccertificate}, a feasible exact stage solution with the cubic wall is already banded Fritz--John, so the check confirms feasibility and the accuracy of the exact solves rather than adding separate evidence.

\begin{table}[htbp]
\centering
\caption{Statistics of all PSK runs.  Goal error is the largest final robot--goal distance on the exact chain (red: goal not reached); min slack is the smallest exact hard-constraint slack; stop $T$ is the iteration at which the stopping rule ended the run, in units of $10^5$ ($\ast$: cap reached without stalling).}
\label{tab:app_cases}
\resizebox{\textwidth}{!}{\begin{tabular}{clrrrr}
\toprule
scene & model & $K$ & goal err. & min slack & stop $T$ ($10^5$)\\
\midrule
\multirow{6}{*}{\shortstack[c]{Corner\\swap}} & \multirow{3}{*}{kinematic} & 4 & 0.121 & 0.047 & 0.3\\
 &  & 8 & 0.057 & 0.027 & 10\\
 &  & 16 & 0.037 & 0.035 & 30\\
 & \multirow{3}{*}{dynamic} & 4 & 0.125 & 0.037 & 3\\
 &  & 8 & 0.059 & 0.003 & 30\\
 &  & 16 & 0.039 & 0.038 & 30$^\ast$\\
\midrule
\multirow{6}{*}{\shortstack[c]{Eight-robot\\swap}} & \multirow{3}{*}{kinematic} & 4 & 0.061 & 0.055 & 10\\
 &  & 8 & 0.044 & 0.050 & 30\\
 &  & 16 & 0.029 & 0.058 & 30$^\ast$\\
 & \multirow{3}{*}{dynamic} & 4 & 0.058 & 0.021 & 1\\
 &  & 8 & 0.046 & 0.009 & 10\\
 &  & 16 & 0.044 & 0.038 & 3\\
\bottomrule
\end{tabular}%
\hspace{1.5em}%
\begin{tabular}{clrrrr}
\toprule
scene & model & $K$ & goal err. & min slack & stop $T$ ($10^5$)\\
\midrule
\multirow{6}{*}{\shortstack[c]{Sealed\\corridor}} & \multirow{3}{*}{kinematic} & 4 & \textcolor{red}{1.678} & 0.039 & 30\\
 &  & 8 & \textcolor{red}{0.892} & 0.037 & 3\\
 &  & 16 & 0.098 & 0.057 & 1\\
 & \multirow{3}{*}{dynamic} & 4 & \textcolor{red}{1.832} & 0.028 & 30$^\ast$\\
 &  & 8 & \textcolor{red}{0.906} & 0.049 & 10\\
 &  & 16 & 0.285 & 0.020 & 30$^\ast$\\
\midrule
\multirow{6}{*}{\shortstack[c]{Zigzag\\channel}} & \multirow{3}{*}{kinematic} & 4 & \textcolor{red}{6.100} & 0.035 & 3\\
 &  & 8 & \textcolor{red}{5.827} & 0.019 & 30$^\ast$\\
 &  & 16 & 0.105 & 0.016 & 30$^\ast$\\
 & \multirow{3}{*}{dynamic} & 4 & \textcolor{red}{6.079} & 0.019 & 3\\
 &  & 8 & \textcolor{red}{5.847} & 0.026 & 1\\
 &  & 16 & 0.334 & 0.029 & 1\\
\bottomrule
\end{tabular}
}
\end{table}

\begin{figure}[htbp]
\centering
\includegraphics[width=0.75\textwidth]{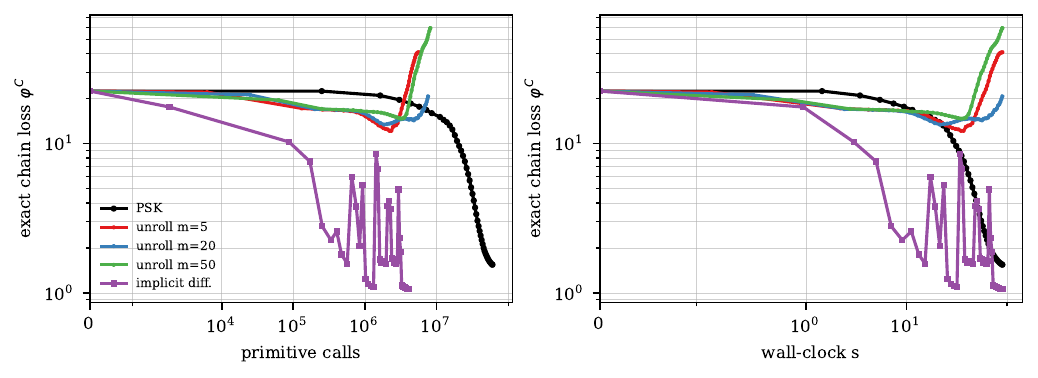}\\
\includegraphics[width=0.75\textwidth]{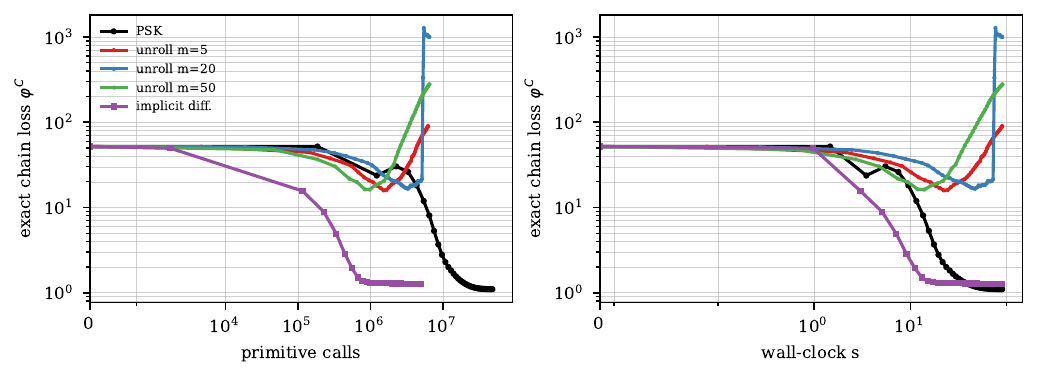}\\
\includegraphics[width=0.75\textwidth]{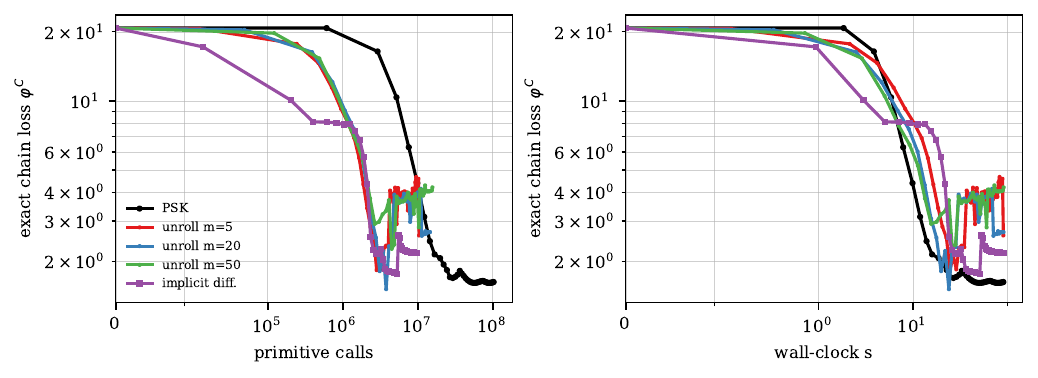}\\
\includegraphics[width=0.75\textwidth]{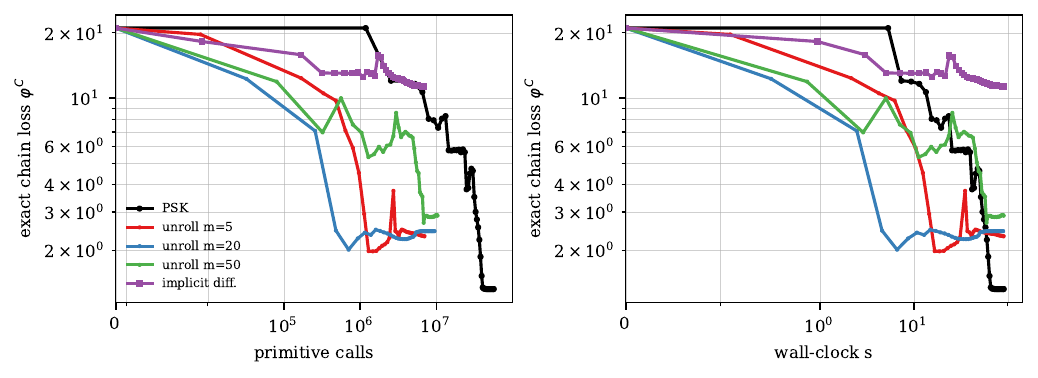}
\caption{Exact chain loss versus first-order work (left) and elapsed time (right) for the dynamic model with $K=16$ (top to bottom: corner swap, eight-robot swap, sealed corridor, zigzag channel).  PSK ends below every baseline in three of the four; on the corner swap, implicit differentiation ends lower.  All methods start from the same neutral control at zero work and zero time; PSK's work and time include its initialization, which appears as its initial flat segment.  The horizontal axes are linear near zero and logarithmic beyond.}
\label{fig:app_e5}
\end{figure}

\paragraph{Baseline comparisons.}  We compare PSK with two baselines.  \emph{Unrolled differentiable simulation} runs $m_u\in\{5,20,50\}$ gradient steps of size $\gamma_u$ per stage from the previous state, backpropagates the gradient of the loss through all $Km_u$ steps, and takes projected gradient steps of size $\alpha_u$ on the control; we tune $\gamma_u$ and $\alpha_u$ by grid search on $30$-second runs and keep the parameters leading to the lowest final $\varphi^C$.  \emph{Implicit differentiation} solves every stage of the surrogate chain exactly by damped Newton steps (gradient norm at most $10^{-10}$), computes the gradient of $\varphi^C$ by the adjoint recursion with one stage-Hessian solve per stage, and takes a projected gradient step; we perform the same grid search to find the best projected gradient step size.  For a fair comparison, every method starts from the same neutral control, optimizes the same loss on the same surrogate (band, wall, and control box), and runs for \factTimeBudget{} seconds in a single-threaded process; PSK's budget includes its initialization, so all methods start at zero time and zero work.  We sample each run's control over time, evaluate the exact chain loss $\varphi^C$ at each sample, and compare methods at matched elapsed time on all \factNeFive{} cases with a feasible chain to the goals, and at matched first-order work for the dynamic model with $K=16$; each method runs once per case.  First-order work counts primitive calls as in \prettyref{def:lbunrolled}, where a stage call costs $1+N_h$ queries.  PSK makes $c_{\rm warm}$ calls for its warm start and $c_{\rm ph}$ per iteration (\prettyref{sec:lbmbzr}).  Unrolling makes one stage call per unrolled step and one more for its backward pass.  Implicit differentiation makes two stage calls (gradient and Hessian) per Newton step and three per stage for the final check and the adjoint; we do not count line-search evaluations.  As in $c_{\rm ph}$, the baseline counts include one task-loss gradient and one projection per iteration.  \prettyref{tab:app_e5} lists the exact chain loss at \factTimeBudget{} seconds and the time per iteration for all \factNeFive{} cases, and \prettyref{fig:app_e5} plots the loss against first-order work and elapsed time for the dynamic model with $K=16$ in all four scenes.  PSK reaches a lower $\varphi^C$ than the best tuned unrolling depth in \factUnrollWins{} of \factNeFive{} cases; the exception is the kinematic zigzag channel with $K=16$, where PSK spends part of the budget on initialization and then improves slowly.  Measured by the peak bytes that automatic differentiation saves during one iteration from the neutral control, 50-step unrolling, our deepest depth, stores \factMemRatioMin--\factMemRatioMax$\times$ as much as PSK.  Implicit differentiation reaches a lower $\varphi^C$ than PSK in \factImplicitLower{} of \factNeFive{} cases, by up to \factImplicitMaxMarginPct\% of PSK's loss.  Each of its iterations, however, is one to two orders of magnitude slower than a PSK iteration, and on the corridor with $K=16$ its final loss is \factHardRatioMin--\factHardRatioMax$\times$ that of PSK.  In three of the four cases of \prettyref{fig:app_e5}, PSK ends below every baseline; on the corner swap, implicit differentiation ends lower.

\begin{table}[htbp]
\centering
\caption{Exact chain loss $\varphi^C$ at matched elapsed time (lower is better) and time per iteration; bold marks the lowest loss and the lowest time in each row.  ``unroll'' is the best of the three tuned unrolling depths, with its depth in parentheses; times per iteration are medians over repeated short timing runs from the neutral control, and ``Newton/it.'' is the mean number of Newton iterations per implicit-differentiation step in these runs, summed over the stages.  An entry of 0.0 means that no wall was active, so every stage's Newton start, its unconstrained minimizer, already met the tolerance.}
\label{tab:app_e5}
\resizebox{\textwidth}{!}{\begin{tabular}{clrrrrrrrr}
\toprule
& & & \multicolumn{3}{c}{exact chain loss $\varphi^C$ at 90\,s} & \multicolumn{4}{c}{ms / iteration}\\
\cmidrule(lr){4-6}\cmidrule(lr){7-10}
scene & model & $K$ & PSK & unroll (best $m_u$) & implicit & PSK & unroll & implicit & Newton/it.\\
\midrule
\multirow{6}{*}{\shortstack[c]{Corner\\swap}} & \multirow{3}{*}{kinematic} & 4 & 2.55 & 4.70 (5) & \textbf{2.53} & \textbf{6.2} & 20.1 & 257 & 3.6\\
 &  & 8 & 1.63 & 2.26 (50) & \textbf{1.44} & \textbf{6.2} & 251.4 & 221 & 3.6\\
 &  & 16 & \textbf{1.36} & 2.15 (5) & 3.52 & \textbf{6.4} & 83.4 & 271 & 5.8\\
 & \multirow{3}{*}{dynamic} & 4 & 2.76 & 28.68 (5) & \textbf{2.70} & \textbf{7.2} & 24.0 & 177 & 0.0\\
 &  & 8 & 1.70 & 16.40 (20) & \textbf{1.50} & \textbf{6.2} & 146.4 & 263 & 11.2\\
 &  & 16 & 1.54 & 19.92 (20) & \textbf{1.06} & \textbf{6.5} & 298.4 & 269 & 32.0\\
\midrule
\multirow{6}{*}{\shortstack[c]{Eight-robot\\swap}} & \multirow{3}{*}{kinematic} & 4 & 2.79 & 15.24 (50) & \textbf{2.71} & \textbf{2.9} & 115.7 & 192 & 0.0\\
 &  & 8 & 2.41 & 5.49 (50) & \textbf{1.69} & \textbf{3.1} & 238.9 & 180 & 5.4\\
 &  & 16 & 2.47 & 4.47 (50) & \textbf{1.10} & \textbf{3.5} & 469.8 & 179 & 4.8\\
 & \multirow{3}{*}{dynamic} & 4 & \textbf{2.73} & 23.29 (5) & 2.78 & \textbf{3.6} & 14.2 & 162 & 0.0\\
 &  & 8 & \textbf{1.75} & 33.07 (50) & 1.79 & \textbf{4.0} & 281.8 & 168 & 6.8\\
 &  & 16 & \textbf{1.10} & 89.09 (5) & 1.26 & \textbf{3.1} & 59.4 & 169 & 2.2\\
\midrule
\multirow{4}{*}{\shortstack[c]{Sealed\\corridor}} & \multirow{2}{*}{kinematic} & 8 & 1.72 & 2.85 (50) & \textbf{1.70} & \textbf{6.2} & 255.6 & 223 & 34.4\\
 &  & 16 & \textbf{0.85} & 2.09 (50) & 1.49 & \textbf{6.4} & 519.7 & 298 & 77.6\\
 & \multirow{2}{*}{dynamic} & 8 & \textbf{2.01} & 3.19 (20) & \textbf{2.01} & \textbf{6.6} & 152.7 & 224 & 30.4\\
 &  & 16 & \textbf{1.63} & 2.68 (20) & 2.18 & \textbf{7.4} & 313.2 & 316 & 60.6\\
\midrule
\multirow{2}{*}{\shortstack[c]{Zigzag\\channel}} & \multirow{1}{*}{kinematic} & 16 & 10.68 & \textbf{2.72} (5) & 8.95 & \textbf{5.8} & 88.0 & 301 & 76.2\\
 & \multirow{1}{*}{dynamic} & 16 & \textbf{1.32} & 2.32 (5) & 11.32 & \textbf{6.1} & 105.0 & 307 & 85.6\\
\bottomrule
\end{tabular}
}
\end{table}

\begin{figure}[htbp]
\centering
\includegraphics[width=\textwidth]{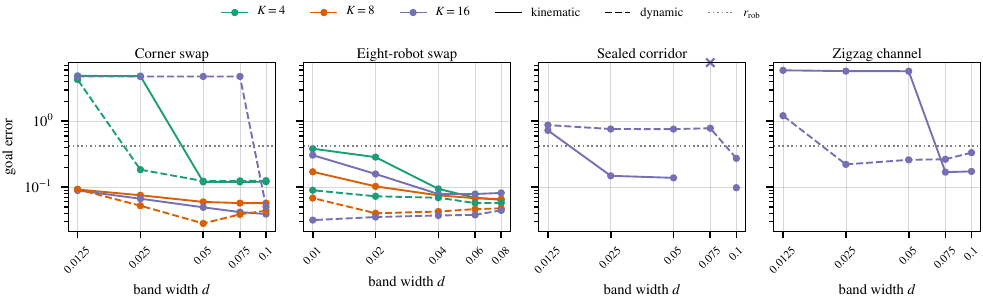}
\caption{Final goal error against band width $d$ under a fixed iteration budget, one panel per scene (color: $K$; solid: kinematic; dashed: dynamic).  The rightmost point is the reference band; the dotted line marks $r_{\rm rob}$, below which the goals are reached; $\times$ marks a run that diverges.}
\label{fig:app_band}
\end{figure}

\paragraph{Ablation study: band width.}  For each case that reaches its goals, we rerun PSK with the band widths $d$, $3d/4$, $d/2$, $d/4$, and $d/8$, where $d$ is the reference band, under a fixed iteration budget equal to the reference run's stopping iteration, capped at $3\cdot10^5$.  The point at $d$ is also rerun under this budget, so its goal error can differ from \prettyref{tab:app_cases}.  To keep the walls equally strong, we scale the wall height $\Gamma$ in proportion to $d$, so the largest wall force $3\Gamma/d$ stays fixed; the wall curvature $6\Gamma/d^2$ then grows like $1/d$, so we also shrink the step sizes $\gamma$ and $c_\alpha$ in proportion to $d$ to keep the iterations stable.  Each run reports its best feasible checkpoint, and \prettyref{fig:app_band} plots the final goal error against $d$.  At half the reference band, \factBandHalfGoal{} of \factNgoal{} cases still reach their goals.  All cases that fail there have $K=16$, and two of them, the dynamic corner swap and the dynamic corridor, already fail at $3d/4$.  At $d/8$, \factBandEighthGoal{} of \factNgoal{} cases reach their goals.  In several cases the goal error grows gradually as the band shrinks; this is expected under a fixed budget, because with step sizes proportional to $d$, each iteration moves the control less.  This effect does not explain the sudden jumps to large goal errors in some runs.  One run (corridor, kinematic, $d=0.075$) diverges.

\paragraph{Ablation study: random start.}  We rerun only the $K=16$ cases, which reach their goals in every scene (\prettyref{tab:app_cases}), with the same parameter settings and stopping rule but four other noise draws.  \prettyref{tab:app_robust} lists the results: \factRobKsixteenGoal{} of \factRobKsixteenTotal{} reruns reach their goals.  All reruns but one keep their exact chain feasible at every checkpoint from $10^3$ until they stop, and their reported controls pass the banded Fritz--John check.  Of all \factRobKsixteenTotal{} reruns, \factRobKsixteenPass{} meet the stopping rule, and the others reach the cap.  Only \factRobKsixteenMiss{} reruns miss their goals, both with the dynamic model: the exception, on the corner swap, ends far from its goals and has an infeasible exact chain at a checkpoint, and one on the zigzag channel ends \factRobNearMissGoal{} from its goals, just above $r_{\rm rob}$.

\begin{table}[htbp]
\centering
\caption{Reruns of the $K=16$ cases with four other noise draws: fraction that reach their goals, and goal error (mean $\pm$ standard deviation over the reruns).}
\label{tab:app_robust}
\begin{tabular}{clcc}
\toprule
scene & model & goals reached & goal err.\\
\midrule
\multirow{2}{*}{\shortstack[c]{Corner\\swap}} & \multirow{1}{*}{kinematic} & 4/4 & 0.04 $\pm$ 0.00\\
 & \multirow{1}{*}{dynamic} & 3/4 & 1.23 $\pm$ 2.38\\
\midrule
\multirow{2}{*}{\shortstack[c]{Eight-robot\\swap}} & \multirow{1}{*}{kinematic} & 4/4 & 0.03 $\pm$ 0.00\\
 & \multirow{1}{*}{dynamic} & 4/4 & 0.05 $\pm$ 0.01\\
\bottomrule
\end{tabular}%
\hspace{1.5em}%
\begin{tabular}{clcc}
\toprule
scene & model & goals reached & goal err.\\
\midrule
\multirow{2}{*}{\shortstack[c]{Sealed\\corridor}} & \multirow{1}{*}{kinematic} & 4/4 & 0.09 $\pm$ 0.01\\
 & \multirow{1}{*}{dynamic} & 4/4 & 0.29 $\pm$ 0.02\\
\midrule
\multirow{2}{*}{\shortstack[c]{Zigzag\\channel}} & \multirow{1}{*}{kinematic} & 4/4 & 0.10 $\pm$ 0.01\\
 & \multirow{1}{*}{dynamic} & 3/4 & 0.29 $\pm$ 0.12\\
\bottomrule
\end{tabular}

\end{table}

\paragraph{Limitations.}

The corridor with $K=8$ is reachable but not solved from a neutral control.  The goal-reaching runs use the trajectory-smoothness term.  The PSK parameters are practical choices tuned per case, including tuning for speed in two cases, and the stopping rule tests a plateau of the loss rather than stationarity.  The practical wall heights stay below the theoretical $\Gamma_{\rm cb}$, so feasibility of the exact states is checked at the checkpoints, not guaranteed.  The comparisons use one machine and one budget.

\ifshownotation
\section{Notation}\label{sec:notation}

\footnotesize
\setlength{\LTpre}{0pt}
\setlength{\LTpost}{0pt}
\setlength{\tabcolsep}{3pt}
\setlength{\aboverulesep}{0pt}
\setlength{\belowrulesep}{0pt}
\renewcommand{\arraystretch}{1.25}
\rowcolors{2}{white}{gray!12}
\begin{longtable}{@{}>{\raggedright\arraybackslash}p{0.18\textwidth}>{\raggedright\arraybackslash}p{0.43\textwidth}>{\raggedright\arraybackslash}p{0.21\textwidth}>{\raggedright\arraybackslash}p{0.12\textwidth}@{}}
\toprule
\rowcolor{gray!24}
Symbol & Definition & Meaning & Ref. \\
\midrule
\endfirsthead
\toprule
\rowcolor{gray!24}
Symbol & Definition & Meaning & Ref. \\
\midrule
\endhead
\bottomrule
\endfoot
\bottomrule
\endlastfoot

\rowcolor{gray!24}
\multicolumn{4}{@{}l}{\textbf{Problem data}}\\
\(K,\ \cX,\ y_0\) & \(K\geq1,\ x\in\cX\subseteq\R^{n_x},\ y_0\in\R^{n_y}\) & Depth, global control set, and initial state. & \prettyref{sec:problem}\\
\(\SelMat_i=\mathbf S_i\) & \(\SelMat_i\in\R^{n_{u,i}\times n_x}\); stage \(i\) reads \(\SelMat_i x\). & Fixed selector for the control coordinates used by stage \(i\). & \prettyref{sec:problem}\\
\(a,\ g_i,\ h_l,\ N_h\) & \(a=(x,p)\) with incoming state \(p\); \(g_i(x,p,y)\) depends on \(x\) through \(\SelMat_i x\), while \(h_l(p,y)\leq0\) are shared constraints. & Stage-\(i\) energy and shared hard constraints. & \eqref{icml:klevel}\\
\(\operatorname{sl}\) & \(\operatorname{sl}(p,y):=\min_l\{-h_l(p,y)\}\). & Constraint slack. & \eqref{eq:nsslack}\\
\(s_i^B,\ \widehat y_i^B,\ \varphi^B\) & \(s_i^B(x,p):=\argmin_y\{g_i(x,p,y):h_l(p,y)\le0\ \forall l\}\); \(\widehat y_i^B=s_i^B(x,\widehat y_{i-1}^B)\), \(\varphi^B=F(x,\widehat y_1^B,\ldots,\widehat y_K^B)\). & Stage-indexed constrained map and chain, and unindexed implicit loss. & \eqref{icml:klevel}, \eqref{eq:traj}\\
\(f_i,\ s_i,\ f_i^\star,\ R_i\) & \(s_i(a):=\argmin_yf_i(a,y)\), \(f_i^\star(a):=\min_yf_i(a,y)\), \(R_i(a,y):=f_i(a,y)-f_i^\star(a)\). & Stage-\(i\) smooth energy, minimizer, value, and gap. & \eqref{eq:stage}, \eqref{eq:gap}\\
\(w,\ \cW,\ a_i\) & \(w=(x,y_1,\ldots,y_K)\in\cW:=\cX\times\R^{Kn_y}\), \(a_i(w):=(x,y_{i-1})\). & Trajectory and stage parameter. & \eqref{eq:w}\\
\(\widehat y_i,\ \varphi\) & \(\widehat y_i=s_i(x,\widehat y_{i-1})\), \(\varphi=F(x,\widehat y_1,\ldots,\widehat y_K)\). & Heterogeneous smooth exact chain and implicit loss. & \eqref{eq:stage}\\
\(F,\ L_F,\ G_F,\ \Flo\) & \(F(w):=F(x,y_1,\ldots,y_K)\), \(\operatorname{Lip}(\nabla F)\leq L_F\), \(\|\nabla F\|\leq G_F\), \(F\geq\Flo\). & Task loss and bounds. & \eqref{eq:w}, \prettyref{ass:ns}, \ref{ass:A}\\
\(\varsigma_i,\bar\varsigma_K,R_X,R_{\SelMat,K},\)\newline \(\bar\varsigma_\infty,R_{\SelMat,\infty}\) & \(\varsigma_i:=\|\SelMat_i\|\), \(\bar\varsigma_K:=\max_{i\leq K}\varsigma_i\), \(R_X:=\max_{x\in\cX}\|x\|\), \(R_{\SelMat,K}:=\max_{i\leq K,x\in\cX}\|\SelMat_i x\|\leq\bar\varsigma_KR_X\); the infinity versions take \(\sup_{K\geq1}\) of the finite-prefix bounds. & Finite-prefix and all-stage selector and selected-control envelopes. & \prettyref{sec:inst_main}, \prettyref{prop:nsrobot}, \ref{prop:dynrobot}\\
\(\mathcal U\);\ (i)--(vi) & \(\mathcal U\subseteq\R^{n_x+n_y}\times\R^{n_y}\) is the convex numerical domain of \prettyref{ass:A}, which holds globally if \(\mathcal U=\R^{n_x+n_y}\times\R^{n_y}\); its items are compact control, stage derivative bounds, strong convexity, outer loss, oracles, and \(C^{2,1}\) bounds. & Smooth-stage assumption, its items, and domain. & \prettyref{sec:P}, \prettyref{ass:A}\\

\rowcolor{gray!24}
\multicolumn{4}{@{}l}{\textbf{Penalties and residuals}}\\
\(r_i(w)\) & \(r_i(w):=\|y_i-s_i(a_i(w))\|\). & Stage solution distance. & \eqref{eq:ri}\\
\(\cL_\rho,\ \Phi_\rho,\ \Phib;\ z,z_i^\star,\rho\) & \(\cL_\rho=F+\rho\sum_i[f_i(a_i,y_i)-f_i(a_i,z_i)]\), where \(z=(z_1,\ldots,z_K)\) are the state copies (\(\cL_\rho\) is maximized at \(z_i^\star(w)=s_i(a_i(w))\)) and \(\rho>0\) is the penalty parameter; \(\Phi_\rho:=\max_z\cL_\rho=F+\rho\sum_iR_i(a_i,y_i)\), \(\Phib:=\Phi_\rho+\iota_\cW\). & Saddle function, state copies, penalty function, and constrained extension. & \eqref{eq:saddle}, \eqref{eq:primal}, \eqref{eq:Phibar}\\
\(\mathcal G_{\alpha,\mathcal Z}^{\mathfrak f}(u)\) & \(\alpha^{-1}\|u-\PiX_{\mathcal Z}(u-\alpha\nabla\mathfrak f(u))\|\). & Scalar projected-gradient residual. & \eqref{eq:G}\\
\(\cG_\alpha^{\mathfrak f},\ \Gphi_\alpha,\ \bar\alpha,\ \epsilon\) & \(\cG_\alpha^{\mathfrak f}:=\mathcal G_{\alpha,\cX}^{\mathfrak f}\), \(\Gphi_\alpha(x)=\cG_\alpha^\varphi(x)\); \(\bar\alpha>0\) is the reference step for evaluating stationarity, and \(x\) is \(\epsilon\)-stationary if \(\Gphi_{\bar\alpha}(x)\leq\epsilon\). & Control-set shorthand, smooth implicit residual, reference step, and accuracy. & \eqref{eq:G}, \prettyref{sub:smooth}\\
\(\cG_\alpha^{\varphi^C}\) & \(\cG_\alpha^{\varphi^C}(x)=\mathcal G_{\alpha,\cX}^{\varphi^C}(x)\). & Constant-band implicit residual. & \eqref{eq:nsGphi}\\
\(\cG_{*,n_x}^{\mathfrak f},\ \cG_*^{\mathfrak f}\) & \(\cG_{*,n_x}^{\mathfrak f}(x):=\mathcal G_{\alpha_*,\cX_{n_x}}^{\mathfrak f}(x)\); suppress \(n_x\) when clear. & Lower-bound residual at the reference step. & \eqref{icml:Gstar}\\
\(\cG_\alpha^\rho\) & \(\cG_\alpha^\rho(w):=\mathcal G_{\alpha,\cW}^{\Phi_\rho}(w)\). & Penalized scalar residual. & \eqref{icml:Grho}\\
\((\cG_\alpha^\rho(w))^2\) & \(\alpha^{-2}\|x-\PiX_\cX(x-\alpha\nabla_x\Phi_\rho)\|^2+\sum_i\|\nabla_{y_i}\Phi_\rho\|^2\). & Product-space residual identity. & \eqref{icml:Gprod}\\

\rowcolor{gray!24}
\multicolumn{4}{@{}l}{\textbf{Shared smooth constants and PSK}}\\
\(\mu,\ L_{yy},\ L_{ay},\ L_{aa},\ L_f\) & Common finite-family envelopes for strong convexity and derivative bounds of the \(f_i\) on \(\mathcal U\). & Uniform stage regularity. & \prettyref{ass:A}, \eqref{eq:La}\\
\(L_H,\ H_i,\ J_i\) & \(H_i:=\nabla^2_{yy}f_i\), \(J_i:=\partial_a\nabla_yf_i\), with uniform Hessian-Lipschitz modulus \(L_H\). & Stage-\(i\) second-order data. & \prettyref{A:C2}, \eqref{eq:HJ}\\
\(M,\ L_\star,\ \chi\) & \(M:=L_{ay}/\mu\), \(L_\star:=L_{aa}+L_{ay}M\), \(\chi:=L_{yy}/\mu\). & Stage-map modulus, value-gradient modulus, and condition number. & \eqref{eq:rhofree}\\
\(m,\ \gamma,\ q,\ \widetilde q\) & \(m\geq1\), \(0<\gamma\leq L_{yy}^{-1}\), \(q:=(1-\gamma\mu)^{1/2}\), \(\widetilde q:=q^m\). & Inner budget, inner step, and contractions. & \eqref{eq:rhofree}, \prettyref{alg:psk}, \prettyref{sub:smooth}\\
\(W_i,\ \kappa_0\) & \(W_i:=\sum_{k=0}^{K-i}M^k\), \(\kappa_0:=(K+1)(L_{ay}\widetilde q)^2\). & Depth sums and certificate weight. & \eqref{eq:depthsums}, \eqref{eq:kappa0}\\
\(c_\alpha,c_\alpha^{\max},c_\Psi,\)\newline \(\omega_1,\omega_2,\omega_3,\)\newline \(\beta,\beta_{\max};\ \rho_0,\rho_t,\alpha_t\) & Outer step and cap, potential, drift coefficients, penalty growth and cap, with \(\rho_t=\rho_0(1+\beta t)^{1/3}\) and \(\alpha_t=c_\alpha/\rho_t\). & Smooth-run constants and schedule. & \eqref{eq:amax}, \eqref{eq:rhoconsts}, \eqref{eq:omega}, \eqref{eq:betamax}\\
\(\mathfrak H_i,\mathfrak H_2^{\rm all},D_i^{\rm der},\)\newline \(\vartheta_i^{\rm der},\mathfrak B_i;\)\newline \(c_{\rm adj},c_\varphi,c_{\rm b1},c_{\rm b2}\) & Depth-weighted trajectory-distance sums and their \(\ell_2\) norm, derivative-distance and derivative-modulus factors, adjoint-bias sums, and the constants converting penalized stationarity to implicit stationarity. & Conversion constants. & \eqref{eq:convconsts}\\
\(C_E(B),C_\rho,C^\star(B)\) & Residual, reciprocal-penalty, and implicit-rate maps at budget \(B\). & Smooth-rate constants. & \eqref{eq:Cs}\\
\(\mathsf T_{i,a},b_{i,t},\)\newline \(b_{i,t}^{(k)},\xi_{i,t}^{(k)}\) & \(\mathsf T_{i,a}(y):=y-\gamma\nabla_yf_i(a,y)\); \(b_{i,t}\) is the stage-\(i\) tracker, \(b_{i,t}^{(k)}\) its \(k\)th inner iterate, and \(\xi_{i,t}^{(k)}\) its \(k\)th inner gradient. & Tracker step and registers. & \eqref{eq:T}, \prettyref{alg:psk}\\
\(\hnabla_t,\ \delta_t\) & \(\hnabla_t=\nabla_w\cL_{\rho_t}(w_t,b_{t+1})\), with blocks \(\hnabla_t^x,\hnabla_t^{y_i}\); \(\delta_t:=\hnabla_t-\nabla\Phi_{\rho_t}(w_t)\), with blocks \(\delta_t^x,\delta_t^{y_i},\delta_t^y\). & Search direction and direction error. & \prettyref{alg:psk}, \eqref{icml:blocks}, \eqref{eq:epsdef}\\
\(e_{i,t},\widehat e_{i,t+1},\Delta w_t\) & \(e_{i,t}:=\|b_{i,t}-s_i(a_i(w_t))\|\), \(\widehat e_{i,t+1}:=\|b_{i,t+1}-s_i(a_i(w_t))\|\), \(\Delta w_t:=w_{t+1}-w_t\). & Tracker errors and outer increment. & \eqref{eq:errs}\\
\(\Xi_t,\ \Ecal_t\) & \(\Xi_t:=\sum_i e_{i,t}^2\), \(\Ecal_t:=\cG_{\alpha_t}^{\rho_t}(w_t)\). & Aggregate tracking error and scalar residual. & \prettyref{sec:statements}, \eqref{icml:Gprod}\\
\(T,\ C_t,\ t^\star\) & \(T\geq1\) is the number of outer iterations; \(C_t\) is the computable PSK certificate; \(t^\star\in\argmin_{\lfloor T/2\rfloor\leq t<T}C_t\). & Horizon, certificate, and output index. & \prettyref{alg:psk}\\
\(\Psi_t,U_t,B_0\) & \(\Psi_t:=\Phi_{\rho_t}(w_t)+c_\Psi\rho_t\Xi_t\), with \(U_t\) and \(B_0\) as displayed. & Potential, decrement, and budget. & \eqref{eq:PsiUB}\\
\((\mathrm H_\rho)\) & Conditions on \(m,\gamma,c_\alpha,\beta,\rho_0\) and on the schedule \(\rho_t,\alpha_t\). & Hypothesis of the Lyapunov inequality. & \eqref{eq:Hrho}, \prettyref{thm:lyap}\\

\rowcolor{gray!24}
\multicolumn{4}{@{}l}{\textbf{Constant-band (cb) theory}}\\
\(d,\ \Gamma,\ \psi_{d,\Gamma}\) & \(d>0\) is the band width, \(\Gamma>0\) its height, and \(\psi_{d,\Gamma}\) is the displayed cubic wall. & Smooth finite wall. & \eqref{eq:nswall}\\
\(f_i^C,\ s_i^C\) & \(f_i^C:=g_i+\sum_l\psi_{d,\Gamma}(-h_l)\), \(s_i^C:=\argmin_y f_i^C\). & Stage-\(i\) constant-band surrogate and map (exact center). & \eqref{eq:nscbsurrogate}, \prettyref{sec:ns}\\
\(\widehat y_i^C,\ \varphi^C\) & \(\widehat y_i^C:=s_i^C(x,\widehat y_{i-1}^C)\), \(\varphi^C:=F(x,\widehat y_1^C,\ldots,\widehat y_K^C)\). & Heterogeneous constant-band exact chain and exact chain loss. & \eqref{eq:nscbchain}\\
\(d\text{-banded Fritz--John}\) & Exact-state certificate in \prettyref{def:nscert}. & Banded Fritz--John notion. & \prettyref{def:nscert}\\
\(\varepsilon_{\rm KKT}\text{-KKT}\) & Approximate KKT point in \prettyref{def:nskkt}. & Appendix-only approximate KKT notion. & \prettyref{def:nskkt}\\
\(\vartheta,\ \lambda,\ \lambda_{i,l}(x)\) & \(\vartheta\) and \(\lambda\) are Fritz--John and KKT multipliers; \(\lambda_{i,l}(x):=-\psi'_{d,\Gamma_{\rm cb}}(-h_l(\widehat y_{i-1}^C,\widehat y_i^C))\). & Exact-state certificate multipliers. & \prettyref{def:nscert}, \ref{def:nskkt}, \eqref{icml:nsmult}\\
(i)--(v) & Items of \prettyref{ass:ns}: compact control, outer loss, all-stage energy envelopes, smooth constraints, and outgoing convexity. & Base constrained-stage class. & \prettyref{ass:ns}\ifshowremarks, \prettyref{rem:nsguniform}\fi\\
\(\mathcal D;\ L_{yy,\mathcal D}^g,L_{ay,\mathcal D}^g,\)\newline \(L_{aa,\mathcal D}^g,L_{f,\mathcal D}^g,L_{H,\mathcal D}^g\) & On every bounded convex \(\mathcal D\subseteq\R^{n_x+n_y}\times\R^{n_y}\), all-stage bounds on \(\nabla^2_{yy}g_i\), \(\partial_a\nabla_yg_i\), \(\nabla^2_{aa}g_i\), \(\nabla^2g_i\), and \(\operatorname{Lip}(\nabla^2g_i)\). & Base-class derivative envelopes. & \prettyref{ass:ns}\\
\(0_K,\mathrm{F}_K,\)\newline \(\mathrm{XT}_K,\mathrm{NI}_K\) & The four items of \prettyref{ass:nscb}, the finite-horizon constant-band class. & Constant-band class. & \prettyref{ass:nscb}\\
\(\Ball_r(c)\) & \(\Ball_r(c):=\{u:\|u-c\|\leq r\}\). & Closed ball of radius \(r\) about \(c\). & \prettyref{ass:nscb}\\
\(\mathcal S_i,\mathcal Y_i,r_{\rm in}\) & \(\mathcal S_i\subseteq\mathcal Y_i\), the \(\mathcal Y_i\) are nested, and \(\Ball_{r_{\rm in}}(s_i^C(x,p))\subseteq\mathcal Y_i\). & Exact cores, numerical tubes, and inward radius. & \eqref{eq:nscbnested}, \eqref{eq:nscbonestep}\\
\(\tau_{\rm cb},\sigma_*,\)\newline \(d,b_{\rm cb}\) & \(d:=(\tau_{\rm cb}+\sigma_*)/2\), \(b_{\rm cb}:=\tau_{\rm cb}/d\). & Exact margin, comparison slack, band width, and band fraction. & \prettyref{ass:nscb}, \eqref{eq:nscbdata}\\
\(\Upsilon_{\rm cb},\Lambda_{\rm cb},\)\newline \(\Gamma_{\rm cb}\) & \(\Lambda_{\rm cb}:=\Upsilon_{\rm cb}/(\sigma_*-d)\), \(\Gamma_{\rm cb}:=d\Upsilon_{\rm cb}/[3(\sigma_*-d)(1-b_{\rm cb})^2]\) for \(N_h>0\). & Force bound, multiplier cap, and wall height. & \eqref{eq:nscbdata}\\
\(\mathcal D_{\rm cb},\mathcal W_{\rm box},\)\newline \(\mathcal B_i,\mathcal W_{\rm cb}\) & Common derivative box, box of tubes \(\mathcal W_{\rm box}:=\cX\times\mathcal Y_1\times\cdots\times\mathcal Y_K\), stagewise fiber \(\|y-s_i^C(x,p)\|\leq r_{\rm in}\), and localized trajectory tube. & Constant-band localization sets. & \eqref{eq:nscbbox}, \prettyref{sec:ns}\\
\(\gamma,\rho_0,\)\newline \(\beta,c_\alpha\) & \(\gamma:=(2L_{yy})^{-1}\), \(c_\alpha\) from \eqref{eq:nscalpha}, \(\beta:=\beta_{\max}\), \(\rho_0\) from \eqref{eq:nsrho0}; \(\rho_t=\rho_0(1+\beta t)^{1/3}\), \(\alpha_t=c_\alpha/\rho_t\). & Constant-band PSK schedule. & \prettyref{sec:nsschedule}, \eqref{eq:nscalpha}, \eqref{eq:nsrho0}\\
\(R_{{\rm tube},K},D_K,\)\newline \(F_{\rm hi},B_{\rm cb}\) & \(R_{{\rm tube},K}:=\max_{y\in\mathcal Y_K}\|y\|\), \(D_K:=\operatorname{diam}\mathcal Y_K\), \(F_{\rm hi}:=\max_{\mathcal W_{\rm box}}F\), \(B_{\rm cb}:=F_{\rm hi}-\Flo+2\). & Horizon localization data. & \prettyref{sec:ns} and \ref{sec:nsschedule}\\
\(r_{\rm cb},\)\newline \(r_{\rm warm},m_0\) & Localization radius, warm-start radius, and warm-start budget. & Bounded-anchor initialization. & \eqref{eq:nslocalization}, \eqref{eq:nswarmradius}, \eqref{eq:nswarmsteps}\\
\(r_{i,t},\ e_{i,t}^{(k)},\ w_t(\theta)\) & \(r_{i,t}=r_i(w_t)\), \(e_{i,t}^{(k)}:=\|b_{i,t}^{(k)}-s_i^C(a_i(w_t))\|\), \(w_t(\theta):=w_t+\theta\Delta w_t\). & Outer-run residuals and segment. & \prettyref{sec:nsschedule}\\
\(C_{\rm cb}^\star\) & The map \(C^\star\) from \eqref{eq:Cs} under the specialization \eqref{eq:nsinstantiate}. & Constant-band rate map. & \eqref{eq:nsinstantiate}, \prettyref{thm:nsrecover}\\

\rowcolor{gray!24}
\multicolumn{4}{@{}l}{\textbf{Lower-bound model}}\\
\(\Psi_{\rm c},\Phi_{\rm c},\bar f_N\) & Smooth cutoff, Gaussian primitive, and Carmon first-order zero chain. & Zero-chain construction. & \eqref{eq:lbcdhs}, \prettyref{sec:lbzerochain}\\
\(L_1,L_2\) & Absolute bounds on \(\|D^2\bar f_N\|\) and \(\|D^3\bar f_N\|\). & Zero-chain smoothness constants. & \prettyref{lem:lbzerochain}\\
\(\Gamma,\sigma_{\rm rel},\)\newline \(\mathfrak m,c_{\mathfrak q}\) & \(\Gamma:=128\), \(\sigma_{\rm rel}:=1/100\), \(\mathfrak m:=3\), \(c_{\mathfrak q}:=1/(4\sigma_{\rm rel}^{K-2})\); \(\mathfrak m\) is the tilt coefficient and \(c_{\mathfrak q}\) the signal scale. & Fixed relay constants. & \prettyref{sec:lbinstance}\\
\(\theta\) & \(\theta:=\sqrt\epsilon\). & Relay angle scale. & \prettyref{sec:lbinstance}\\
\(P_\zeta,\chi_\zeta,\zeta\) & \(P_\zeta(t):=1-10t^3+15t^4-6t^5\) for \(0\leq t\leq1\); \(\chi_\zeta(\varrho):=\begin{cases}1,&|\varrho|\leq1,\\ P_\zeta((|\varrho|-1)/4),&1<|\varrho|<5,\\0,&|\varrho|\geq5\end{cases}\); \(\zeta(\varrho):=\int_0^\varrho\chi_\zeta(s)\,\mathrm ds\). & Limiter and its cutoff derivative. & \prettyref{sec:lbinstance}, \eqref{eq:lbzetabounds}\\
\(u,v,\varrho;\)\newline \(p_u,p_v,p_\varrho;\ \ell\) & \(y=(u,v,\varrho)\), \(p=(p_u,p_v,p_\varrho)\), \(\ell(p):=p_v+p_\varrho\). & Relay coordinates and incoming scalar. & \eqref{eq:lbinstance}\\
\(h_+,h_-,\)\newline \(g_i=g,F\) & The two affine relay walls, diagonal tilted strongly convex stage family with signal \(\mathfrak q\), and task loss \(F=\tanh(v_K)\). & \(K\)-level relay. & \eqref{eq:lbinstance}\\
\(f_{i,d}^C,s_{i,d}^C,\)\newline \(\widehat y_{i,d}^C,\varphi_d^C\) & Stage-\(i\) width-\(d\), height-\(\Gamma\) surrogate and map, exact chain, and exact chain loss; on the relay \(f_{i,d}^C=f_d^C\), \(s_{i,d}^C=s_d^C\). & Shrinking-band family. & \eqref{eq:lbwallrule}, \eqref{eq:lbwallchain}\\
\(\theta_K,\bar c_K,\kappa_d\) & Constants of the amplification lemma; \(\nabla\varphi_d^C(x)=\kappa_d(x)\nabla\mathfrak q(x)\) with \(|\kappa_d|\geq\bar c_K\theta^{-(K-1)}\). & Depth-amplified derivative gain. & \prettyref{lem:lbamplifier}, \eqref{eq:lbdepthgain}\\
\(c_r,c_{\rm len}\) & \(c_r:=2/\sqrt{\bar c_K}\), \(c_{\rm len}:=c_{\mathfrak q}/24\). & Zero-chain radius and length scales. & \prettyref{sec:lbinstance}\\
\(r_{\rm zc},A_{\rm zc},\)\newline \(\widetilde N_\theta,N_{\rm zc}\) & \(r_{\rm zc}:=c_r\theta^{(K+1)/2}\), \(A_{\rm zc}:=r_{\rm zc}^3\), \(\widetilde N_\theta:=c_{\rm len}\theta^K/r_{\rm zc}^3\), \(N_{\rm zc}:=\lfloor \widetilde N_\theta\rfloor\). & Signal scaling and control dimension. & \eqref{eq:lbscaling}, \eqref{eq:lblength}\\
\(\mathfrak q,\ \mathsf I_\epsilon,\ \varphi_\epsilon^C\) & \(\mathfrak q(x):=A_{\rm zc}[\bar f_{N_{\rm zc}}(x/r_{\rm zc})-\bar f_{N_{\rm zc}}(0)]\); \(\mathsf I_\epsilon\) is the relay with this signal, and \(\varphi_\epsilon^C\) its exact chain loss at \(d=\epsilon\). & Signal, hard instance, and its loss. & \prettyref{sec:lbinstance}, \eqref{icml:lbqbounds}, \eqref{eq:lbqbox}\\
\(\alpha_*,\ \cX_{n_x}\) & \(\alpha_*>0\) is the fixed reference step and \(\cX_{n_x}:=[-1,1]^{n_x}\). & Lower-bound stationarity data. & \prettyref{sec:lbmbzr}, \eqref{icml:Gstar}\\
\(\mathsf B,\mathsf O,\)\newline \(\mathsf o^\tau,\mathsf z^\tau\) & \(\mathsf B:=\{x,0,\ldots,K\}\) indexes trajectory blocks; \(\mathsf O:=\{F,\cX\}\cup\{(g_i,i):1\leq i\leq K\}\cup\{(h_l,i):1\leq l\leq N_h,\ 1\leq i\leq K\}\); \(\mathsf o^\tau\) and \(\mathsf z^\tau\) are the query label and point. & Oracle labels and block slots. & \eqref{icml:oracleset}, \eqref{eq:lbsignature}, \prettyref{sec:lbmbzr}\\
\(\mathcal O_F^{\mathsf I},\mathcal O_{g_i,i}^{\mathsf I},\)\newline \(\mathcal O_{h_l,i}^{\mathsf I},\mathcal O_\cX^{\mathsf I};\)\newline \(\mathsf r^\tau\) & Stage-labeled primitive first-order and projection oracles and their reply. & Oracle model. & \eqref{eq:lboracles}, \prettyref{sec:lbmbzr}\\
\(\mathsf{info},\mathsf Q_\tau,\)\newline \(\mathsf R_\tau,x^\tau\) & Parameter map, query rules, output rules, and the output after \(\tau\) calls. & Deterministic algorithm. & \prettyref{def:lbmodel}\\
\(\mathfrak A_{\rm zr},\mathsf S_{\mathsf b}^\tau\) & Class of multi-block zero-respecting algorithms and their revealed block supports. & Lower-bound algorithm class. & \prettyref{def:lbmodel}\\
\(A_{\rm PSK},c_{\rm warm},\)\newline \(c_{\rm ph}\) & PSK as a primitive-query algorithm; \(c_{\rm warm}:=(1+N_h)Km_0\), \(c_{\rm ph}:=K(m+2)(1+N_h)+2\). & Zero-respecting PSK and call counts. & \prettyref{def:lbunrolled}\\
\(\mathcal F_{1,2,2}\) & \(\mathcal F_{1,2,2}(\Delta,L_F,G_F,\mu,L_H^g,L_H^h)\), with fixed \(n_y,N_h,y_0,\Gamma\) and control set \(\cX_{n_x}\); it satisfies \prettyref{ass:ns} (\prettyref{lem:lbclassns}). & Lower-bound instance class. & \prettyref{def:lbclass}\\
\(\operatorname{Calls}_\epsilon\) & \(\operatorname{Calls}_\epsilon(A,\mathsf I)\) is the first call count with residual at most \(\epsilon\); \(\operatorname{Calls}_\epsilon(\mathfrak A,\mathcal F)\) is the minimax complexity. & Residual primitive-call complexity. & \prettyref{def:lbcomplexity}\\
\(c_K,\epsilon_K\) & For \(0<\epsilon\leq\epsilon_K\), at least \(c_K\epsilon^{-(K+3)/4}\) primitive calls are required. & Shrinking-band lower-bound constants. & \prettyref{thm:lbmain}, \ref{thm:lbpsk}\\
\(d,\tau_{\rm cb},\sigma_*,\)\newline \(\Upsilon_{\rm cb},\Lambda_{\rm cb},\Gamma_{\rm cb}\) & Relay specialization: \(\tau_{\rm cb}=d/2\), \(\sigma_*=3d/2\), \(\Upsilon_{\rm cb}=48\), \(\Lambda_{\rm cb}=96/d\), \(\Gamma_{\rm cb}=128\). & Per-width constant-band data. & \eqref{eq:lbcbdata}\\

\rowcolor{gray!24}
\multicolumn{4}{@{}l}{\textbf{Kinematic multi-robot navigation}}\\
\(N_{\rm rob},r_{\rm rob},\)\newline \(N_{\rm obs},p^{\rm goal},\delta_F\) & Robot count and radius, number of segment obstacles, goal positions, and pseudo-Huber scale \(\delta_F\). & Navigation and task-loss inputs. & \prettyref{sub:motivating}, \eqref{icml:navF}\\
\(\sigma_0,\ r_{\rm tail}\) & \(\sigma_0>0\) is the initial-clearance margin, and \(r_{\rm tail}>0\) is the exact-tail radius. & Persistent clearance and smoothing parameters. & \prettyref{lem:nsjam}, \prettyref{sec:instances}\\
\(j,\ S_j,\ o_{j,0},\)\newline \(o_{j,1},R_O\) & \(S_j:=\operatorname{conv}\{o_{j,0},o_{j,1}\}\), \(R_O:=\max_{j,k}\|o_{j,k}\|\). & Obstacle index, endpoints, and radius bound. & \prettyref{sec:instances}, \ref{sec:inst_aux}\\
\(\cE,\cE_{\rm p},\cE_{\rm o}\) & \(\cE=\cE_{\rm p}\cup\cE_{\rm o}\), with pair edges \(\{r,r'\}\) and obstacle edges \((r,j)\). & Navigation edge sets. & \prettyref{sub:motivating}, \prettyref{sec:instances}\\
\(\mathbf B_e,S_e,R_e\) & \(\mathbf B_ey=y_r-y_{r'}\), \(S_e=\{0\}\) for \(e=\{r,r'\}\); \(\mathbf B_ey=y_r\), \(S_e=S_j\) for \(e=(r,j)\); \(R_e:=\max_{v\in S_e}\|v\|\). & Edge geometry. & \prettyref{sec:inst_smooth}, \ref{sec:inst_aux}\\
\(\PiX_S,\operatorname{dist}(\cdot,S),\)\newline \(\sigma_S,\Sigma_S,N_S\) & For nonempty compact convex \(S\subset\R^{n_S}\): Euclidean projection and distance; \(\sigma_S(n):=\max_{v\in S}\langle n,v\rangle\), \(\Sigma_S(n;y):=\min_{v\in S}\langle n,y-v\rangle\), \(N_S(y):=(y-\PiX_Sy)/\operatorname{dist}(y,S)\) for \(y\notin S\). & Convex-set geometry. & \prettyref{sec:instances}, \ref{sec:inst_smooth}\\
\(\mathfrak d,\mathfrak d_e,\mathfrak d_0\) & \(\mathfrak d:=\operatorname{dist}(z,S)\) for a segment \(S\); \(\mathfrak d_e:=\operatorname{dist}(p_r,S_j)\) for \(e=(r,j)\); \(\mathfrak d_0:=2r_{\rm rob}+\tau_{\rm cb}\). & Distances to segments and their persistent lower bound. & \prettyref{lem:segproj}, \ref{lem:nsjam}\\
\(\nu\) & \(\displaystyle \nu(u):=u\left[\|u\|^2+\frac{(r_{\rm tail}^2-\|u\|^2)_+^4}{r_{\rm tail}^6}\right]^{-1/2}\). & Exact-tail mollified normal. & \eqref{icml:nu}\\
\(\eta>0,\ \phi,\ t_\eta,\)\newline \(c_\eta,u_\eta,\nu_\eta\) & For endpoints \(o_0,o_1\) and a query point \(z\), \(\phi(t):=\tfrac12\|z-o_0-t(o_1-o_0)\|^2-\eta\log t-\eta\log(1-t)\); \(t_\eta:=\argmin_{t\in(0,1)}\phi(t)\), \(c_\eta:=o_0+t_\eta(o_1-o_0)\), \(u_\eta:=z-c_\eta\), \(\nu_\eta(z;o_0,o_1):=\nu(u_\eta)\). & Barrier-smoothed segment-normal family. & \eqref{icml:barrier}, \prettyref{lem:segsmooth}, \ref{lem:segproj}\\
\(\operatorname{clr}_e,\operatorname{clr}\) & \(\operatorname{clr}_e(y):=\operatorname{dist}(\mathbf B_ey,S_e)-2r_{\rm rob}\), \(\operatorname{clr}(y):=\min_{e\in\cE}\operatorname{clr}_e(y)\). & True edge slack and true clearance. & \prettyref{sec:inst_smooth}\\
\(\mathcal R_{>0},\mathcal R_{\ge c},\)\newline \(n_e,\nabla\operatorname{clr}_e\) & \(\mathcal R_{>0}:=\{y:\operatorname{clr}(y)>0\}\), \(\mathcal R_{\ge c}:=\{y:\operatorname{clr}(y)\geq c\}\) for \(c>0\); \(\nabla\operatorname{clr}_e(y):=\mathbf B_e^\top N_{S_e}(\mathbf B_ey)\); at a predecessor \(p\in\mathcal R_{>0}\), \(n_e(p):=N_{S_e}(\mathbf B_ep)\). & Clear set, clearance superlevel set, exact edge normal, and clearance gradient. & \eqref{eq:kinexact}, \prettyref{sec:inst_aux}\\
\(h_e,h_{e,k},\)\newline \(N_h,\operatorname{sl}\) & Exact linearized constraints \eqref{eq:kinexact}; the analysis uses their smoothed versions \eqref{eq:nspair} and \eqref{eq:nsobstacle}, one per pair edge and two per obstacle edge; \(N_h=|\cE_{\rm p}|+2|\cE_{\rm o}|\); \(\operatorname{sl}(p,y):=\min_l\{-h_l(p,y)\}\), with \(\operatorname{sl}\equiv+\infty\) if \(N_h=0\). & Kinematic constraints and their slack. & \eqref{eq:kinexact}, \eqref{eq:nspair}, \eqref{eq:nsobstacle}\\
\(\mathcal J,\lambda^{\rm jam}\) & \(\mathcal J\) is the positive jam-value set; \(\lambda^{\rm jam}\in\R_+^{\cE}\) is its normalized stress. & Jam levels and stress. & \eqref{eq:nsjamfinite}\\
\(\operatorname{act},\mathcal N\) & \(\operatorname{act}(y):=\{e:\operatorname{clr}_e(y)=\operatorname{clr}(y)\}\), and \(\mathcal N\) is the product of edge unit spheres. & Active edges and normal tuples. & \prettyref{sec:inst_aux}\\
\(\delta_{\rm jam},\Delta_{\rm jam},\)\newline \(Z_{\rm jam},\bar y(p)\) & \(\delta_{\rm jam}\) is the positive dual gap, \(\Delta_{\rm jam}:=\delta_{\rm jam}/2\); \(\bar y(p)\) is the least-norm comparison point and \(Z_{\rm jam}:=\sup\|\bar y(p)\|\). & Uniform jam-regeneration data. & \prettyref{sec:inst_aux}\\
\(\eta_0\) & \(\eta_0:=[\Delta_{\rm jam}\mathfrak d_0/(2(Z_{\rm jam}+R_O))]^2\) when \(N_{\rm obs}\geq1\), and \(+\infty\) otherwise. & Maximum mollifier parameter used by regeneration. & \prettyref{sec:inst_aux}\\
\(\bar h,C_{\rm wall},\)\newline \(R_{\rm ctr},R_{\rm kin}\) & \(\bar h:=2r_{\rm rob}+\max\{\sqrt2R_{\SelMat,\infty},R_{\SelMat,\infty}+R_O\}\); \(C_{\rm wall},R_{\rm ctr},R_{\rm kin}\) are defined in \eqref{eq:nsrobotR}. & Upper bound on the constraint values at the trial state \(\SelMat_ix\), wall-value bound, center radius, and kinematic tube radius. & \prettyref{prop:nsrobot}, \eqref{eq:nsrobotR}\\
\(\Upsilon_{\rm kin}\) & \(\Upsilon_{\rm kin}:=(\mu(Z_{\rm jam}+R_{\SelMat,\infty})+1)^2/(4\mu)\), independent of \(K\). & Kinematic comparison-force bound. & \prettyref{prop:nsrobot}\\

\rowcolor{gray!24}
\multicolumn{4}{@{}l}{\textbf{Dynamic multi-robot navigation}}\\
\(p^{\pos},p^{\vel},\)\newline \(y^{\pos},y^{\vel}\) & \(p=(p^{\pos},p^{\vel})\), \(y=(y^{\pos},y^{\vel})\in\R^{2n_q}\), \(n_q=2N_{\rm rob}\); kinematic model: \(y^{\pos}:=y\). & Incoming and outgoing position--velocity blocks. & \prettyref{sub:motivating}, \prettyref{sec:instances}\\
\(\SelMat_i x,m_{\rm mass},\Delta t\) & Selected stacked force, mass \(m_{\rm mass}>0\), and time step \(\Delta t>0\). & Dynamic-stage inputs. & \eqref{eq:dynobjective}\\
\(\mu_{\rm dyn},u_i^{\pos},u_i^{\vel}\) & Strong-convexity modulus and the stage-\(i\) semi-implicit Euler center. & Dynamic conditioning. & \eqref{eq:dynspectrum}, \eqref{eq:dyncenter}\\
\(h_l^{\rm dyn},\operatorname{sl}\) & \(h_l^{\rm dyn}(p,y):=h_l(p^{\pos},y^{\pos})\), \(\operatorname{sl}(p,y):=\min_l\{-h_l^{\rm dyn}(p,y)\}\). & Lifted collision geometry. & \eqref{eq:dynwalls}, \prettyref{prop:dynrobot}\\
\(\mathsf P(R),\mathsf V(R),\)\newline \(\kappa_{\rm dyn},T_{\rm ex}(R)\) & Selected-control position and velocity center bounds, comparison amplification, and exact transfer bound. & Exact dynamic transfer data. & \eqref{icml:dynexdata}\\
\(R_{{\rm ex},i},R_{{\rm num},i}\) & Recursive radii of the exact cores and numerical tubes. & Finite-horizon dynamic radii. & \eqref{eq:dynexactradii}, \eqref{eq:dynnumradii}\\
\(G_{\rm dyn},\Upsilon_{\rm dyn}\) & Dynamic gradient bound and comparison-force bound; both depend on \(K\). & Dynamic wall data. & \eqref{eq:dynforceconstants}\\
\(\bar h(R),\mathsf C(R),\)\newline \(T_{\rm num}(R)\) & Upper bound on the constraint values at the Euler center, wall-value bound, and numerical transfer bound. & Numerical tube data. & \eqref{icml:dynnumdata}, \eqref{eq:dynnumtransfer}\\
\end{longtable}
\normalsize

\fi

\end{document}